\documentclass[10pt,journal,compsoc]{IEEEtran}
\usepackage{times}
\usepackage{soul}
\usepackage{url}
\usepackage[hidelinks]{hyperref}
\usepackage[utf8]{inputenc}
\usepackage[small]{caption}
\usepackage{graphicx}
\usepackage{amsmath}
\usepackage{booktabs}
\usepackage{algorithm}
\usepackage{algorithmic}
\usepackage{amsthm}
\usepackage{epstopdf}
\usepackage{amssymb}
\usepackage{subfigure}
\usepackage{float}
\usepackage{color}
\usepackage{cite}
\usepackage{multirow}
\usepackage{hyperref}
\ifCLASSINFOpdf
\else
\fi

\begin{document}
\title{Region-wise Generative Adversarial Image Inpainting for Large Missing Areas}

\author{
Yuqing Ma, Xianglong~Liu, Shihao Bai, Lei Wang, Aishan Liu, Dacheng Tao, Edwin Hancock

\thanks{Y. Ma, X. Liu, S. Bai, L. Wang, and A. Liu are with the State Key Lab of Software Development Environment, Beihang University, Beijing 100191, China. X. Liu is also with Beijing Advanced Innovation Center for Big Data-Based Precision Medicine (Corresponding author: Xianglong Liu, xlliu@nlsde.buaa.edu.cn)}
\thanks{D. Tao is with the UBTECH Sydney Artificial Intelligence Centre and the School of Information Technologies, Faculty of Engineering and Information Technologies, The University of Sydney, Darlington, NSW 2008, Australia}% <-this % stops a space
\thanks{E. Hancock is with the Department of Computer Science, University of York, York, U.K.}
}


\IEEEtitleabstractindextext{%
\begin{abstract}
Recently deep neutral networks have achieved promising performance for filling large missing regions in image inpainting tasks. They usually adopted the standard convolutional architecture over the corrupted image, leading to meaningless contents, such as color discrepancy, blur and artifacts. Moreover, most inpainting approaches cannot well handle the large contiguous missing area cases. To address these problems, we propose a generic inpainting framework capable of handling with incomplete images on both contiguous and discontiguous large missing areas, in an adversarial manner. From which, region-wise convolution is deployed in both generator and discriminator to separately handle with the different regions, namely existing regions and missing ones. Moreover, a correlation loss is introduced to capture the non-local correlations between different patches, and thus guides the generator to obtain more information during inference. With the help of our proposed framework, we can restore semantically reasonable and visually realistic images. Extensive experiments on three widely-used datasets for image inpainting tasks have been conducted, and both qualitative and quantitative experimental results demonstrate that the proposed model significantly outperforms the state-of-the-art approaches, both on the large contiguous and discontiguous missing areas. 
\end{abstract}

}

\maketitle

\section{Introduction}
Image inpainting (i.e., image completion or image hole-filling), synthesizing visually realistic and semantically plausible contents in missing regions, has attracted great attentions in recent years. It can be widely applied in many tasks \cite{Barnes:2009:PAR,newson2014video,park2017transformation,simakov2008summarizing}, such as photo editing, image-based rendering, computational photography, etc. In recent decades, there have been many image inpainting methods proposed for generating desirable contents in different ways. For instance, context encoders \cite{pathak2016context} first exploit GANs to restore images, using a channel-wise fully connected layer to propagate information between encoder and decoder. To perceptually enhance image quality, several studies \cite{yang2017high,song2018contextual,wang2018image} attempted to extract features using a pre-trained VGG network to reduce the perceptual loss \cite{Johnson2016Perceptual} or style loss \cite{gatys2015texture}. More recently, \cite{liu2018image,yu2018free,nazeri2019edgeconnect} further concentrated on irregular missing regions and achieved satisfying performance especially for the highly structured images. 

\begin{figure}[tp!]
    \vspace{0.02in}
    \centering
    \begin{subfigure}
        \centering
        \includegraphics[width=2cm]{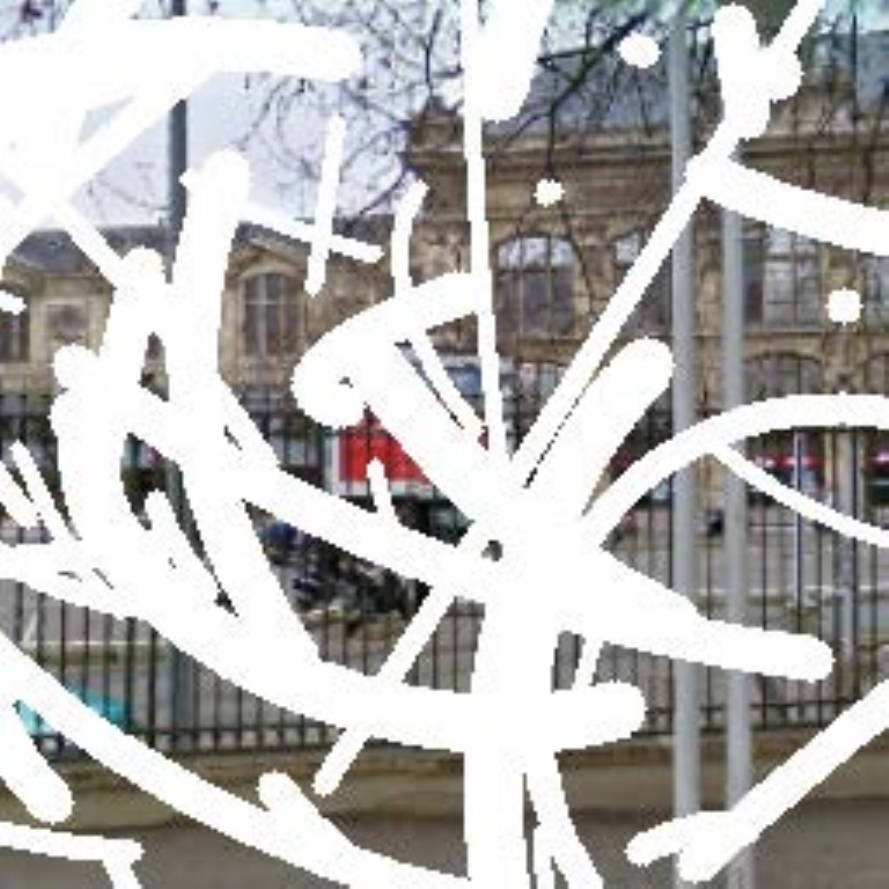}
        \includegraphics[width=2cm]{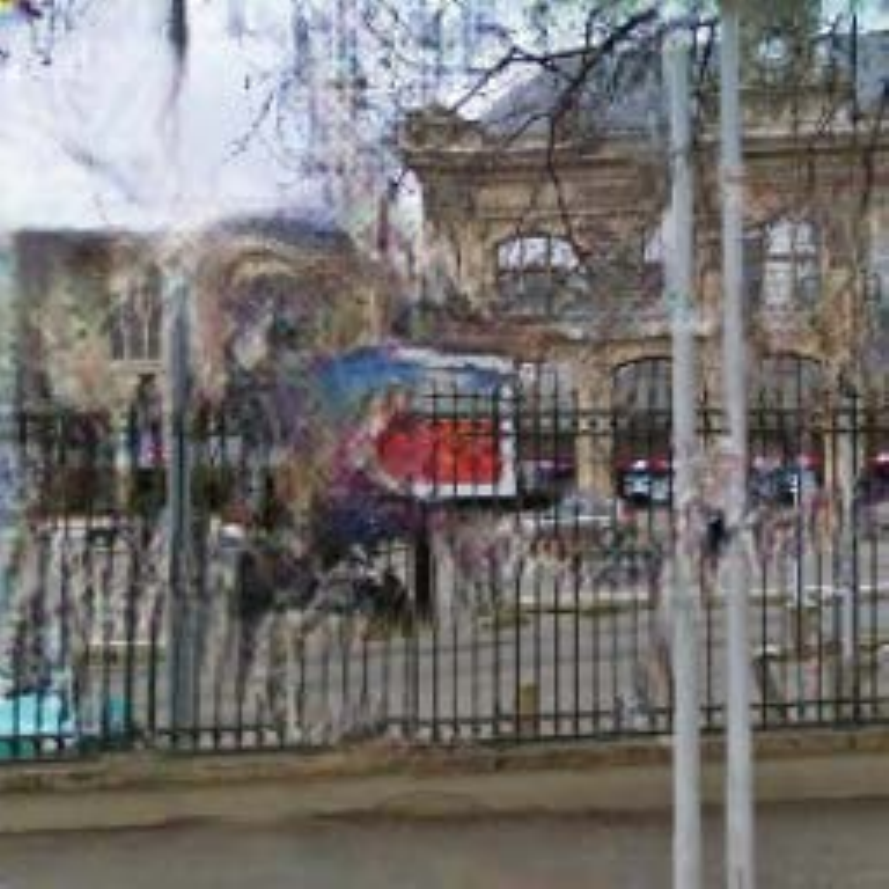}
        \includegraphics[width=2cm]{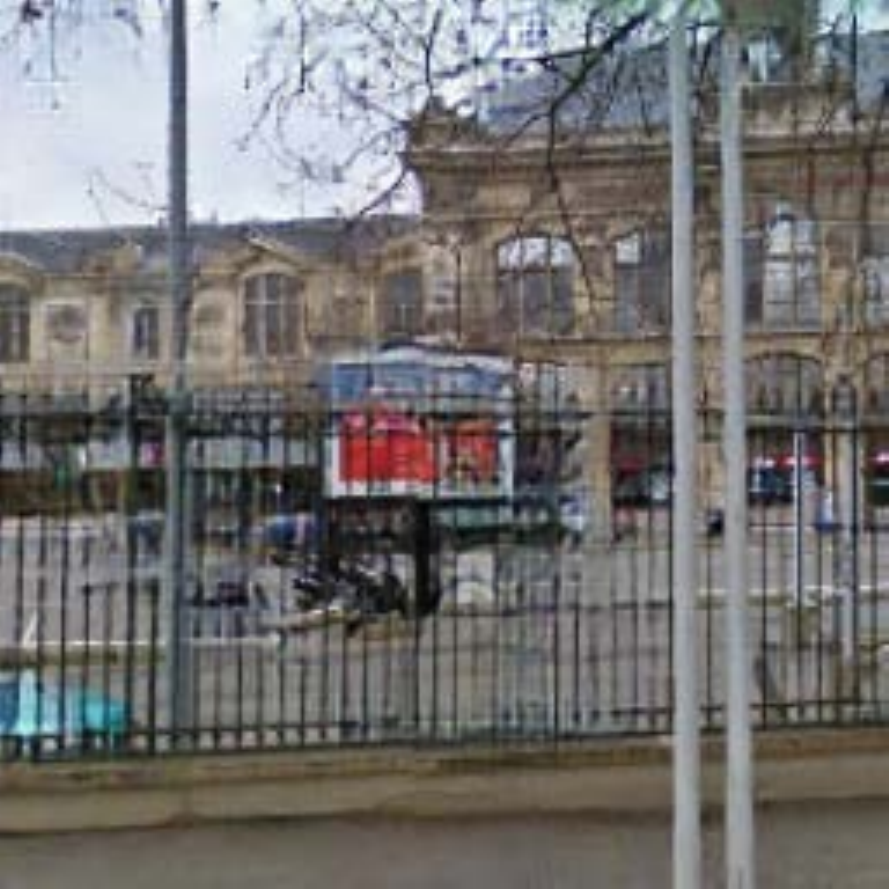}
        \includegraphics[width=2cm]{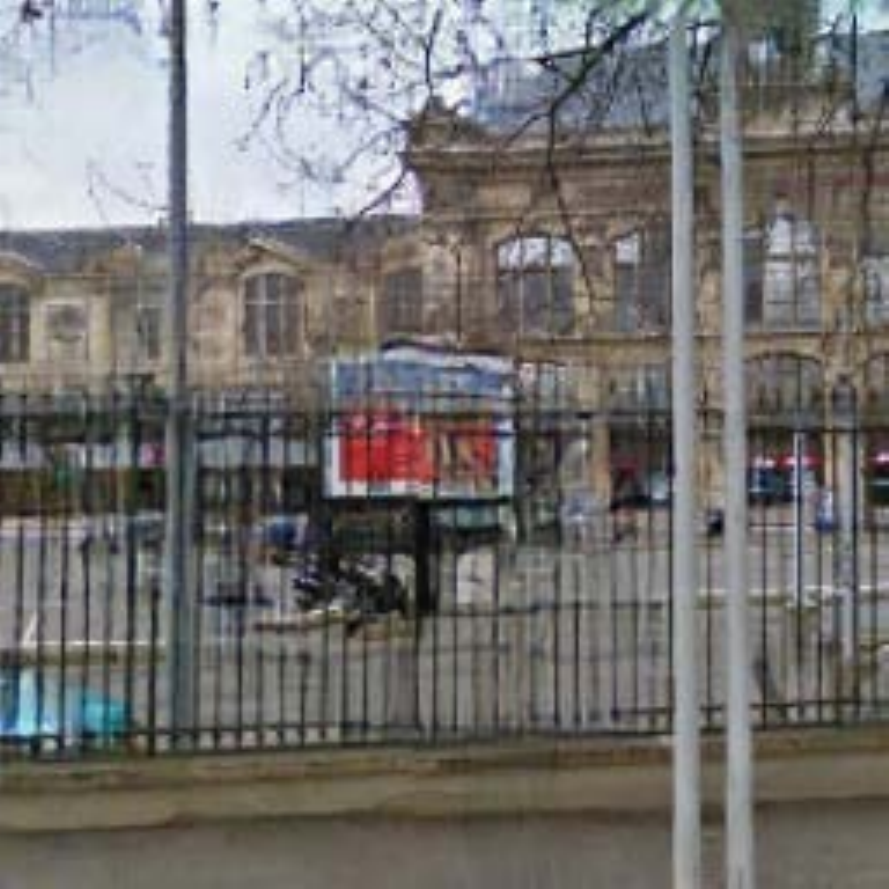}
    \end{subfigure}
    \begin{subfigure}
        \centering
        \vspace{-0.05in}
        \includegraphics[width=2cm]{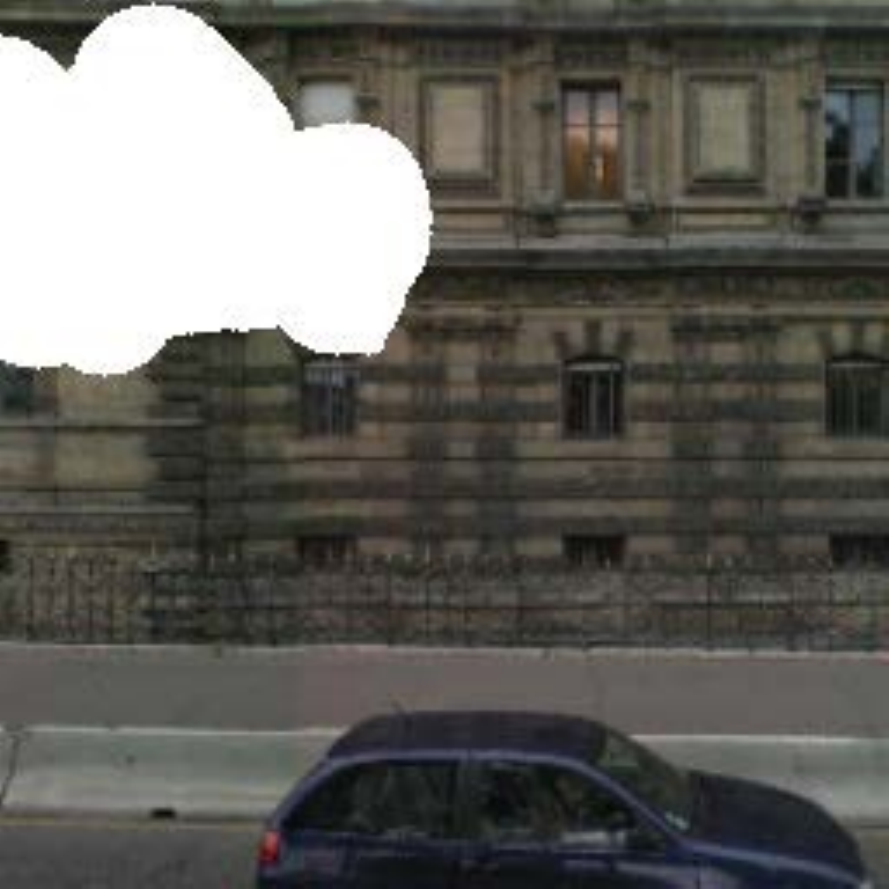}
        \includegraphics[width=2cm]{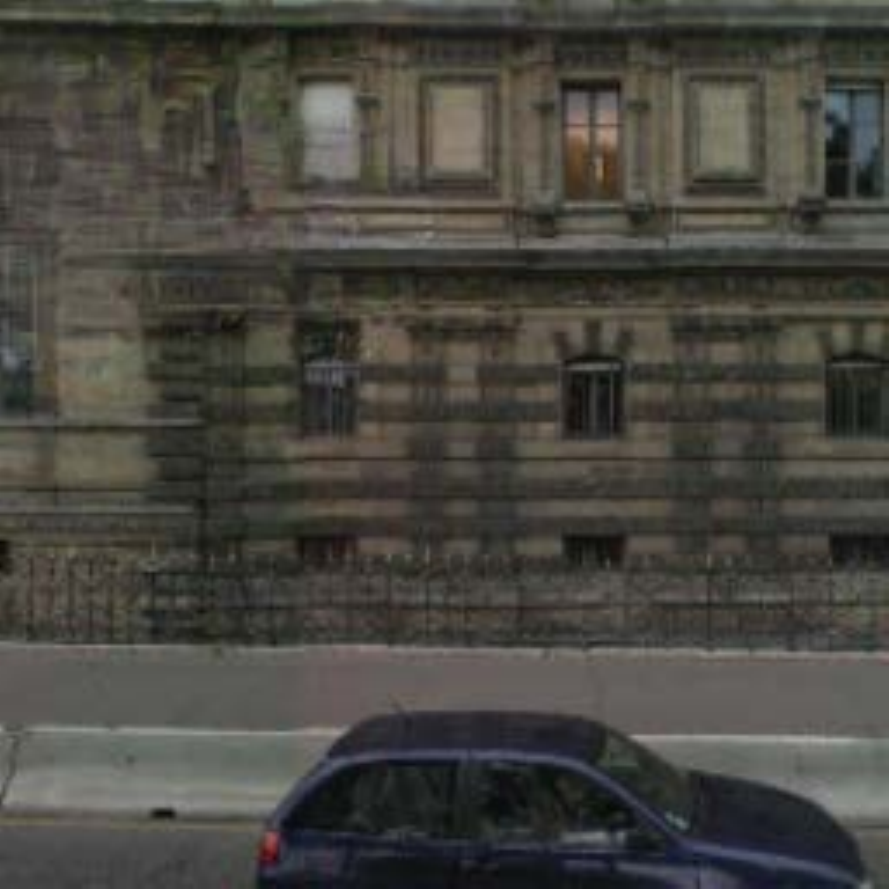}
        \includegraphics[width=2cm]{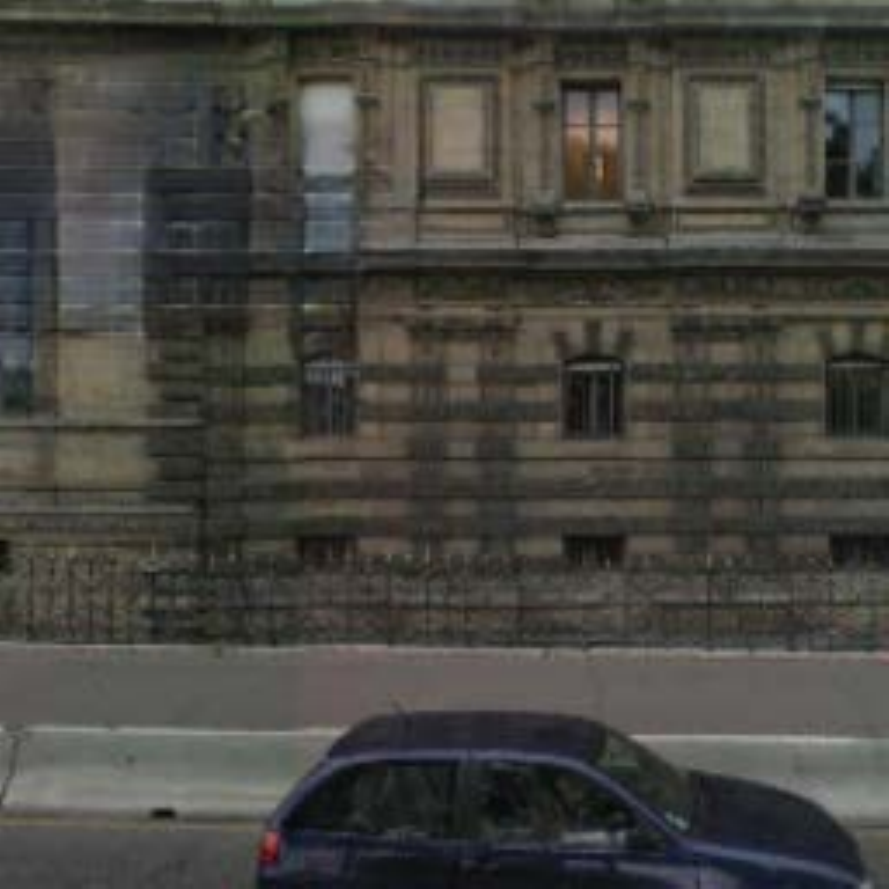}
        \includegraphics[width=2cm]{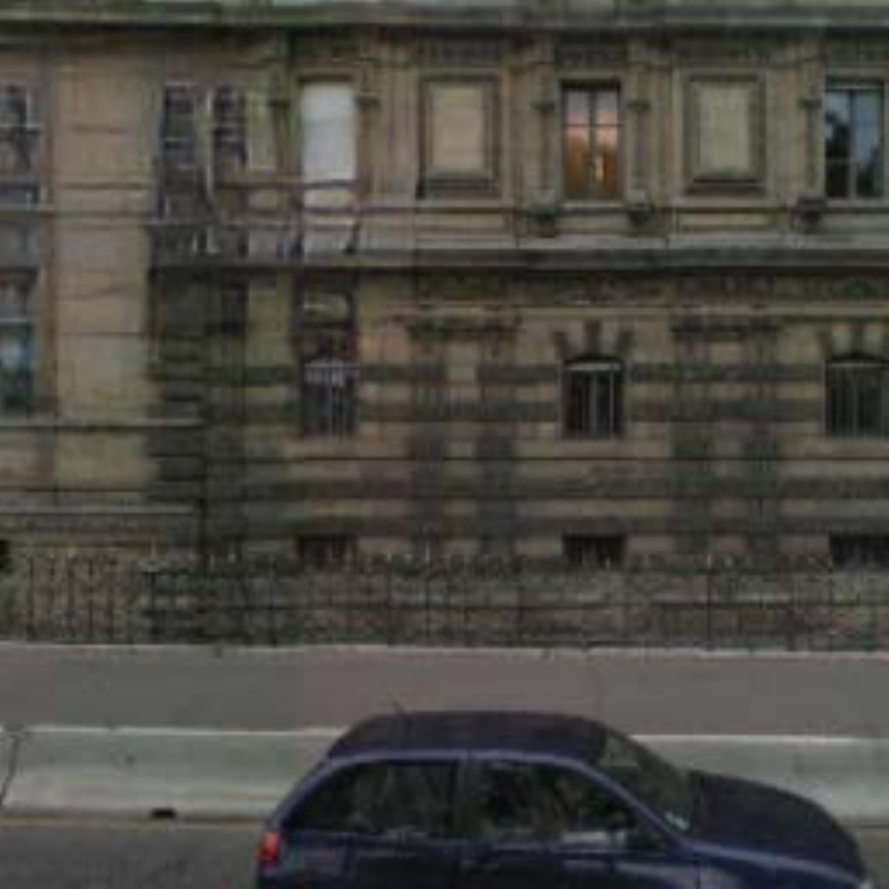}
    \end{subfigure}
    \vspace{-0in}
    \\
    (a) Input                   \hspace{0.8cm}
    (b) EC                      \hspace{0.8cm}
    (c) RED                   \hspace{0.8cm}
    (d) Ours                    
    \vspace{-0.05in}
    \caption{Image inpainting results for large missing areas (discontiguous at the top row, and contiguous at the bottom row), using EdgeConnect (EC) \cite{nazeri2019edgeconnect} , our previous model Region-wise Encoder-Decoder (RED) \cite{ma2019inpainting} and our proposed method on street view image.}
    \label{fig:show}
    \vspace{-0.20in}
\end{figure}

Despite the encouraging progress in image inpainting, most existing methods still face the inconsistency problem, such as distorted structures and blurry textures, especially when the missing regions are large. Figure \ref{fig:show} shows that the very recent method EC \cite{nazeri2019edgeconnect} (the second column) suffers severe artifacts in the consecutive large missing region. This phenomenon is much likely due to the inappropriate convolution operation over the two types of regions, i.e., existing and missing regions. 

Intuitively, different feature representations should be extracted to characterize different types of regions, since there is sufficient content information in existing regions, but none in the missing ones, which needs to be inferred from existing regions. Therefore, directly applying the same convolution filters to generate semantic contents inevitably leads to visual artifacts such as color discrepancy, blur and spurious edge responses surrounding holes. The changeable mask was been proposed in recent work \cite{liu2018image} to handle the difference. However, relying on the same filters for different regions, they still fail to generate favourable results. Meanwhile, many discriminators proposed in inpainting tasks take the whole image as input, and use the same filter to deal with the reconstructed content and inferred content, which inevitably caused an unsatisfied compromise.

\begin{figure*}[tp!]
    \centering 
    \vspace{0.0in}
    \includegraphics[width=0.8\linewidth,]{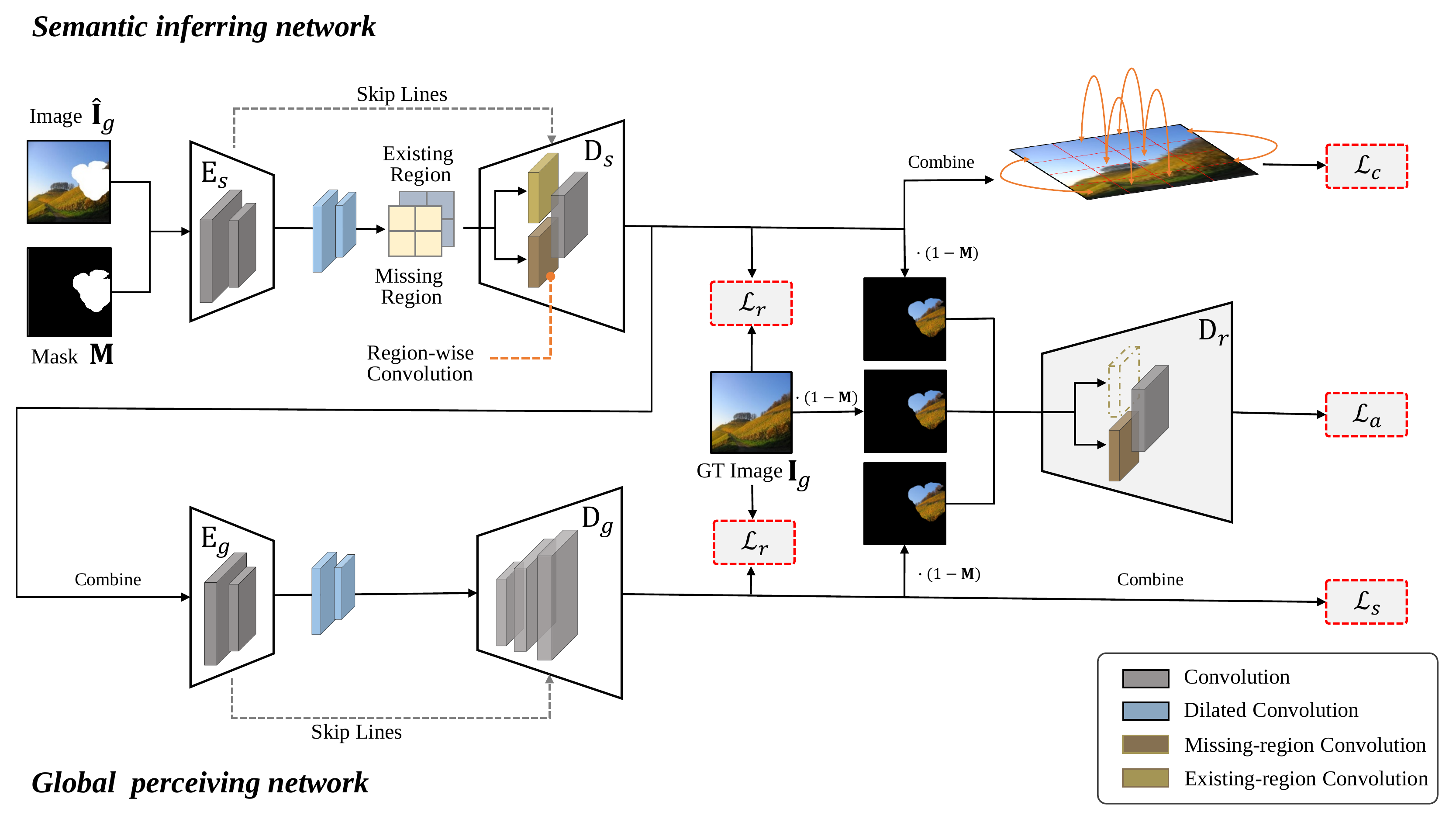}
    \vspace{-0.in}
    \caption{The architecture of our proposed region-wise adversarial image inpainting framework. }%At stage 1, the coarse network infers the semantic contents using region-wise convolution filters and enhances the quality of the composited image based on correlation loss. At stage 2, the fine network employs the style loss together with reconstruction loss over the whole image to perceptually enhance the image quality.
    
    \label{fig:framwork}
    \vspace{-0.15in}
\end{figure*}

In this paper, to generate desirable contents for missing regions, we develop a region-wise generative adversarial framework to handle the different regions in each image. Figure \ref{fig:framwork} shows the architecture of our whole framework. Note that we extend upon our prior conference publication \cite{ma2019inpainting} that mainly concentrated on the discontiguous missing areas using region-wise convolutions and suffered severe artifacts when the large missing areas are contiguous. Figure \ref{fig:show} (a) shows the different types of missing regions, namely discontiguous and contiguous missing regions. For discontiguous missing regions, even though the total missing area is large, but it is still easier to infer the missing information from the surrounding area. However, for large contiguous missing regions, it is hard to infer semantically plausible and visually realistic information. This can be confirmed by the observations in Figure \ref{fig:show} (b) and (c), where both EdgeConnect (EC) \cite{nazeri2019edgeconnect} and our previous model Region-wise Encoder-Decoder (RED) \cite{ma2019inpainting} can infer the semantic content in discontiguous missing areas in the first row, even facing distortion and inconsistency problem. However, they usually fail to infer the semantic information for contiguous missing areas (the second row in Figure \ref{fig:show}).

We attempt to address these problems for both contiguous and discontiguous large missing regions using one generic framework in this paper. Our adversarial inpainting framework utilizes the region-wise generative adversarial networks to accomplish the image inpainting task. The region-wise generator integrating two consecutive encoder-decoder networks, first infers the missing semantic contents roughly and captures the correlations between missing regions and existing regions guided by correlation loss, and further adversarially enhances the naturally visual appearance via region-wise discriminator. The key contributions of this paper can be summarized as follows:
\begin{itemize}
\item A generic inpainting framework is proposed to handle the images with both contiguous and discontiguous large missing areas at the same time, in an adversarial learning manner, consisting of region-wise generator and region-wise discriminator.

\item To locally handle features in different regions, the region-wise generator employs and integrates a region-wise convolution in semantic inferring networks, accomplishing the inpainting task at pixel level based on the $\ell_1$ reconstruction loss.

\item To model non-local correlation between existing regions and missing regions, the correlation loss guides the region-wise generator to infer semantic contents and generate more detailed information.
\item To further eliminate large area artifacts and obtain visually realistic generated contents, the framework introduces a region-wise discriminator and utilizes the adversarial loss and the popular style loss to enhance the image quality respectively from local and global perspective.

\item Extensive experiments on various popular datasets, including face (CelebA-HQ \cite{karras2017progressive}), and natural scenes (Paris StreetView \cite{doersch2012makes} and Places2 \cite{zhou2018places}), demonstrate that our proposed method can significantly outperform other state-of-the-art approaches in image inpainting for both discontiguous and contiguous missing areas.
\end{itemize}

The remaining sections are organized as follows. Section \ref{sec:related work} discusses the related work for image inpainting. In Section \ref{sec:3} we introduce our inpainting framework and the formulation. Comprehensive experiments over three popular datasets are presented in Section \ref{sec:4}. Finally, we conclude in Section \ref{sec:5}.

\section{Related Work} \label{sec:related work}
%\subsection{Image Inpainting}
Until now, there have been many methods proposed for generating desirable contents in different ways, including the traditional methods using handcrafted features and the deep generative models. We mainly focus on the deep models and introduce three different types of deep methods in detail.

\subsection{Traditional Methods}
Traditional approaches can be roughly divided into two types: diffusion-based and patch-based. The former methods propagate background data into missing regions by following a diffusive process typically modeled using differential operators ~\cite{ballester2000filling,esedoglu2002digital}. Patch-based methods ~\cite{kwatra2005texture,barnes2009patchmatch} fill in missing regions with patches from a collection of source images that maximize the patch similarity. These methods have good effects on the completion of repeating structured images. However, they are usually time-consuming and besides they cannot hallucinate semantically plausible contents for challenging cases where inpainting regions involve complex, non-repetitive structures, e.g., faces, objects, etc.

\subsection{Deep Generative Methods}
The development of deep neural networks \cite{al2009handwriting, Krizhevsky2012ImageNet} significantly promoted the progress of computer vision tasks \cite{redmon2016you,He2017Mask, howard2017mobilenets, xie2017aggregated}. Generative models \cite{goodfellow2014generative, kingma2013auto} are widely used in many areas for its strong ability of modeling and learning approximately true dataset distributions, including image generation \cite{mirza2014conditional, Arjovsky2017Wasserstein, karras2019style}, representation learning \cite{radford2015unsupervised}, image retrieval \cite{song2018binary,ma2018progressive}, object detection \cite{li2017perceptual}, video applications \cite{mathieu2015deep, tulyakov2018mocogan} and image translation \cite{isola2017image, zhu2017unpaired, choi2018stargan, wang2018high, ledig2017photo, wang2018esrgan, jo2019sc}. In general, the framework for solving image inpainting task using deep generative methods mainly consists of an encoder and a decoder. The encoder aims to capture and extract the context of an image into a compact latent feature representation. After that, the decoder uses the extracted representation to produce the missing image content, thus combine the original existing regions and the generating content in order to obtain the final restoring results.

\subsubsection{Synthesising Realistic Contents}
Inspired by the prevalence of GANs \cite{goodfellow2014generative}, many works force the restored contents to be consistent with existing regions using the adversarial loss. By picking a particular mode from the distribution of complete images, Context Encoders \cite{pathak2016context} attempted to get ``hints'' from pixels near the missing areas of the images. The authors connected the encoder and decoder through a channel-wise fully-connected layer, which allows each unit in the decoder to reason about the entire image content. After that, Semantic Inpainting \cite{yeh2017semantic} was proposed to solve image inpainting task by treating it as a constrained image generation problem. Based on that observation, the authors first trained a generative adversarial network on real images, then they recovered the encoding of the corrupted image to the ``closest'' intact one  while being constrained to the real image manifold. %The authors formulated the process of recovering the encoding as an optimization problem with a weighted context loss to condition on the corrupted image and an adversarial loss to penalize unrealistic images. 
Meanwhile, in the work of GLCIC \cite{iizuka2017globally}, global and local context discriminators were utilized to distinguish real images from completed ones. Among them, the global discriminator looked at the entire images to assess its coherency and completeness, while the local discriminator only concentrated on a small area centered at the completed region to ensure the local consistency of the generated patches.

\subsubsection{Inferring High Frequency Details}
It is hard to restore the low-level information, such as texture, illumination if we only use adversarial loss. Thus, several studies attempted to not only preserve contextual structures but also produce high frequency details. These methods are classified into optimization-based approaches and exemplar-based approaches.

\noindent\textbf{Optimization-based Approach} This type of method usually utilizes regularization to guide the whole framework to produce high frequency details which can be computed and extracted from pretrained-VGG network. Yang et al. \cite{yang2017high} proposed a multi-scale neural patch synthesis approach based on joint optimization of image content and texture constraints. They first trained a holistic content network and fed the output into a local texture network to compute the texture loss which penalizes the differences of the texture appearance between the missing and existing regions. Wang et al. \cite{wang2018image} further proposed an implicit diversified MRF regularization method which extracts features from a pre-trained VGG to enhance the diversification of texture pattern generation process in the missing region. %Meanwhile, They also adopted a relative distance measure to model the relation among patches of different regions and penalized the inpainting results where many generated patches are similar to one existing patches.

\noindent \textbf{Exemplar-based Approach} It is believed that the missing part is the spatial rearrangement of the patches in the existing region, and thus the inpainiting or completion process can be regarded as a searching and copying process using the existing regions from the exterior to the interior of the missing part.
Based on the above assumption, Contextual-based Image Inpainting \cite{song2018contextual} and Shift-Net \cite{yan2018shift} were proposed by designing a ``patch-swap'' layer and a ``shift-connection'' layer respectively which high-frequency texture details from the existing regions to the missing regions are propagated. Similarly, Yu et al. \cite{yu2018generative} introduced CA which adopted a two-stage coarse-to-fine network. The first part made an initial coarse prediction, the model further computed the similarity between existing patches through a contextual attention layer and restored the patches predicted by the coarse network.

\subsubsection{Filling Irregular Holes}
As discussed above, previous approaches mainly focus on rectangular shape holes which were often assumed to be at the center part of an image. Obviously, this kind of assumption contains strong limitations which may lead to overfitting to the rectangular holes, meanwhile ultimately limits the utilities in more widely used applications. Thus, several strategies have been proposed to fill irregular holes.
Liu et al. \cite{liu2018image} first proposed a partial convolutional layer, which consists of a masked and a re-normalized convolution operation to be conditioned on only valid pixels, followed by a mask-update step.
Meanwhile, Yu et al. \cite{yu2018free} introduced a gated convolution, which generalizes partial convolution by providing a learnable dynamic feature selection machanism for each channel at each spatial location across all layers.
More recently, a two-stage adversarial model that consists of an edge generator followed by an image completion network was proposed in \cite{nazeri2019edgeconnect}. According to the paper, the edge generator hallucinated edges of the missing region of the image, meanwhile the image completion network filled in the missing regions using hallucinated edges as a priori.
Moreover, Zhang et al. \cite{Zheng2019Pluralistic} came up with a probabilistically principled strategy named PIC to deal with the problem, which mainly contains two parallel paths including reconstructive path and generative path. Thus, PIC is able to resolve non-unique ground truth problem, with diverse information filled in the missing region.

\section{The Approach} \label{sec:3}
In this section, we elaborate the details of our adversarial inpainting framework. We will first introduce the whole framework which utilizes region-wise generative adversarial network to accomplish image inpainting task. The region-wise generator recovers the missing information with two consecutive networks based on region-wise convolution and non-local correlation, while region-wise discriminator further enhances the image quality adversarially in a region-wise manner. Finally, the whole formulation and optimization strategies will be provided.

\subsection{The Adversarial Inpainting Framework}
The state-of-the-art image inpainting solutions often ignore either the difference or the correlation between the existing and missing regions, and thus suffer from the inferior content quality for restoring large missing areas. What's worse, they cannot fill appropriate content and show particularly bad performance when facing large contiguous missing region. To simultaneously address these issues, we introduce our generic region-wise generative adversarial inpainting framework which is suitable for both discontiguous and contiguous missing cases as follows:

The region-wise generator recovers the semantic content in missing regions and pursues visually realistic, consisting of semantic inferring networks and global perceiving networks.
The semantic inferring networks focus on dealing with the differences and correlations between different regions, using region-wise convolution and non-local operation, respectively. We will further discuss them in Section \ref{sec:inferring} and Section \ref{sec:modelling}. The global perceiving networks consider the two different regions together using a style loss over the whole image, which perceptually enhance the image quality in a region-wise manner. Note that we take the two different regions into consideration over the whole image only to pursue the semantic and appearance information filled in missing region. 

Specifically, as shown in Figure \ref{fig:framwork}, the region-wise generator takes the incomplete image $\mathbf{\hat{I}}_{g}$ and a binary mask as input, and attempts to restore the complete image close to ground truth image $\mathbf{I}_{g}$, where $\mathbf{M}$ indicates the missing regions (the mask value is $0$ for missing pixels and $1$ for elsewhere), $\mathbf{\hat{I}}_{g}=\mathbf{I}_{g} \odot \mathbf{M}$ and $\odot$ denotes dot product. To accomplish this goal, the semantic inferring networks infer the semantic contents from the existing regions. Encoder $\mathrm{E}_s$ extracts semantic features from $\mathbf{\hat{I}}_{g}$, decoder $\mathrm{D}_s$ composing of the proposed region-wised convolutional layers is employed after encoder $\mathrm{E}_s$ to restore the semantic contents for different regions, and generates the predicted image $\mathbf{I}^{(1)}_p=\mathrm{D}_s\left(\mathrm{E}_s(\mathbf{\hat{I}}_g)\right)$. After feeding the composited image $\mathbf{I}_c^{(1)}=\mathbf{\hat{I}}_g+\mathbf{I}^{(1)}_p\odot (\mathbf{1}-\mathbf{M})$ to global perceiving encoder $\mathrm{E}_g$, global perceiving decoder $\mathrm{D}_g$ further globally and perceptually synthesizes the refined image $\mathbf{I}^{(2)}_p=\mathrm{D}_g\left(\mathrm{E}_g(\mathbf{I}^{(1)}_c)\right)$. Still, the composited image $\mathbf{I}_c^{(2)}=\mathbf{\hat{I}}_g+\mathbf{I}^{(2)}_p\odot (\mathbf{1}-\mathbf{M})$ could be obtained. 

With the region-wise image generation, the region-wise discriminator is further introduced to stress the importance of inferred regions in restoring results, which means, we still need to consider the difference between the two types of region. It is worth noting that, undesired artifacts only exist in inferred regions, which means there is no need to penalize the existing regions. In fact, focusing on the whole images, with existing regions involved, inevitably exerts bad influence on inferred regions. Therefore, here we only feed the inferred regions into region-wise discriminator. The region-wise discriminator forces the appearance of inferred content to approximate the true images, and further visually enhances the image quality. The inferred content of the predicted image and refined image $\mathbf{I}^{(1)}_p \odot (\mathbf{1}-\mathbf{M})$, $\mathbf{I}^{(2)}_p \odot (\mathbf{1}-\mathbf{M})$ are fed into discriminator $\mathrm{D}_r$ to adversarially enhance the capability of the region-wise generator. With the region-wise generator that distinguishingly deals with the different types of regions to infer the missing information, and the region-wise discriminator that adversarially guides the networks to enhance the image quality, our framework could accomplish the inpainting tasks and produce visually realistic and semantically reasonable restored images. We finally have the visually and semantically realistic inpainting result $\mathbf{I}_c^{(2)}$ close to the ground truth image $\mathbf{I}_g$. More details will be presented in Section \ref{sec:eliminating}. 

In the following part of this section, we will present the key components and the corresponding techniques of our framework.

\subsection{Generating Region-wise Contents}\label{sec:inferring}
For image inpainting tasks, the input images are composed of both existing regions with the valid pixels and the missing regions (masked regions) with invalid pixels in mask to be synthesized. During the inpainting process, the existing regions are basically reconstructing themselves which is easy to accomplish, while the missing region should be inferred from existing region and kept semantically reasonable and visually realistic from both local and global perspectives. That is to say, different learning operations should be conducted on these two types of regions. Only relying on the same convolution filter, we can hardly synthesis the appropriate features over different regions, which in practice usually leads to the visual artifacts such as color discrepancy, blur and obvious edge responses surrounding the missing regions. Motivated by this observation, we first propose region-wise convolutions to seperately handle with the different regions using different convolution filters, in avoid of compromise between the two different learning operations. 

Specifically, let $\mathbf{W}, \mathbf{\hat{W}}$ be the weights of the region-wise convolution filters for existing and missing regions respectively, and $\mathbf{b}, \mathbf{\hat{b}}$ correspond to the biases. $\mathbf{x}$ is the feature for the current convolution (sliding) window belonging to the whole feature map $\mathbf{X}$. Then, the region-wise convolutions at every location can be formulated as follows:
\begin{equation}
\mathbf{x}'= 
\begin{cases}
\mathbf{\hat{W}}^\top \mathbf{x}+\mathbf{\hat{b}}, \quad \mathbf{x} \in \mathbf{X} \odot (1-\mathbf{M}) \\
\mathbf{W}^\top \mathbf{x}+\mathbf{b}, \quad \mathbf{x} \in \mathbf{X} \odot  \mathbf{M} \\
\end{cases}
\end{equation}
This means that for different types of regions, different convolution filters will be learnt for feature representation respectively for inferring and reconstruction. 

In practice, we can accomplish region-wise convolutions through separating the two types of regions by channel according to masks which are resized proportionally as feature maps down-sampled through convolution layers. Thus, the information in different regions can be learned separately and transmitted consistently across layers.

\noindent\textbf{Reconstruction Loss} We employ $\ell_1$ reconstruction loss over the two predicted images generated by region-wise generator, to promise the reconstruction of existing regions and generation for missing regions inferred from existing regions. Note that, although we only need the inferred contents for missing regions, the framework should have a better understanding of existing regions and infer the missing information from both local and global perspectives. Thus, it is essential to reconstruct the existing information as well. The reconstruction loss is defined as follows:
\begin{equation}
\mathcal{L}_{r}=\left\|{\mathbf{I}^{(1)}_{p}-\mathbf{I}_{g}}\right\|_1 + \left\|{\mathbf{I}^{(2)}_{p}-\mathbf{I}_{g}}\right\|_1.
\end{equation}
The reconstruction loss is useful for region-wise convolution filters to learn to generate meaningful contents for different regions, especially for semantic inferring networks.

\subsection{Inferring Missing Contents via Correlations}\label{sec:modelling}
The reconstruction loss treats all pixels independently without consideration of their correlation, and thus the framework generates a coarse predicted image. However, the inferred missing contents are similar to surrounding existing regions, which is hard to achieve semantically meaningful and visually realistic. This is mainly because the convolution operations are skilled in processing local neighborhoods whereas fail to model the correlation between distant positions.

To address this problem and further guide the region-wise convolutions to infer semantic contents from the existing regions, a non-local correlation loss is adopted following prior studies \cite{wang2018non, zhang2018self}. Traditional non-local operation computes the response at a position as a weighted sum of the features at all positions in the input feature map during the feed-forward process. It can capture long-distance correlation between patches inside an image at the expense of a large amount of calculations. Therefore, it is not appropriate for large feature maps in our generative models, where the smallest feature map is $128\times128$. Besides, we prefer to build the same correlations between different patches just as ground-truth image, which is hard to accomplish only guided by reconstruction loss. Therefore, in this paper, we introduce the correlation loss to model the non-local correlations and further guide the region-wise convolution to infer the missing information according to such correlations.

Formally, given an image $\mathbf{I}^{(1)}_c$, $\Psi(\mathbf{I}^{(1)}_{c})$ denotes the  ${c} \times {h} \times {w}$ feature map computed by feature extraction method $\Psi$. In practice, in order to index an output position in space dimension easily, we reshape the feature map to the size of ${c} \times {n}$, where ${n} = {h} \times {w}$.  Correspondingly, $\Psi^i(\mathbf{I}_{c})$ is the $i$-th column in the reshaped feature map $\Psi(\mathbf{I}_{c})$, where $i=1,\ldots,{n}$, of length ${c}$. Then, a pairwise function ${f}_{ij}$ can be defined as a non-local operation, which generates a ${n} \times {n}$ gram matrix evaluating the correlation between position $i$ and $j$:
\begin{equation}
{f}_{ij}(\mathbf{I}^{(1)}_{c})= \left(\Psi^i(\mathbf{I}^{(1)}_{c})\right)^\top\left(\Psi^j(\mathbf{I}^{(1)}_{c})\right).
\end{equation}
Once we have the non-local correlation, we can bring it into the inpainting framework by introducing a correlation loss.

\noindent \textbf{Correlation Loss} Since the relationship among distant local patches plays a critical role in keeping the semantic and visual consistency between the generated missing regions and the existing ones, we further introduce a correlation loss that can help to determine the expected non-local operation. Namely, for image $\mathbf{I}^{(1)}_{c}$, the correlation loss is defined based on ${f}_{ij}(\cdot)$:
\begin{equation}
\mathcal{L}_{c}=\sigma\sum_{i,j}^{{n}}\left\|{f}_{ij}(\mathbf{I}^{(1)}_{c})-{f}_{ij}(\mathbf{I}_{g})\right\|_1,
\end{equation} 
where $\sigma$ denotes the normalization factor by position. The correlation loss forces the region-wise convolution to infer missing information with semantic details much closer to the realistic image according to semantic-related patches, rather than surrounding ones.

\subsection{Eliminating large area artifacts}\label{sec:eliminating}
With the consideration of both differences and correlations between different regions, the framework could infer the semantically reasonable contents filled in the missing regions. However, it is very common to produce unwanted artifacts in unstable generative models, which could cause visually unrealistic results. Some work \cite{odena2016deconvolution, bau2018gan} tried to analyze the reason behind, but still cannot totally overcome this problem. 

Image generation tasks usually adopt style loss which poses as an effective tool to combat ``checkerboard'' artifacts \cite{Sajjadi_2017_ICCV}. Since our region-wise convolutions and non-local operation are handling with the difference and correlations between local patches, it is reasonable to adopt style loss over the whole image and perceptually enhance the image quality and remove some unpleasant artifacts. Thus, we use style loss to globally and perceptually enhance the image quality.

\noindent \textbf{Style Loss}  After projecting image $\mathbf{I}^{(2)}_c$ into a higher level feature space using a pre-trained VGG, we could obtain the feature map $\Phi_p(\mathbf{I}^{(2)}_p)$ of the $p$-th layer with size ${c}_p \times {h}_p \times {w}_p$, and thus the style loss is formulated as follows:
\begin{equation}\small
%\begin{aligned}
\mathcal{L}_{s}= \sum_{p}\delta_p\left\|\left(\Phi_p(\mathbf{I}^{(2)}_{c})\right)^\top\left(\Phi_p(\mathbf{I}^{(2)}_{c})\right)-\left(\Phi_p(\mathbf{I}_{g})\right)^\top\left(\Phi_p(\mathbf{I}_{g})\right)\right\|_1,
%\end{aligned}
\end{equation}
where $\delta_p$ denotes the normalization factor for the $p$-th selected layer by channel. The style loss focuses on the relationship between different channels to transfer the style for the composited image $\mathbf{I}^{(2)}_{c}$, and thus globally perceiving over the whole image, rather than separately dealing with the different regions. Here, different from the prior work of PConv, we only consider the style loss for the composited image.

Even with the style loss involved, the contents are still covered with lots of artifacts and not as realistic as the completed ground truth. Such phenomenon is particularly obvious when facing large contiguous missing regions. We speculate that there are two main reasons for this phenomenon: First, the encoder tries to capture the information contained by existing regions, and the missing region could still obtain the information from surrounding pixels due to the essence of convolutional operation. For pixels inside the missing region, the pixels near the boundary could soon obtain effective information from existing region. However, the pixels deep inside only obtain few information depending on the distance between them and the boundary. Only with the network deepening can the distant pixels obtain information from effective existing regions, which could be seen as an uneven sample, easily contributing to artifacts. Second, as the network deepening, the nearby pixels obtain more information which might not be accurate, meanwhile the distant pixels obtain much inaccurate information and thus learn undesired features.

To address the issues, we introduce a region-wise discriminator to guide the region-wise generator. After locally handling the two types of regions and globally perceiving over the whole image, we further adopt a region-wise discriminator operating on output of both networks in region-wise generator to eliminate artifacts and force the generated content as realistic as real ground-truth image. 

\noindent \textbf{Adversarial Loss} 
Formally, given $I_{p}^{(1)}$, $I_{p}^{(2)}$, $I_{g}$, we extract the missing regions of each image and concatenate the mask as the input, instead of the whole image. It could help the generator to pay more attention to specific regions. For existing regions containing enough information, it is easy to reconstruct without too much guidance. However, for the inferred contents, the case is contrary. Penalizing these two regions at the same time and still using the same filters, is likely to cause an unwanted compromise and affect the visual appearance of the inferred contents. Thus, we deploy the region-wise discriminator architecture, penalizing input images at the scale of patches, which could further preserve local details. While training the region-wise generator, the generated patches will be considered as real and thus labeled as 1. As the discriminator improves, the generators enhance their ability to generate realistic images. After several iterations, the generative networks and discriminator gradually reach a balance, eliminating the unpleasant artifacts and generate visually realistic inpainting results. We minimize following loss to enhance the output of region-wise generator: 
\begin{equation}
\begin{aligned}
\mathcal{L}_{a}= \alpha&\mathbb{E}(\mathrm{D}_r(\mathbf{I}_g\odot(\mathbf{1}-\mathbf{M}),\mathbf{M}))\\
+&\mathbb{E}(\mathbf{1}-\mathrm{D}_r(\mathbf{I}^{(1)}_p\odot(\mathbf{1}-\mathbf{M}),\mathbf{M}))\\
+&\mathbb{E}(\mathbf{1}-\mathrm{D}_r(\mathbf{I}^{(2)}_p\odot(\mathbf{1}-\mathbf{M}), \mathbf{M})),
\end{aligned}
\end{equation}
where $\alpha$ is a hyper-parameter to define the significance of each part of adversarial loss. We concatenate mask $\mathbf{M}$ to separate inferred contents and existing contents, which seems better than simply concatenating $(\mathbf{1}-\mathbf{M})$. The reason we speculate is that, via defining inferred regions as $\mathbf{1}$ will introduce some noises and affect the final visual appearance. 

%Due to the difficulties for the encoder-decoder networks to generate more local details in an image with style loss, we come up with a strategy. In the first stage, the network learns to coordinate the relationship between different regions and generate detailed information. Then, style loss comes to the stage and is used to perceptually enhance the image quality as a whole. 

\begin{algorithm}
  \renewcommand{\algorithmicrequire}{\textbf{Input:}}
  \renewcommand{\algorithmicensure}{\textbf{Output:}}
  \caption{Training of our proposed framework}
  \label{alg:1}
  \begin{algorithmic}[1]
    % \REQUIRE Training dataset
    % \ENSURE Complete images
        \WHILE {iterations $t < T_{train}$}
        \STATE Sample batch images $\mathbf{I}_{g}$
        \STATE Generate continue binary masks $\mathbf{M}$
            \STATE Construct input images $\mathbf{\hat{I}}_{g} = \mathbf{I}_{g} \odot \mathbf{M}$
        \STATE Predicted by semantic inferring networks and get $\mathbf{I}^{(1)}_{p}=\mathbf{D}_s(\mathbf{E}_s(\mathbf{\hat{I}}_g))$
            \STATE Construct composited images $\mathbf{I}_c^{(1)}=\mathbf{\hat{I}}_g+\mathbf{I}^{(1)}_p\odot(1-\mathbf{M})$
            \STATE Predicted by global perceiving networks and get $\mathbf{I}^{(2)}_{p}=\mathbf{D}_g(\mathbf{E}_g(\mathbf{I}^{(1)}_c))$
            \STATE Construct output images $\mathbf{I}_c^{(2)}=\mathbf{\hat{I}}_g+\mathbf{I}^{(2)}_p\odot (1-\mathbf{M})$
            \STATE Calculate $\mathcal{L}_{c}$ by $\mathbf{I}^{(1)}_{p}$, $\mathcal{L}_{s}$ by $\mathbf{I}^{(2)}_{p}$ , $\mathcal{L}_{r}$ by $\mathbf{I}^{(1)}_{p}$ and $\mathbf{I}^{(2)}_{p}$
            \IF{$t < T_{pretrain}$}
                \STATE Update $\mathbf{E}_s$, $\mathbf{D}_s$, $\mathbf{E}_g$ and $\mathbf{D}_g$ with $\mathcal{L}_{c}$, $\mathcal{L}_{s}$ and $\mathcal{L}_{r}$
            \ELSE
                \STATE Calculate $\mathcal{L}_{a}$ by $\mathbf{I}_g\odot(1-M)$, $\mathbf{I}_p^{(1)}\odot(1-M)$ and $\mathbf{I}_p^{(2)}\odot(1-M)$
                \STATE Update $\mathbf{E}_s$, $\mathbf{D}_s$, $\mathbf{E}_g$ and $\mathbf{D}_g$ with $\mathcal{L}_{c}$, $\mathcal{L}_{s}$, $\mathcal{L}_{r}$ and $\mathcal{L}_{a}$
                \STATE Update $\mathbf{D}_r$ with $-\mathcal{L}_{a}$
            \ENDIF
        \ENDWHILE
  \end{algorithmic}
  \label{algorithm}
\end{algorithm}
\subsection{The Formulation and Optimization}\label{sec:form}
\noindent \textbf{Formulation} To guide the learning of the region-wise generator, we combines the reconstruction, correlation, styles, and adversarial loss as the overall loss $\mathcal{L}$:
\begin{equation}
\mathcal{L}=\mathcal{L}_r + \lambda_1 \mathcal{L}_c + \lambda_2 \mathcal{L}_s + \lambda_3 \mathcal{L}_{a},
\end{equation}
and $\mathcal{L}_{a}$ is maximized only to guide the discriminator to distinguish the generated contents and the real contents. We alternatively train the generators and discriminator, until the loss is convergent. 

\noindent \textbf{Implementation} For our method, we basically develop the model based on the encoder-decoder architecture of CA, discarding its contextual attention module and discriminator but adding the region-wise convolutions. We adopt the idea of patch discriminator and utilize the spectral normalization to stabilize the training, with leaky ReLU used as activation function. Input images are resized to $256\times 256$, and the proportion of irregular missing regions varies from 0 to 40\% in the training process. We empirically choose the hyper-parameters $\lambda_1=10^{-5}$, $\lambda_2=10^{-3}$. $\lambda_3=0$ for previous 20 epochs, $\lambda_3=1$ for later 9 epochs. The $\alpha$ is set as 0.01, which means heavy penalization for inferred contents and thus could better eliminate the artifacts. The initial learning rate is $10^{-4}$ using the Adam optimizer.

We also adopt skip links in our encoder-decoder architecture, which as \cite{liu2018image} claimed, may propagate the noises or mistakes for most inpainting architectures. However, we find that skip links will not suffer the negative effect in our framework due to the region-wise convolutions and thus enable the detailed output from existing regions.
%, on CelebA-HQ and Paris StreetView we train the model with a batch size of 8, and on Places2 we train it with a batch size of 48.

In practice, we exploit the widely-adopted pre-trained VGG network to extract features for the calculation of correlation loss as well as style loss. For the computation of correlation loss, only feature maps extracted by $pool2$ are adopted due to the weak semantic representation capability of $pool1$ and the blur caused by $pool3$ and $pool4$. In order to calculate the style loss, we use the output of $pool1$, $pool2$, and $pool3$ together. In another word, $\Psi(\cdot)=\Phi_p(\cdot)$ when $p=2$.

\noindent \textbf{Optimization} The whole optimization process is described in Algorithm \ref{algorithm}. It follows the standard forward and backward optimization paradigm. In our framework, the reconstruction and adversarial loss work on two consecutive networks in region-wise generator, to respectively guarantee the pixel-wise consistency between the two predicted images and the ground truth, and produce natural visual appearance especially for inferred contents. To capture the relationship among different regions and generate detailed contents, the correlation loss is adopted to guide the training of the semantic inferring networks. Moreover, the style loss helps perceptually enhance the image quality by considering the whole image in global perceiving networks. 
In the forward step, given a ground truth image $\mathbf{I}_g$, we first sample an irregular binary mask $\mathbf{M}$ and subsequently generate the incomplete image $\mathbf{\hat{I}}_{g}$. The region-wise generator takes the concatenation of $\mathbf{\hat{I}}_g$ and $\mathbf{M}$ as the input, and outputs the predicted image $\mathbf{I}^{(1)}_p$ and $\mathbf{I}^{(2)}_p$. In the backward step, in avoid of unstabilized character of generative models, we only adopt $\mathcal{L}_1, \mathcal{L}_c, \mathcal{L}_s$ over the predicted and composited images in previous epochs. After several epochs, we introduce the adversarial loss $\mathcal{L}_a$ to further guide the previous networks. Instead of taking the whole image as input, we specifically highlight the restored information for missing regions, which further enhances the inpainting results.

\begin{figure*}[htp!]
    \centering
    \vspace{0.2in}
    %\vspace{0.05in}
    \centering
\begin{subfigure}
        \centering
        \includegraphics[width=2.5cm]{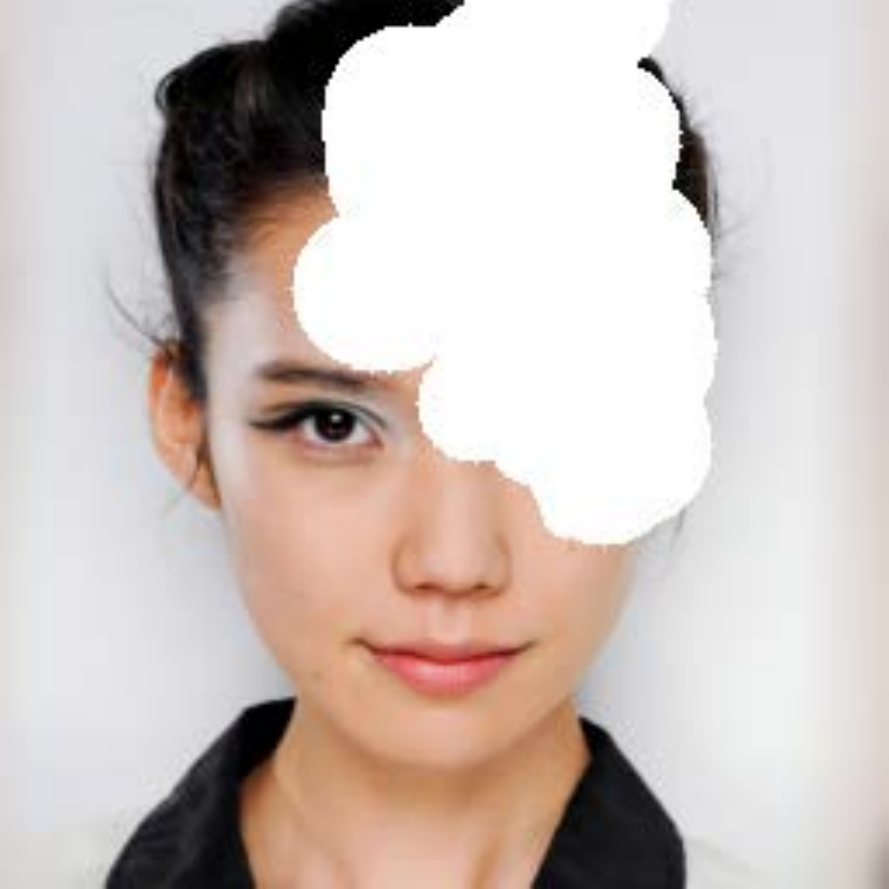}
        \includegraphics[width=2.5cm]{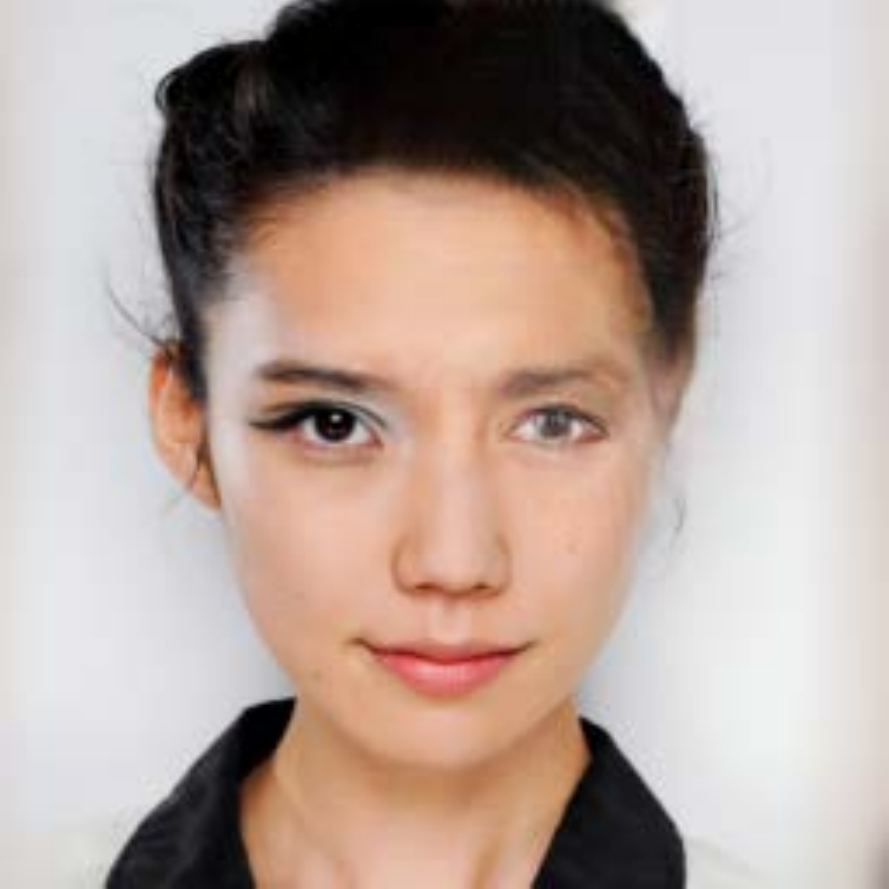}
        \includegraphics[width=2.5cm]{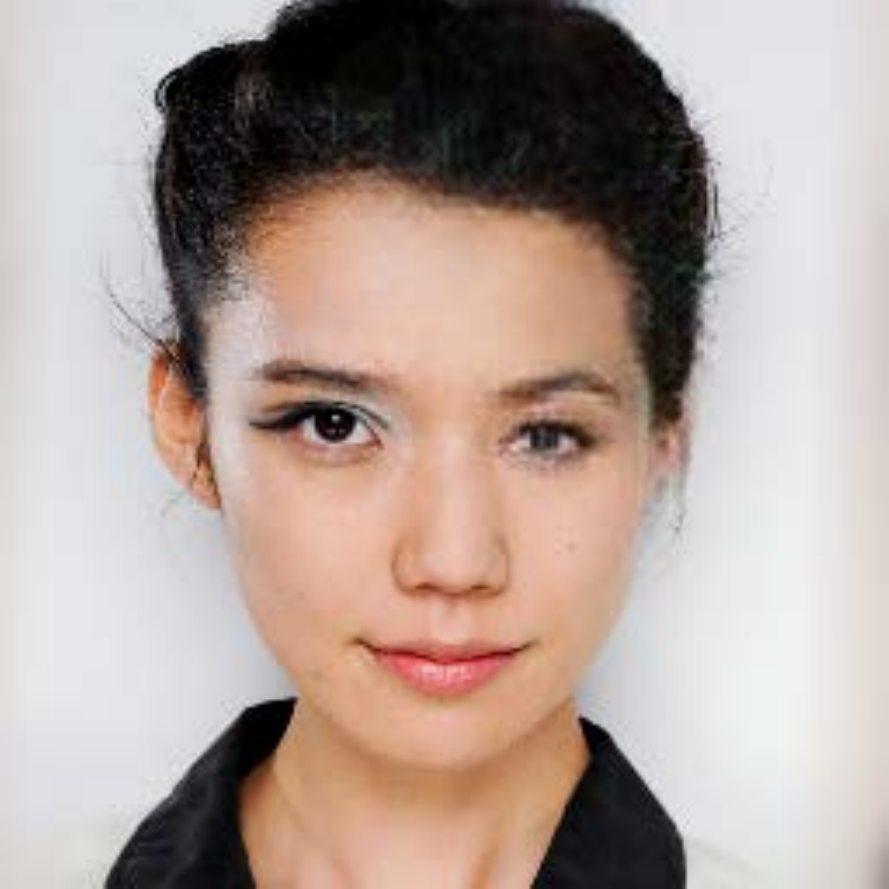}
        \includegraphics[width=2.5cm]{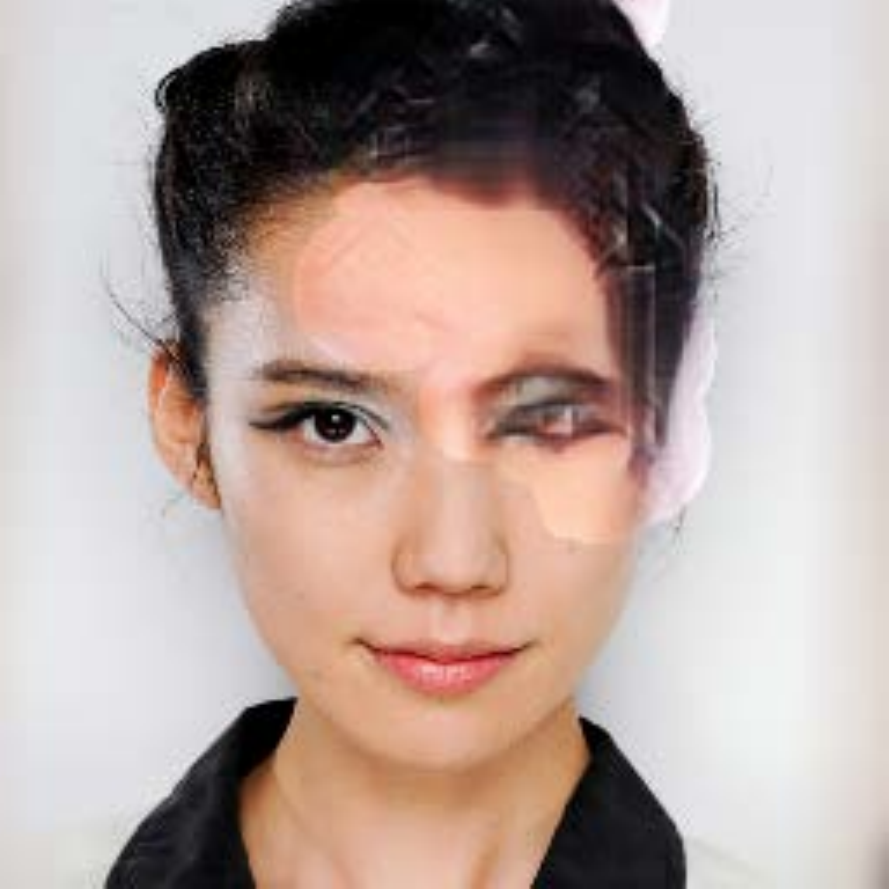}
        \includegraphics[width=2.5cm]{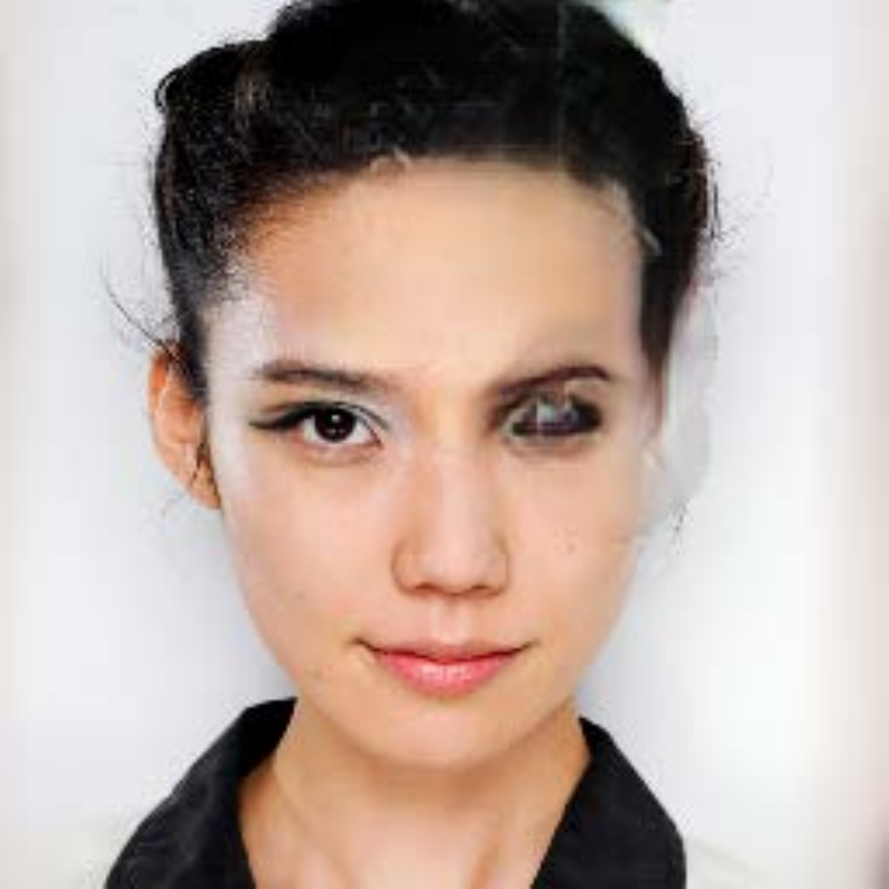}
        \includegraphics[width=2.5cm]{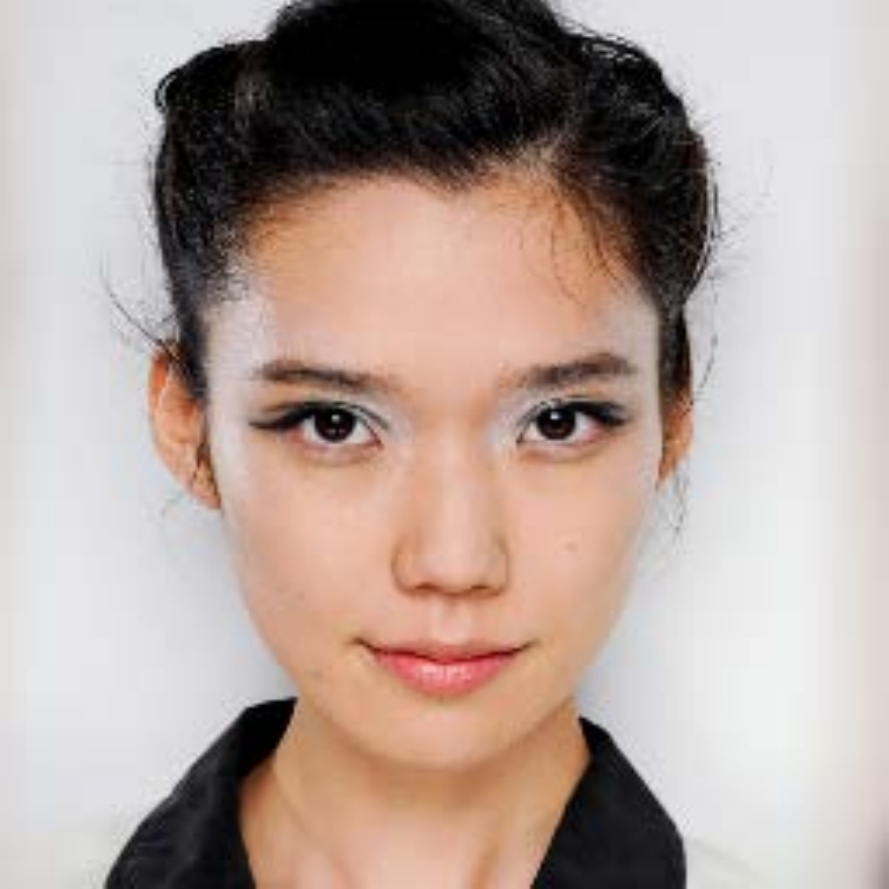}
    \end{subfigure}
    \begin{subfigure}
        \centering
        \vspace{-0.05in}
        \includegraphics[width=2.5cm]{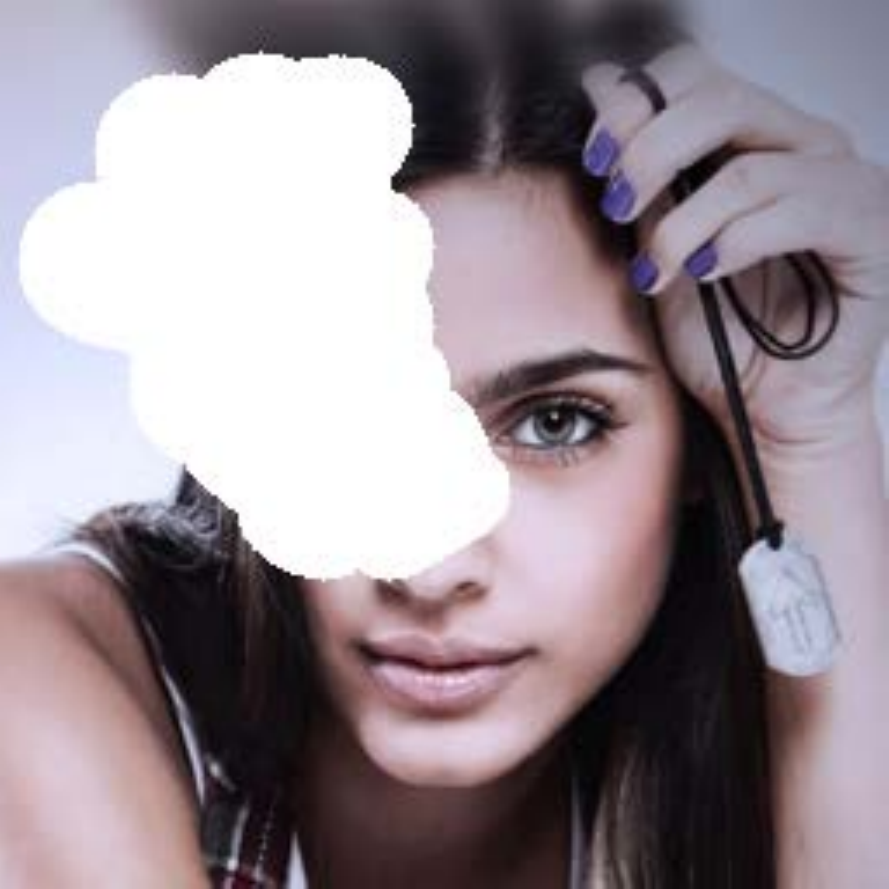}
        \includegraphics[width=2.5cm]{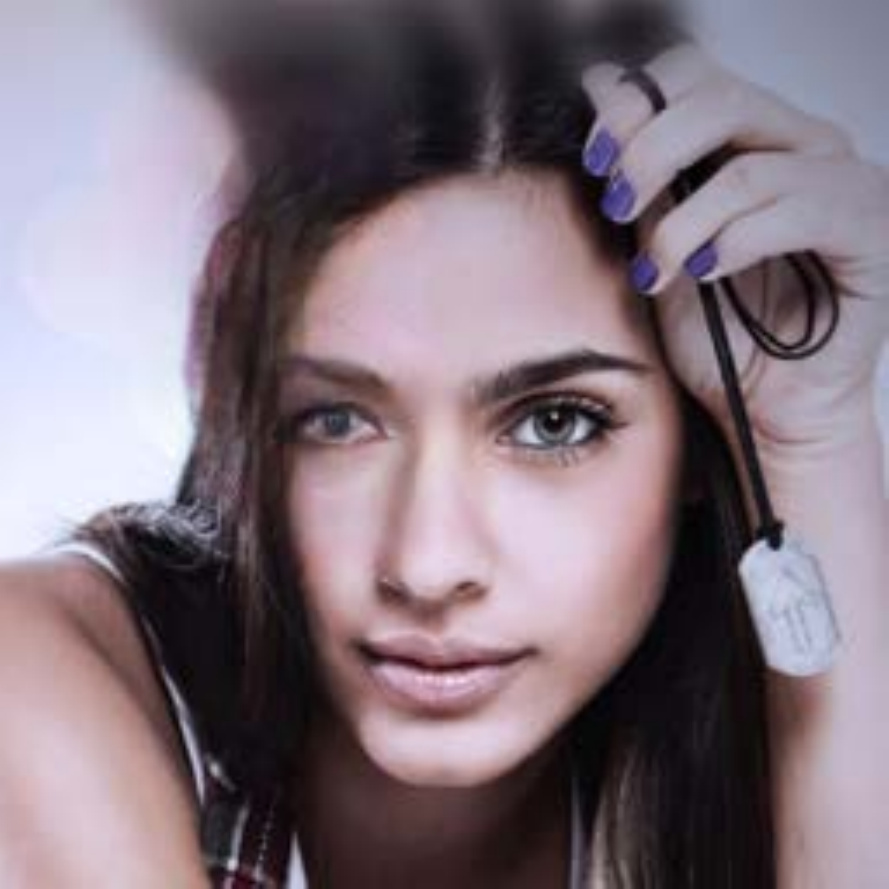}
        \includegraphics[width=2.5cm]{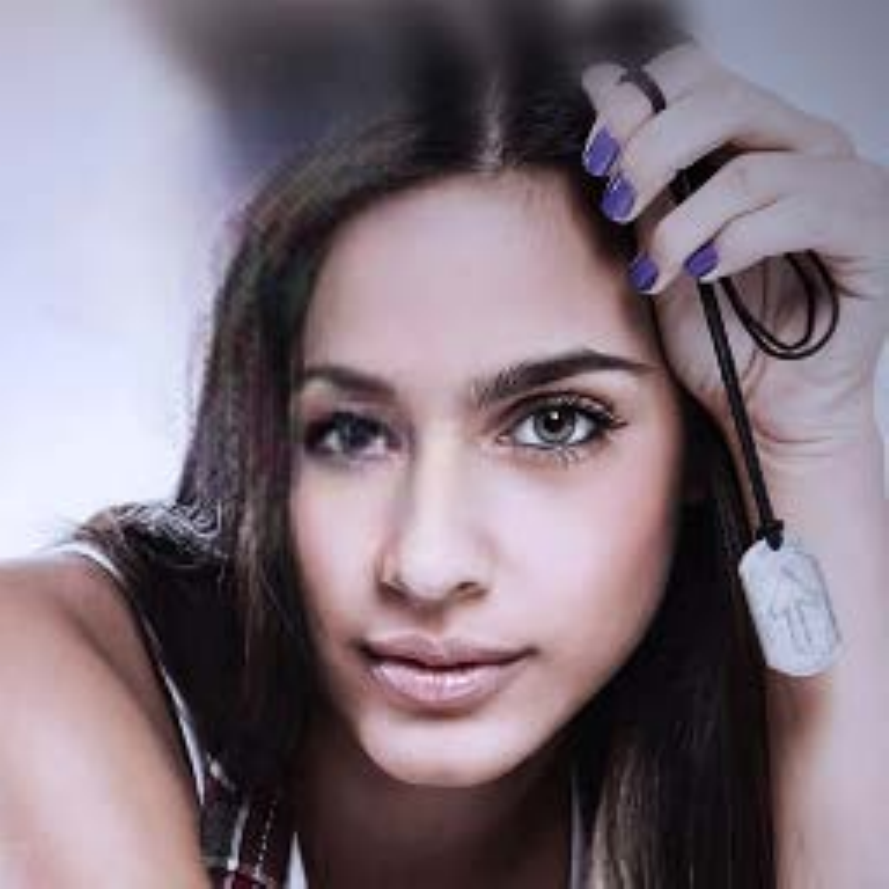}
        \includegraphics[width=2.5cm]{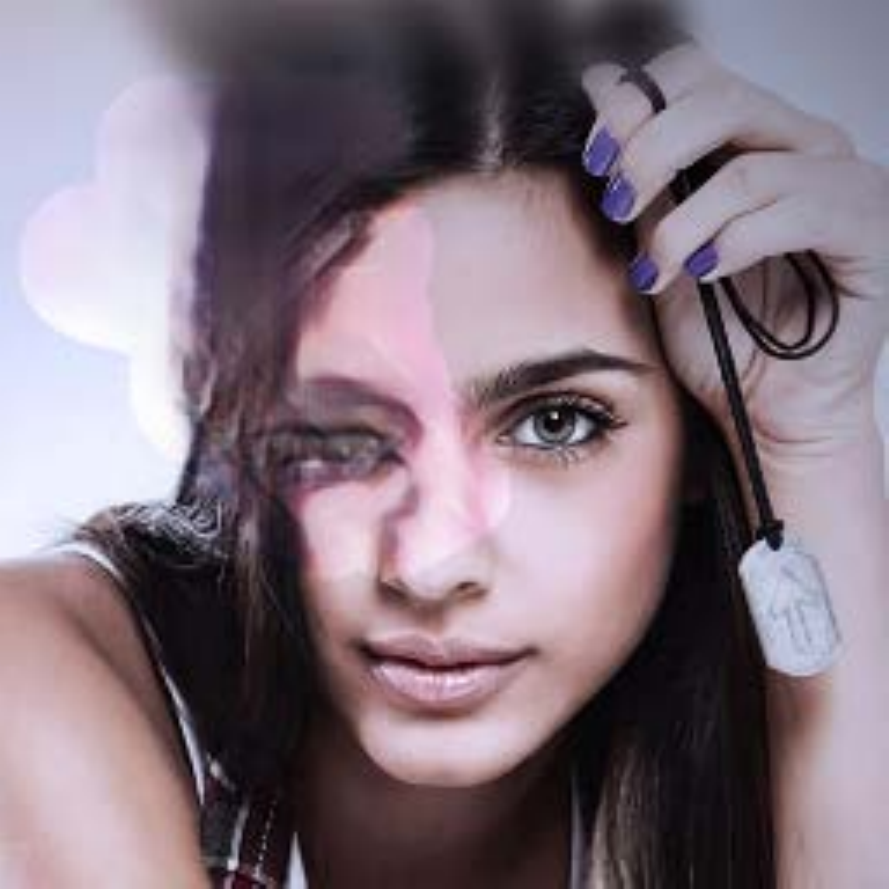}
        \includegraphics[width=2.5cm]{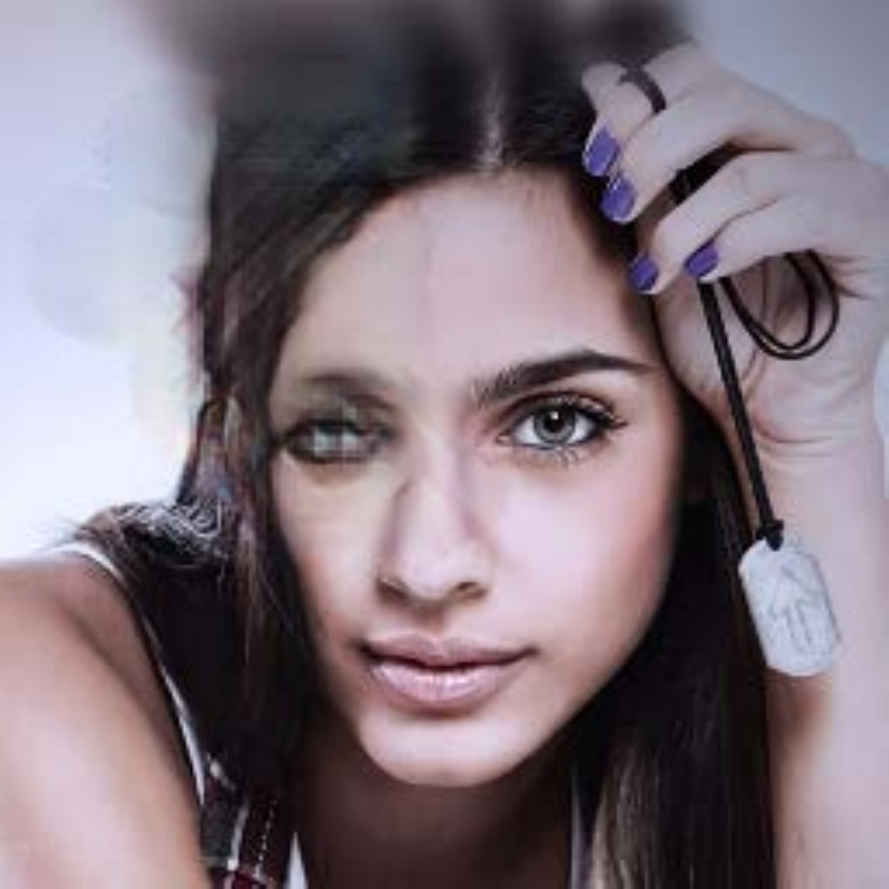}
        \includegraphics[width=2.5cm]{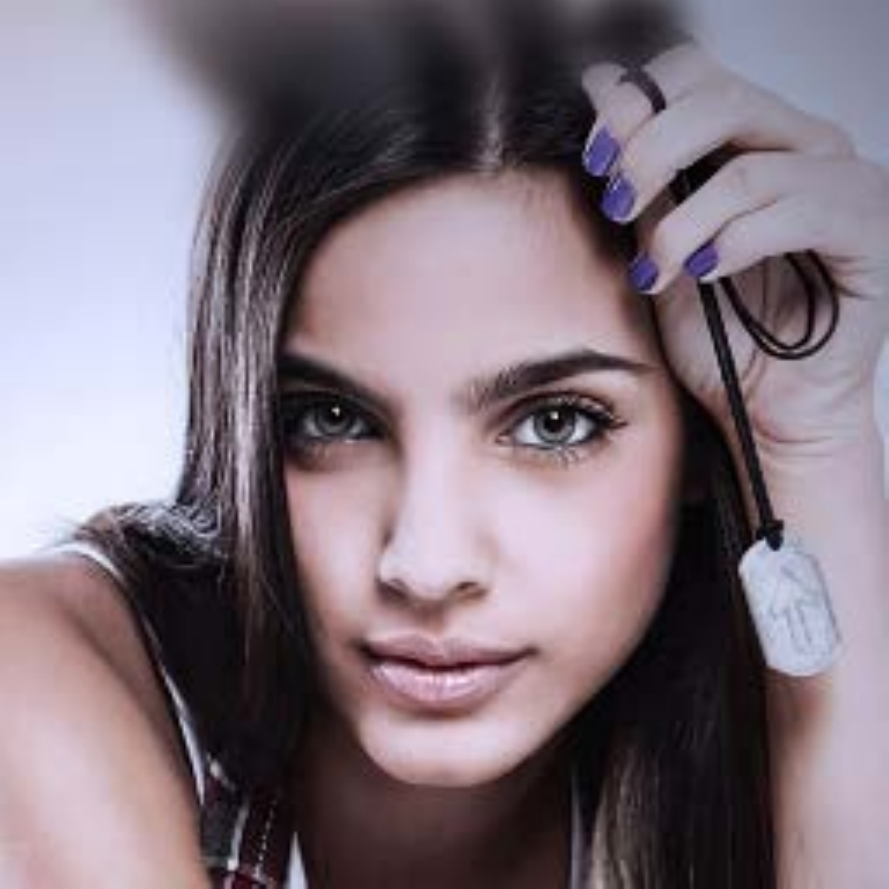}
    \end{subfigure}
    \begin{subfigure}
        \centering
        \vspace{-0.05in}
        \includegraphics[width=2.5cm]{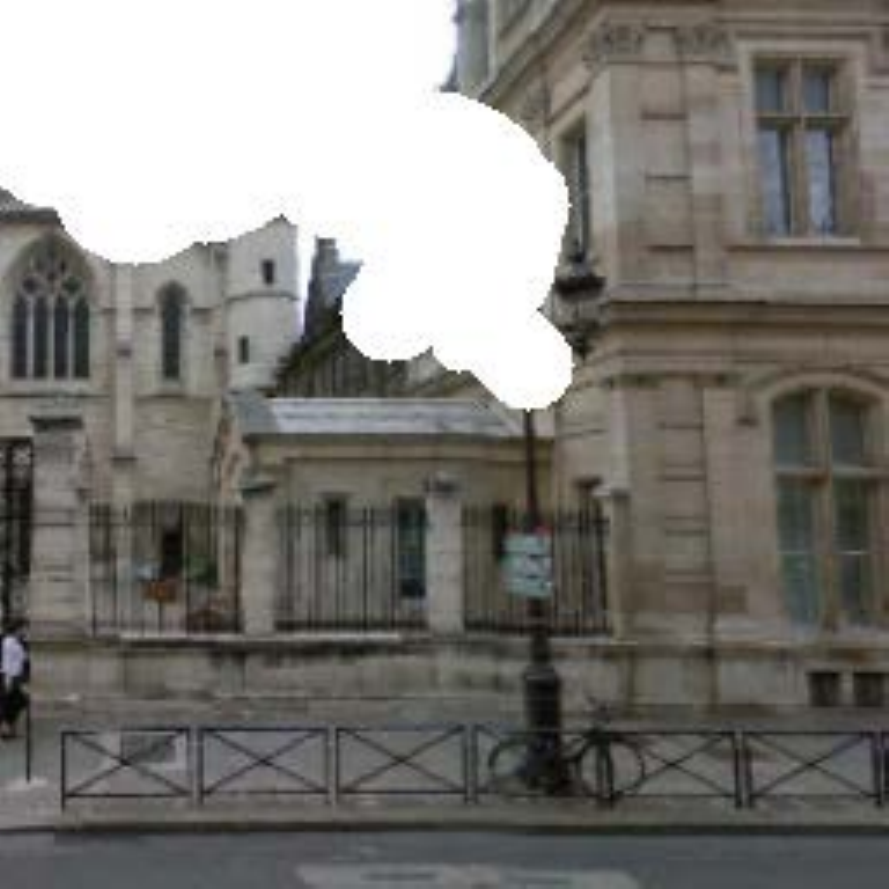}
        \includegraphics[width=2.5cm]{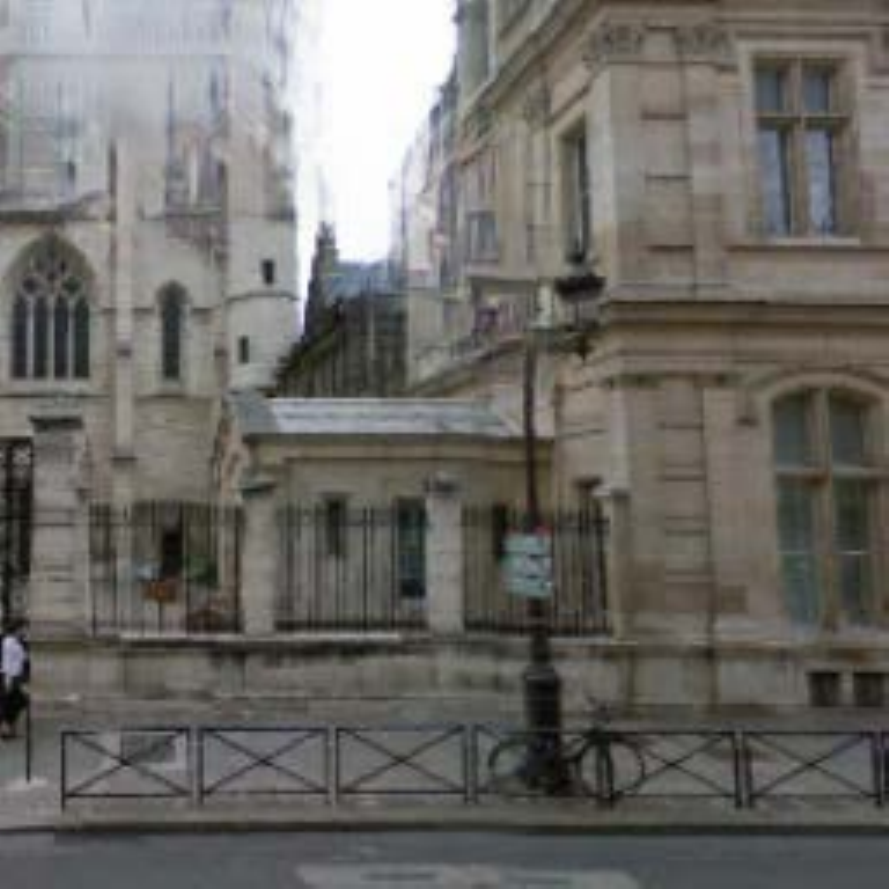}
        \includegraphics[width=2.5cm]{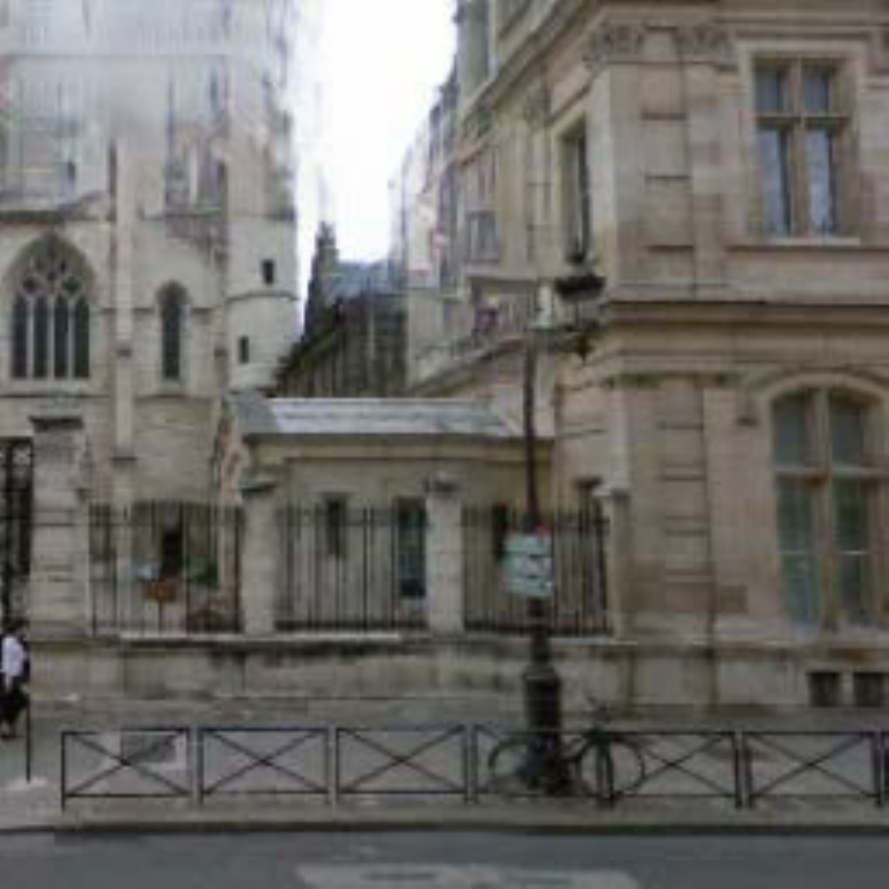}
        \includegraphics[width=2.5cm]{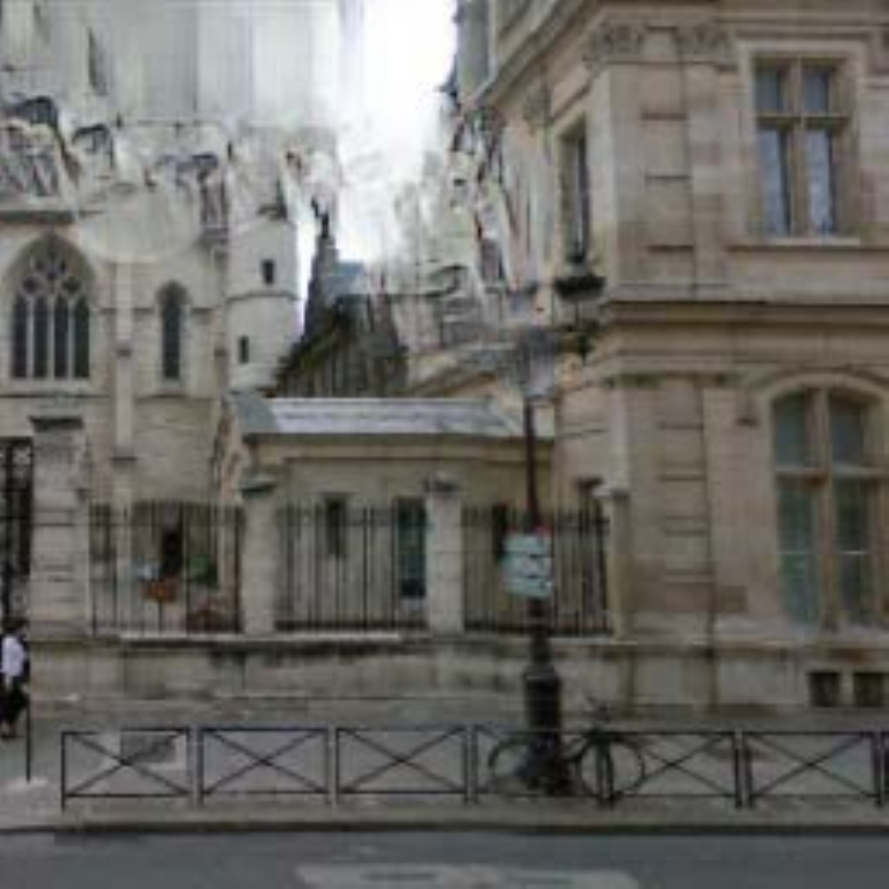}
        \includegraphics[width=2.5cm]{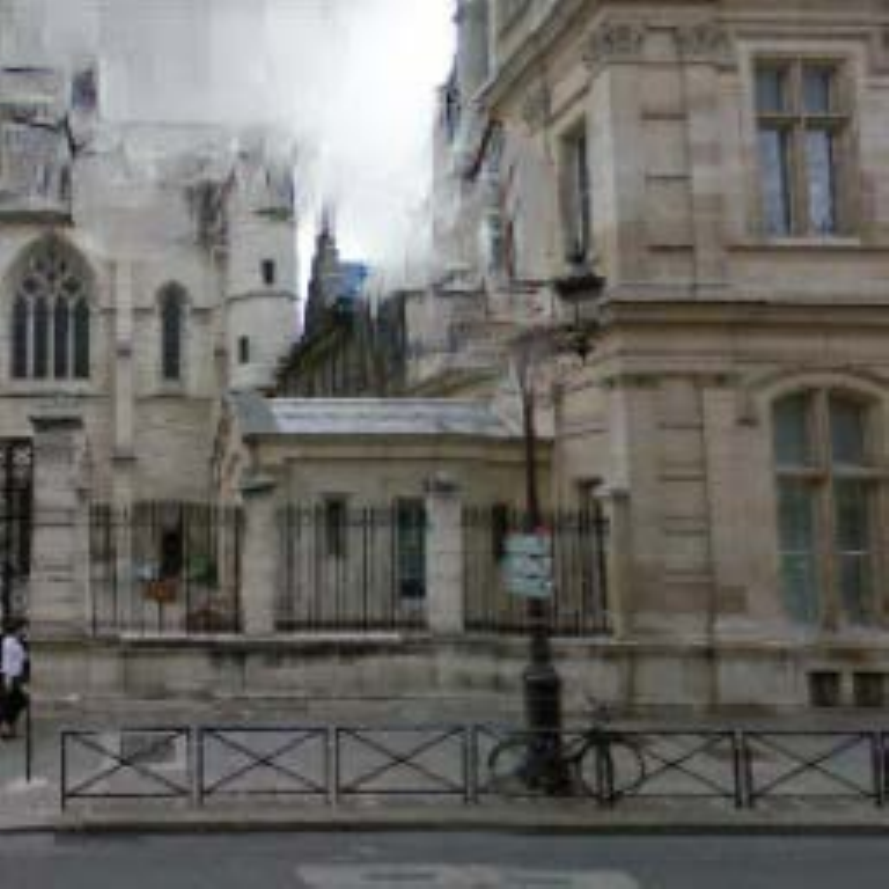}
        \includegraphics[width=2.5cm]{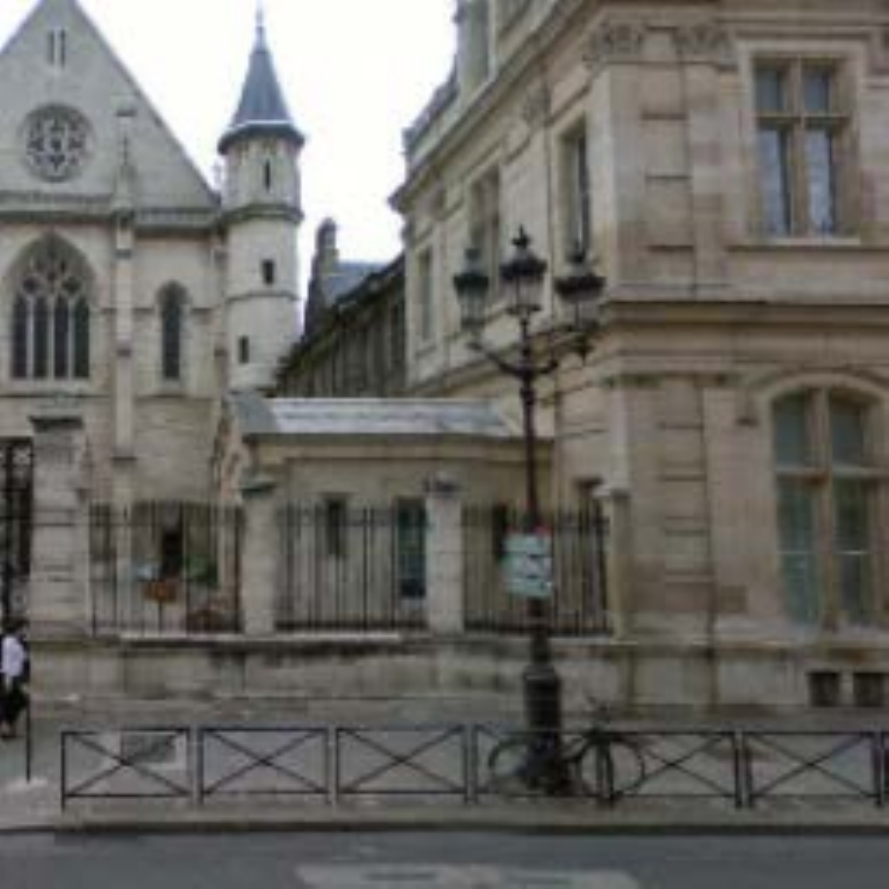}
    \end{subfigure}
    \begin{subfigure}
        \centering
        \vspace{-0.05in}
        \includegraphics[width=2.5cm]{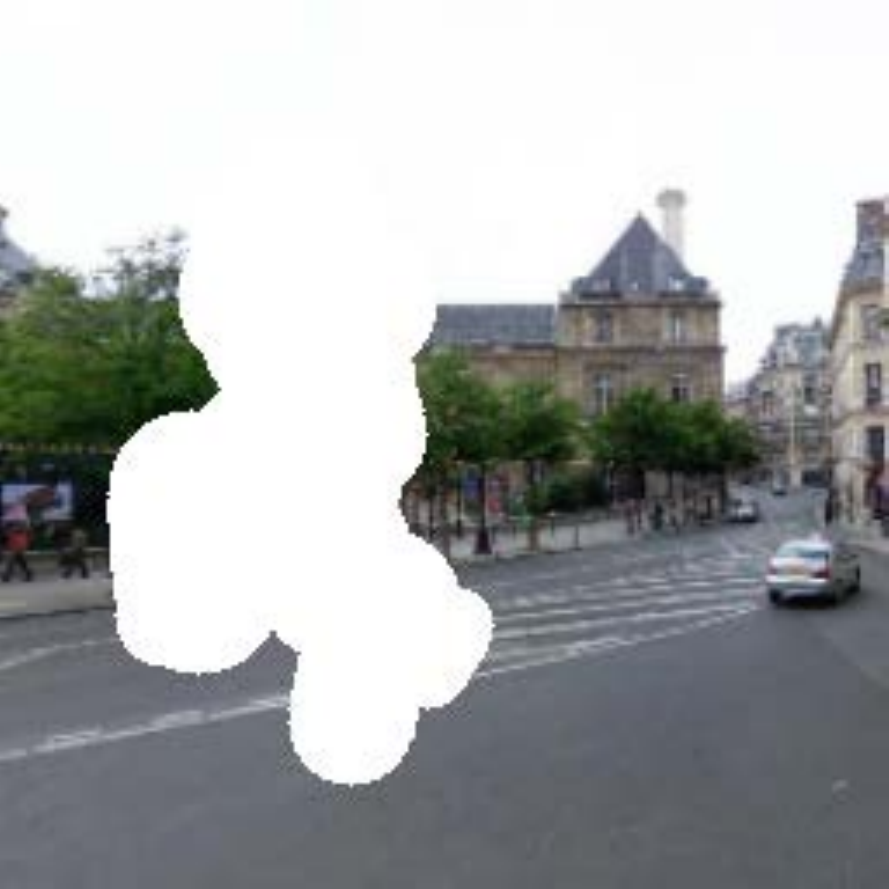}
        \includegraphics[width=2.5cm]{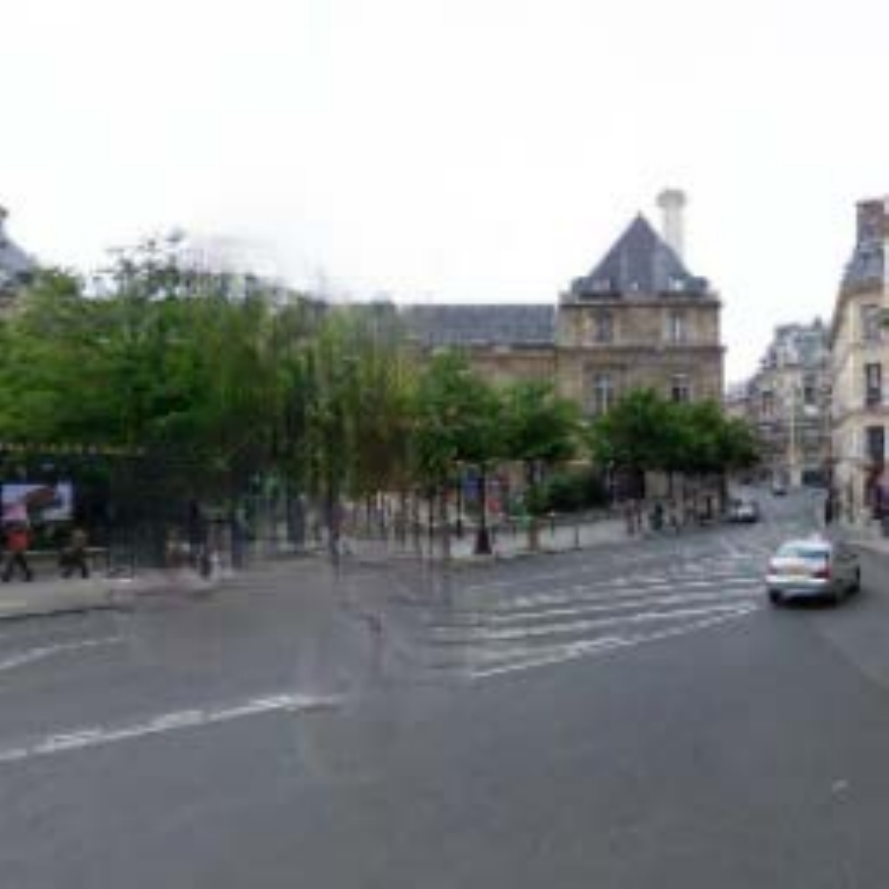}
        \includegraphics[width=2.5cm]{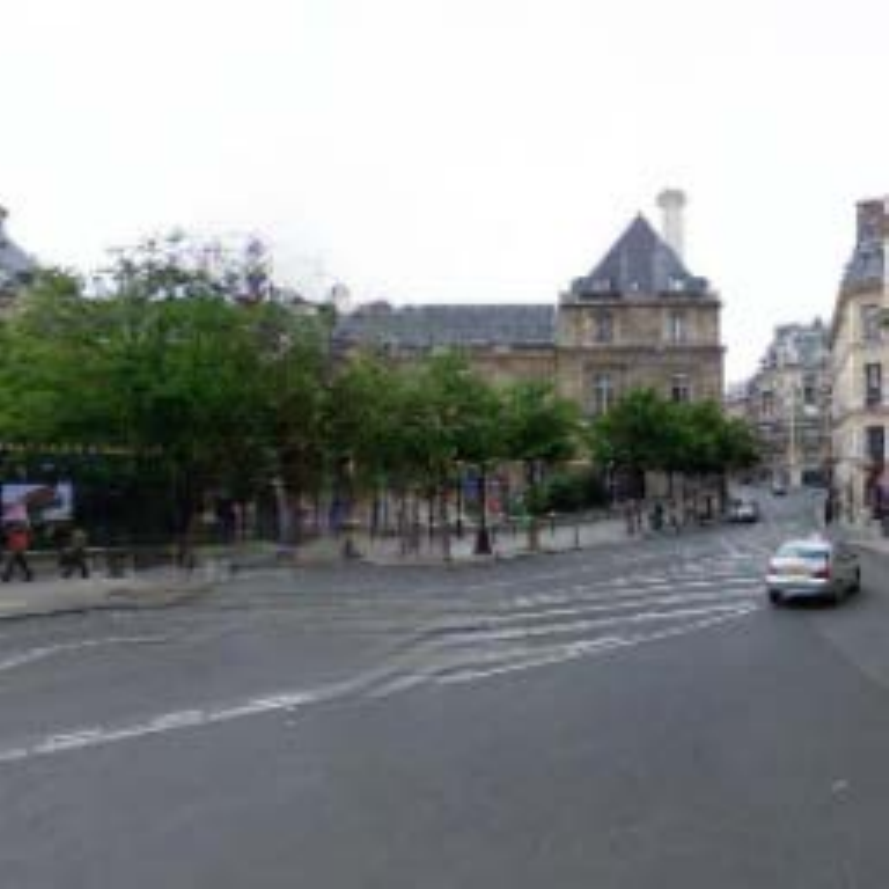}
        \includegraphics[width=2.5cm]{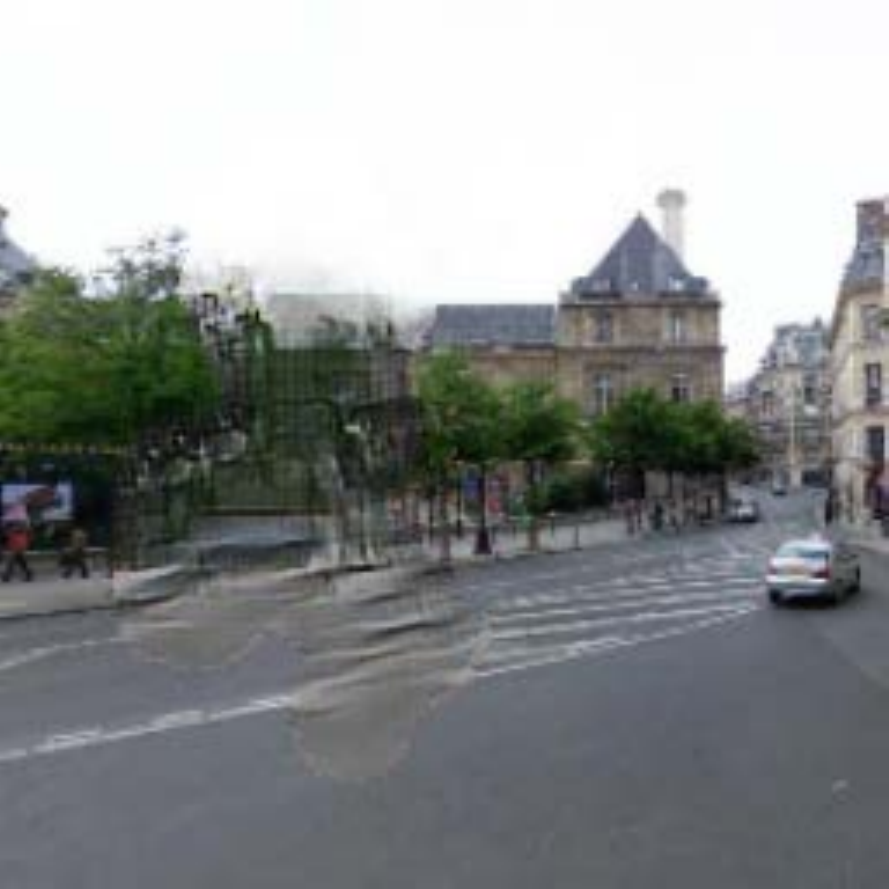}
        \includegraphics[width=2.5cm]{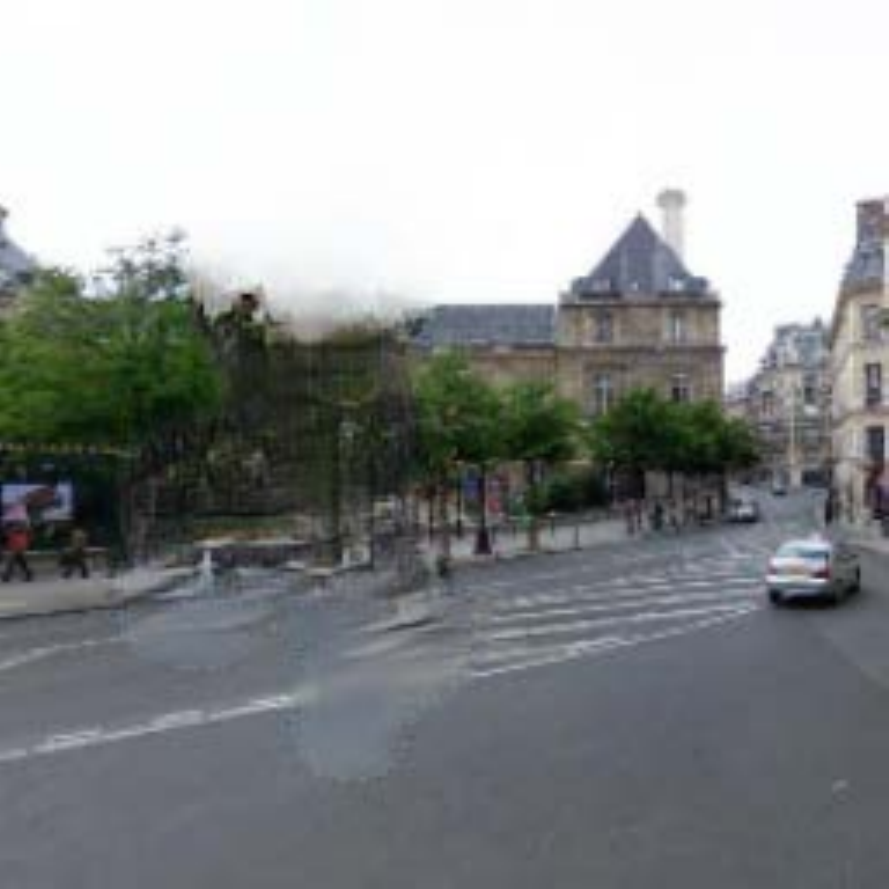}
        \includegraphics[width=2.5cm]{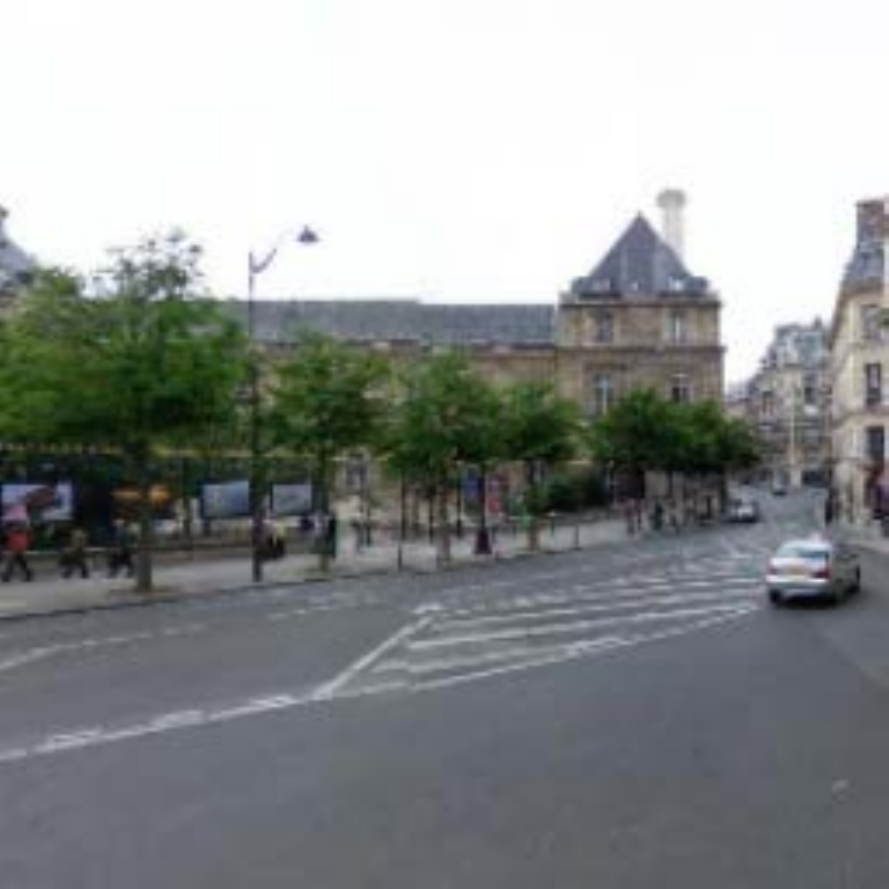}
    \end{subfigure}  
    \centering
    \vspace{-0in}
    \\
    \centering
    (a) Input   \hspace{1.0cm}
    (b) EC w/o $\mathcal{L}_p$     \hspace{0.9cm}
    (c) EC   \hspace{0.8cm}
    (d) Pconv  w/o $\mathcal{L}_p$     \hspace{0.5cm}
    (e) Pconv \hspace{1.2cm}
    (f) GT \hspace{0.5cm}
    \caption{Effect of the perceptual loss $\mathcal{L}_p$ in EC and PConv.}
    \label{fig:loss comparisons on EC}
\end{figure*}

\section{Experiments} \label{sec:4}
In this section, we first analyze the effect of different loss functions, and study the performance caused by each component of our adversarial inpainting framework. Then we evaluate our proposed method visually and quantitatively over several common datasets in image inpainting compared to state-of-the-art methods. Our code is available at \url{https://github.com/vickyFox/Region-wise-Inpainting}.
%More results could be found in the supplementary material\footnote{https://anonymousfiles.io/OIY948mZ/}. %approach visually and quantitatively compared to some state-of-the-art approaches. We will first introduce the datasets and our compared methods, and then show the effectiveness of our method. 

\subsection{Datasets and Protocols} 
We employ the widely-used datasets in prior studies, including  CelebA-HQ
\cite{karras2017progressive}, Places2 \cite{zhou2018places}, and Paris StreetView \cite{doersch2012makes}. CelebA-HQ contains 30k high-resolution face images, and we adopt the same partition as \cite{yu2018generative} did. The Places2 dataset includes 8,097,967 training images with diverse scenes. The Paris StreetView contains 14,900 training images and 100 test images. For both datasets, we adopt the original train, test, and validate splits. 

We compare our method with four state-of-the-art models, namely, Contextual Attention (CA) \cite{yu2018generative}, Partial Convolution (PConv) \cite{liu2018image}, EdgeConnect (EC) \cite{nazeri2019edgeconnect}, and Pluralistic Image Completion (PIC) \cite{Zheng2019Pluralistic}. Among those models, CA are initially designed for regular missing regions, while PConv, EC, PIC and ours focus on irregular holes. We directly apply their released pre-trained models in our experiments. As to PConv, since there is no published codes, we borrow the implementation on github \textcolor{red}{\footnote{https://github.com/MathiasGruber/PConv-Keras}}, and retrain the model following the authors' advice.
 
We compare our model with state-of-the-art both visually and quantitatively. As for quantitative protocols, following \cite{nazeri2019edgeconnect}, we use the following quantitative metrics: 1) $\ell_1$ error, 2) $\ell_2$ error, 3) peak signal-to-noise ratio (PSNR), and 4) structural similarity index (SSIM). These metrics assume pixel-wise independence, and can help to compare the visual appearance of different inpainting images. But in practice, they may assign favorable scores to perceptually inaccurate results. Recent studies \cite{Xu2018An} have shown that metrics based on deep features are closer to those based on human perception. Therefore, we also adopt another metric, namely Frechet Inception Dsitance (FID).

\begin{table}[tp]
\caption{Different loss functions used by different methods.}
\label{tab:loss-used}
\centering
\begin{tabular}{c|c|c|c|c|c |c |c}
     \toprule
     methods &  $\mathcal{L}_r$ & $\mathcal{L}_{p}$ & $\mathcal{L}_{s}$  & $\mathcal{L}_{a}$ & $\mathcal{L}_c$ & $\mathcal{L}_{k}$ & $\mathcal{L}_{f}$\\
     \midrule
     PConv   & \checkmark  & \checkmark & \checkmark  & - & - & - & -\\
     \midrule
     EC      & \checkmark &\checkmark &\checkmark &\checkmark & - & - & \checkmark\\
     \midrule
     PIC     &\checkmark  &- &- & \checkmark& - & \checkmark & -\\
     \midrule
     Ours    & \checkmark & - & \checkmark & \checkmark & \checkmark & - & -\\
     \bottomrule
\end{tabular}
\end{table}

\begin{figure*}[htb]
\label{fig: contribution}
    \vspace{0.02in}
    \centering
    \begin{subfigure}
        \centering
        \includegraphics[width=2.4cm]{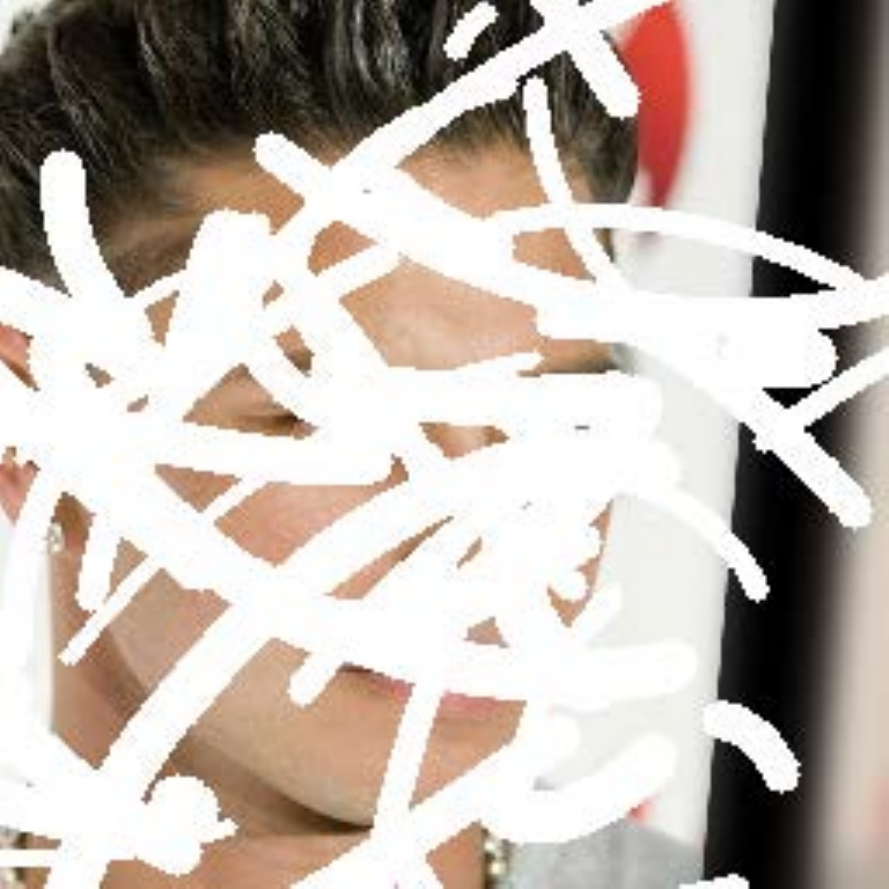}
        \includegraphics[width=2.4cm]{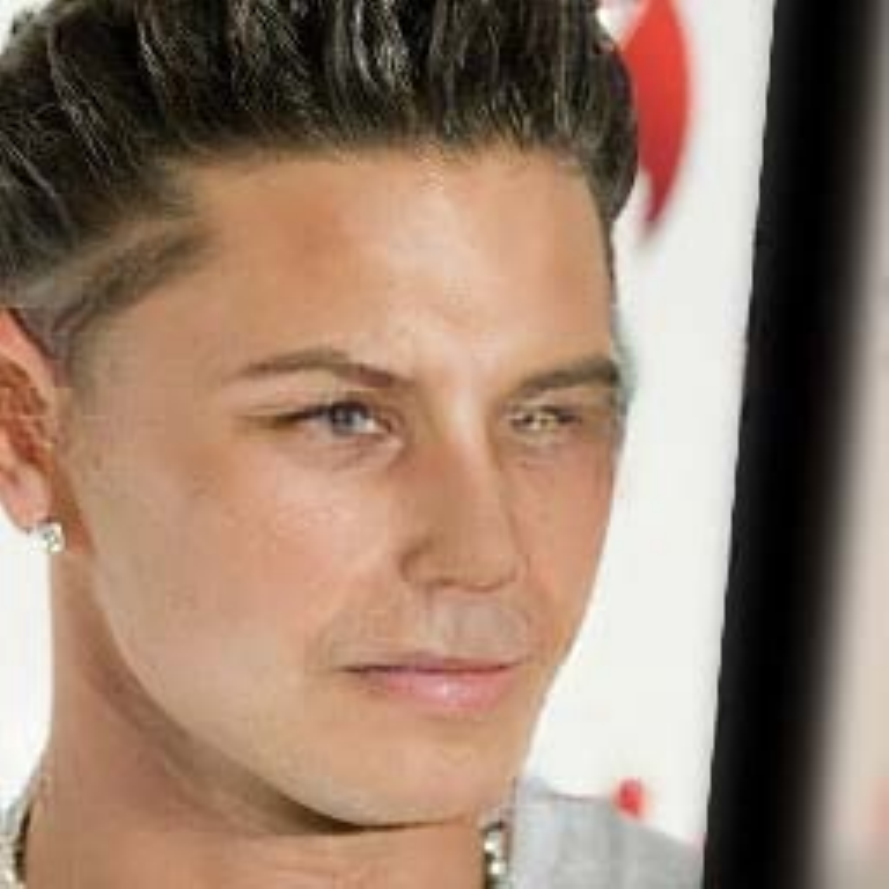}
        \includegraphics[width=2.4cm]{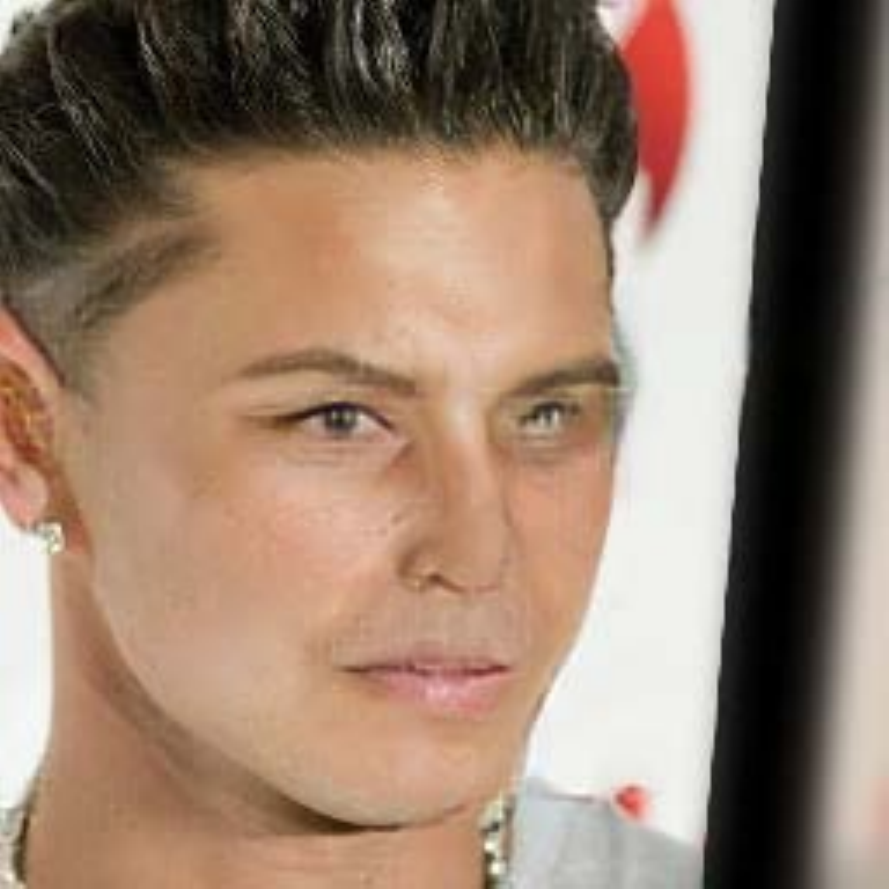}
        \includegraphics[width=2.4cm]{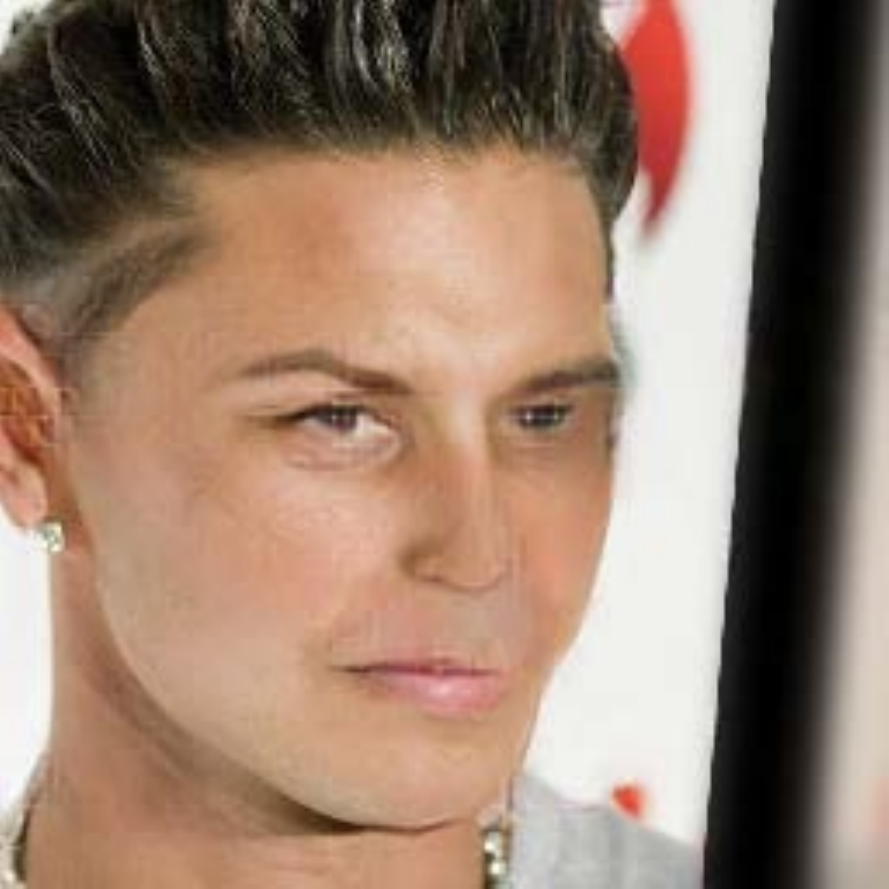}
        \includegraphics[width=2.4cm]{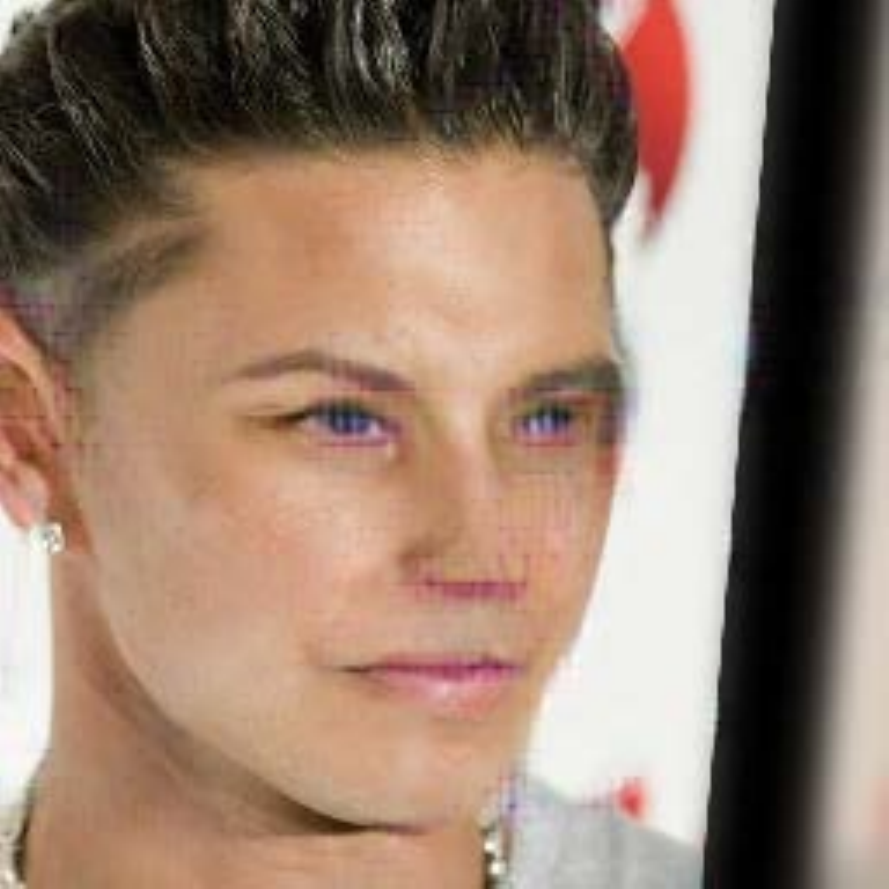}
        \includegraphics[width=2.4cm]{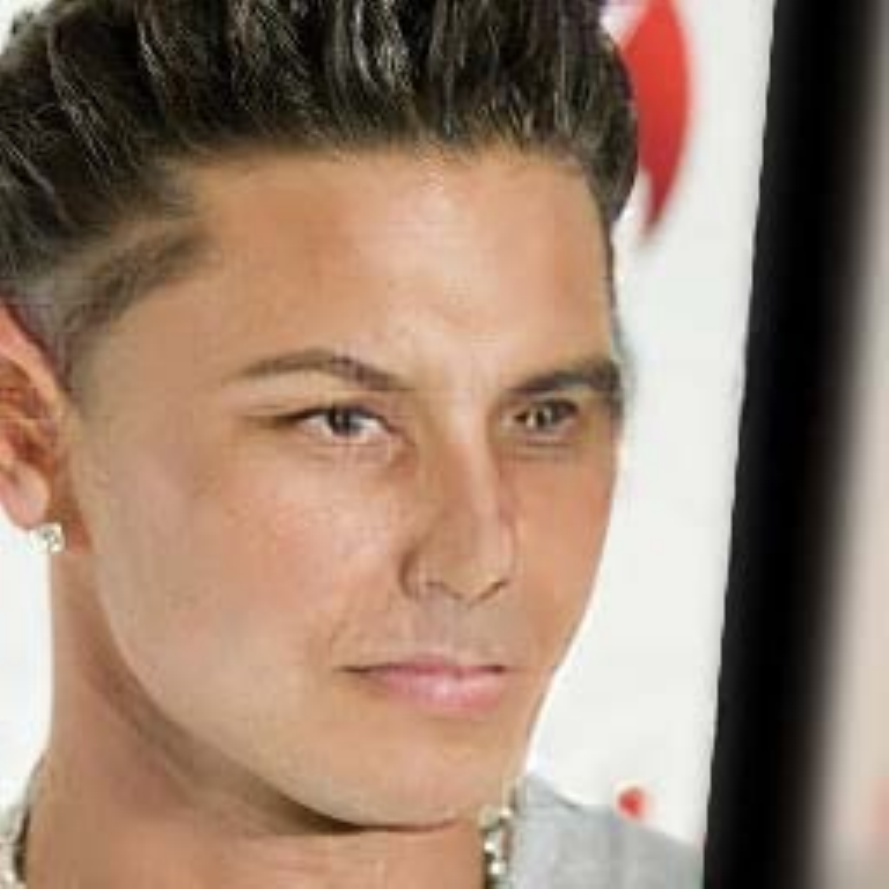}
        \includegraphics[width=2.4cm]{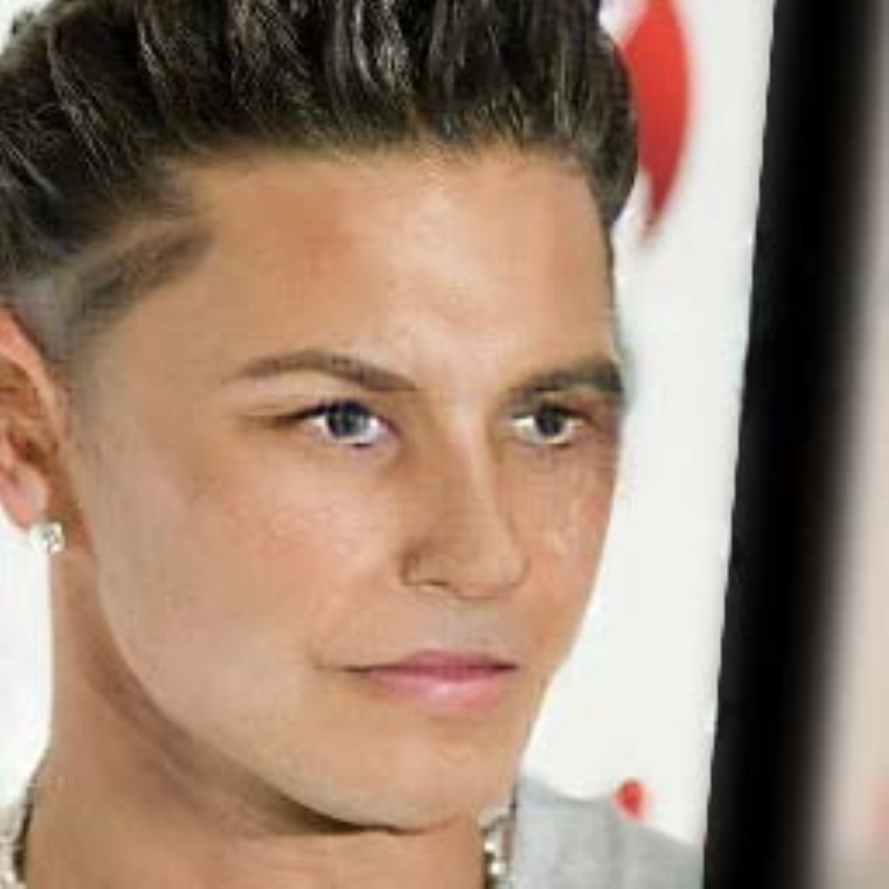}
    \end{subfigure}
    \begin{subfigure}
        \centering
        \vspace{-0.05in}
        \includegraphics[width=2.4cm]{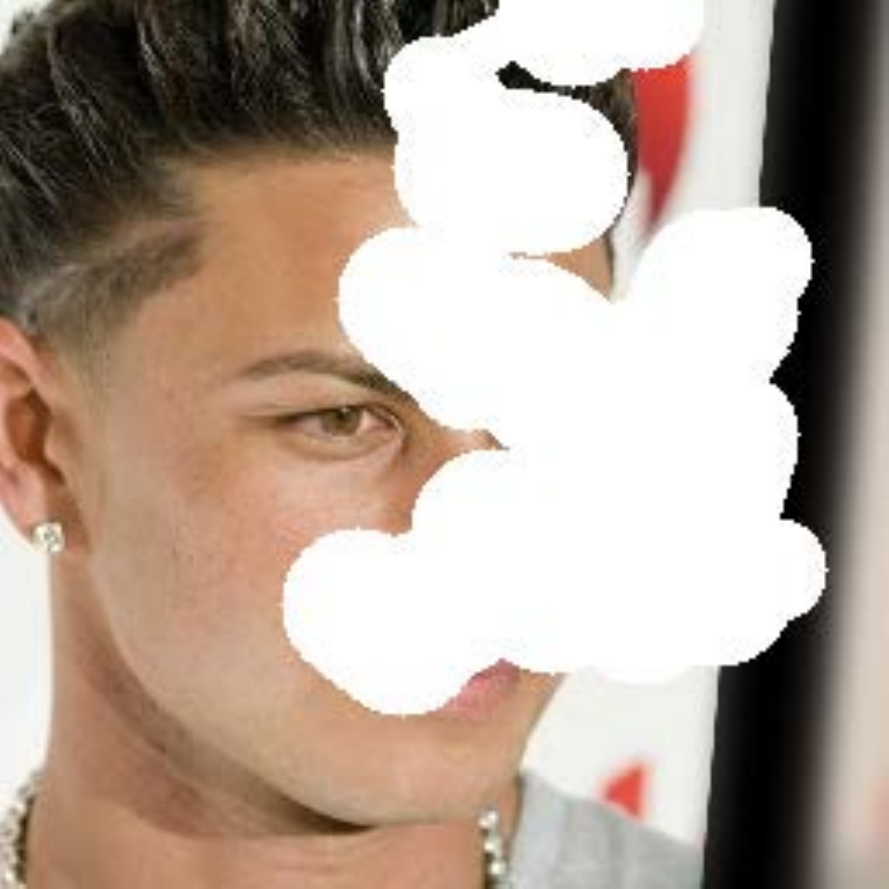}
        \includegraphics[width=2.4cm]{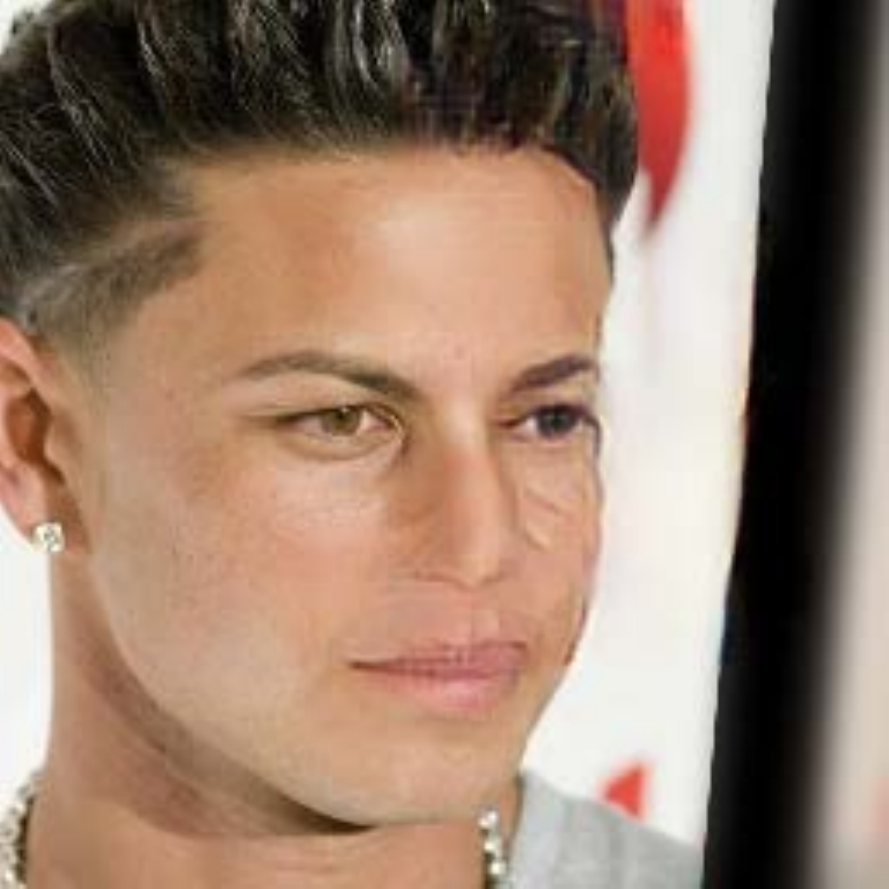}
        \includegraphics[width=2.4cm]{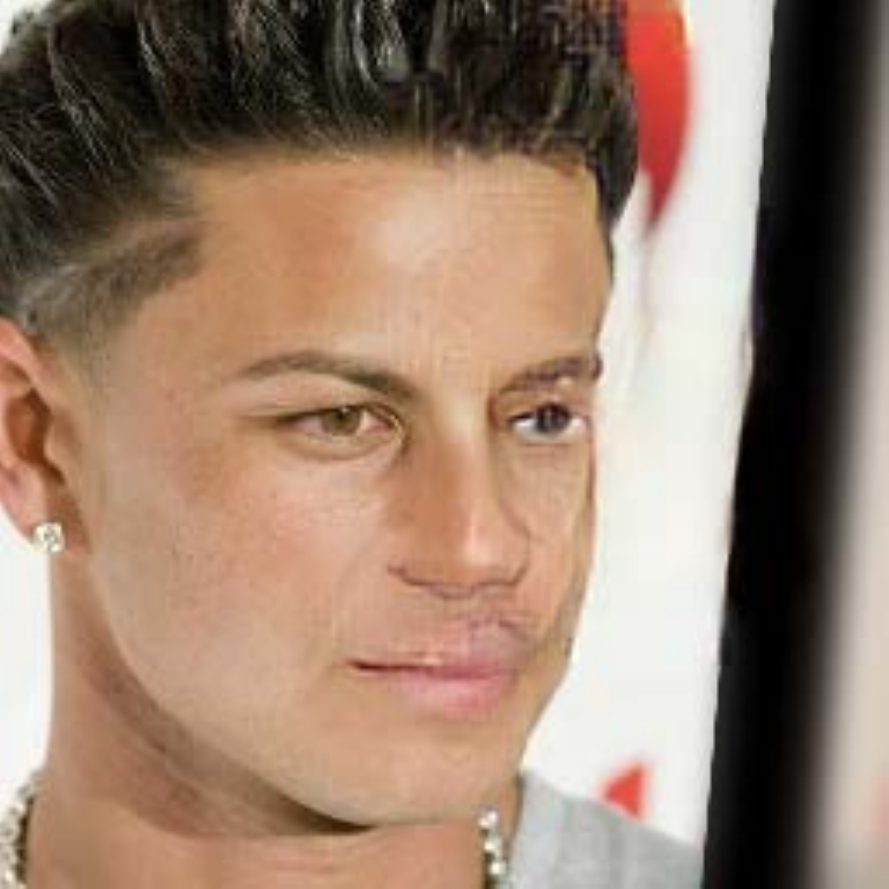}
        \includegraphics[width=2.4cm]{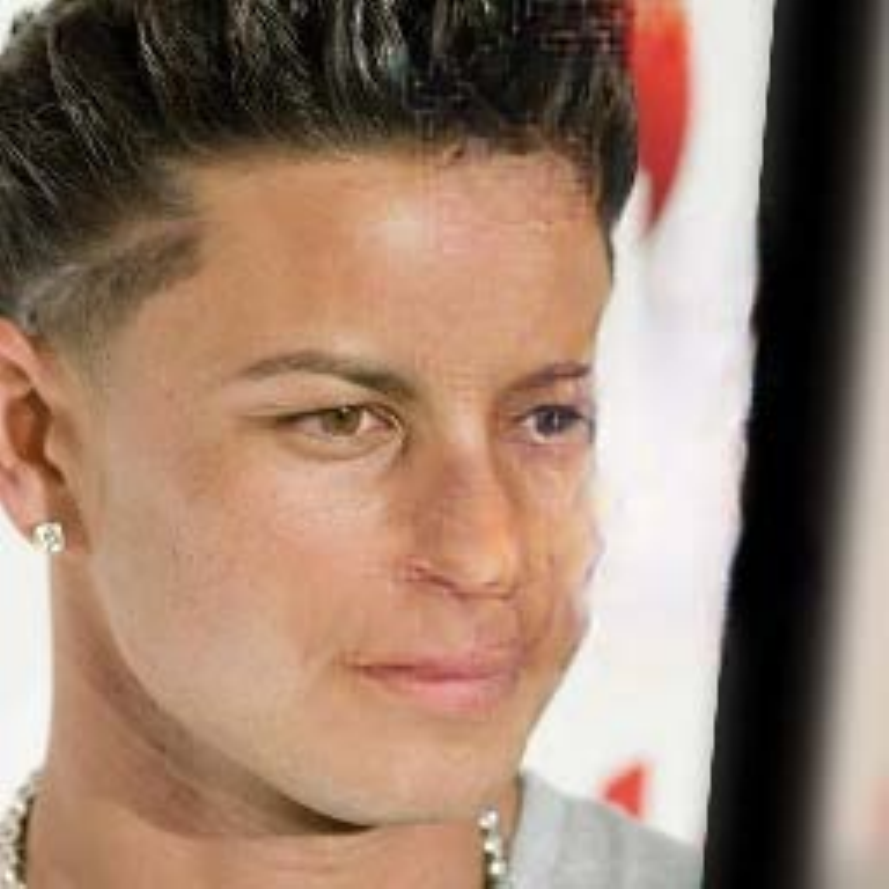}
        \includegraphics[width=2.4cm]{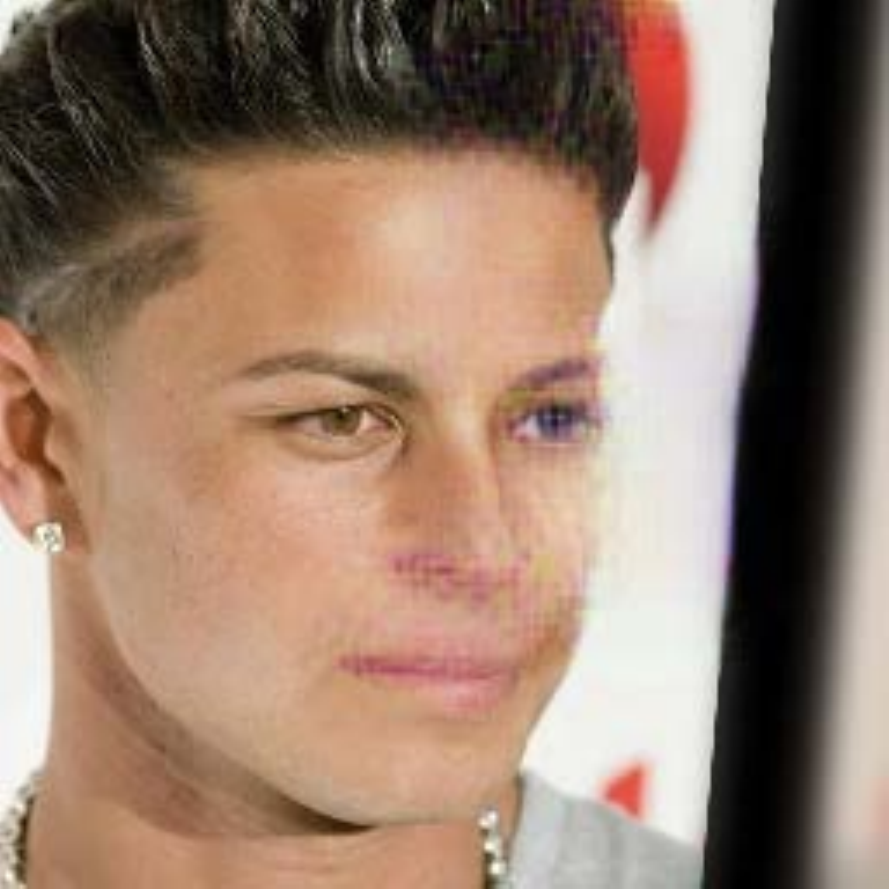}
        \includegraphics[width=2.4cm]{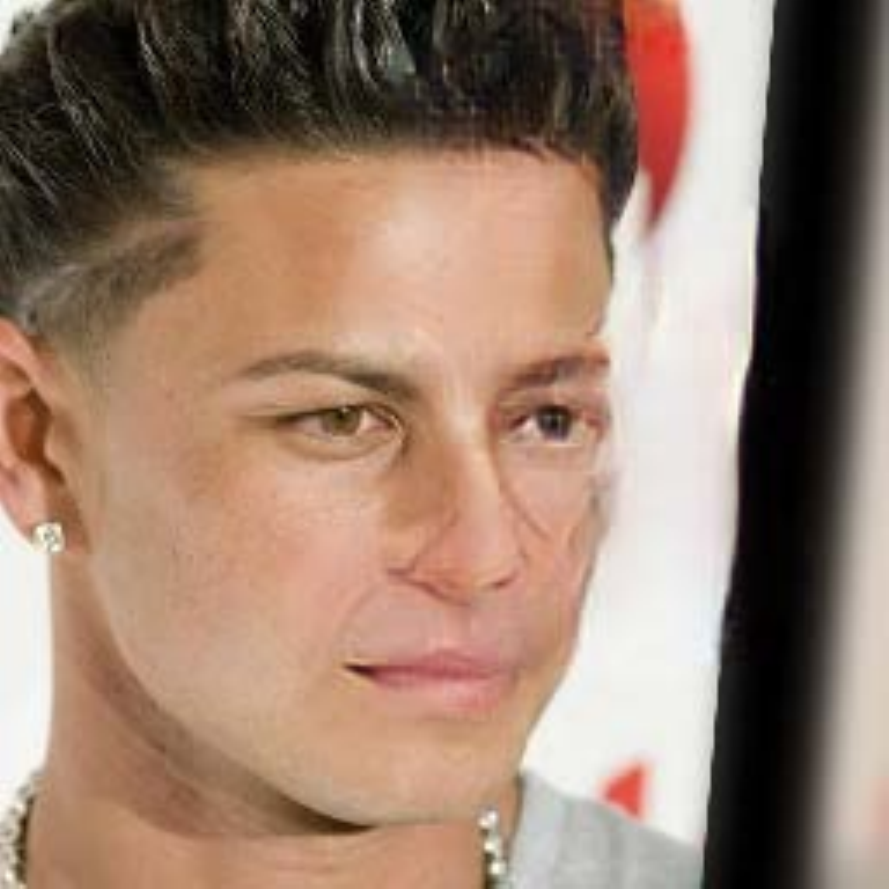}
        \includegraphics[width=2.4cm]{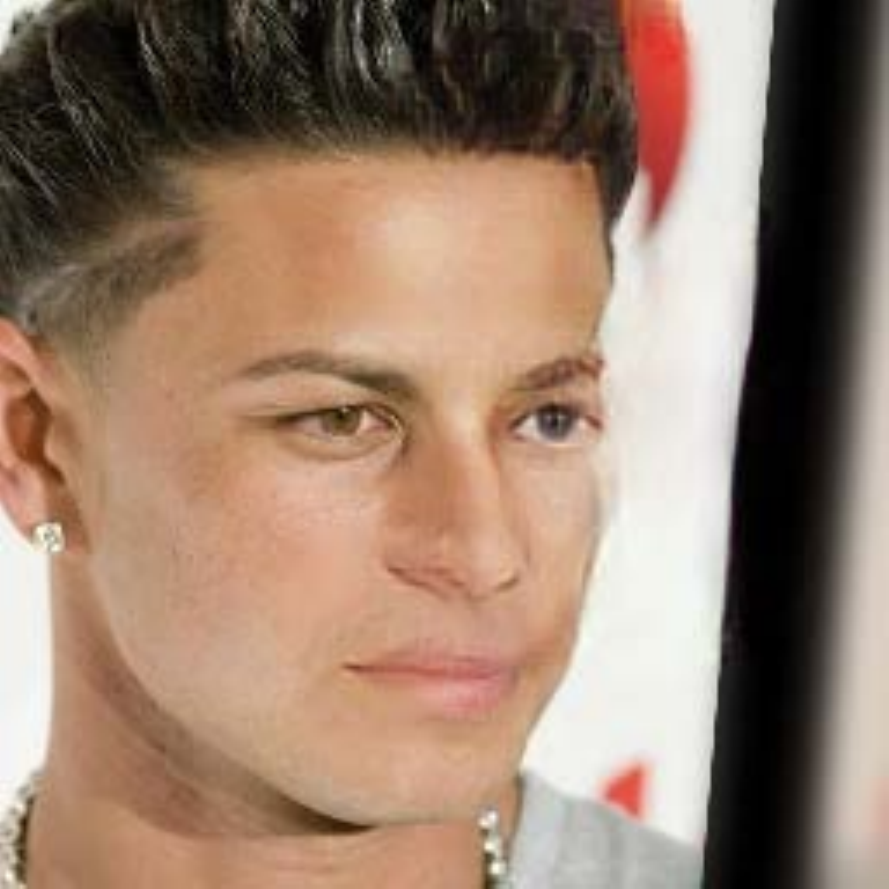}
    \end{subfigure}
    \begin{subfigure}
        \centering
        \vspace{-0.05in}
        \includegraphics[width=2.4cm]{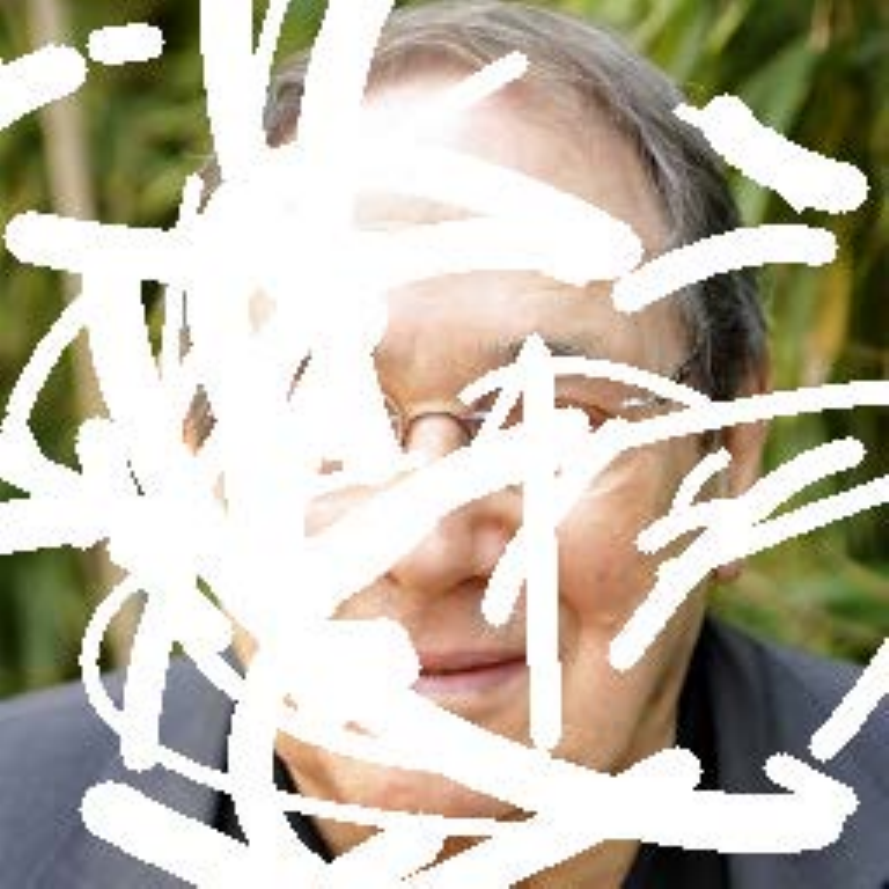}
        \includegraphics[width=2.4cm]{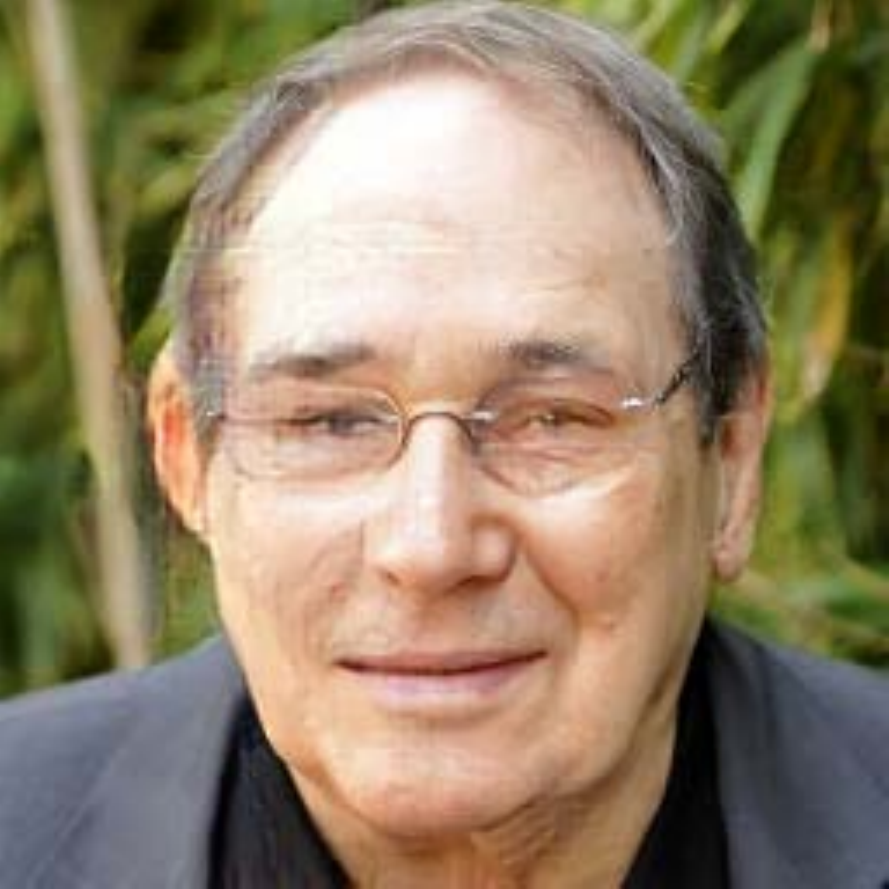}
        \includegraphics[width=2.4cm]{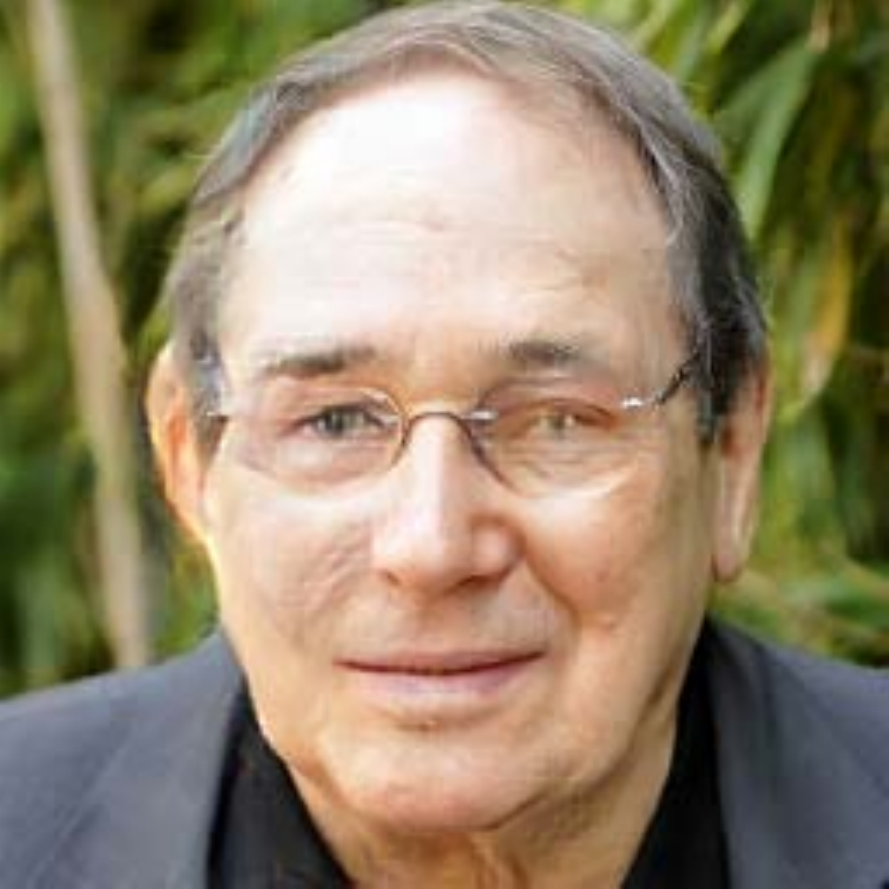}
        \includegraphics[width=2.4cm]{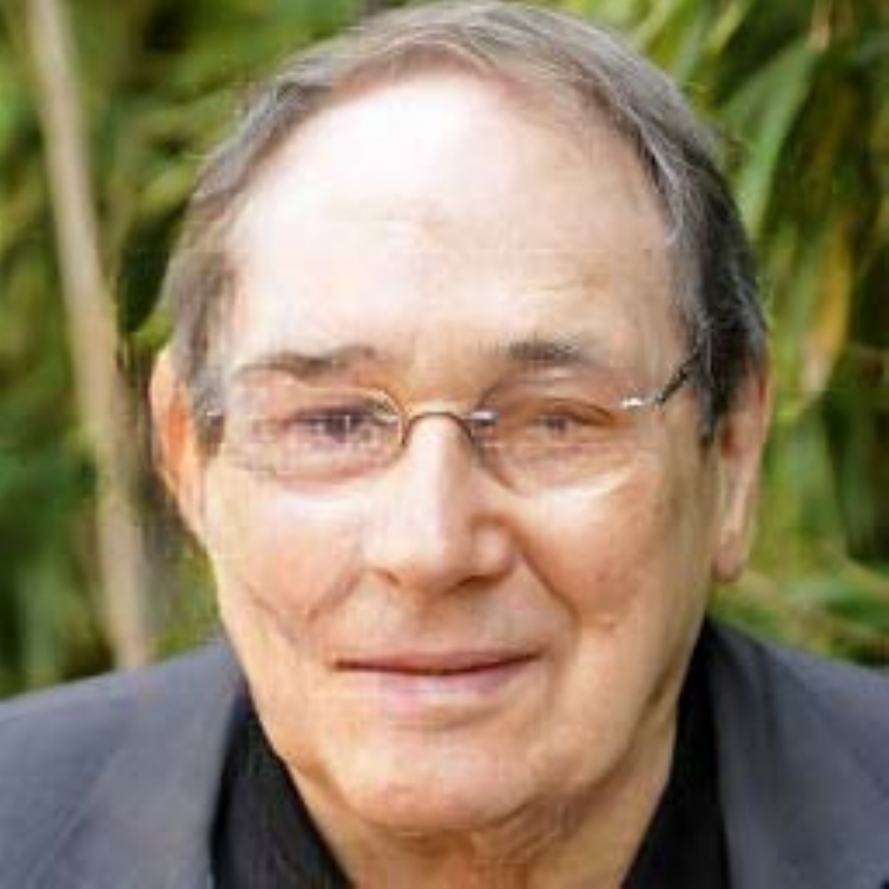}
        \includegraphics[width=2.4cm]{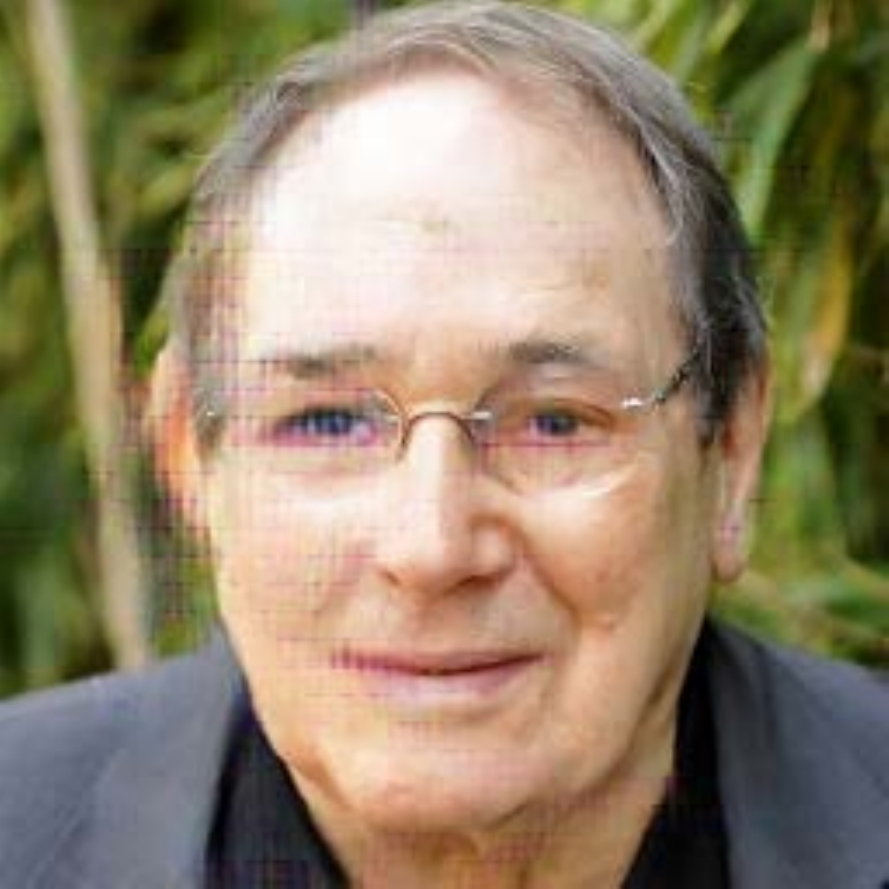}
        \includegraphics[width=2.4cm]{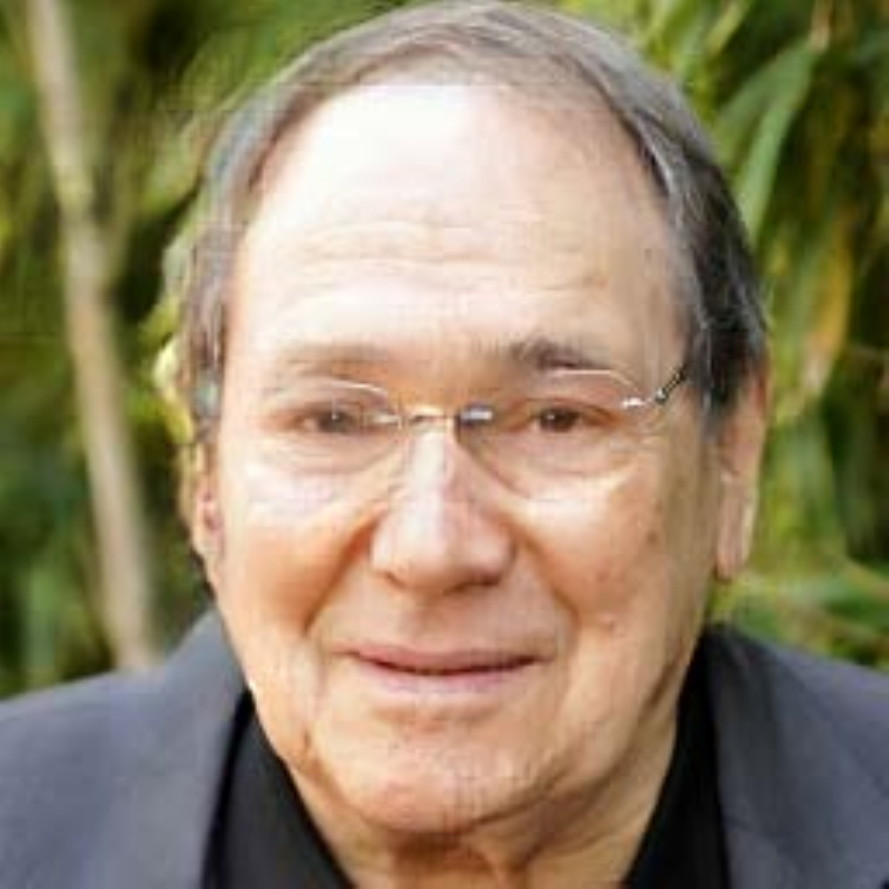}
        \includegraphics[width=2.4cm]{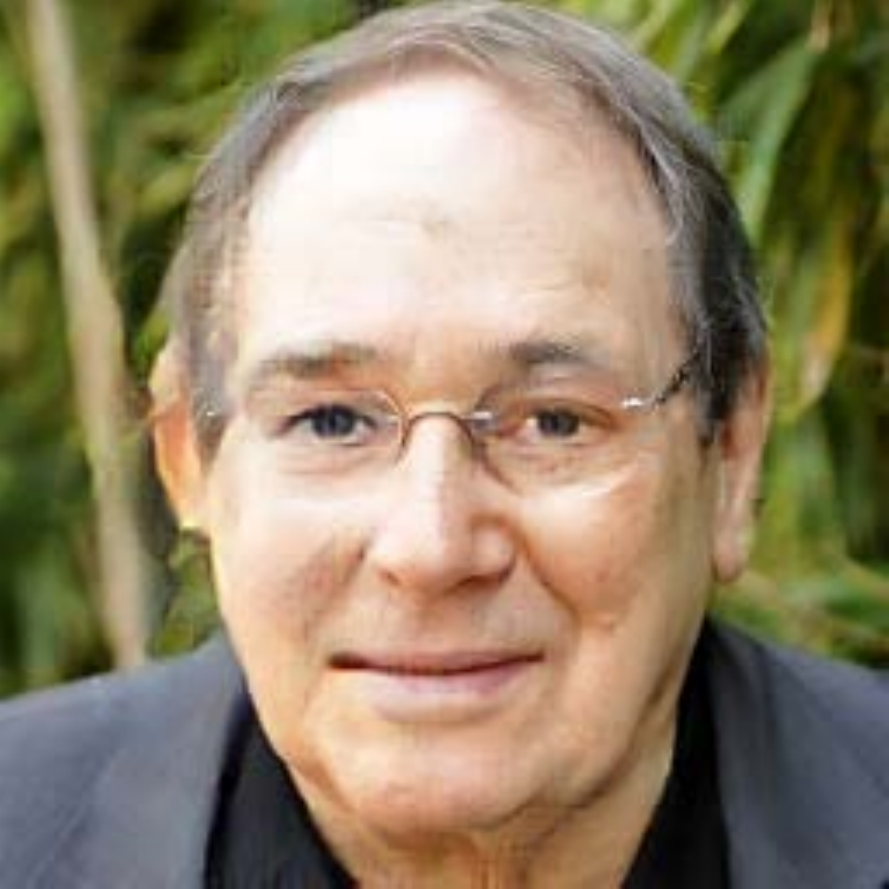}
    \end{subfigure}
    \begin{subfigure}
        \centering
        \vspace{-0.05in}
        \includegraphics[width=2.4cm]{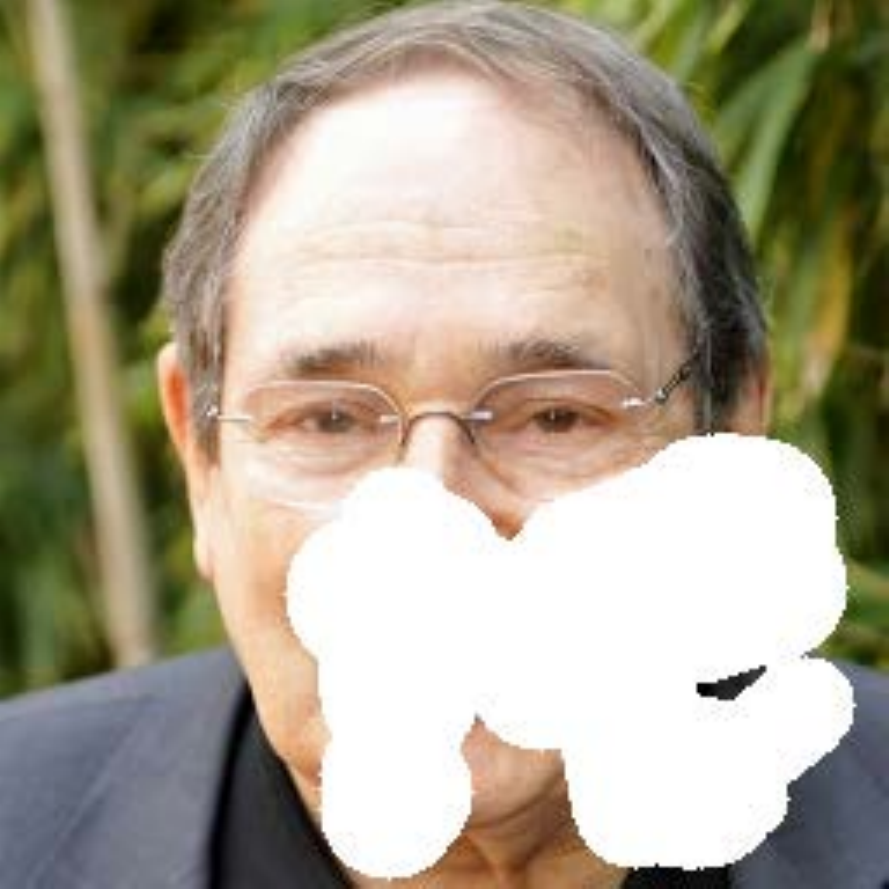}
        \includegraphics[width=2.4cm]{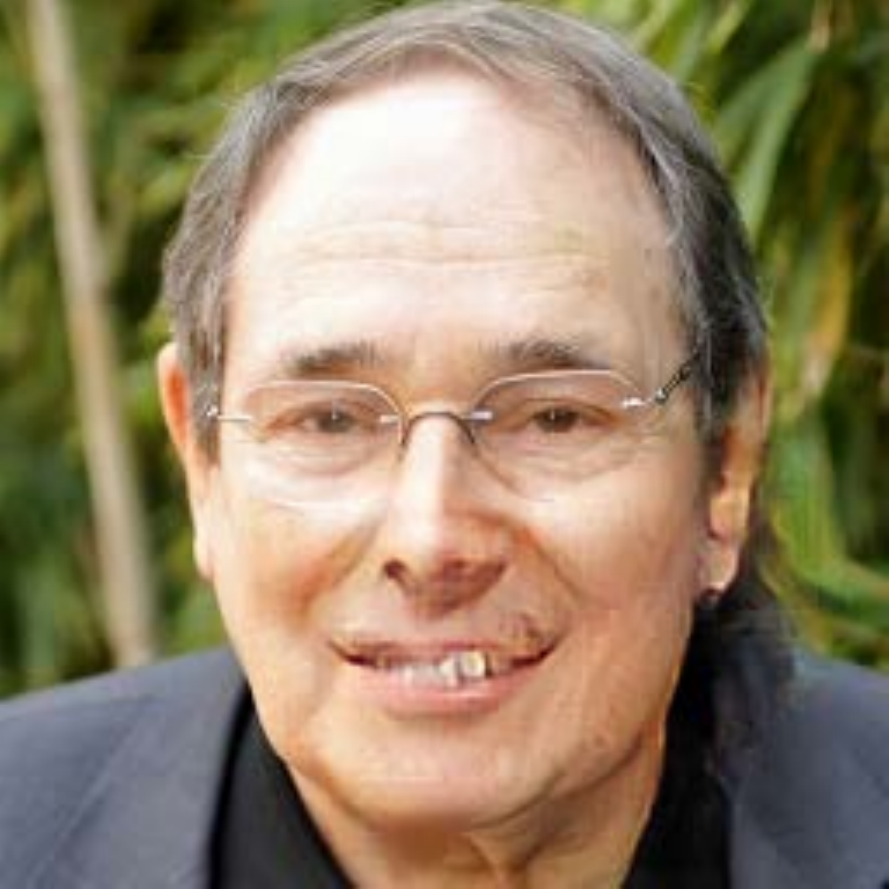}
        \includegraphics[width=2.4cm]{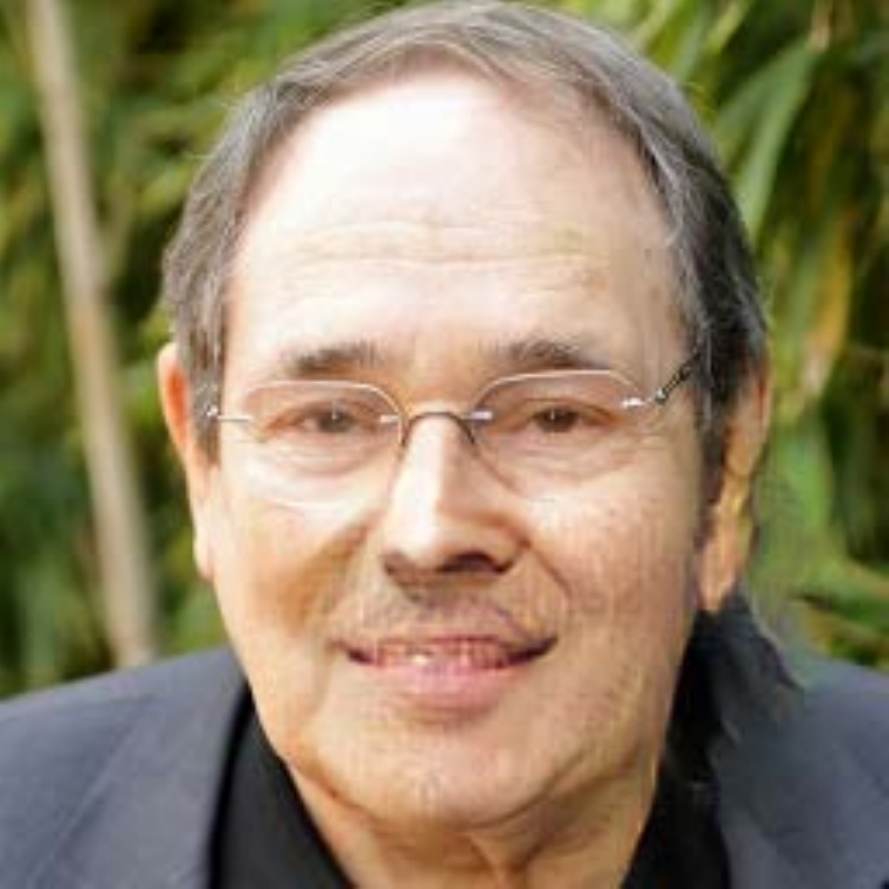}
        \includegraphics[width=2.4cm]{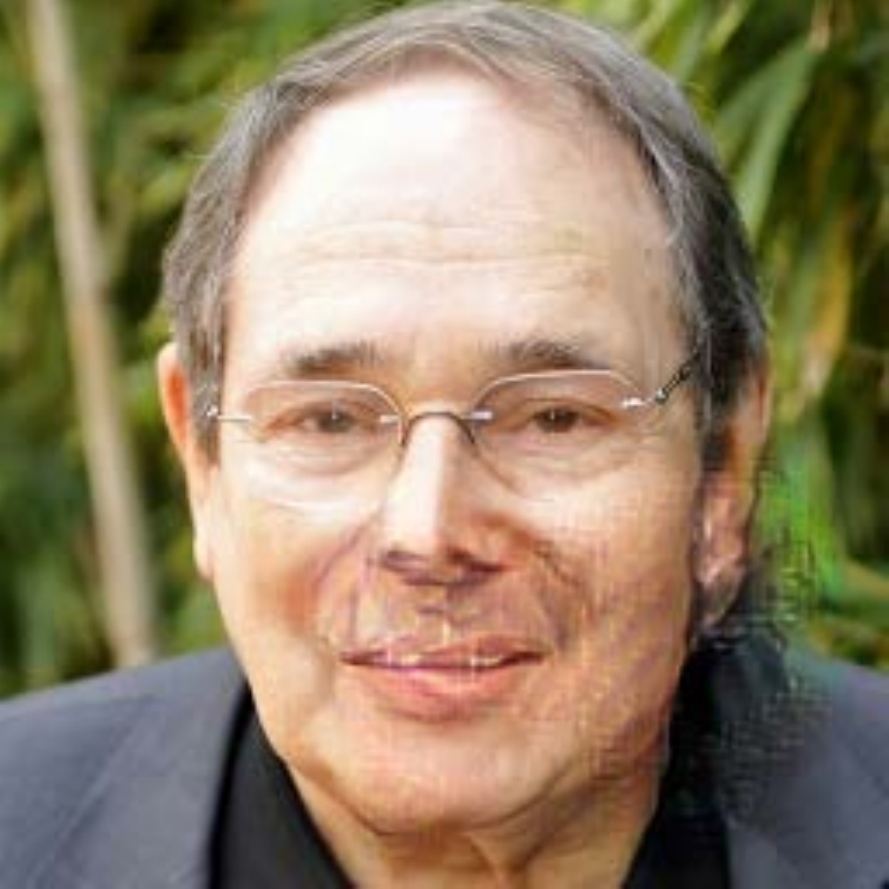}
        \includegraphics[width=2.4cm]{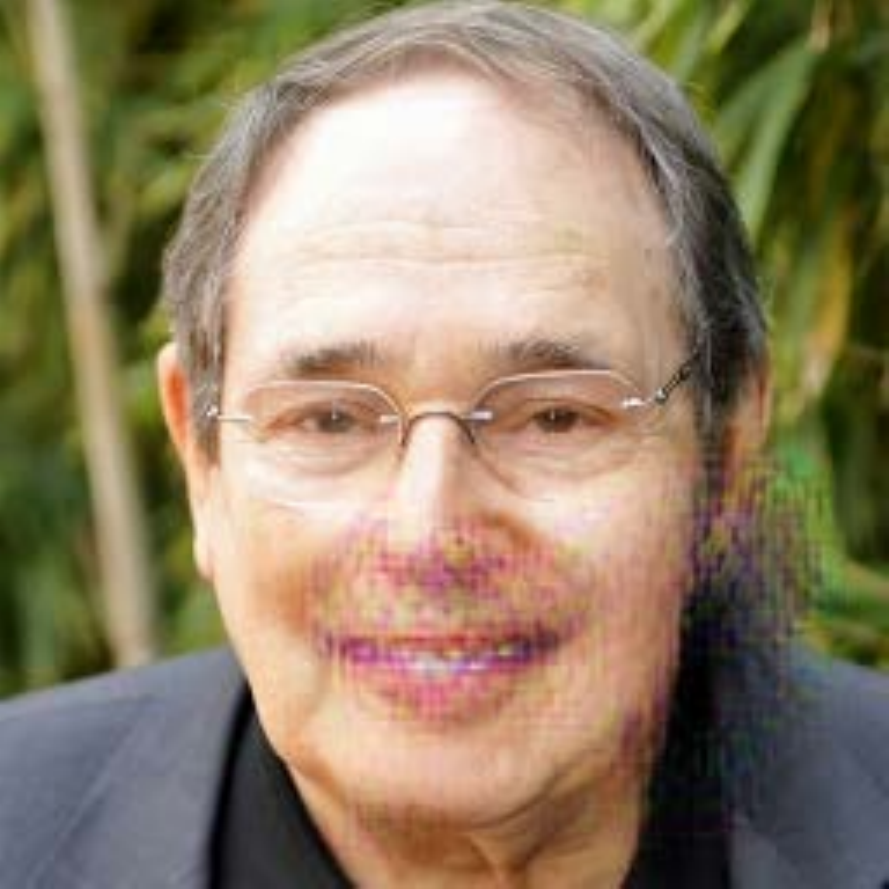}
        \includegraphics[width=2.4cm]{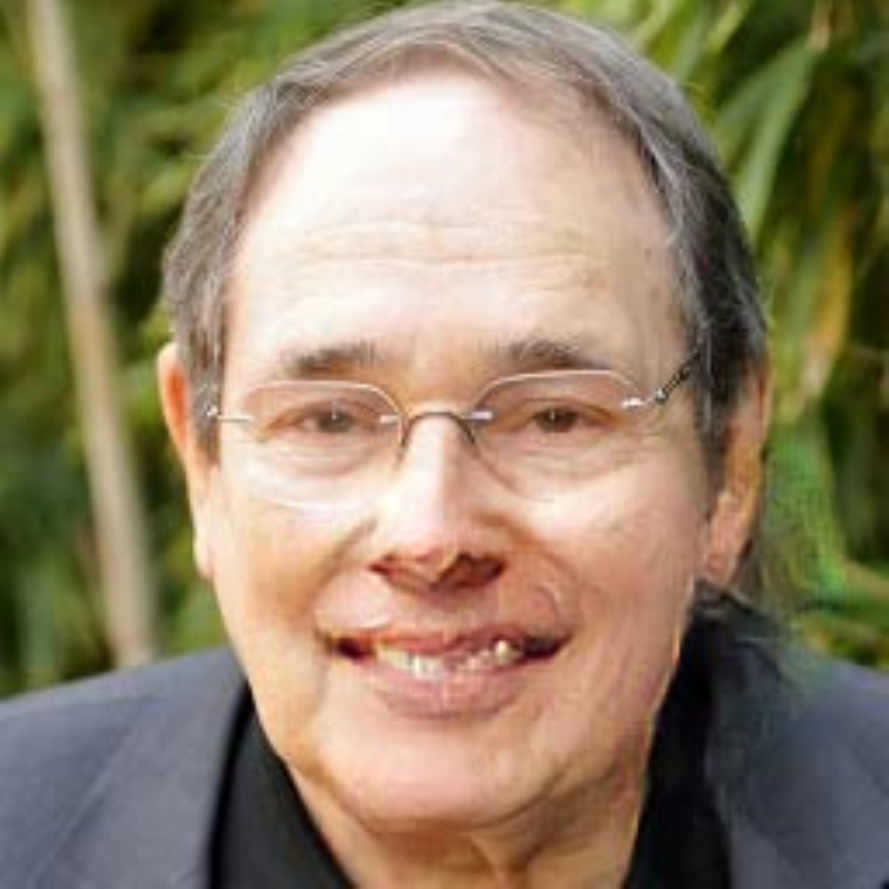}
        \includegraphics[width=2.4cm]{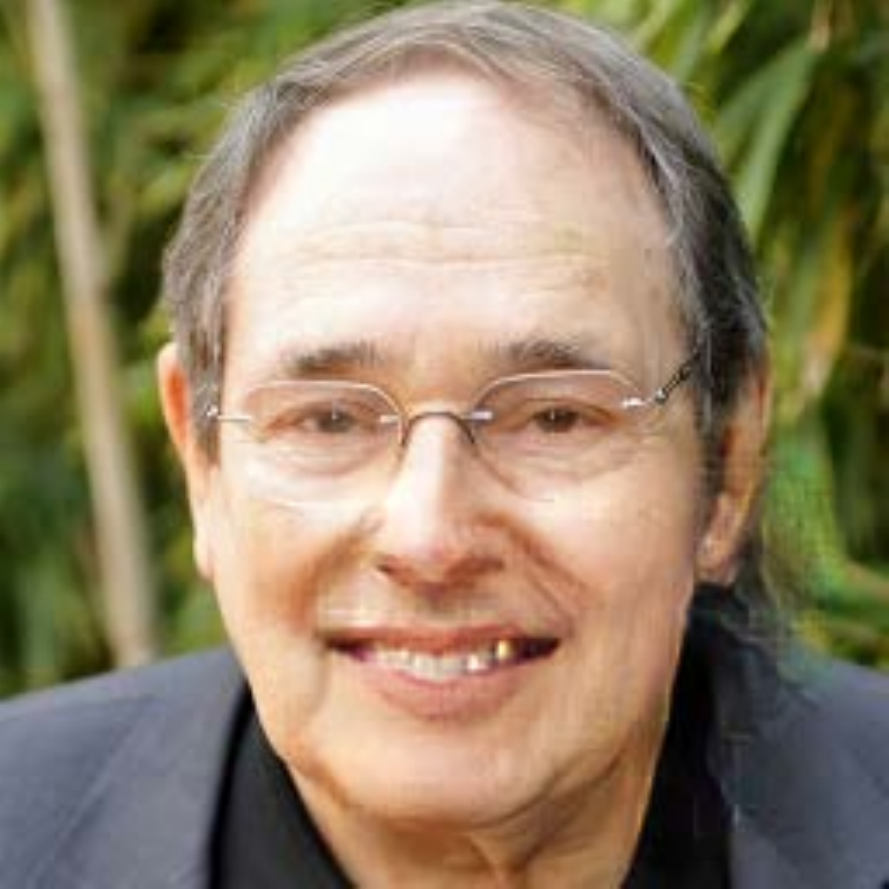}
    \end{subfigure}
    \vspace{-0in}
                                        
    (a) Input                               \hspace{0.6cm}
    (b) standard conv.                 \hspace{0.4cm}
    (c) w/o $\mathcal{L}_c$            \hspace{0.7cm}
    (d) $\mathcal{L}_c + \mathcal{L}_s$  \hspace{1.0cm}
    (e) $\mathbf{I}_c^{(1)}$                     \hspace{1.0cm}
    (f) w/o $\mathcal{L}_a$           \hspace{0.8cm}
    (g) full model                         \hspace{-0.3cm}
    %(h) GT                                 %\hspace{0.2cm}
    \vspace{-0in}
    \\                                          \hspace{0.5cm}
     (PSNR/SSIM)                                \hspace{0.5cm}
     (22.97/0.800)                              \hspace{0.5cm}
     (22.94/0.797)                              \hspace{0.5cm}
     (22.54/0.792)                              \hspace{0.5cm}
     (22.88/0.797)                              \hspace{0.5cm}
     (22.76/0.796)                              \hspace{0.5cm}
     (23.02/0.804)                              \hspace{0.5cm}
    %%(h) GT                                  \hspace{0.2cm}
    \vspace{-0.00in}
    \caption{Results of inpainting on the large contiguous and discontiguous missing areas generated by masking   randomly. (a) the input incomplete images, (b) results using standard convolutions instead of our region-wise convolutions, (c) results of model trained without our correlation loss $\mathcal{L}_c$, (d) results of model trained with $\mathcal{L}_c, \mathcal{L}_s$ at the networks, (e) results of the semantic inferring network,  (f) results of model trained without adversarial loss, namely RED\cite{ma2019inpainting} and (g) results of our full model.}
    \label{fig:ab}
    \vspace{-0.20in}
\end{figure*}

\subsection{Ablation Study}
In this part, we will analyze the effect of different components in our model, proving that our method mainly gains from the region-wise generative adversarial networks architecture. Before that we first study the commonly-used loss functions in several state-of-the-art inpainting models, aiming to explain what they actually do for improving inpainting quality.

\subsubsection{Loss Analysis} 
Table \ref{tab:loss-used} illustrates the widely used loss functions for inpainting, including  reconstruction loss $\mathcal{L}_r$, style loss $\mathcal{L}_s$, perceptual loss $\mathcal{L}_p$, adversarial loss $\mathcal{L}_a$, KL divergence loss $\mathcal{L}_k$ and feature  matching loss $\mathcal{L}_f$. They have been used in the typical models including PConv, EC, PIC and our model. Among all these models, EC proposed a two-stage architecture that divides the image inapinting into two easier tasks, i.e., edge recovery for missing regions and colorization according to surrounding existing regions. Thus, it depends on the adversarial loss $\mathcal{L}_a$ and feature matching loss $\mathcal{L}_f$ to recover the missing edge in the first stage , and restore the color and texture information based on the recovered edges. Removing $\mathcal{L}_a$ and $\mathcal{L}_f$ in the first stage, it is hard for EC to generate any meaningful contents. Thus, we only remove loss functions used in the second stage. Similar to Pconv, removing $\mathcal{L}_p$ seems not make much difference from the full models(see Figure \ref{fig:loss comparisons on EC} (b), (c), and (d), (e)). Besides, without considering the correlations and differences between existing and missing regions, the inpainting results contain obvious inconsistency.
\iffalse
(see Figure \ref{fig:loss comparisons on EC} (b) and (c)). As to PConv, generation of the final inpainting results is guided by reconstruction loss $\mathcal{L}_r$, perceptual loss $\mathcal{L}_p$, and style loss $\mathcal{L}_s$. As shown in Figure \ref{fig:loss comparisons on EC} , removing $\mathcal{L}_p$ seems not make much difference from the full models. Besides, without considering the correlations and differences between existing and missing regions, the inpainting results contain obvious inconsistency.
\fi

We also study the PIC model, which is merely guided by reconstruction loss, adversarial loss and KL divergence loss $\mathcal{L}_{k}$ due to their motivation of generating diverse inpainting results rather than one ground-truth image. Without high-level semantic guidance, such as style loss or perceptual loss, it is difficult to capture the semantic information for complicated structures in images. Therefore, PIC performs well on faces dataset, especially when the missing regions lying on faces. When facing complex structures, such as scene data or the background around the faces, it can hardly approximate the image distribution, and thus generates undesired inpainting contents. 

Different from the prior models like EC, PConv and PIC, our method mainly gains from region-wise generative adversarial networks architecture. We adopt the reconstruction loss to help region-wise convolution learn to represent features distinguishingly at local and pixel level. The correlation loss is introduced to capture the non-local relations between different regions and thus further fill semantically meaningful details in the missing regions. To guarantee the visual appearance of inpainting contents, style loss and adversarial loss are used to eliminate the artifacts and approximate the ground-truth data distribution both from global and local perspectives. 

\subsubsection{Component Analysis} 
To validate the effect of different components in our adversarial image inpainting framework, Figure \ref{fig:ab} respectively shows the inpainting results obtained by our framework, and the framework using different settings: standard convolution filters instead of region-wise ones, removing correlation loss, using $\mathcal{L}_c$ and $\mathcal{L}_s$ at the same networks, only adopting semantic inferring networks without global perceiving, and removing region-wise discriminator (namely RED\cite{ma2019inpainting}). From the results, we can see that without region-wise convolutional layers, the framework can hardly infer the consistent information with existing regions. Furthermore, without considering the non-local correlation, the framework restores the missing regions only according to the surrounding areas. Moreover, using $\mathcal{L}_c, \mathcal{L}_s$ at the same stage will cause artifacts and cannot restore semantic contents. Besides, we can see that only relying on semantic inferring network can restore the semantic information, and the outputs still contain strange artifacts. Without the help of region-wise discriminator, the inpainting results contain some fold-like artifacts. Together with region-wise convolutions, non-local correlation and region-wise discriminator, our framework enjoys strong power to generate visually and semantically close images to the ground truth.

%% Here is the comparison on Places2 datasets with different methods.

\subsection{Comparison with State-of-the-arts}
Now we compare our region-wise generative adversarial method with the state-of-the-art inpainting models, in terms of both qualitative and quantitative evaluations.

\subsubsection{Qualitative Results}
%% Here is the comparison on celeba-HQ and paris datasets with different methods.
\begin{figure*}[htp]
\centering
    \vspace{0.02in}
    \centering
    \begin{subfigure}
        \centering
        \includegraphics[width=2.4cm]{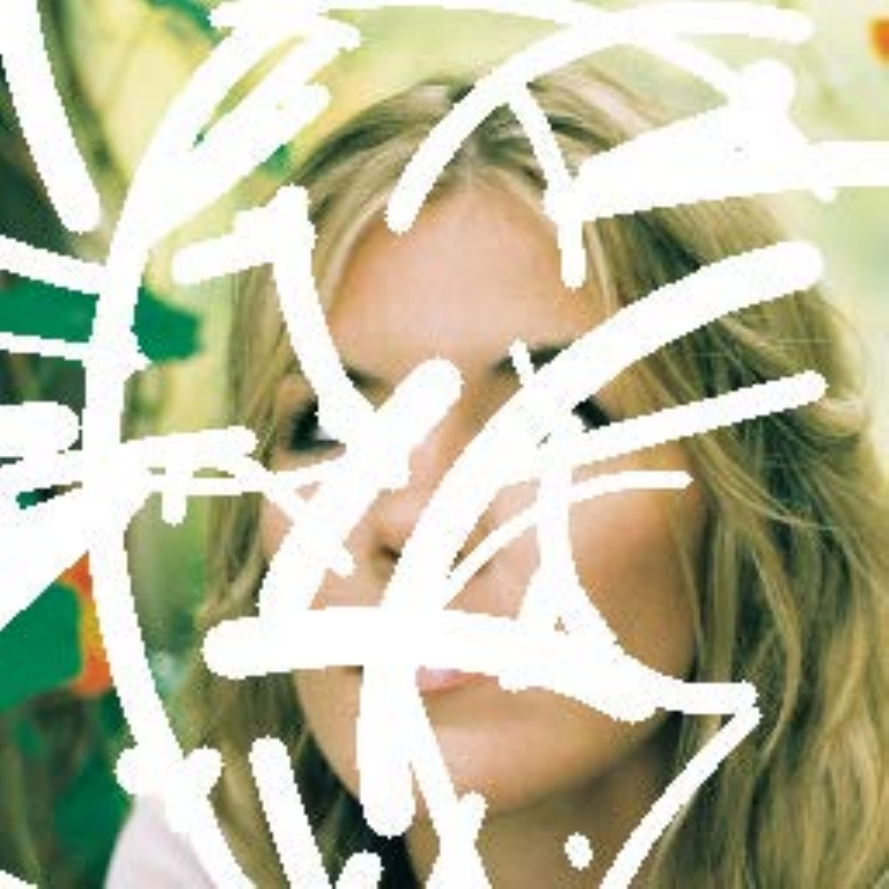}
        \includegraphics[width=2.4cm]{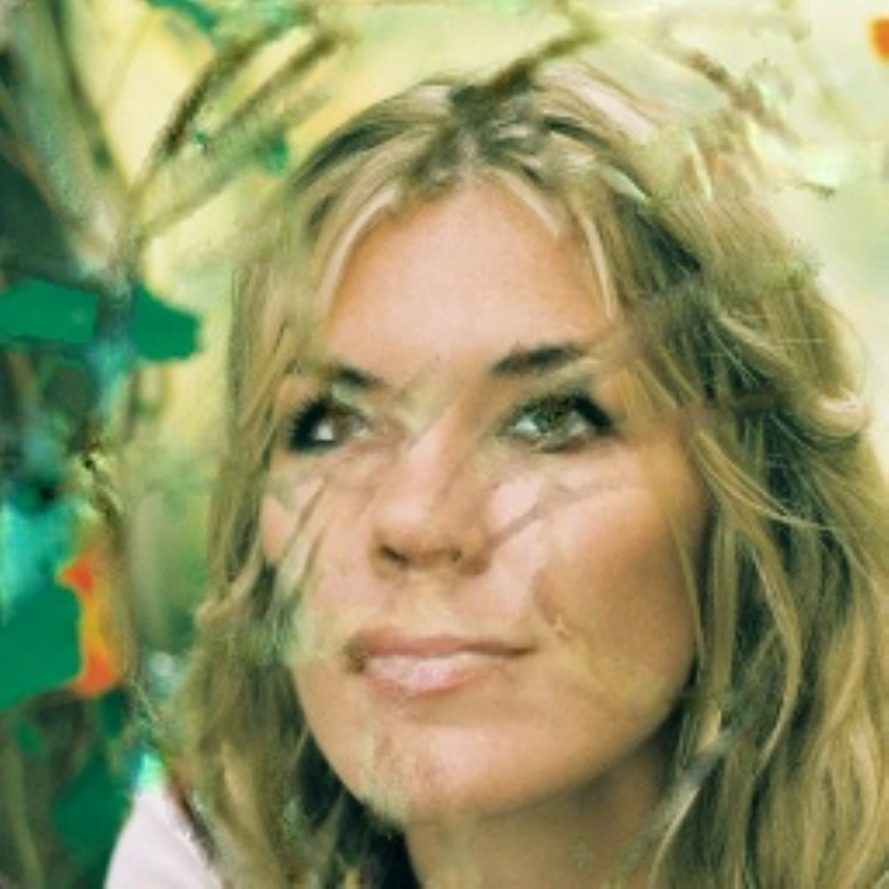}
        \includegraphics[width=2.4cm]{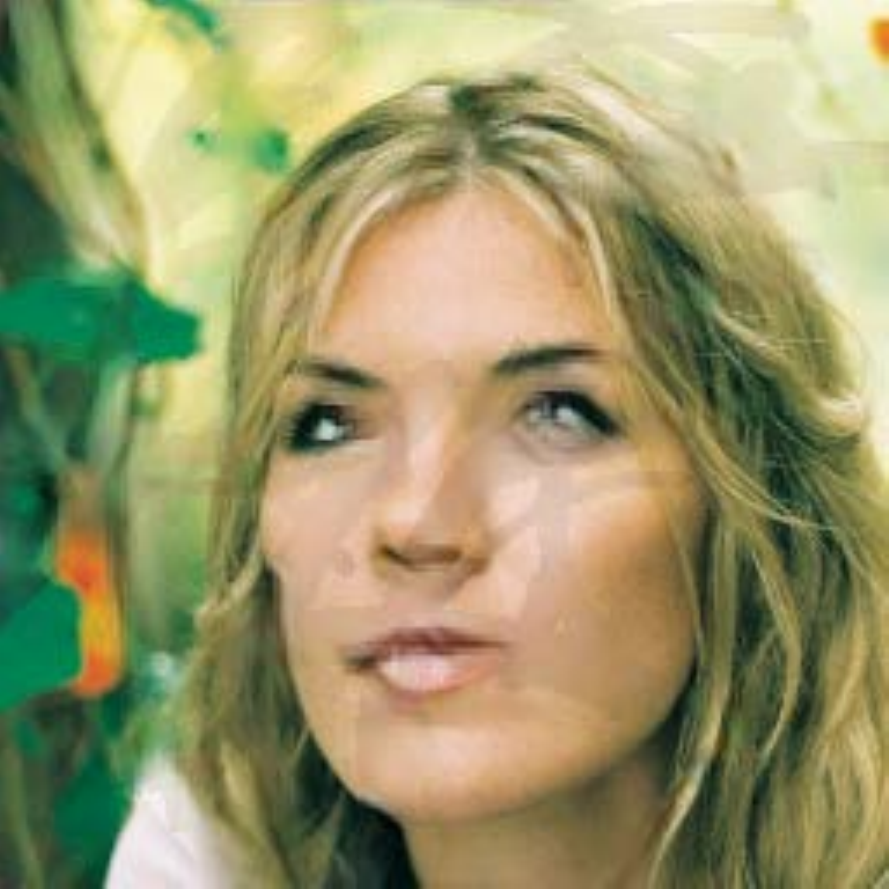}
        \includegraphics[width=2.4cm]{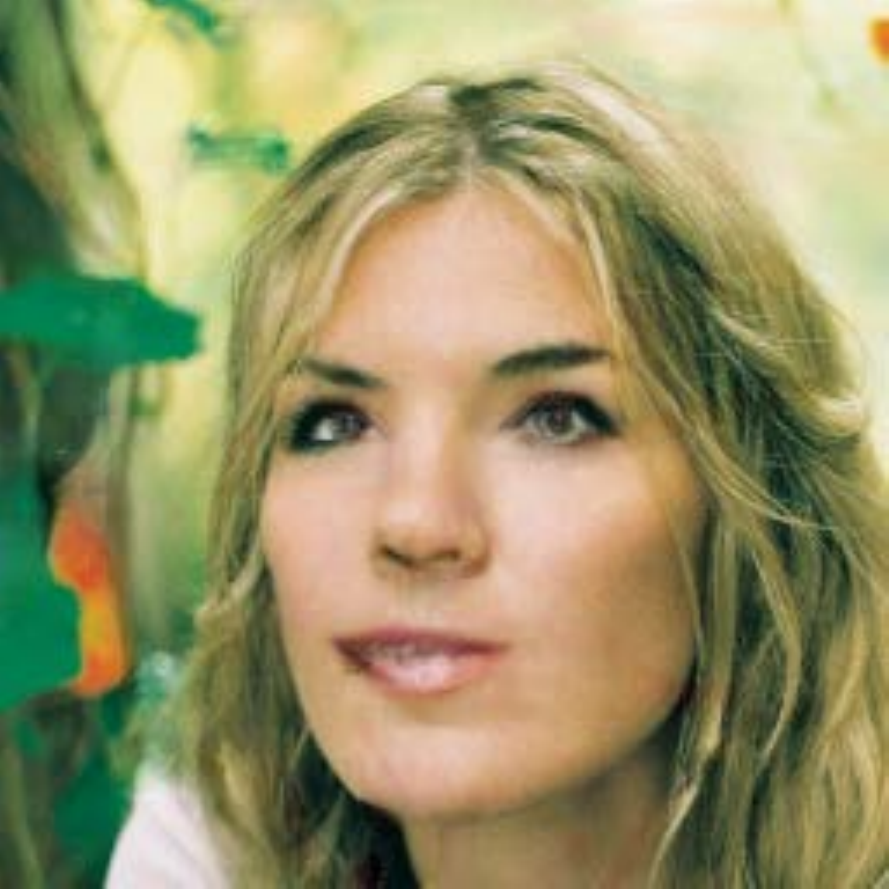}
        \includegraphics[width=2.4cm]{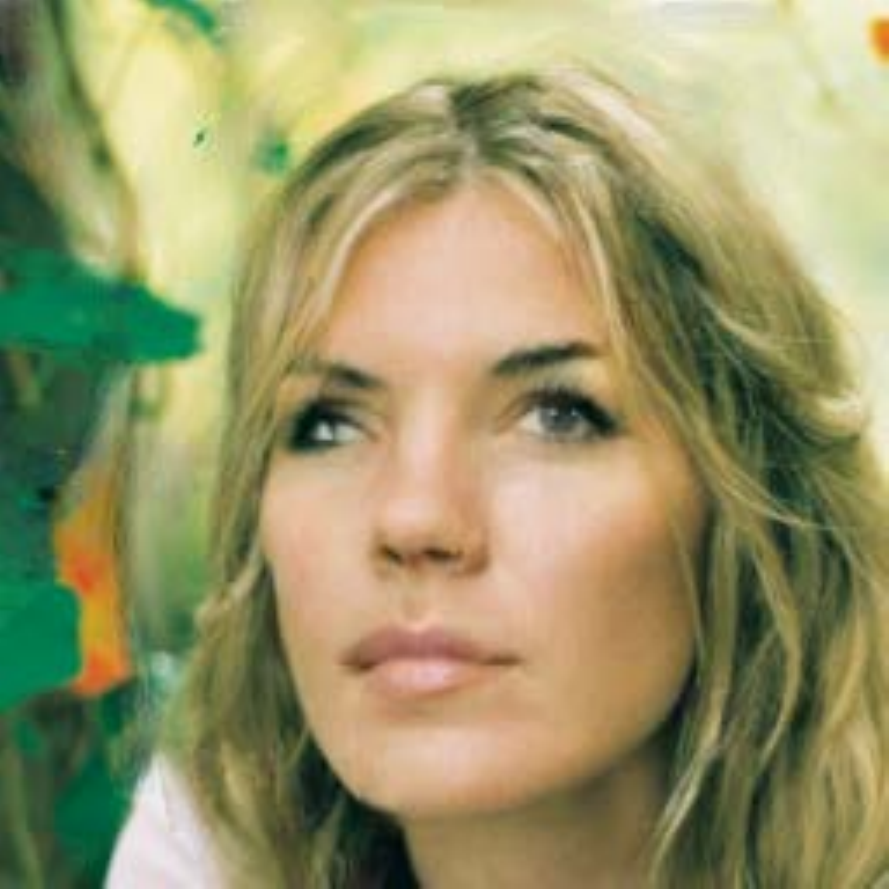}
        \includegraphics[width=2.4cm]{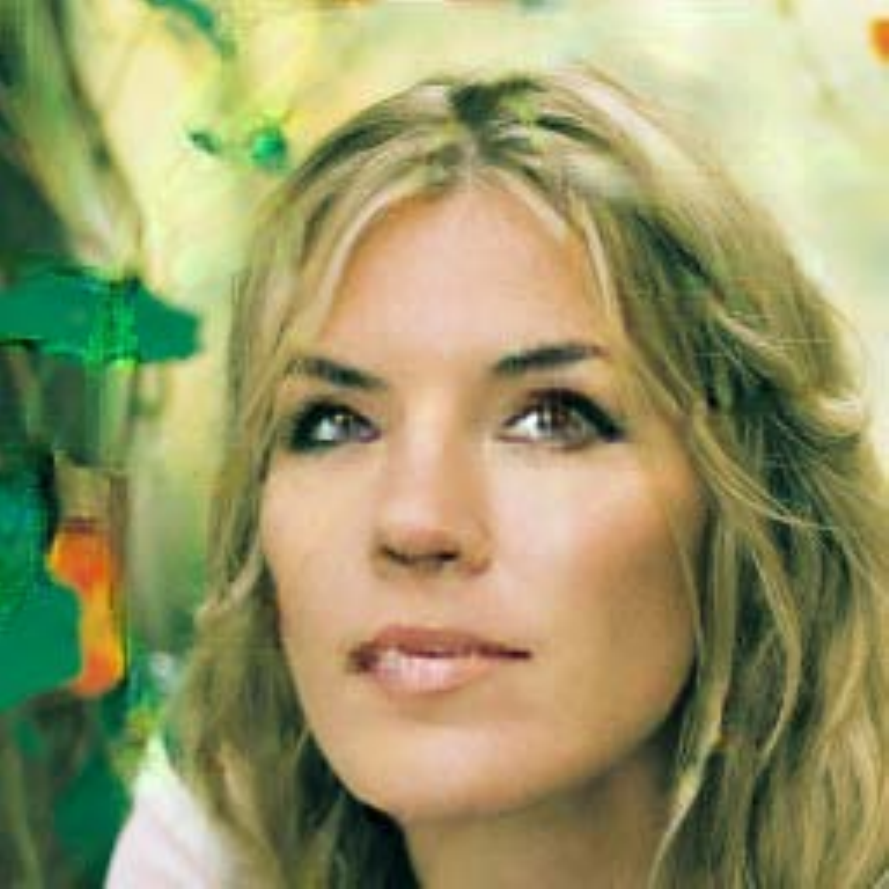}
        \includegraphics[width=2.4cm]{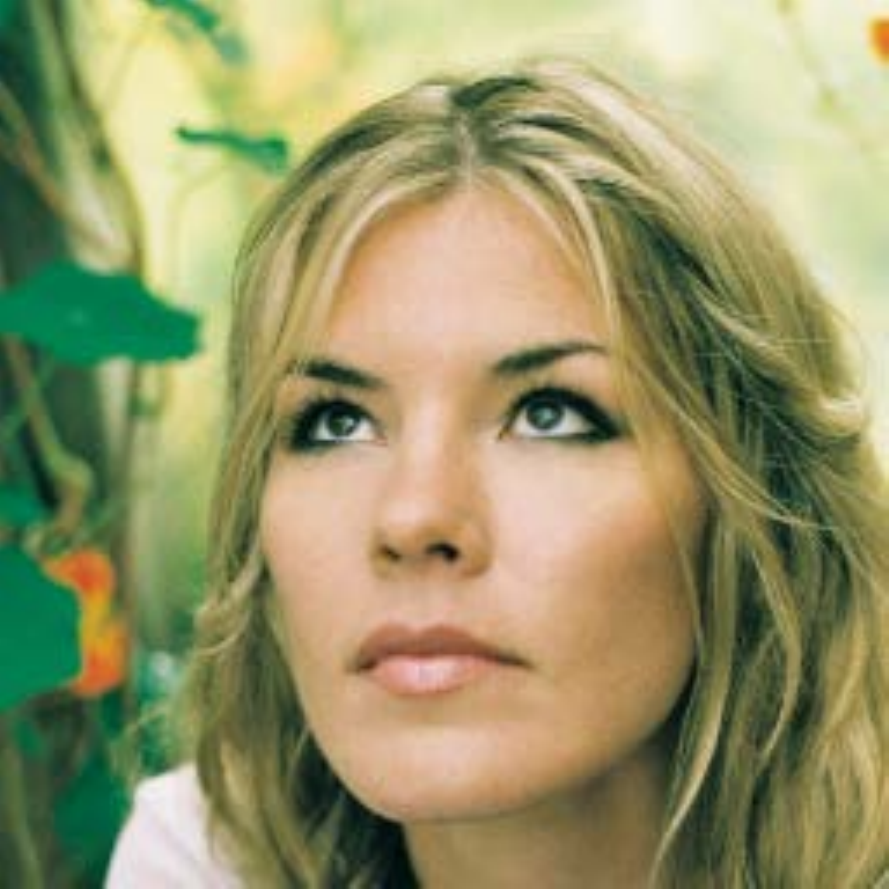}
    \end{subfigure}
    \begin{subfigure}
        \centering
        \vspace{-0.05in}
        \includegraphics[width=2.4cm]{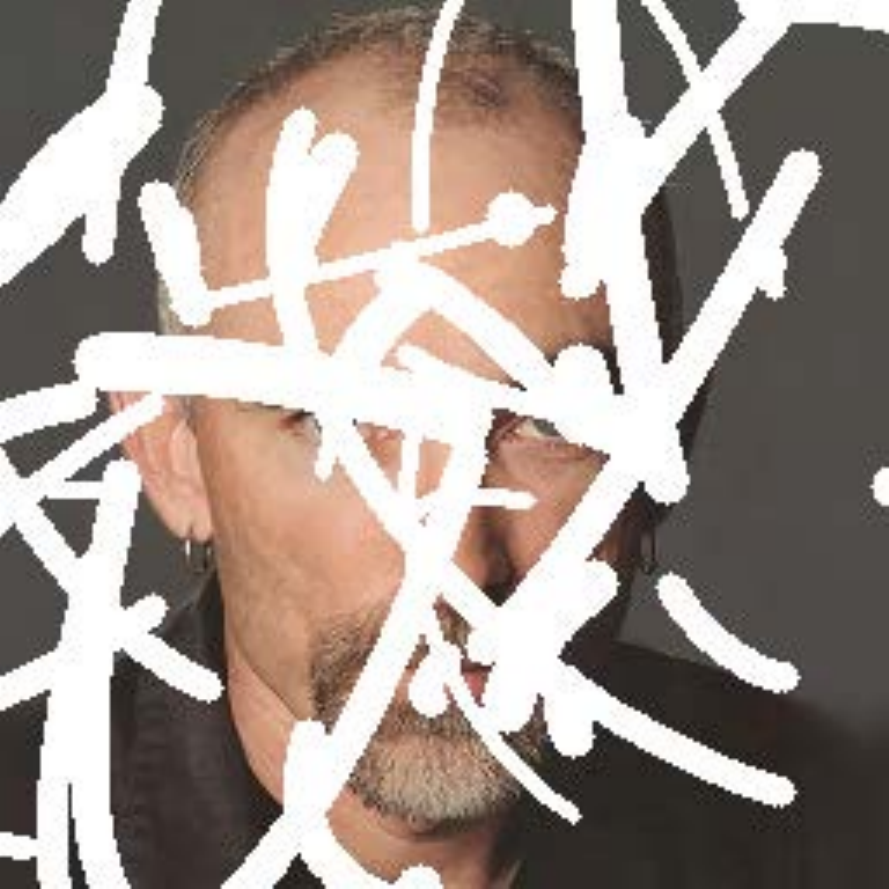}
        \includegraphics[width=2.4cm]{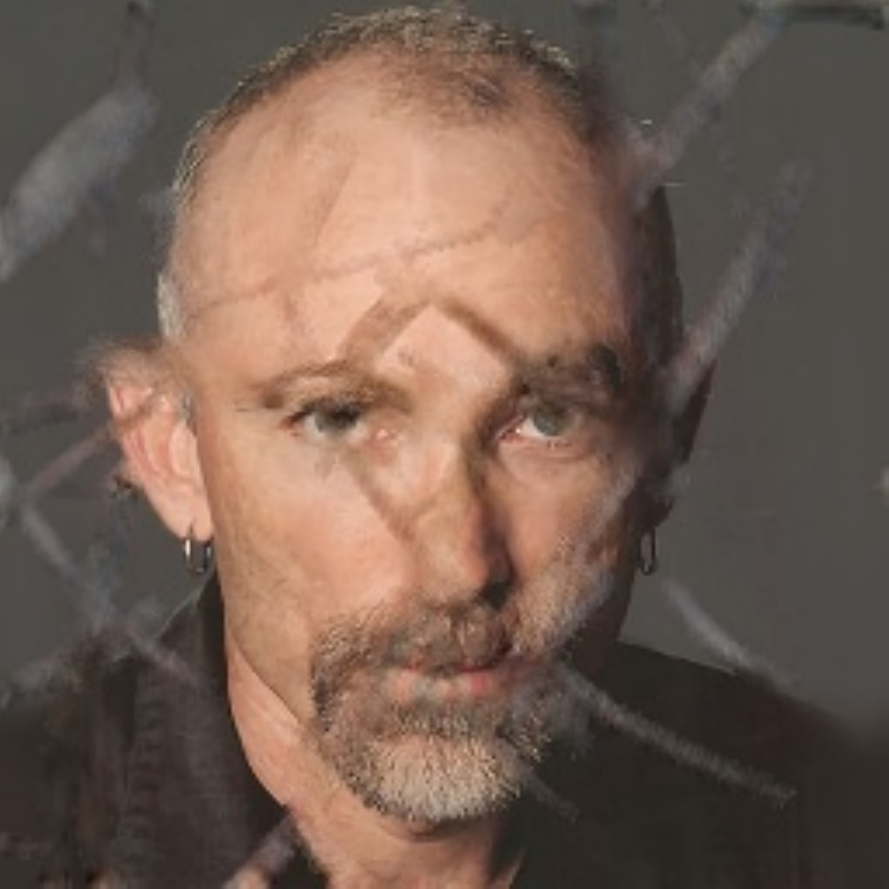}
        \includegraphics[width=2.4cm]{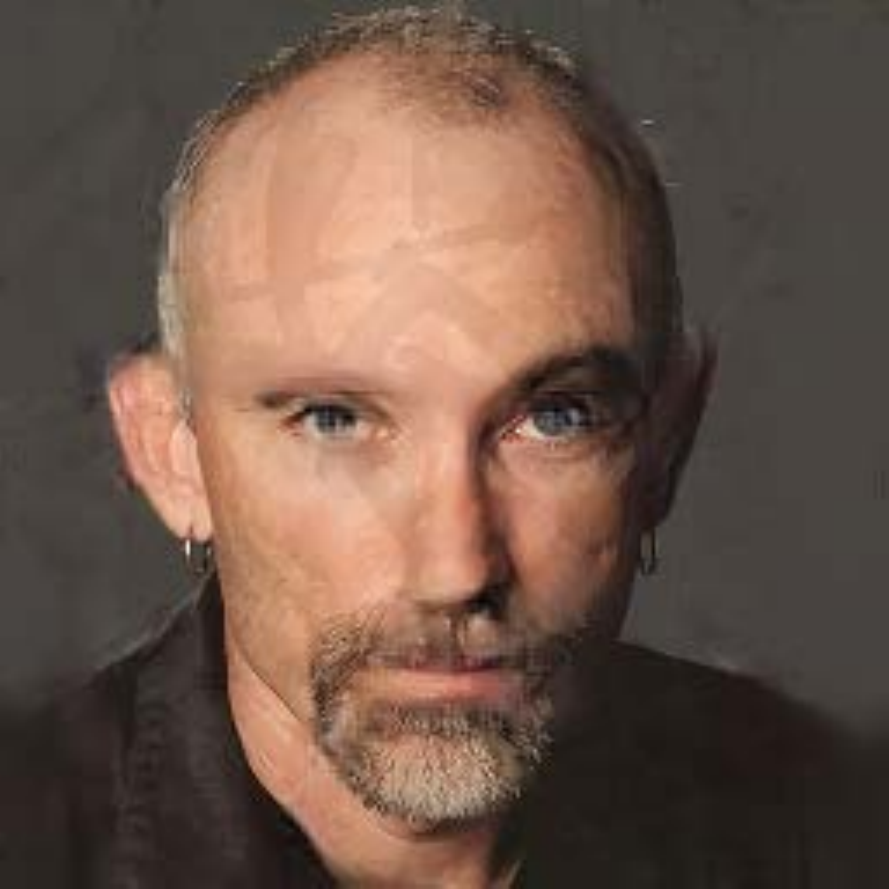}
        \includegraphics[width=2.4cm]{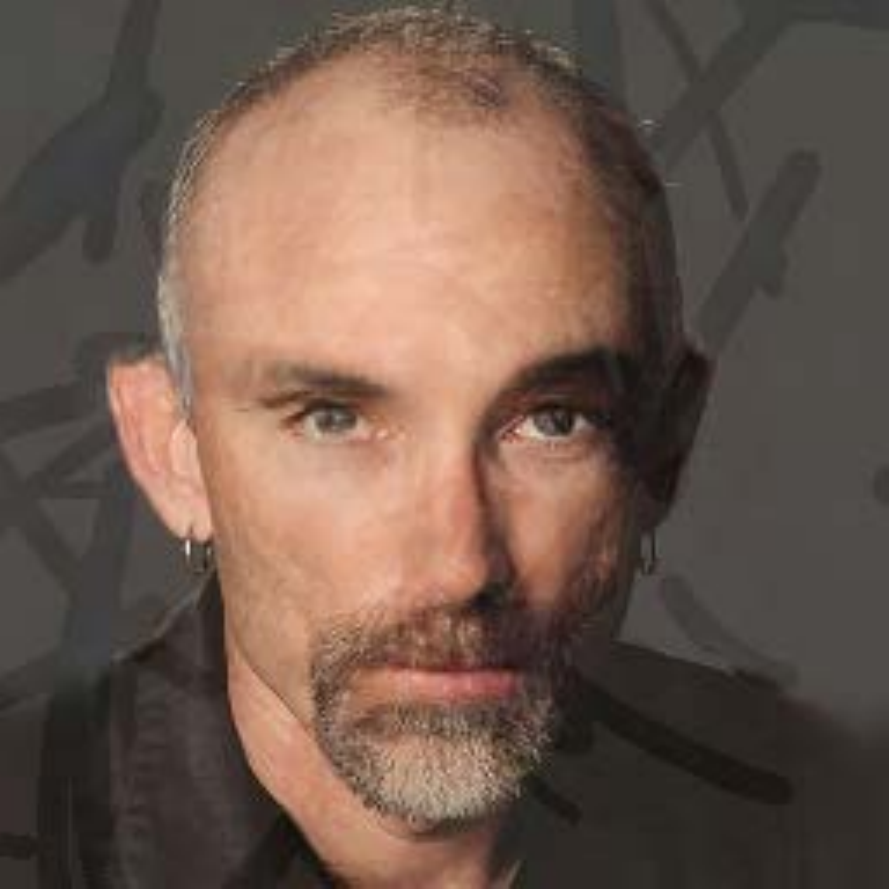}
        \includegraphics[width=2.4cm]{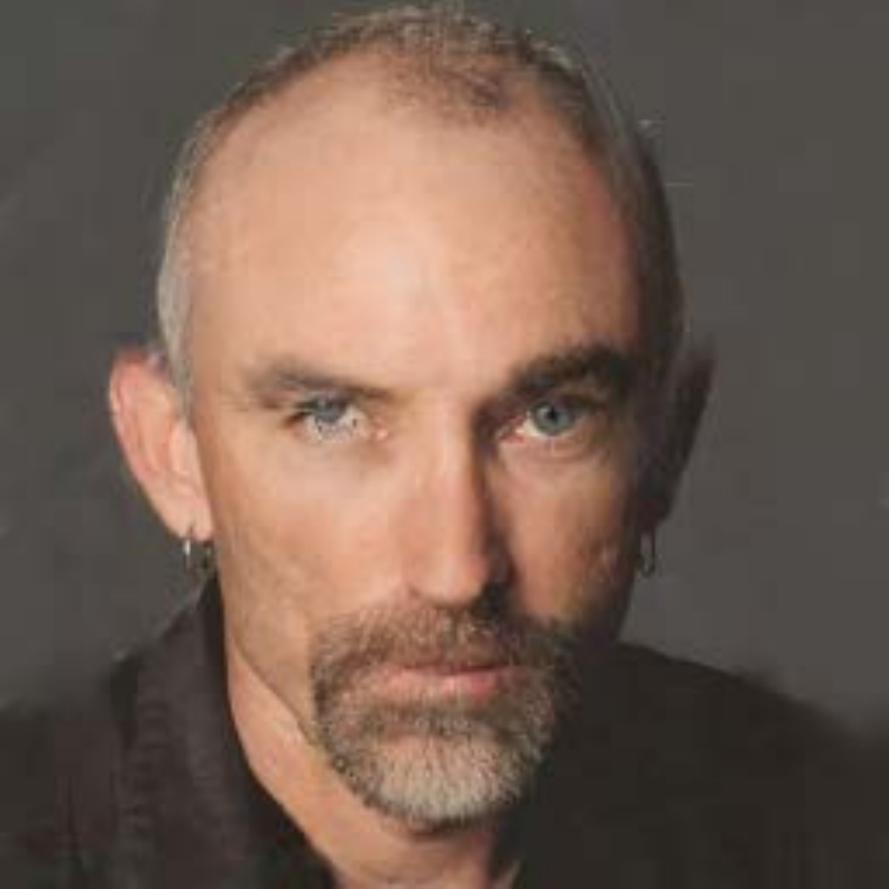}
        \includegraphics[width=2.4cm]{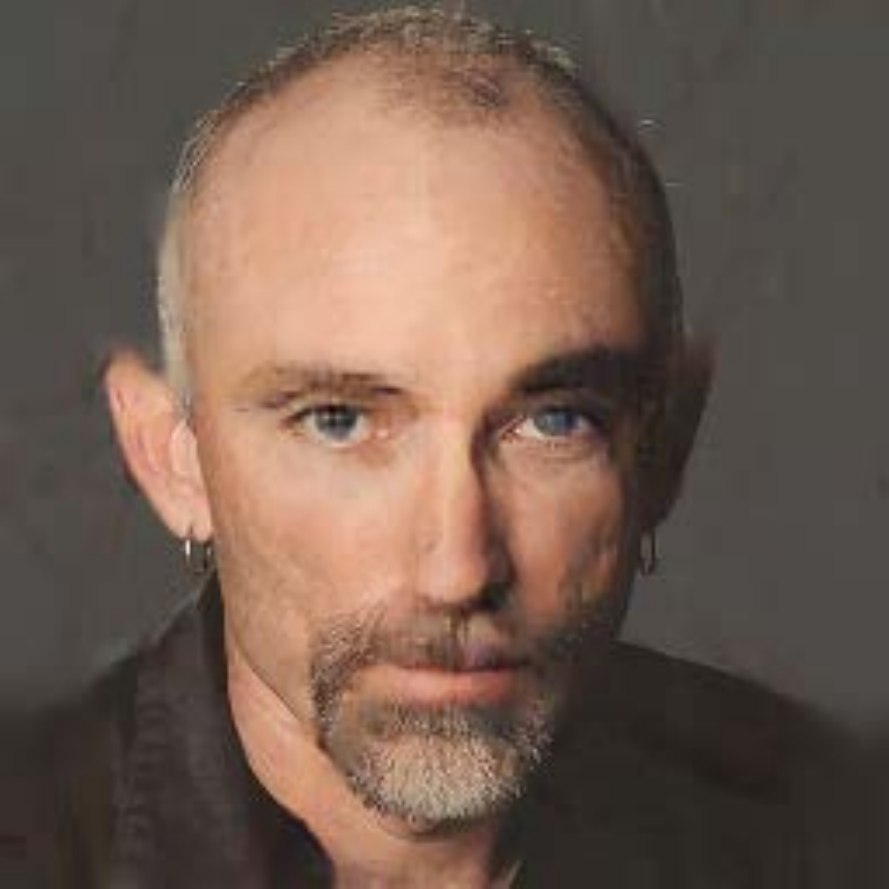}
        \includegraphics[width=2.4cm]{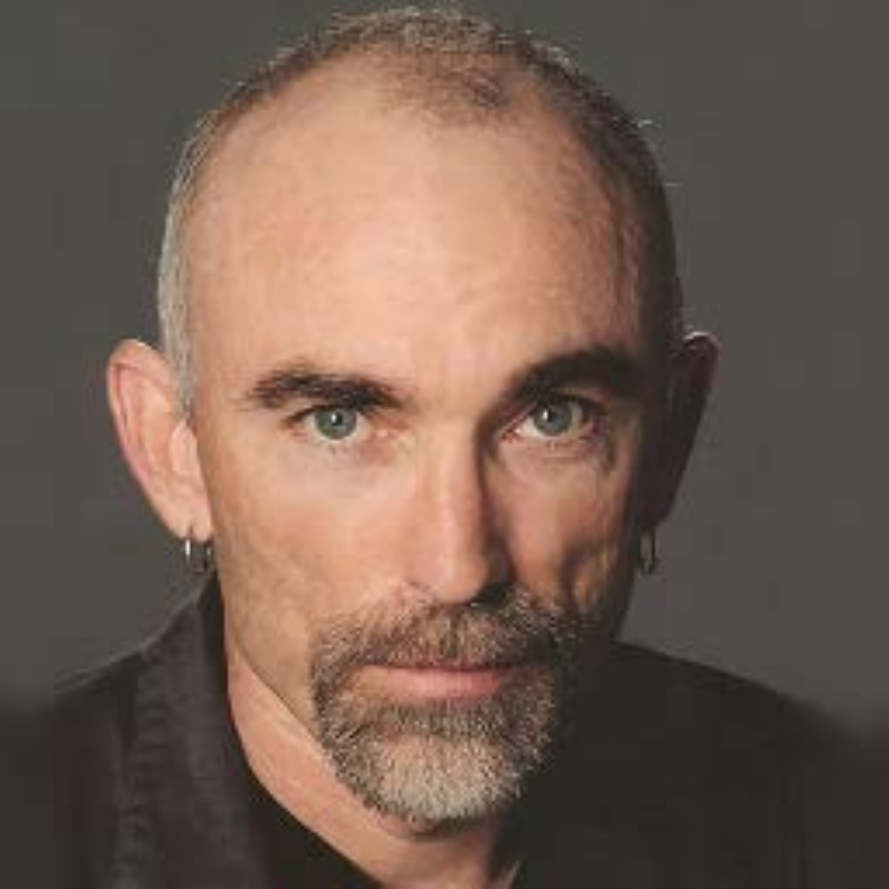} 
    \end{subfigure}
    \begin{subfigure}
        \centering
        \vspace{-0.05in}
        \includegraphics[width=2.4cm]{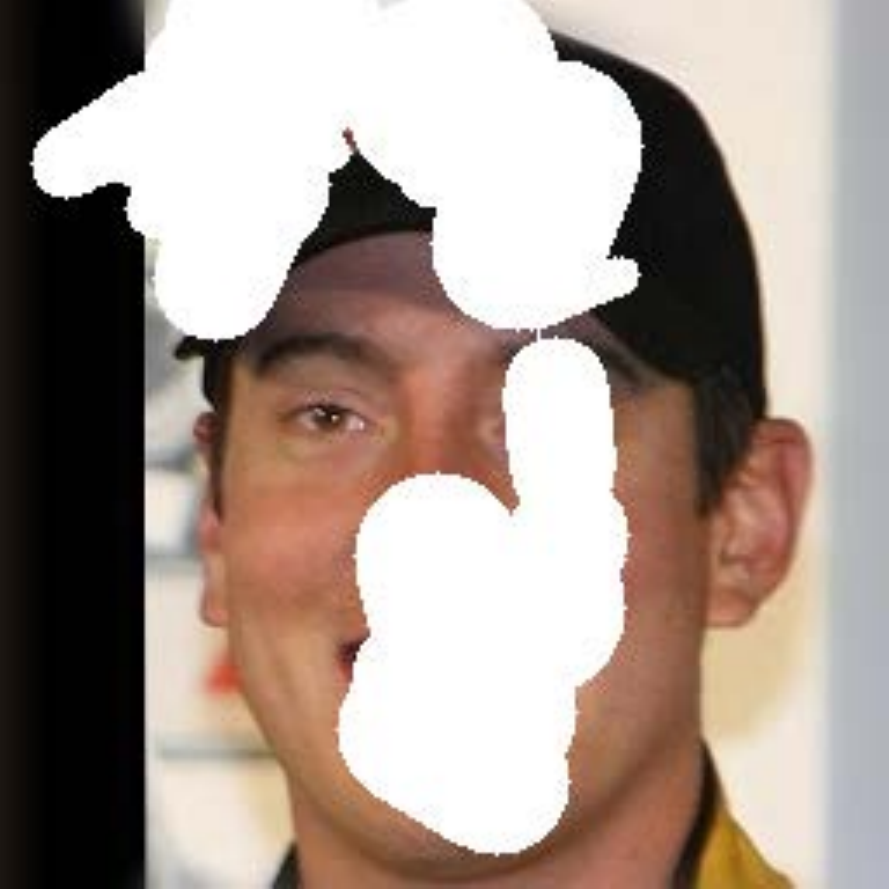}
        \includegraphics[width=2.4cm]{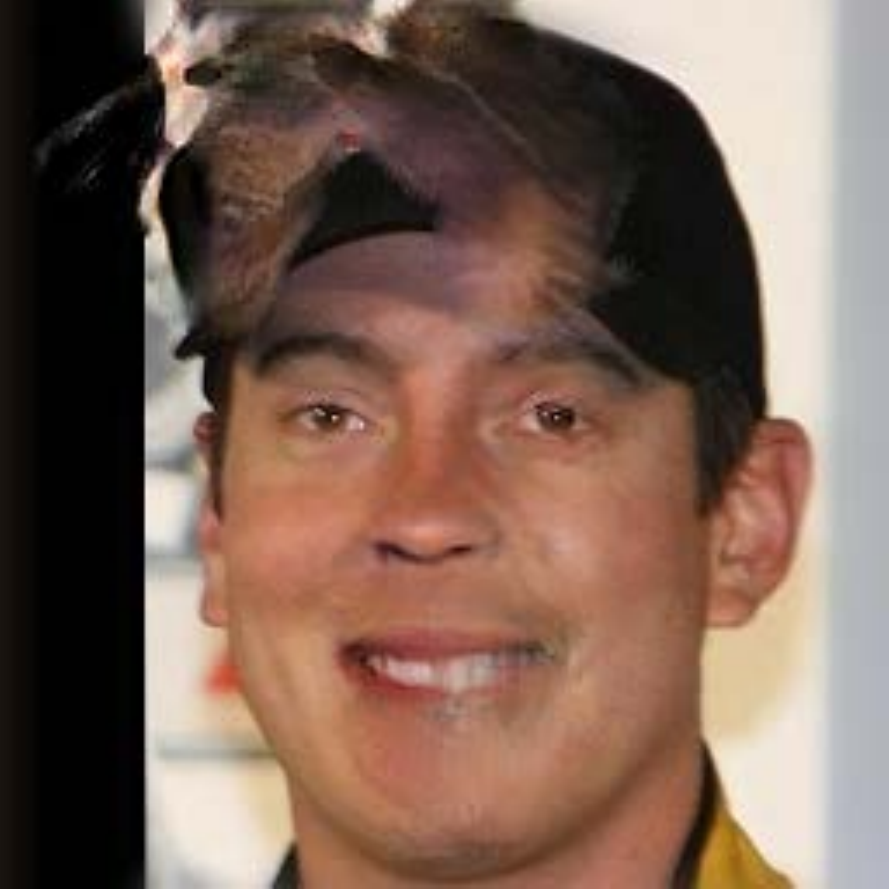}
        \includegraphics[width=2.4cm]{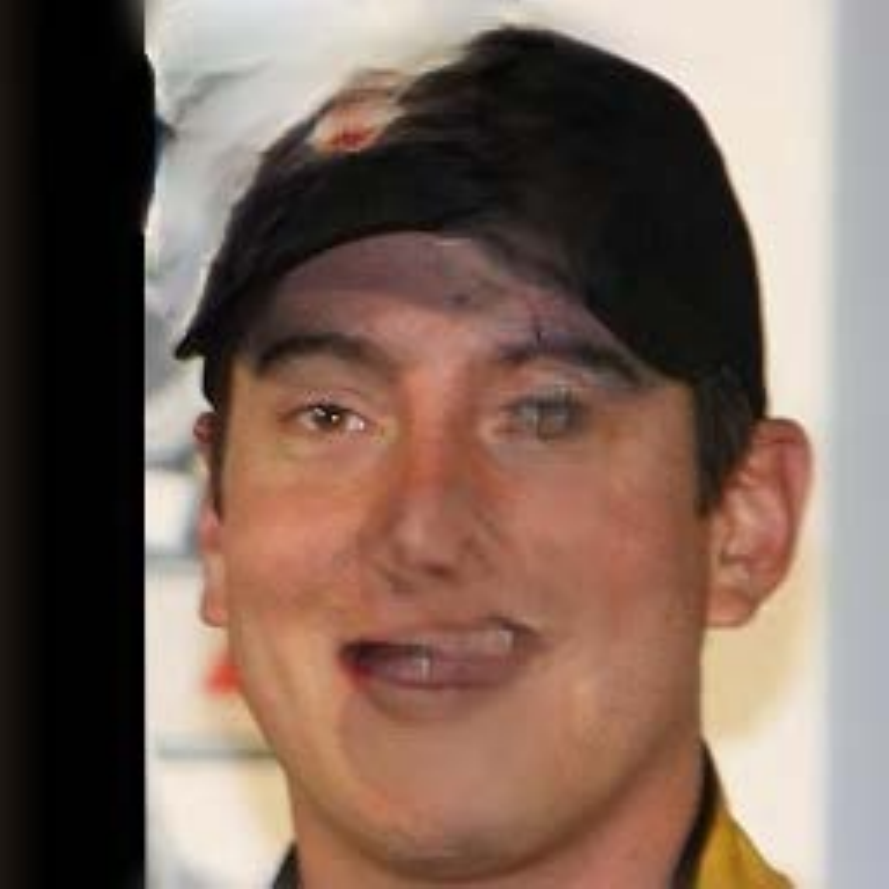}
        \includegraphics[width=2.4cm]{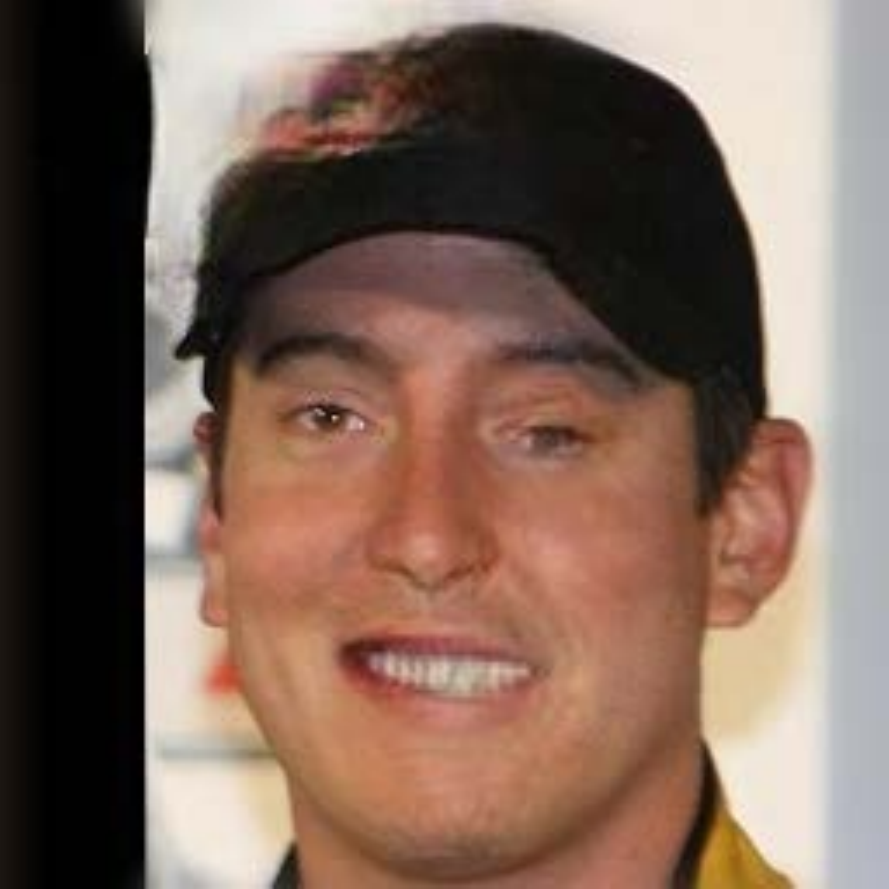}
        \includegraphics[width=2.4cm]{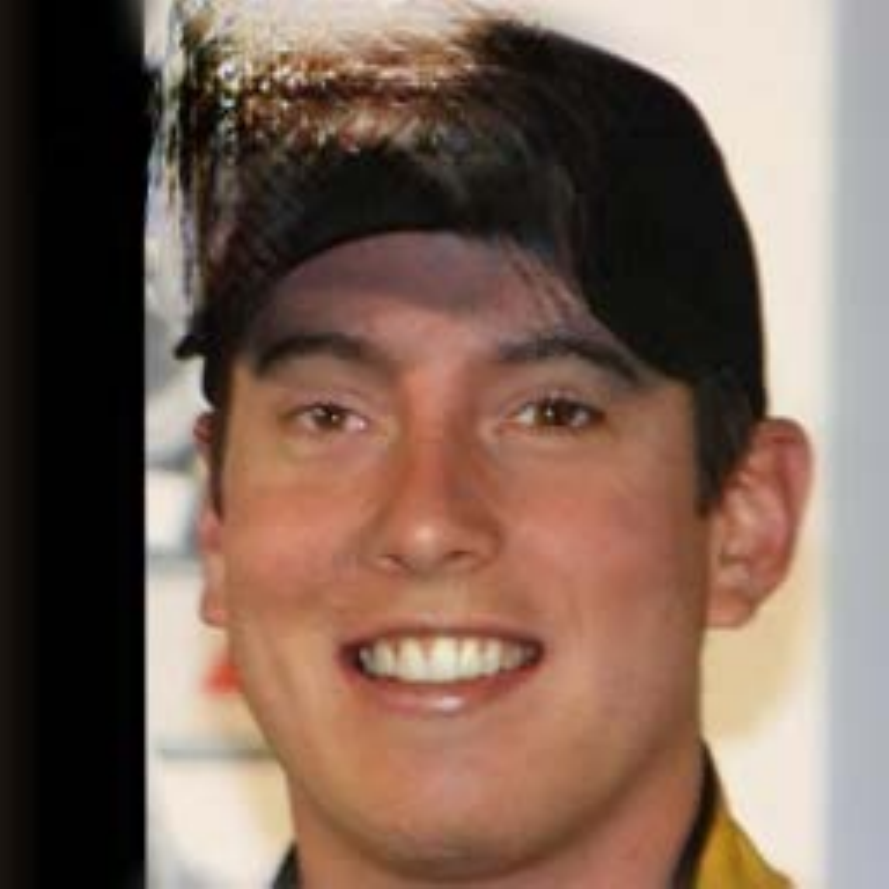}
        \includegraphics[width=2.4cm]{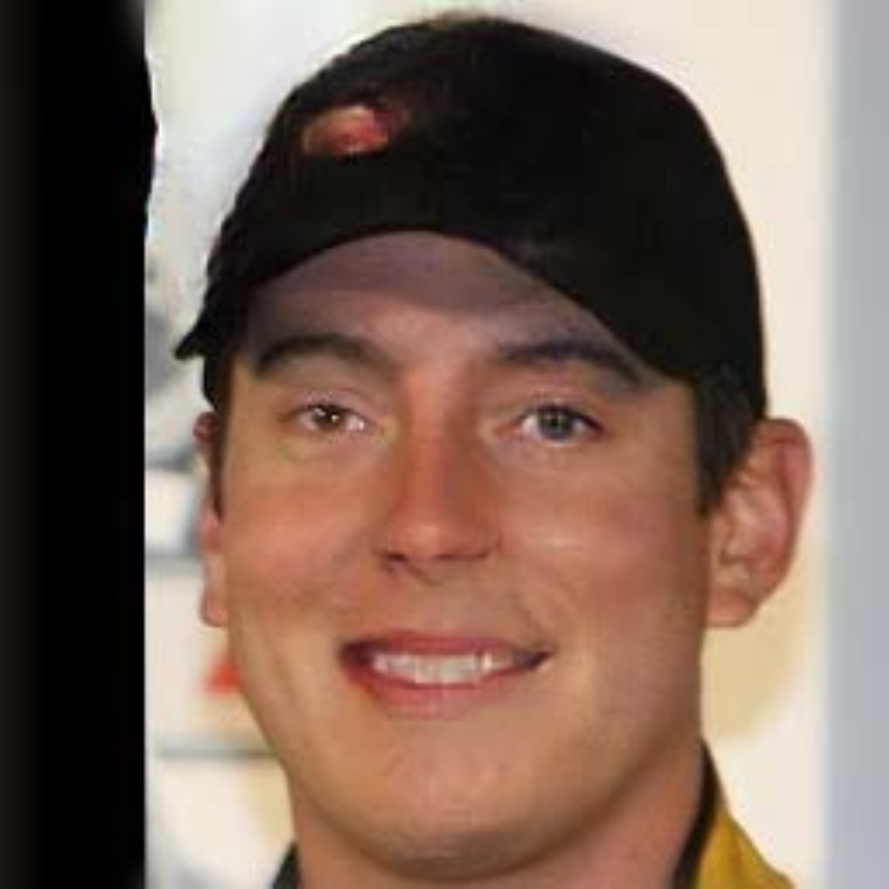}
        \includegraphics[width=2.4cm]{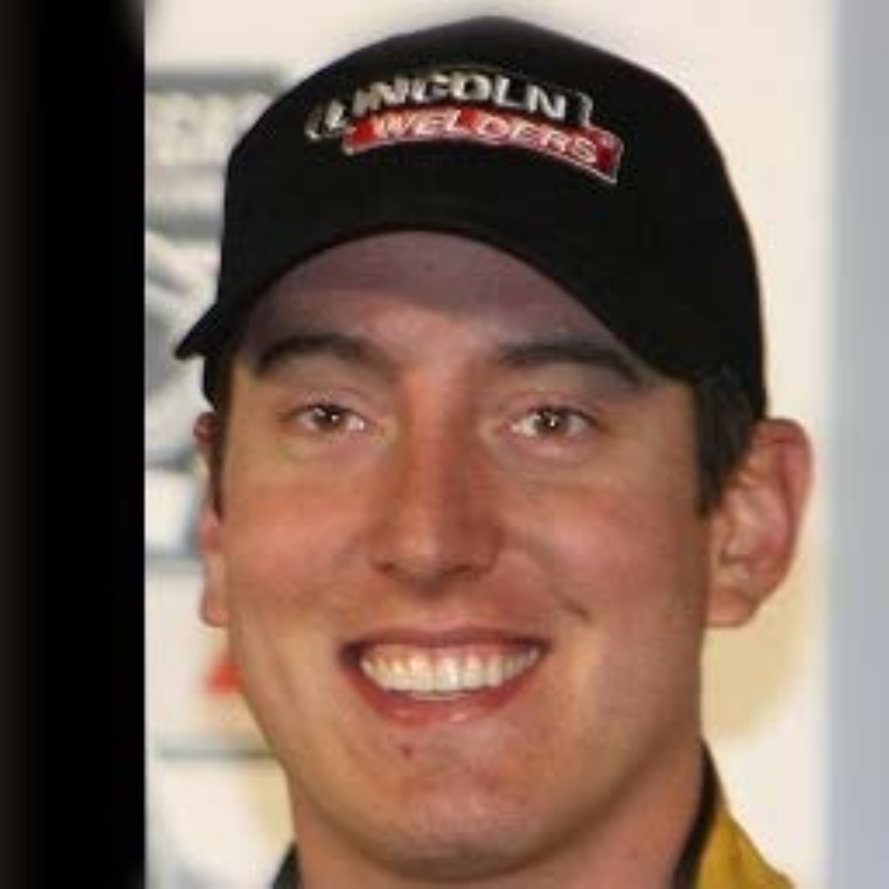}
    \end{subfigure}
    \begin{subfigure}
        \centering
        \vspace{-0.05in}
        \includegraphics[width=2.4cm]{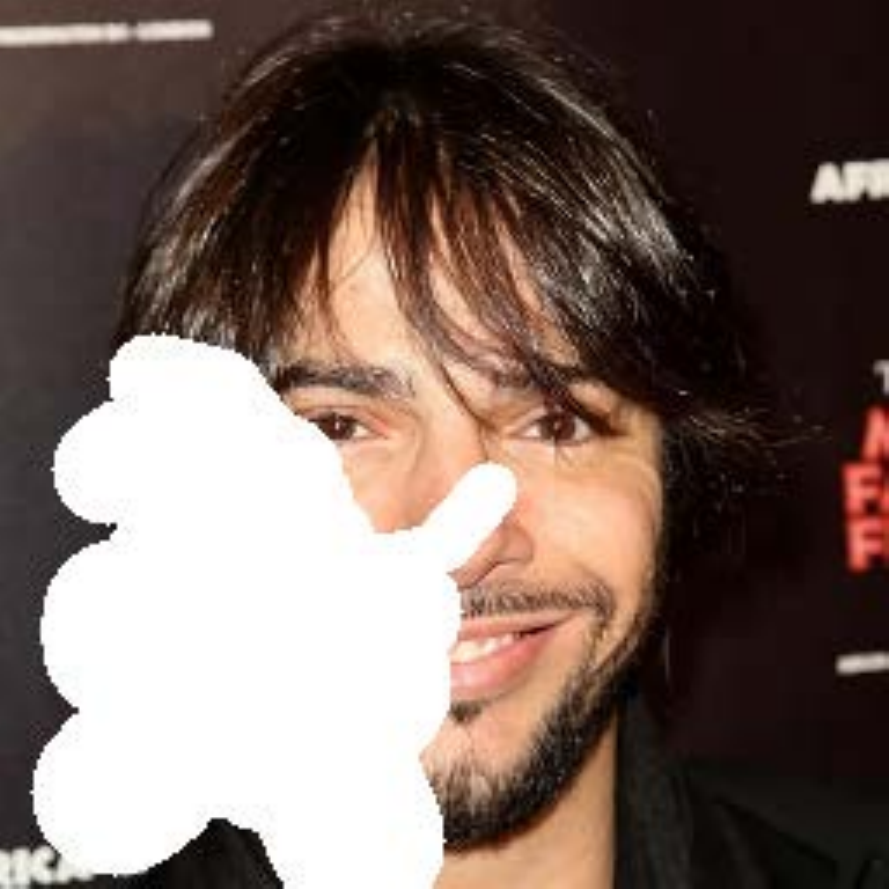}
        \includegraphics[width=2.4cm]{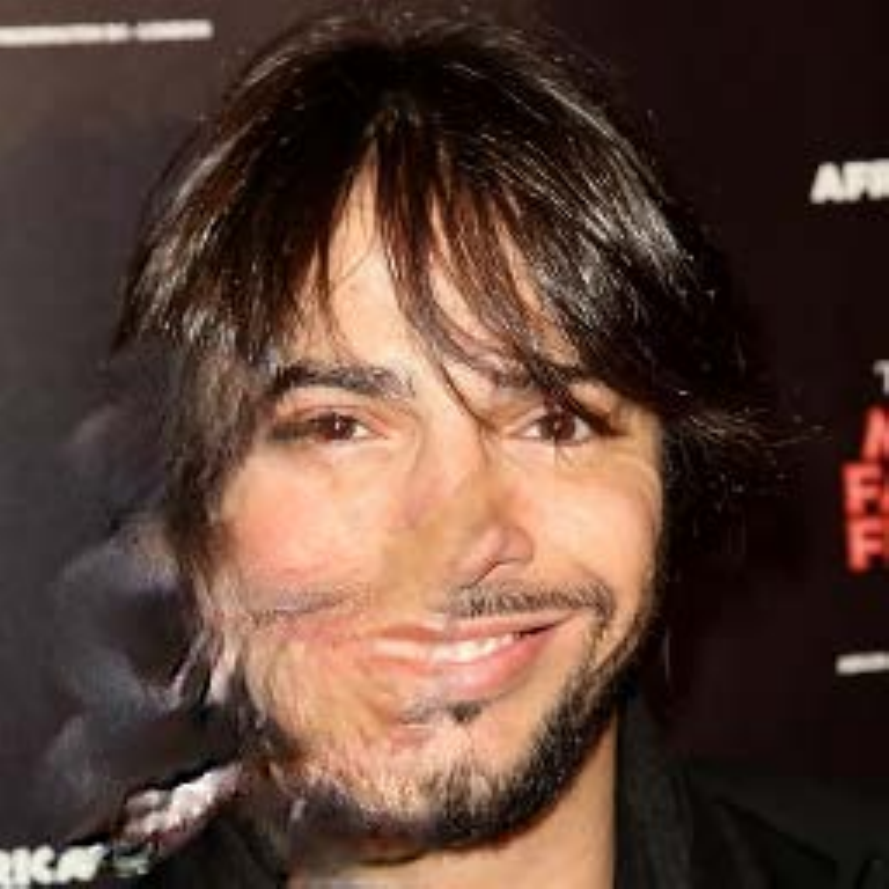}
        \includegraphics[width=2.4cm]{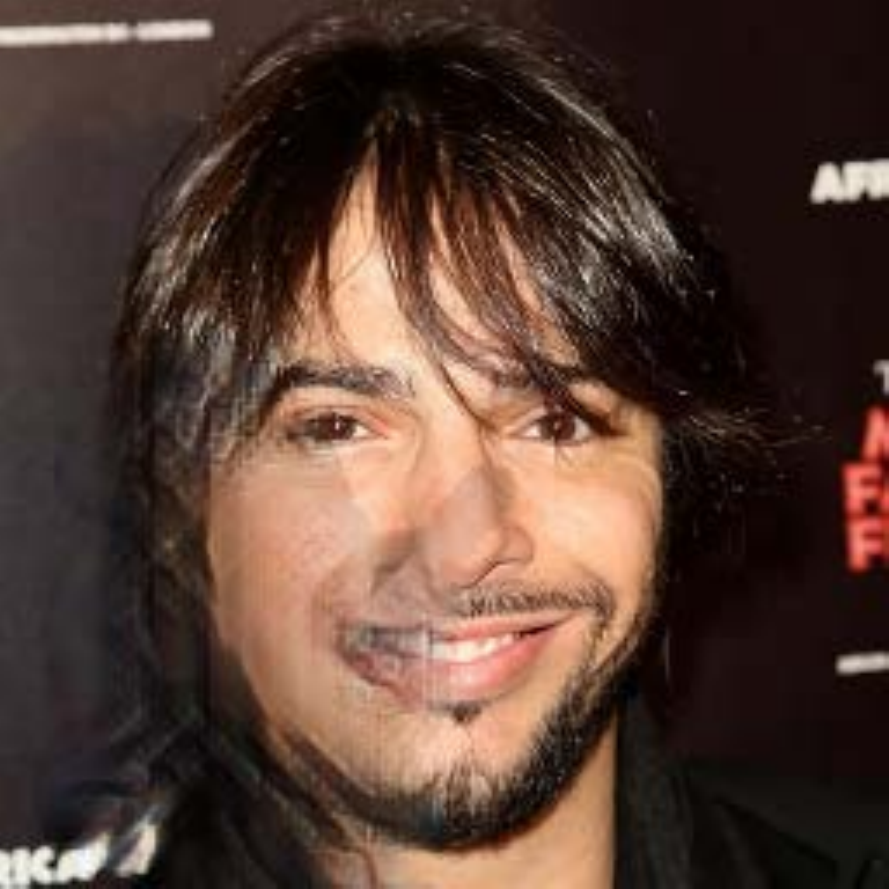}
        \includegraphics[width=2.4cm]{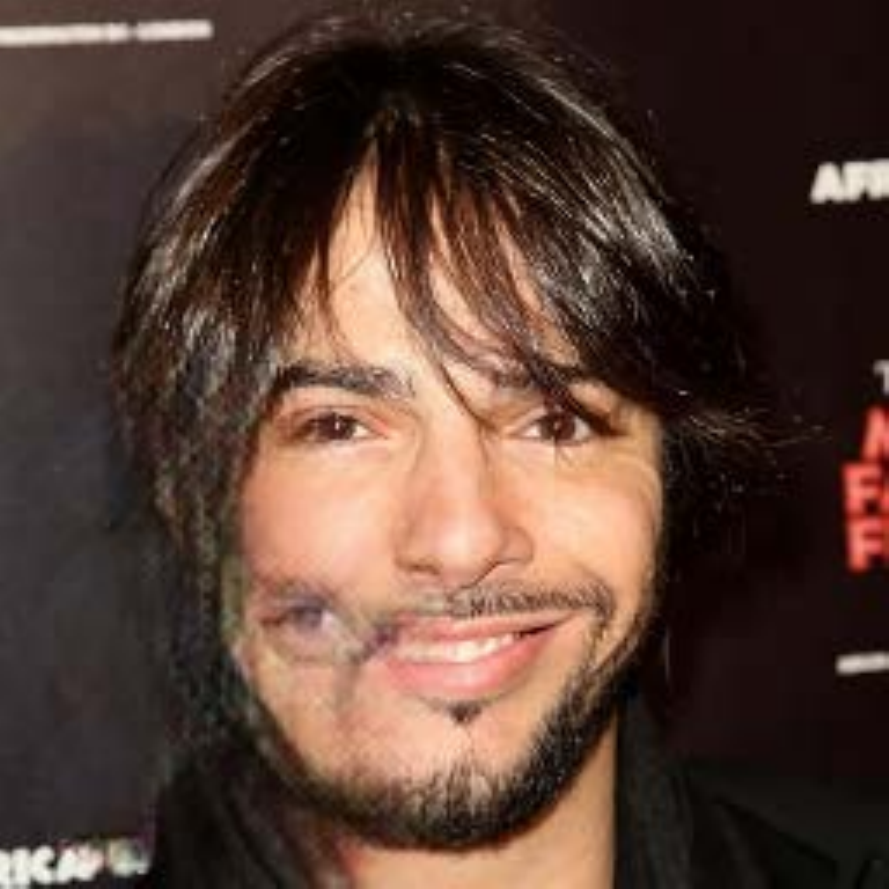}
        \includegraphics[width=2.4cm]{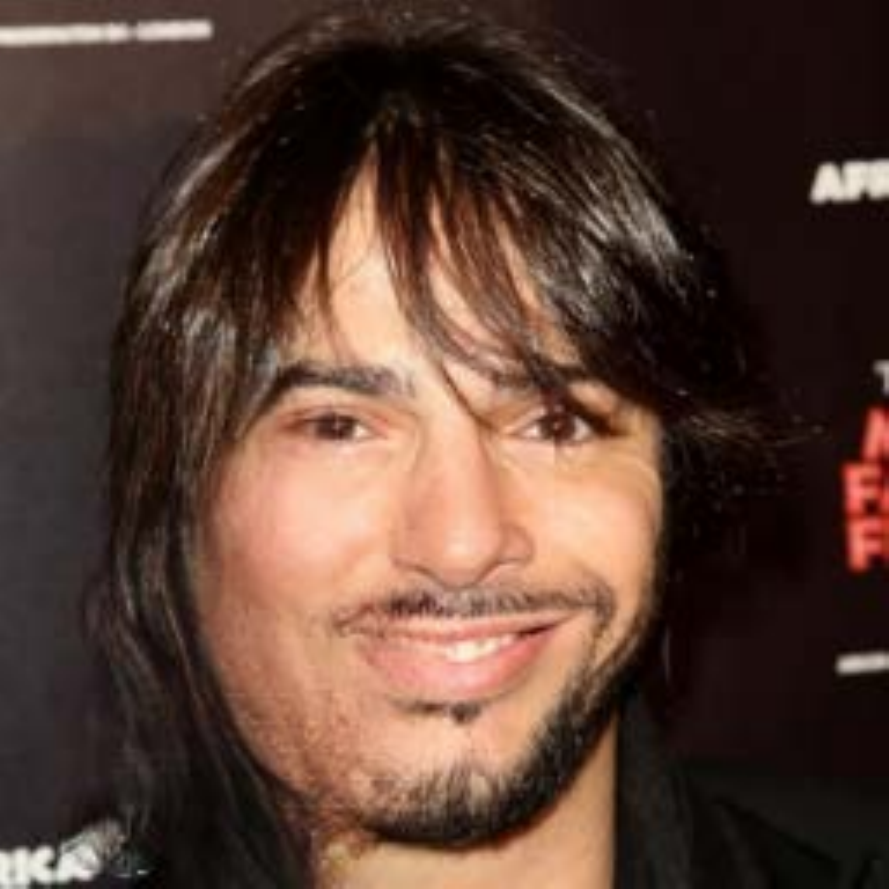}
        \includegraphics[width=2.4cm]{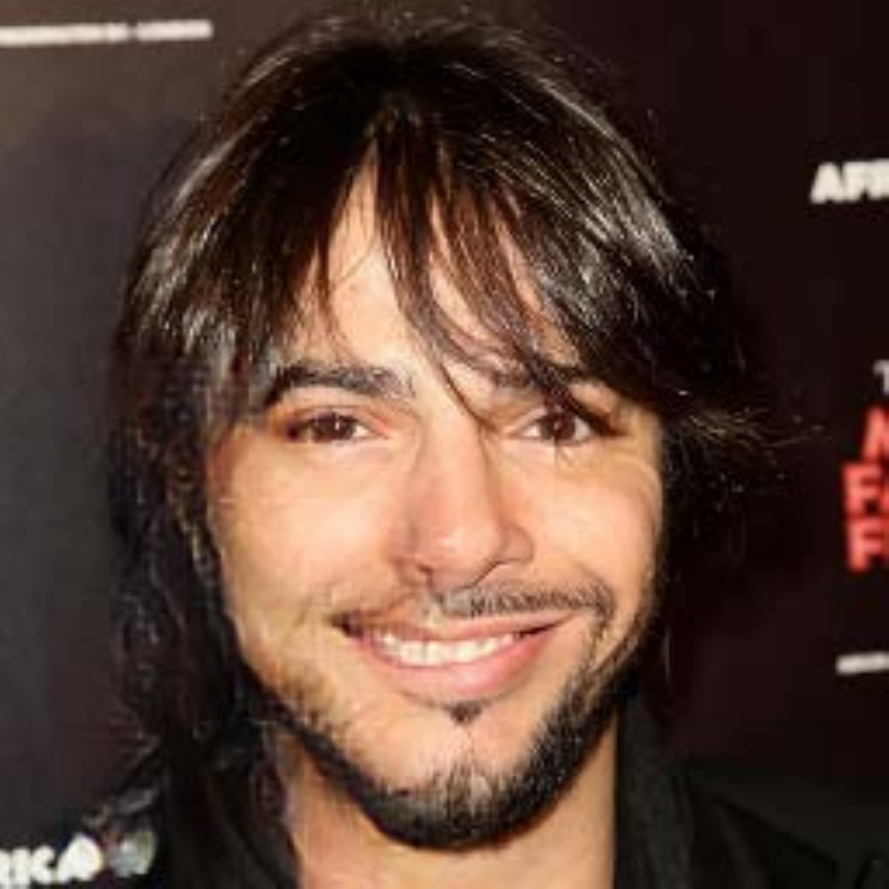}
        \includegraphics[width=2.4cm]{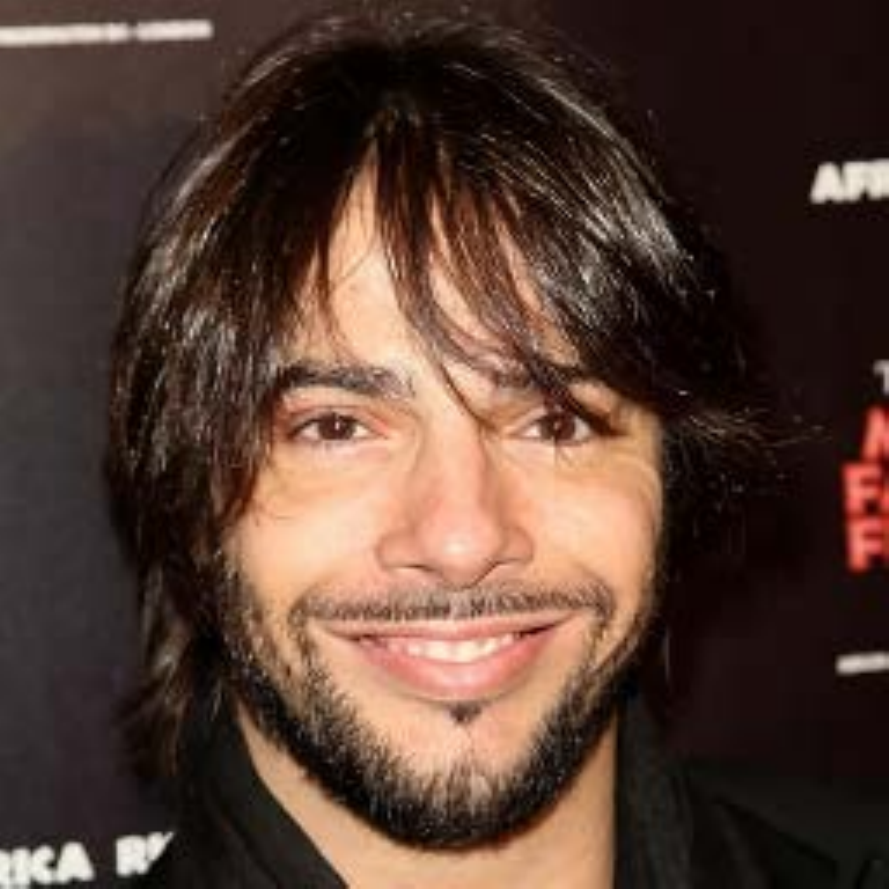}
    \end{subfigure}
    \centering
    \vspace{-0in}
    \\
    \centering
    (a) Input   \hspace{1.0cm}
    (b) CA\cite{yu2018generative}      \hspace{0.7cm}
    (c) Pconv\cite{liu2018image}  \hspace{0.6cm}
    (d) EC\cite{nazeri2019edgeconnect}      \hspace{0.6cm}
    (e) PIC\cite{Zheng2019Pluralistic}     \hspace{0.8cm}
    (f) Ours    \hspace{1.2cm}
    (g) GT      \hspace{0.5cm}
    \caption{Qualitative comparisons between different methods on CelebA-HQ}
    \label{fig:new comparisons on CelebA-HQ}
\end{figure*}

%% Here is the comparison on celeba-HQ and paris datasets with different methods.
\begin{figure*}[htp]
\centering
    \vspace{0.02in}
    \centering
    \begin{subfigure}
        \centering
        \includegraphics[width=2.4cm]{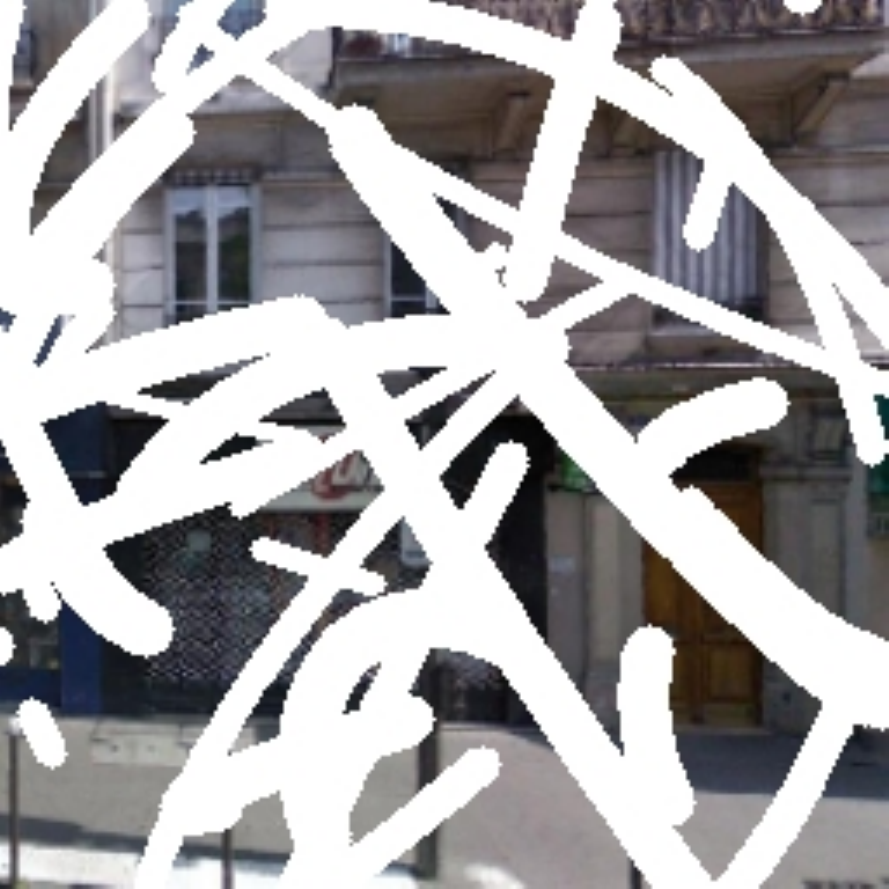}
        \includegraphics[width=2.4cm]{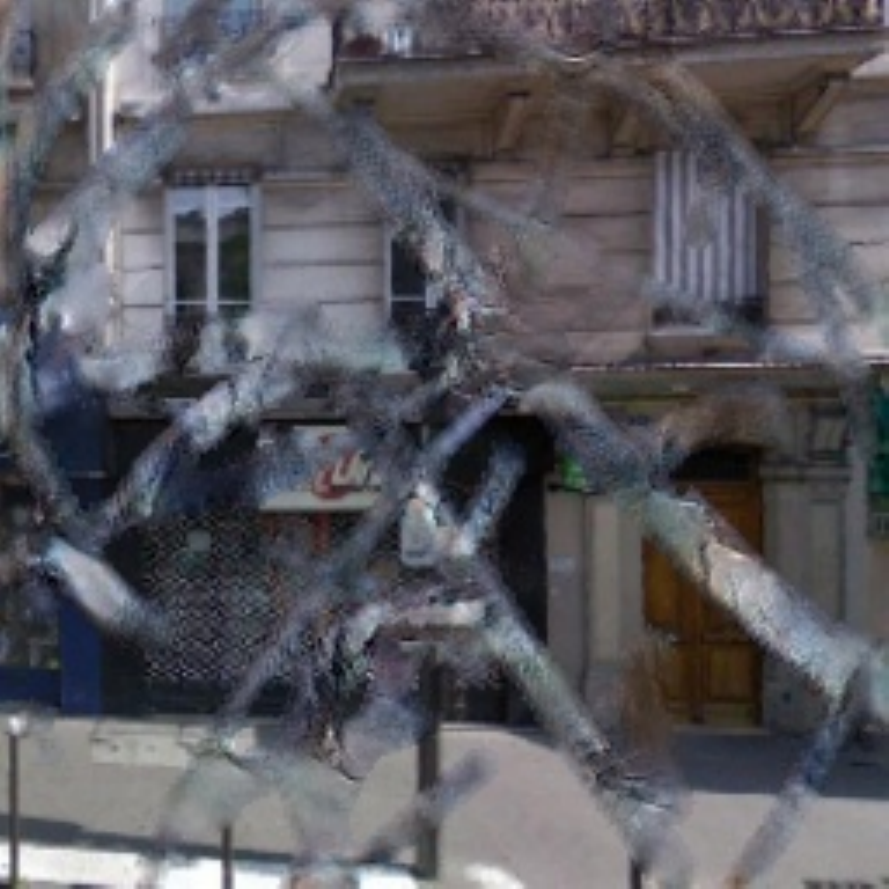}
        \includegraphics[width=2.4cm]{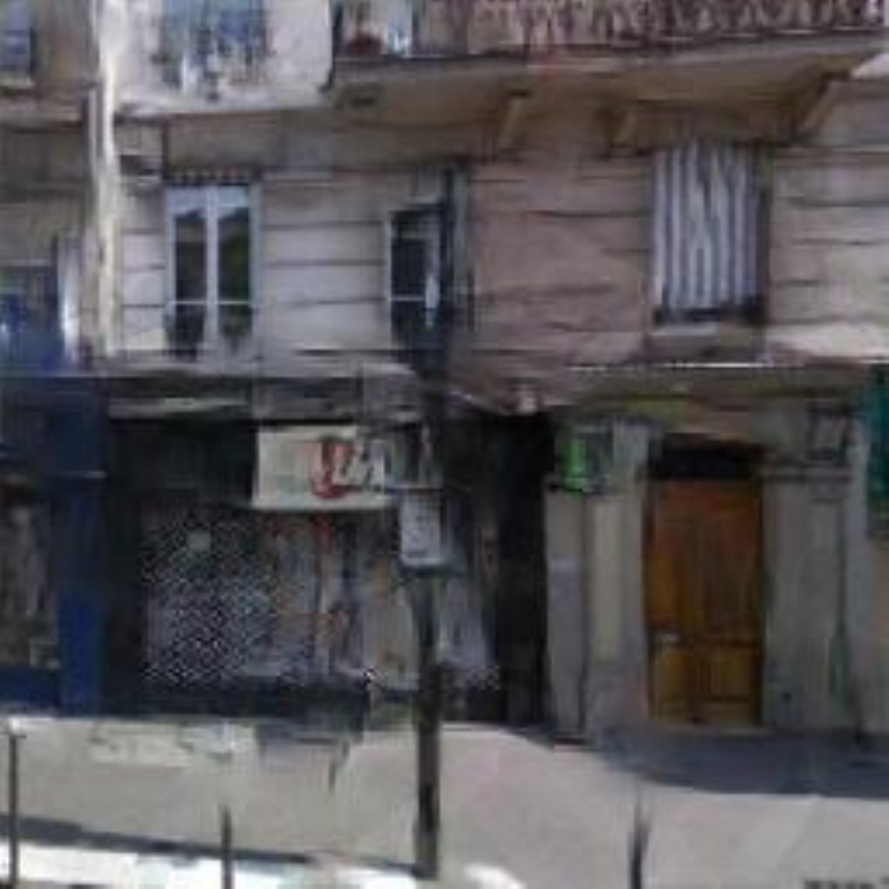}
        \includegraphics[width=2.4cm]{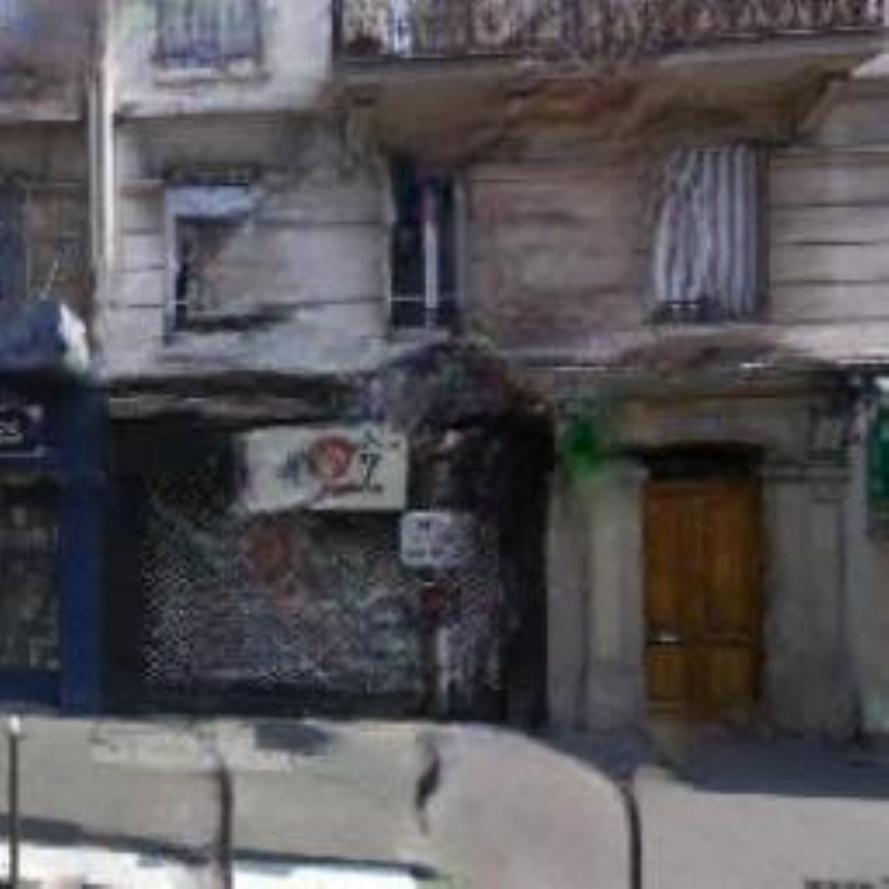}
        \includegraphics[width=2.4cm]{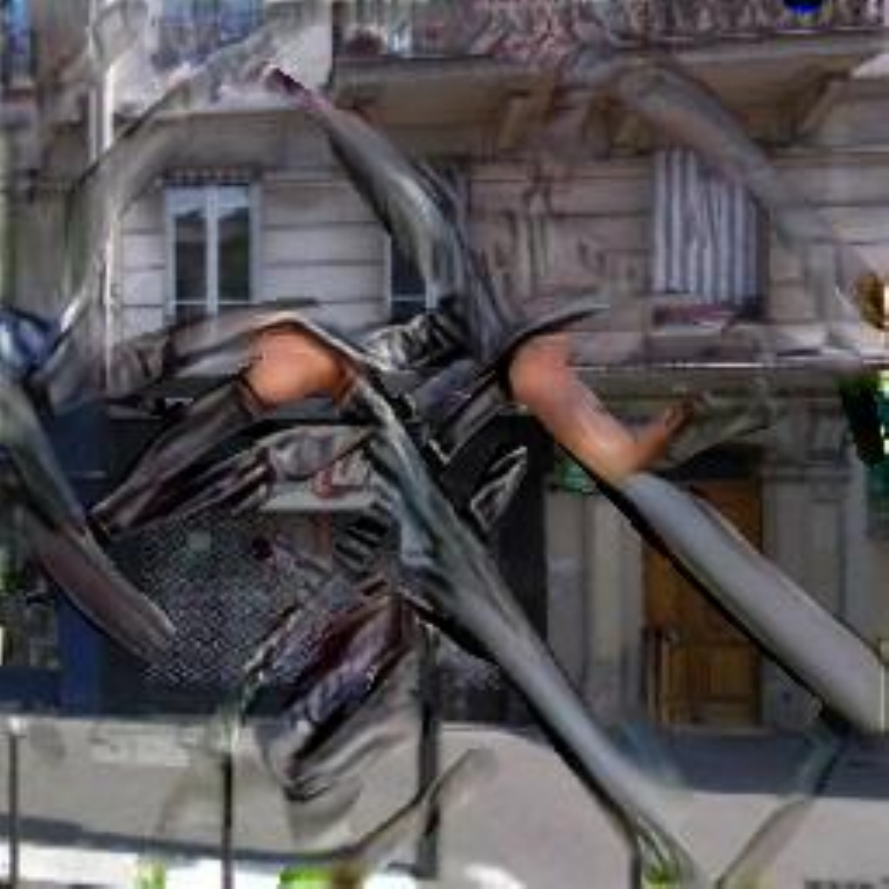}
        \includegraphics[width=2.4cm]{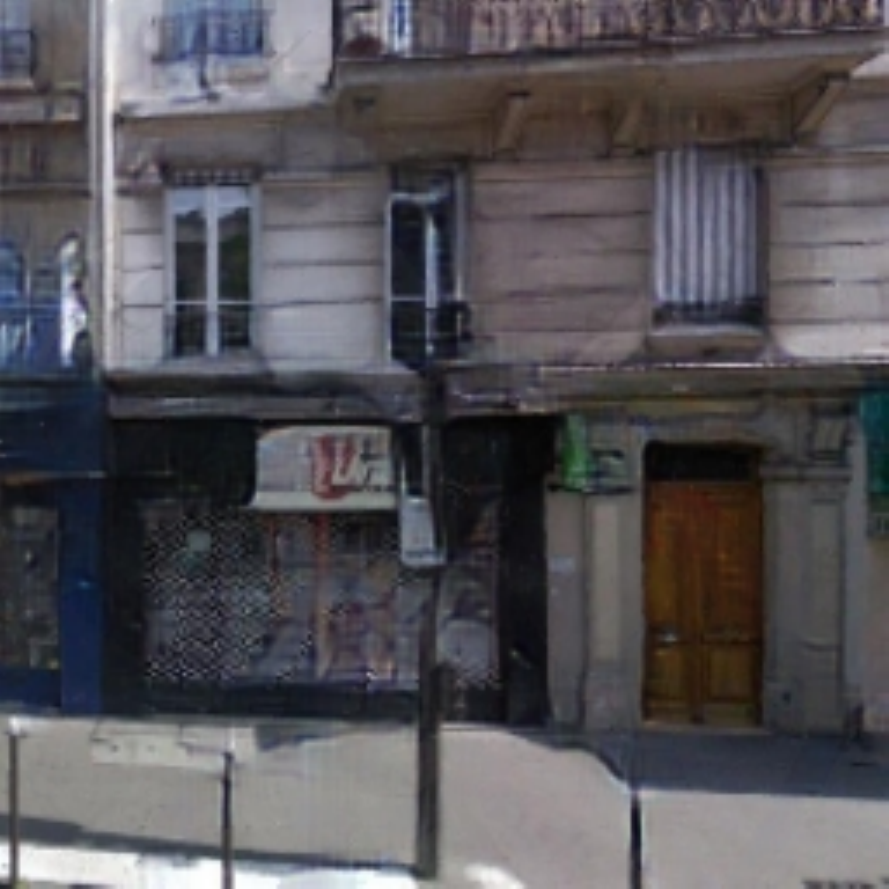}
        \includegraphics[width=2.4cm]{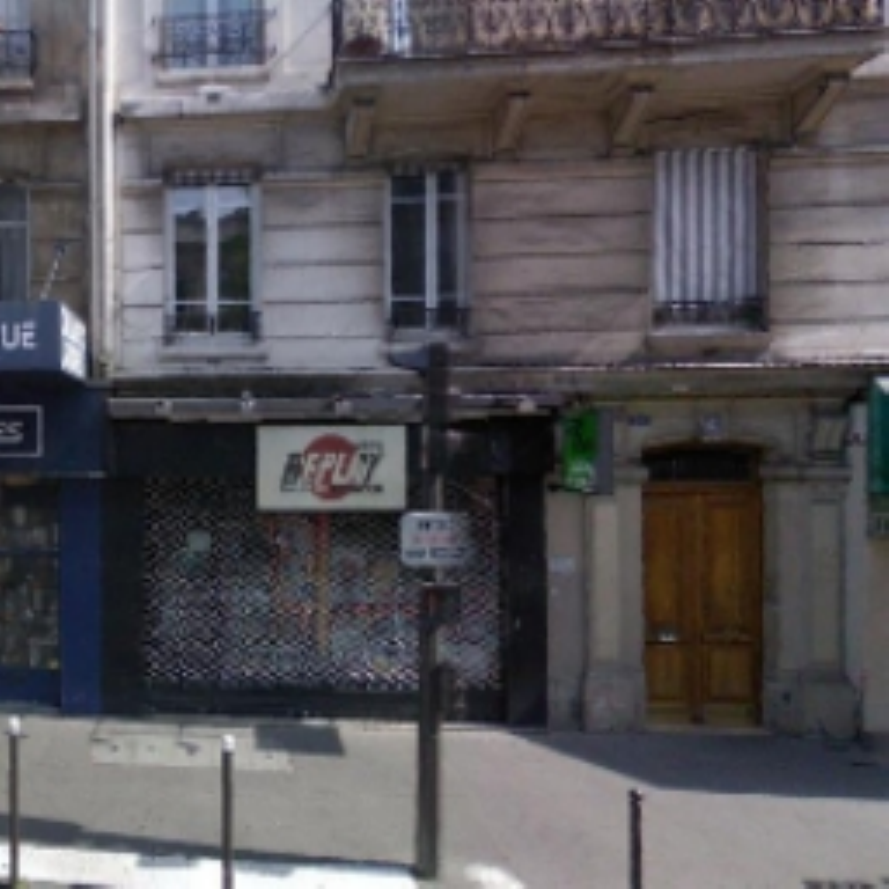}
    \end{subfigure}
    \begin{subfigure}
        \centering
        \vspace{-0.05in}
        \includegraphics[width=2.4cm]{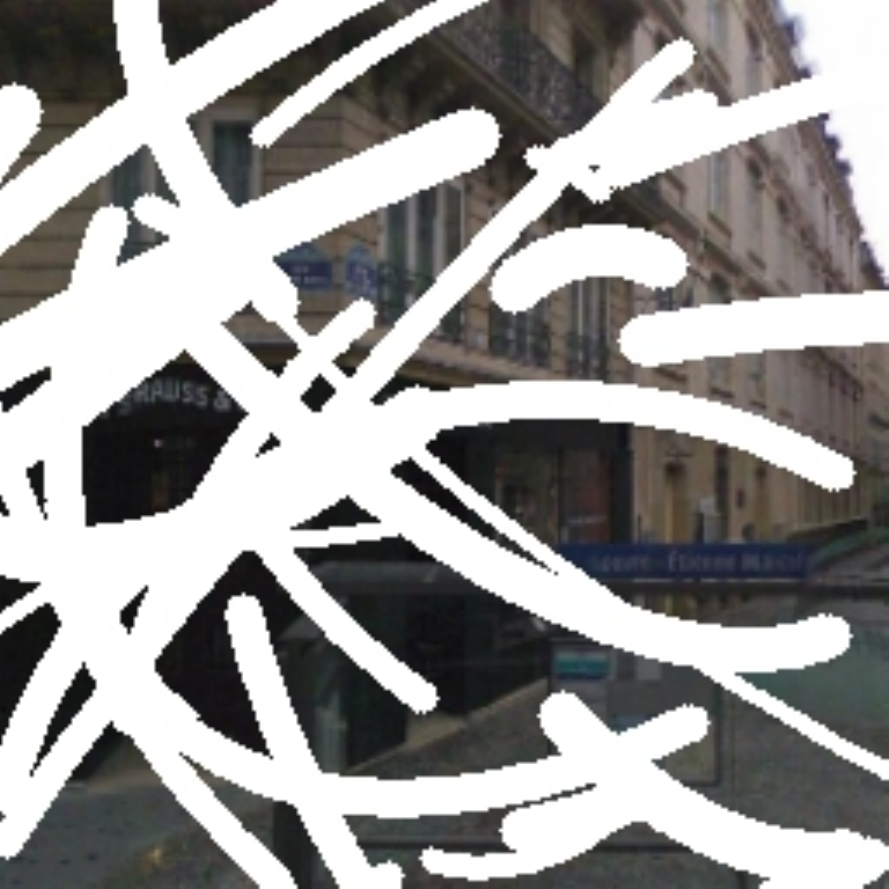}
        \includegraphics[width=2.4cm]{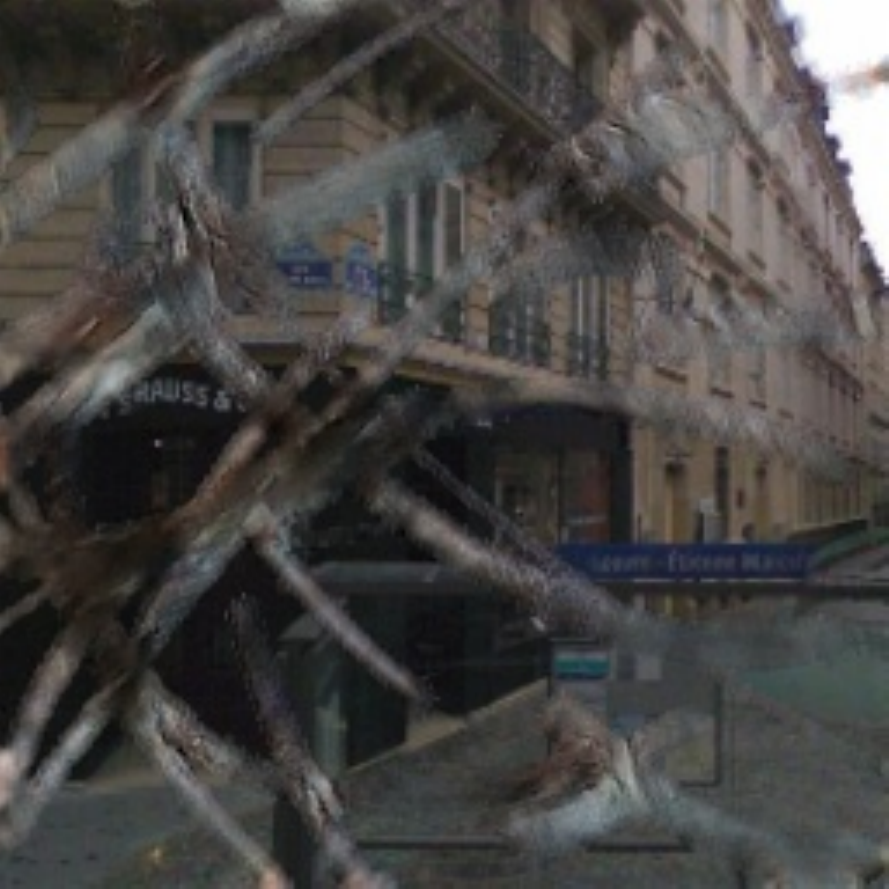}
        \includegraphics[width=2.4cm]{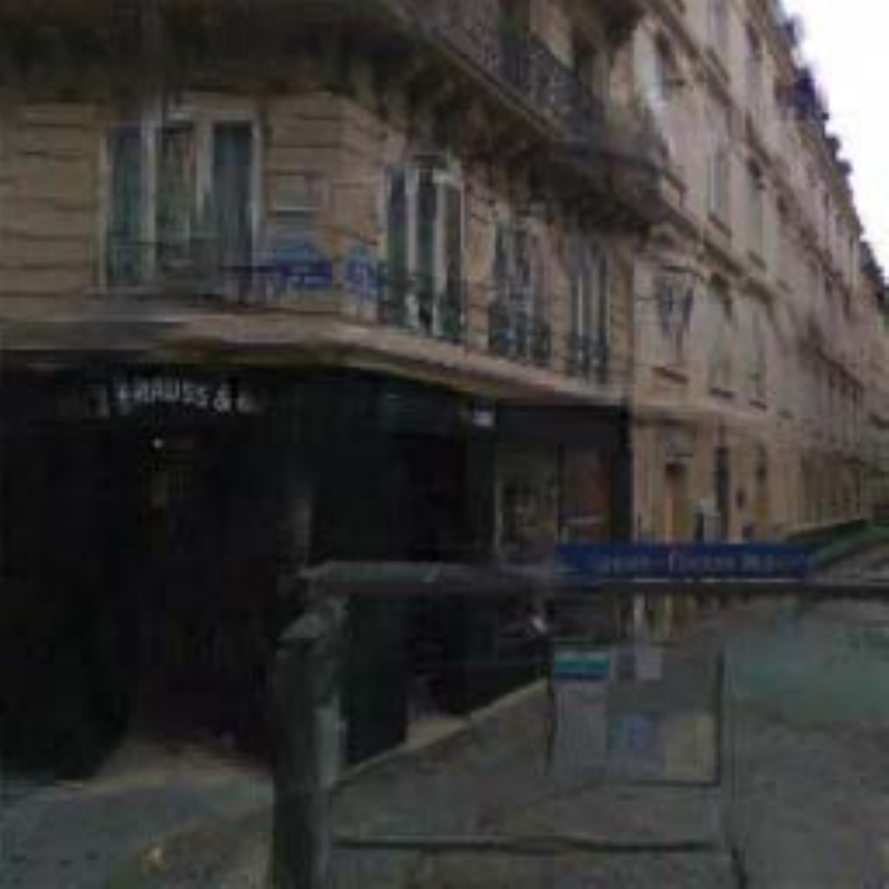}
        \includegraphics[width=2.4cm]{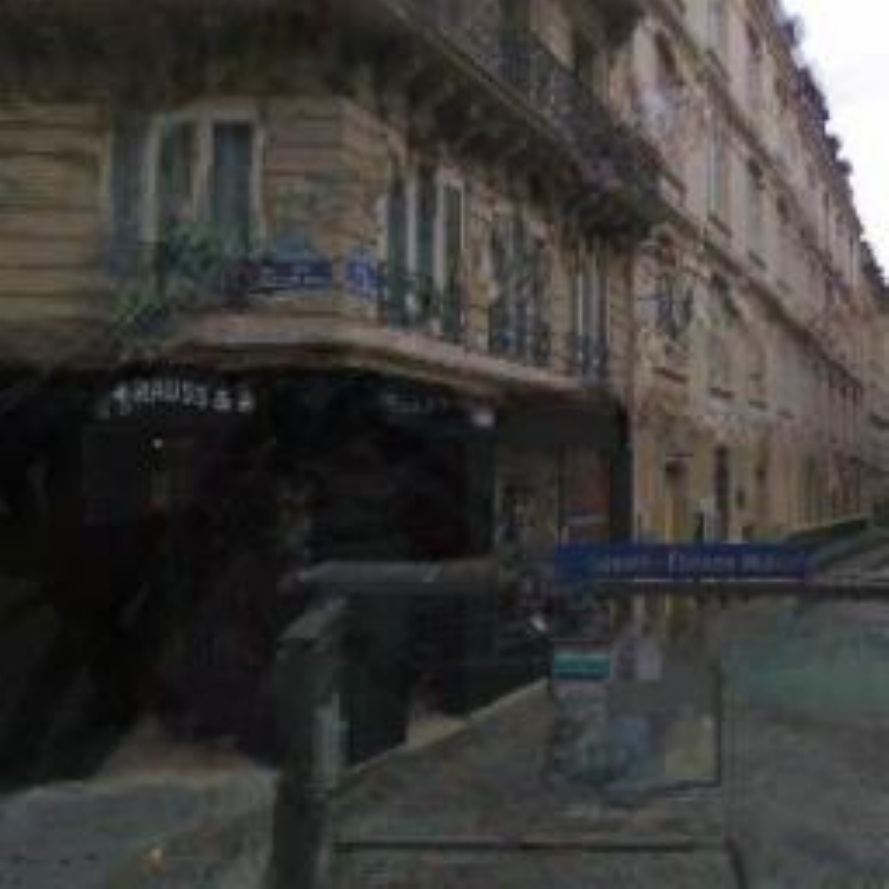}
        \includegraphics[width=2.4cm]{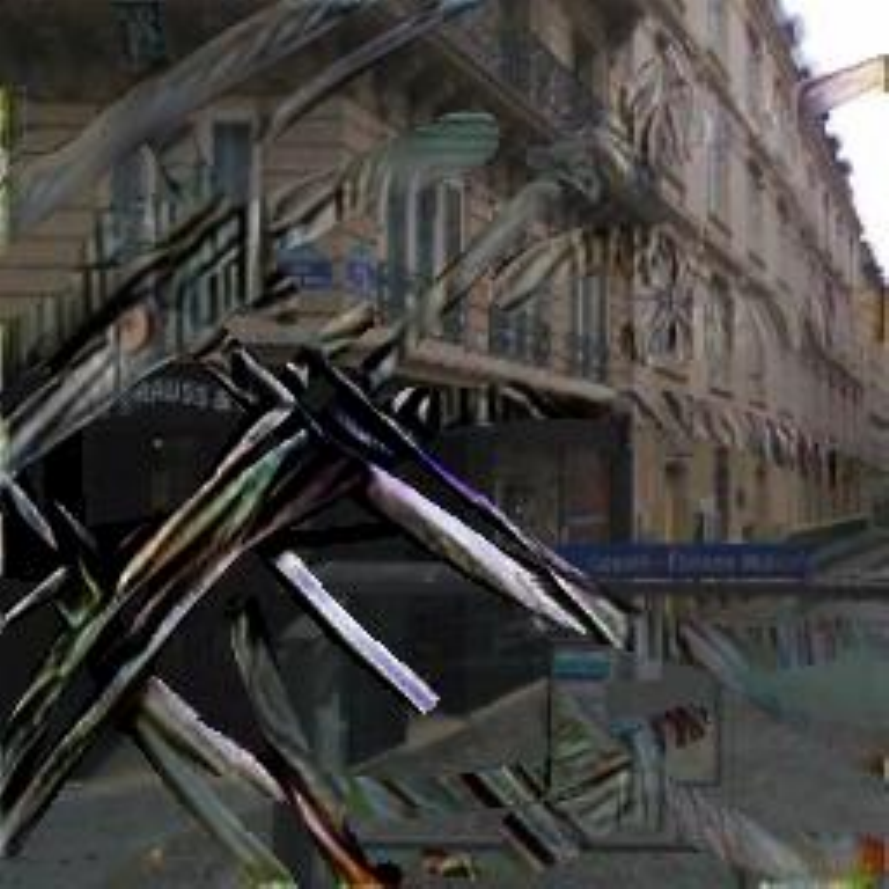}
        \includegraphics[width=2.4cm]{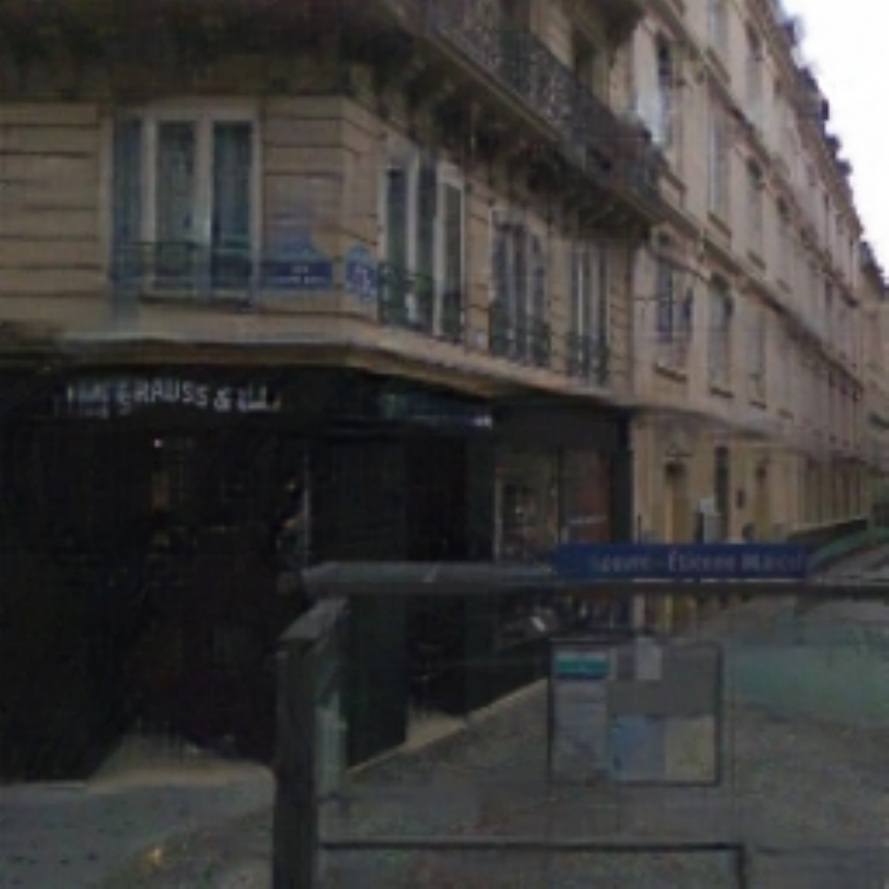}
        \includegraphics[width=2.4cm]{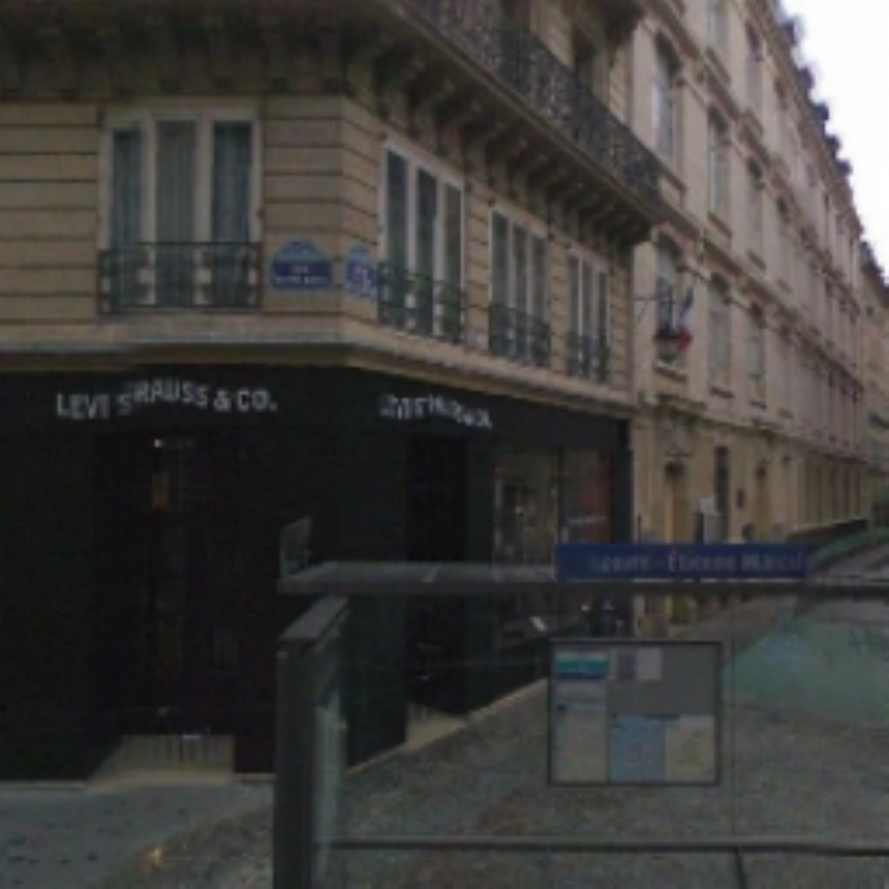}
    \end{subfigure}
    \begin{subfigure}
        \centering
        \vspace{-0.05in}
        \includegraphics[width=2.4cm]{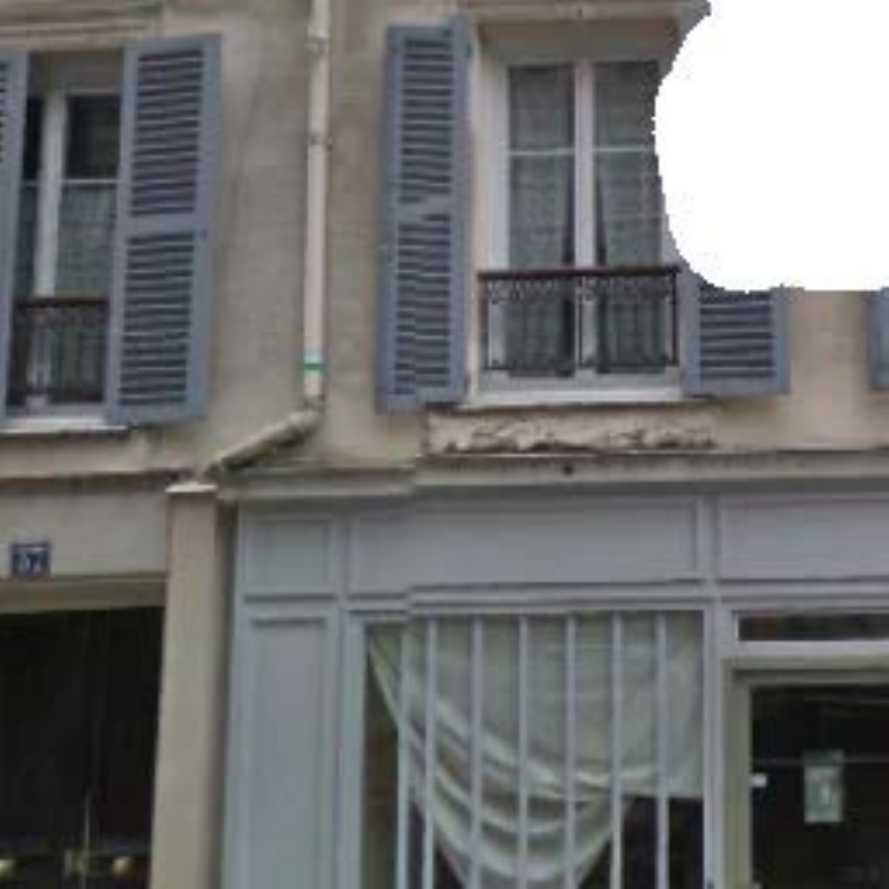}
        \includegraphics[width=2.4cm]{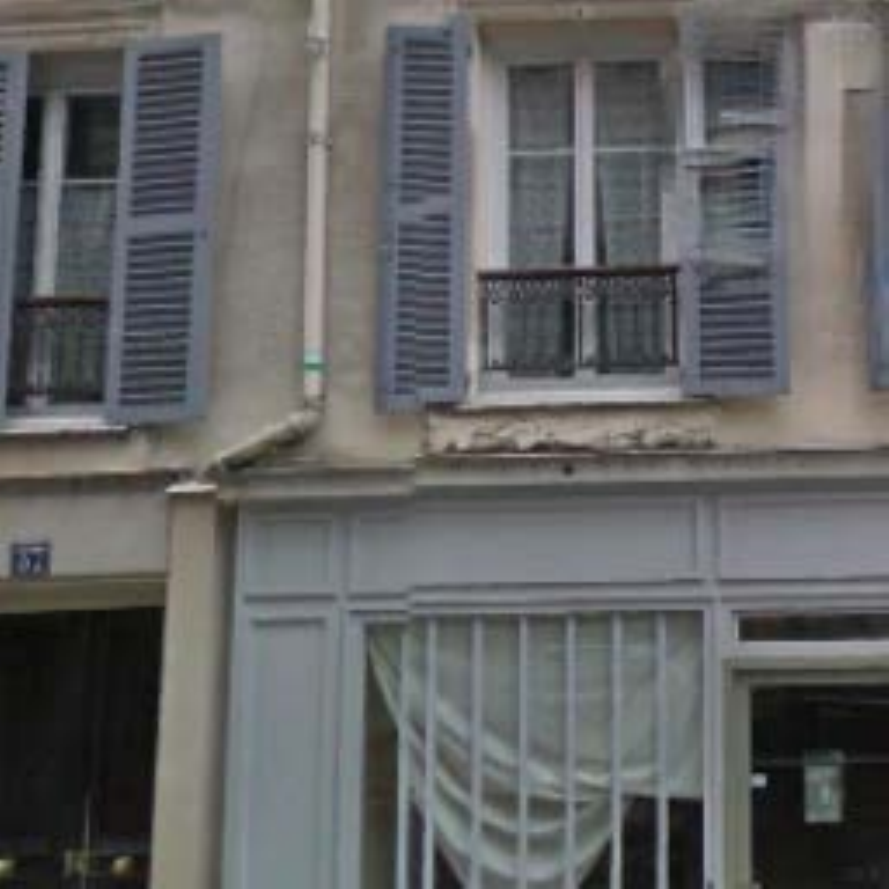}
        \includegraphics[width=2.4cm]{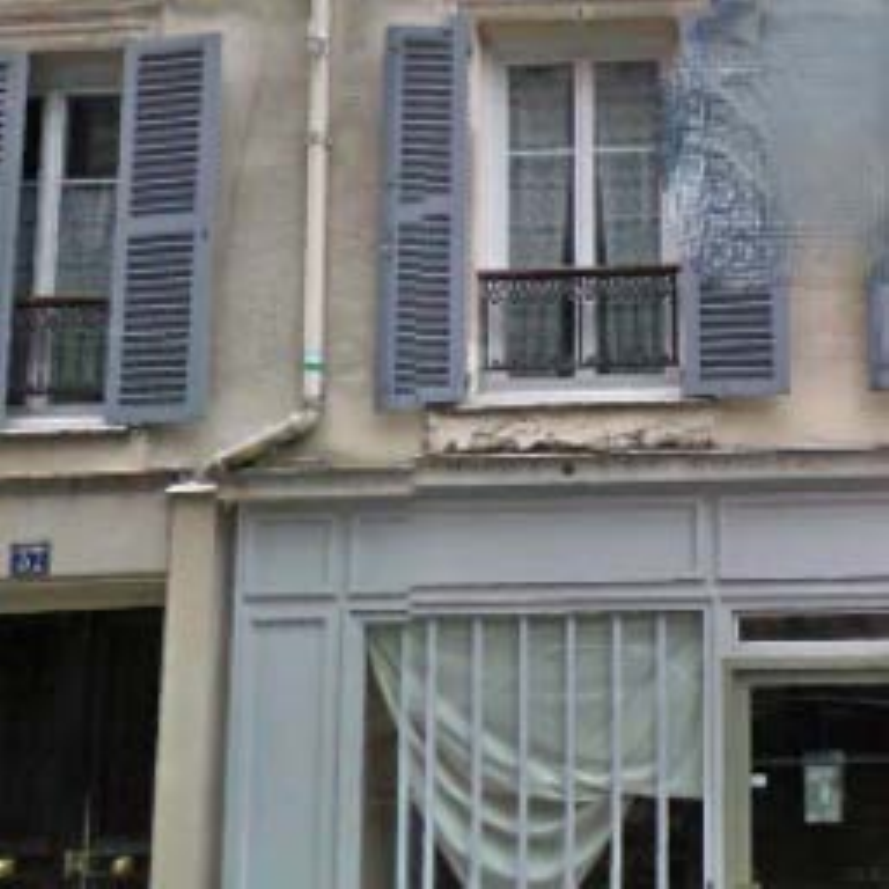}
        \includegraphics[width=2.4cm]{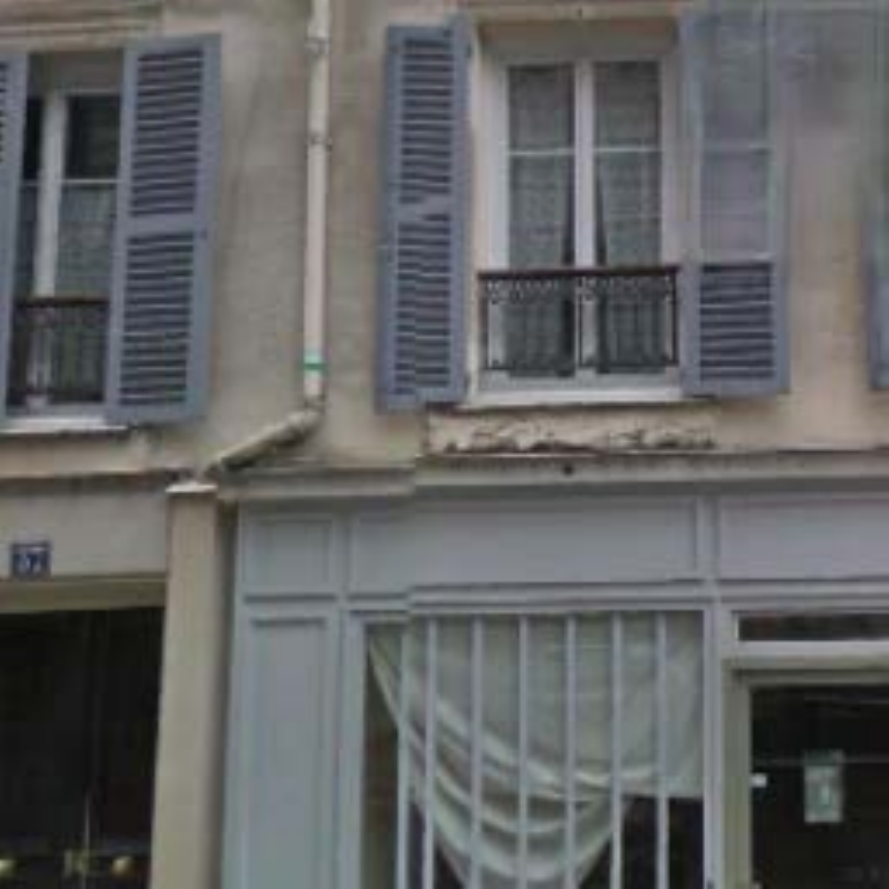}
        \includegraphics[width=2.4cm]{partial_008_im_con_paris.pdf}
        \includegraphics[width=2.4cm]{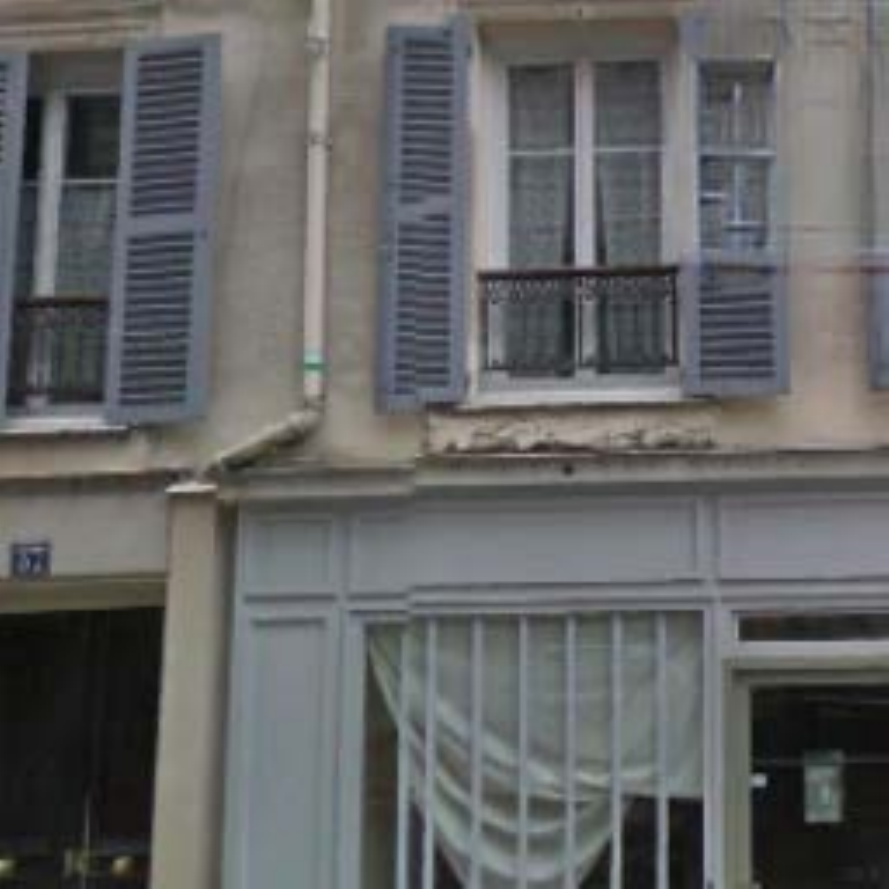}
        \includegraphics[width=2.4cm]{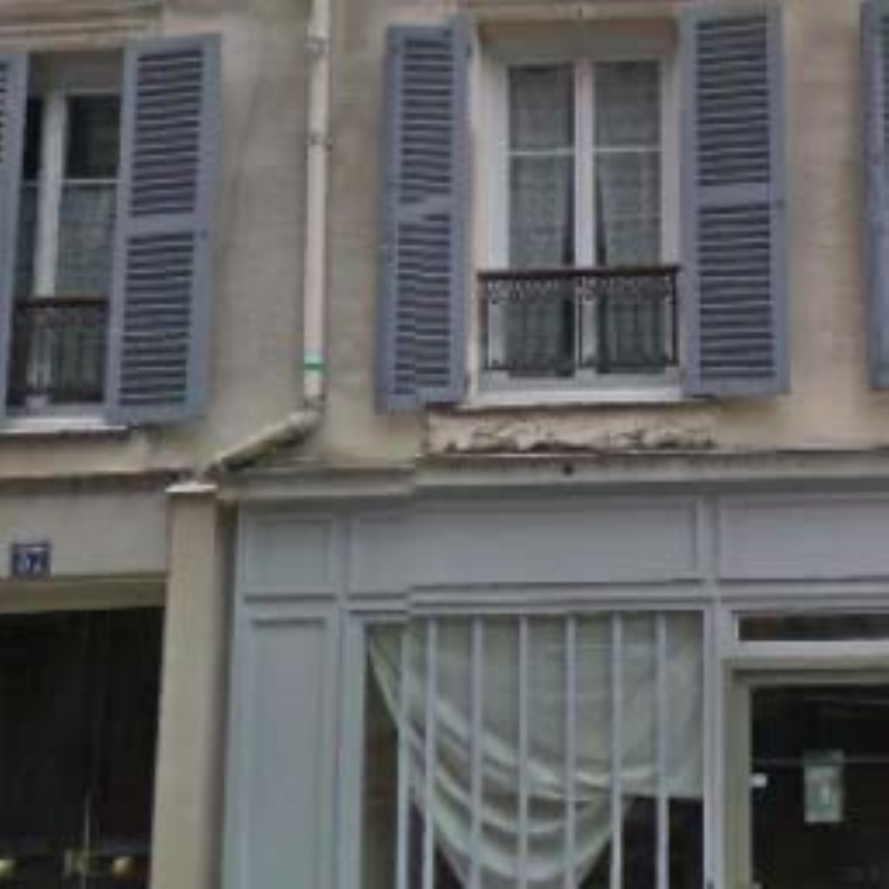}
    \end{subfigure}
    \begin{subfigure}
        \centering
        \vspace{-0.05in}
        \includegraphics[width=2.4cm]{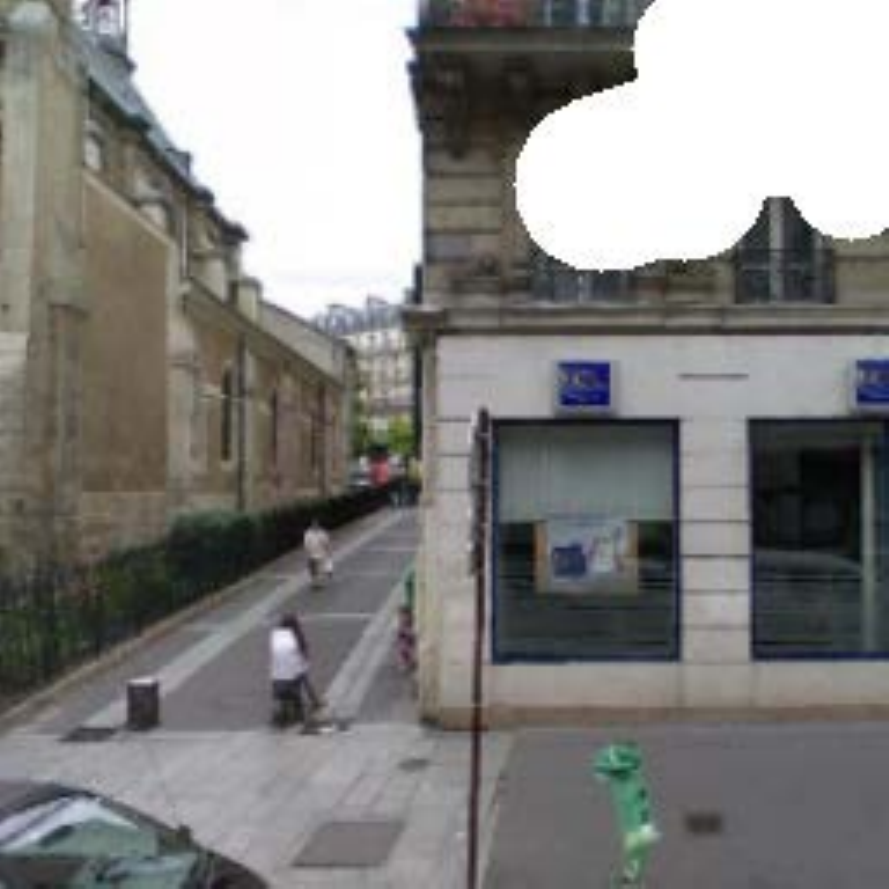}
        \includegraphics[width=2.4cm]{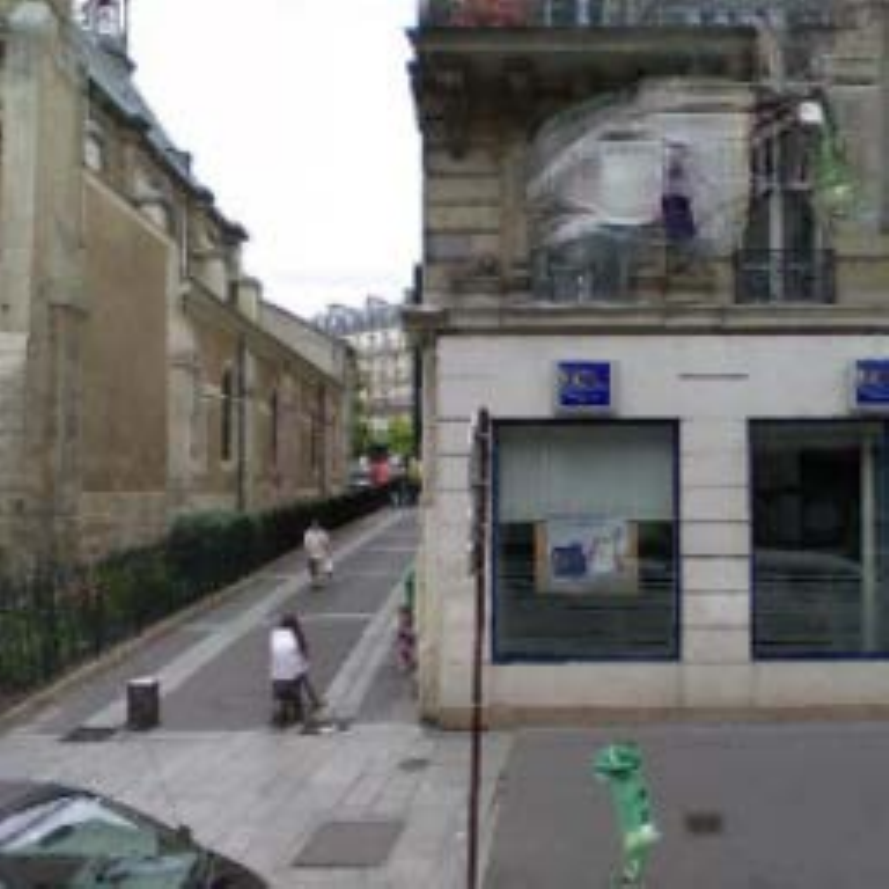}
        \includegraphics[width=2.4cm]{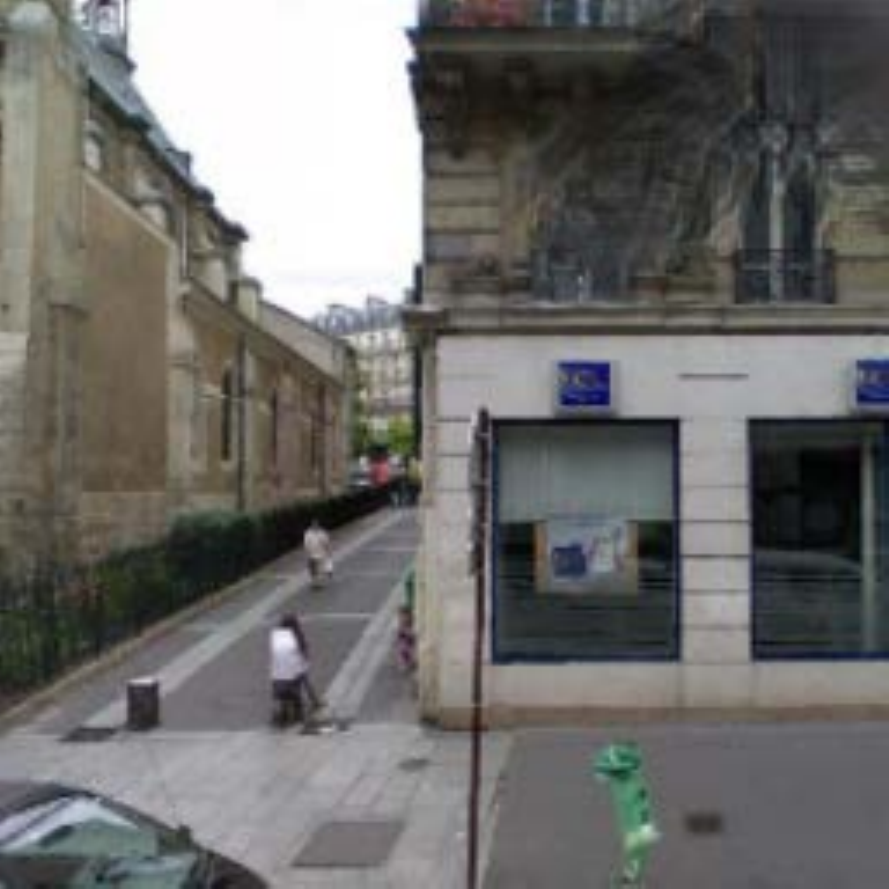}
        \includegraphics[width=2.4cm]{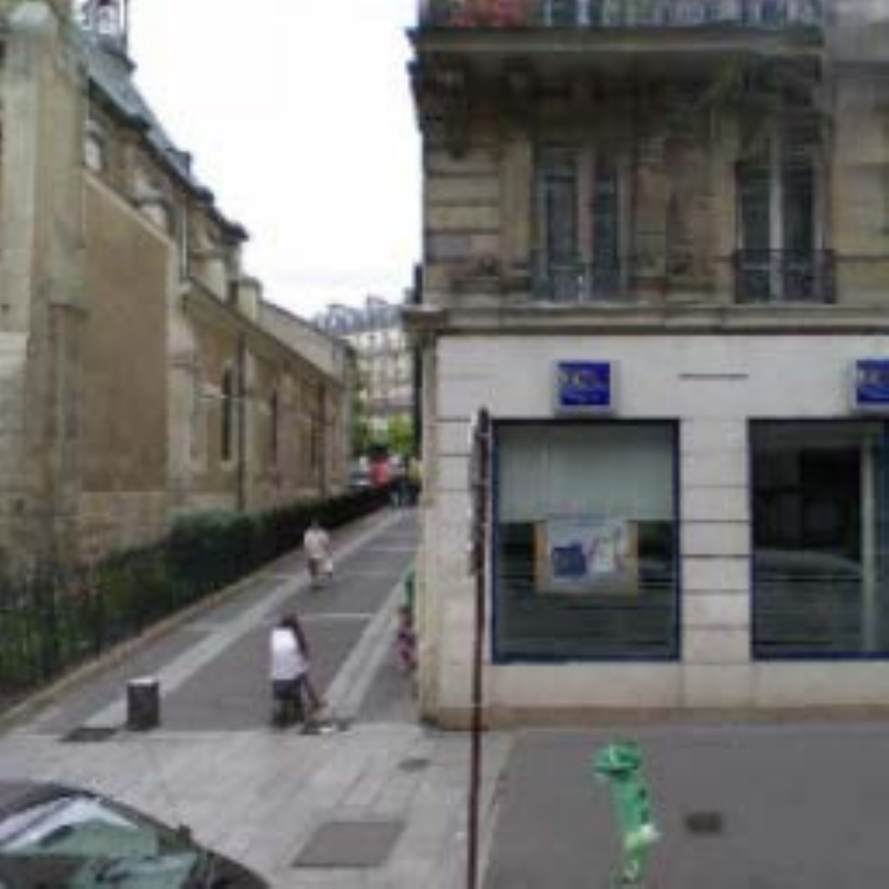}
        \includegraphics[width=2.4cm]{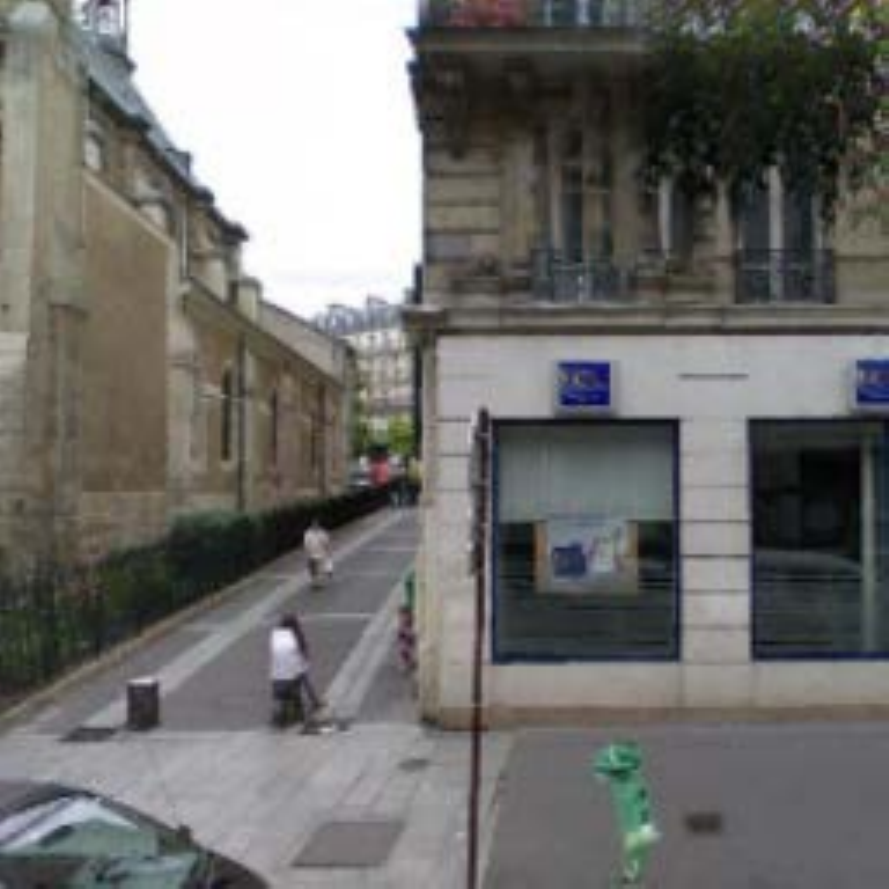}
        \includegraphics[width=2.4cm]{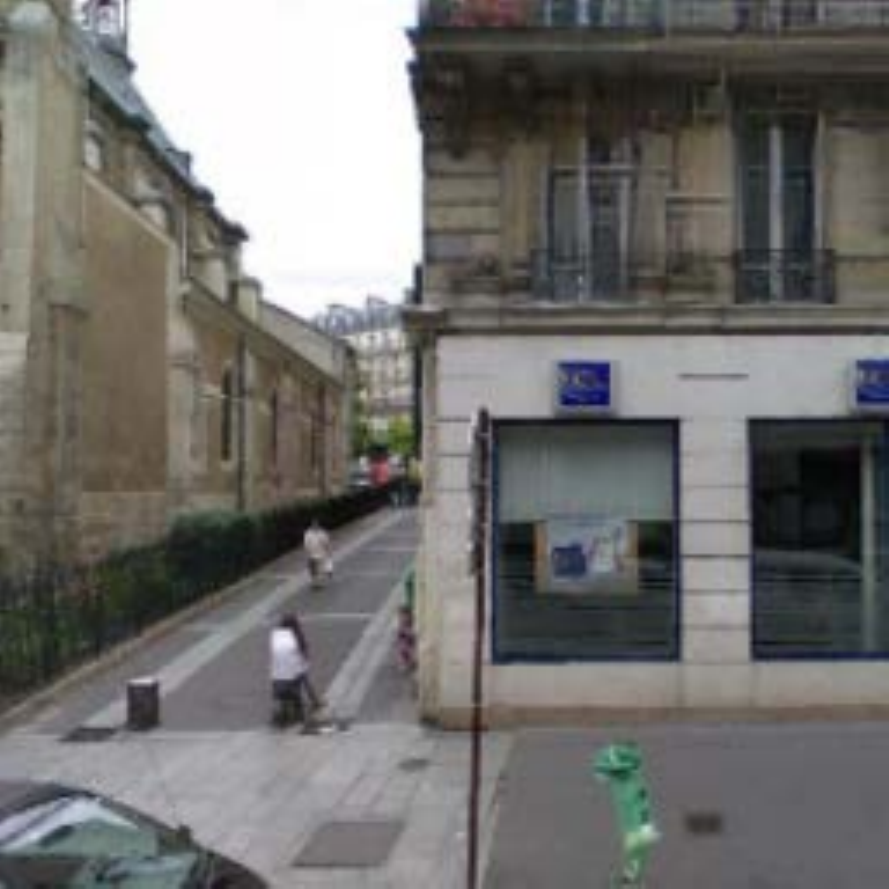}
        \includegraphics[width=2.4cm]{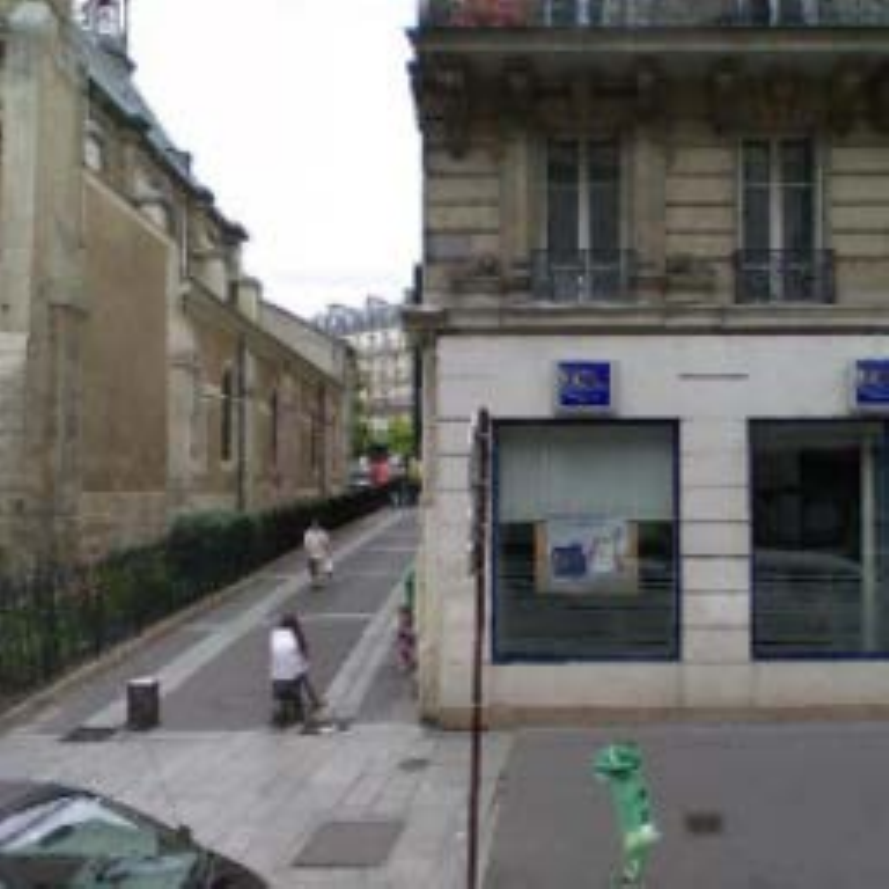}
    \end{subfigure}
    \centering
    \vspace{-0in}
    \\
    \centering
    (a) Input   \hspace{1.0cm}
    (b) CA\cite{yu2018generative}      \hspace{0.7cm}
    (c) Pconv\cite{liu2018image}  \hspace{0.6cm}
    (d) EC\cite{nazeri2019edgeconnect}      \hspace{0.6cm}
    (e) PIC\cite{Zheng2019Pluralistic}     \hspace{0.8cm}
    (f) Ours    \hspace{1.2cm}
    (g) GT      \hspace{0.5cm}
    \caption{Qualitative comparisons between different methods on Paris}
    \label{fig:new comparisons on Paris}
\end{figure*}
\begin{figure*}[htp]
\centering
    \vspace{0.02in}
    \centering
    \begin{subfigure}
        \centering
        \includegraphics[width=2.4cm]{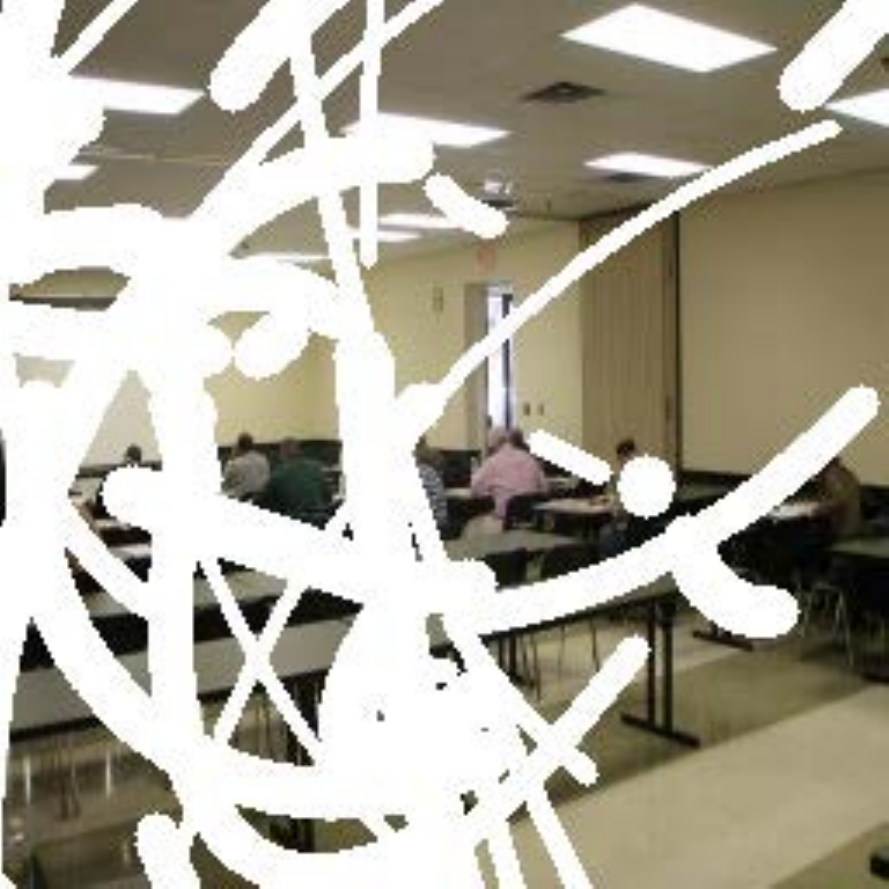}
        \includegraphics[width=2.4cm]{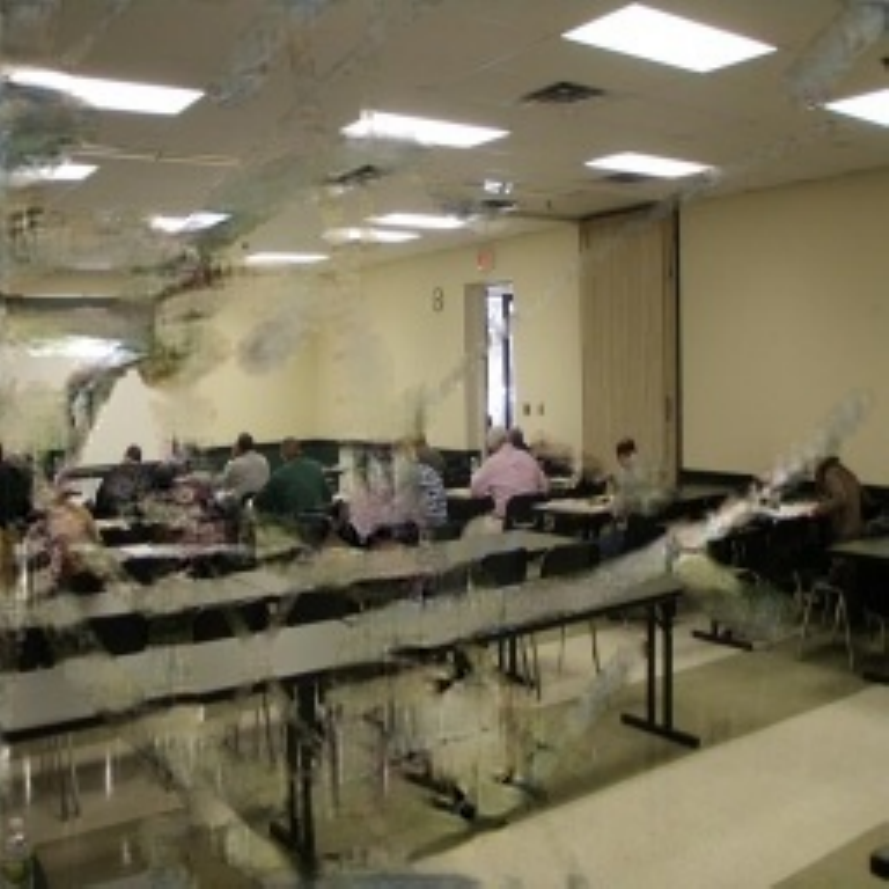}
        \includegraphics[width=2.4cm]{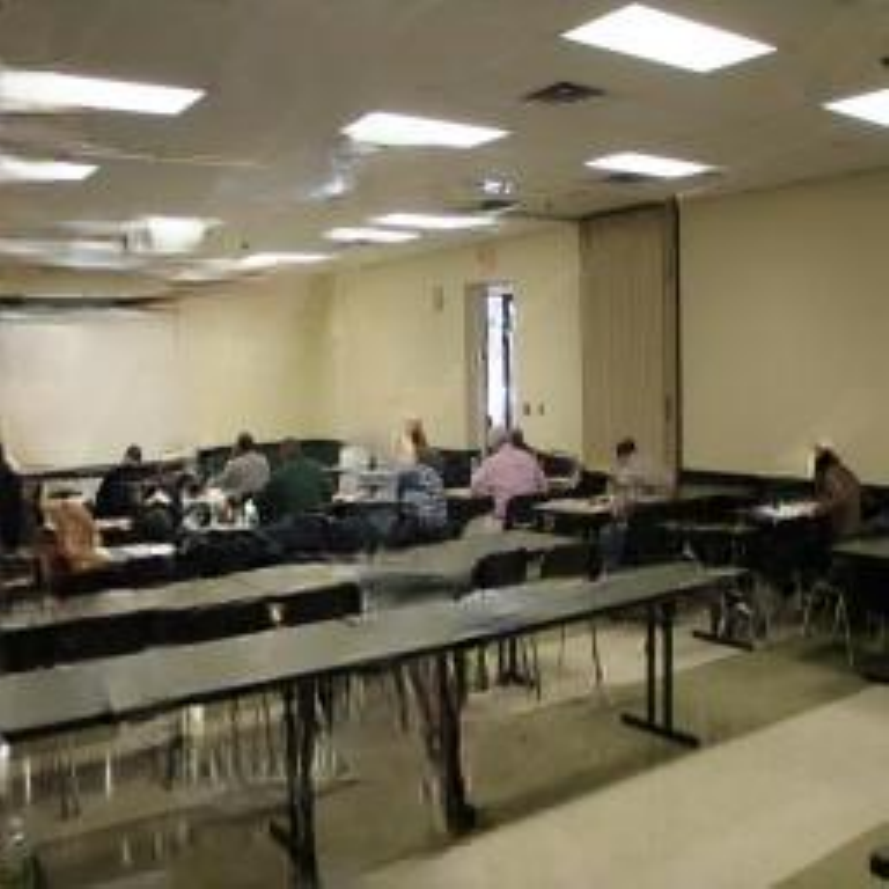}
        \includegraphics[width=2.4cm]{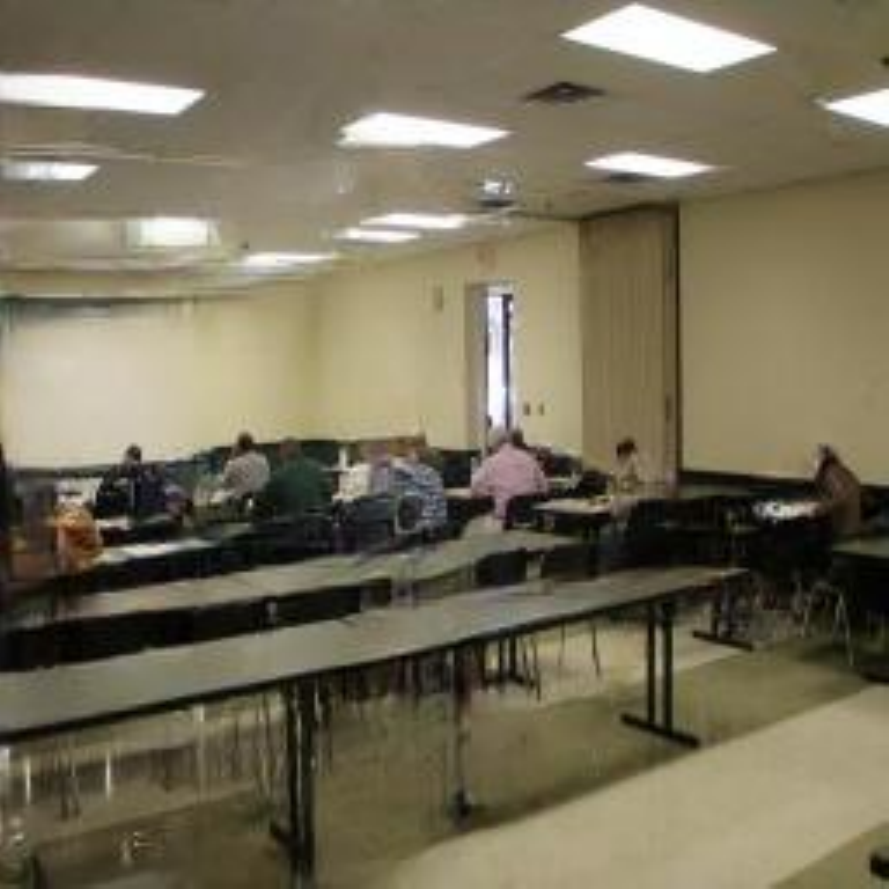}
        \includegraphics[width=2.4cm]{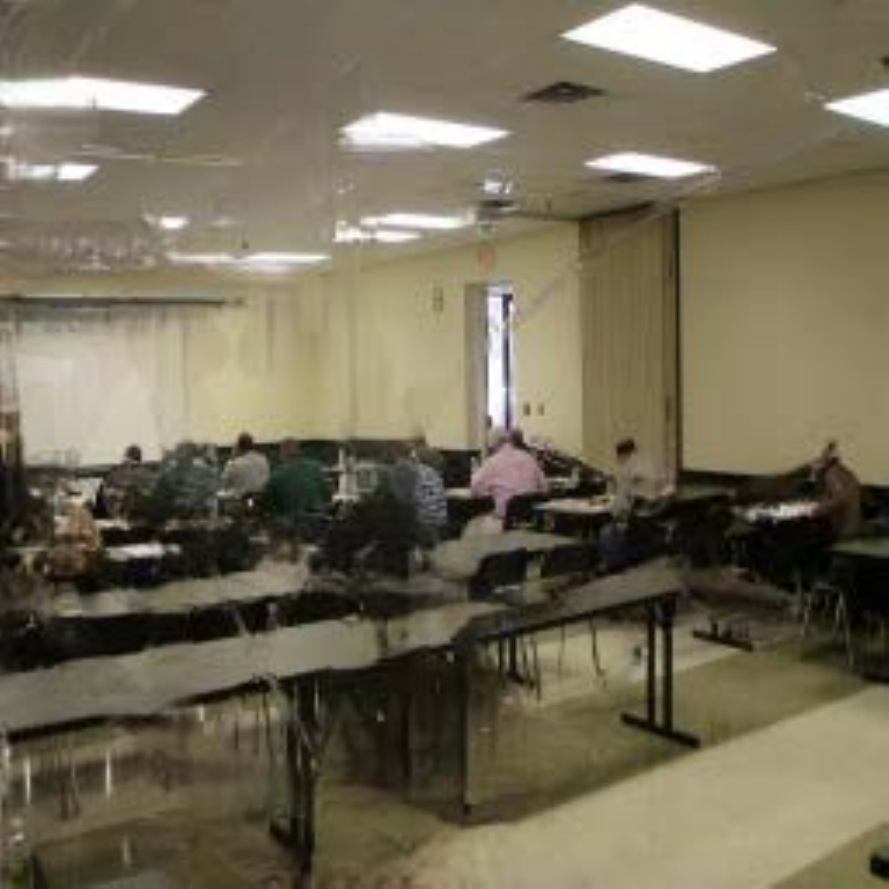}
        \includegraphics[width=2.4cm]{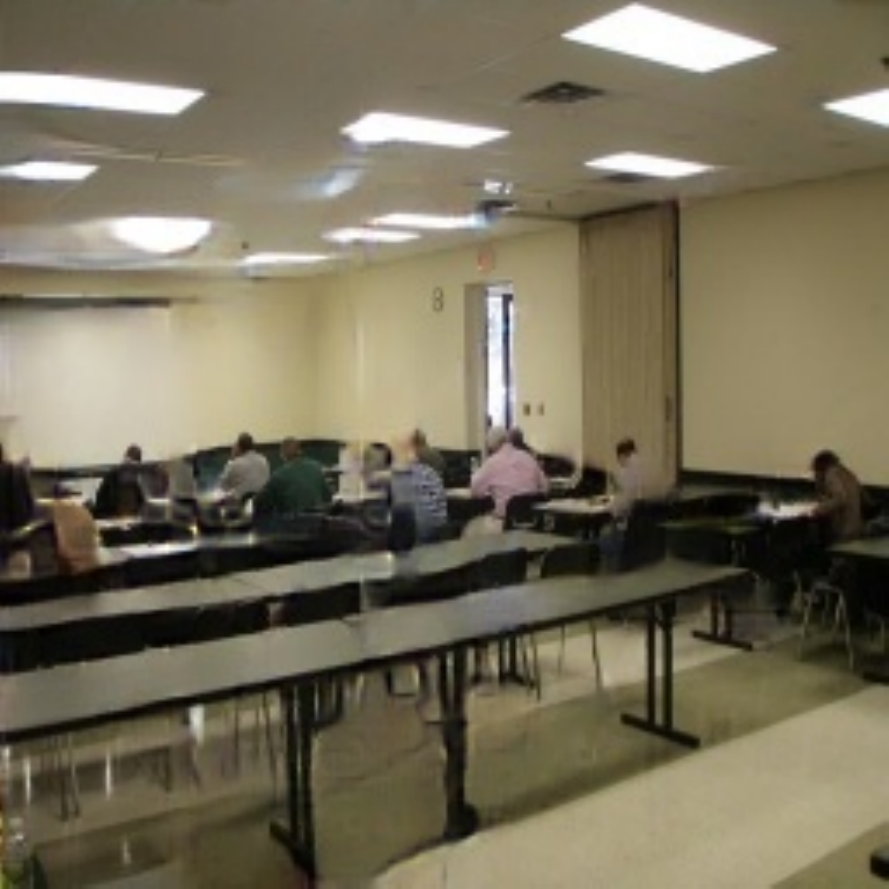}
        \includegraphics[width=2.4cm]{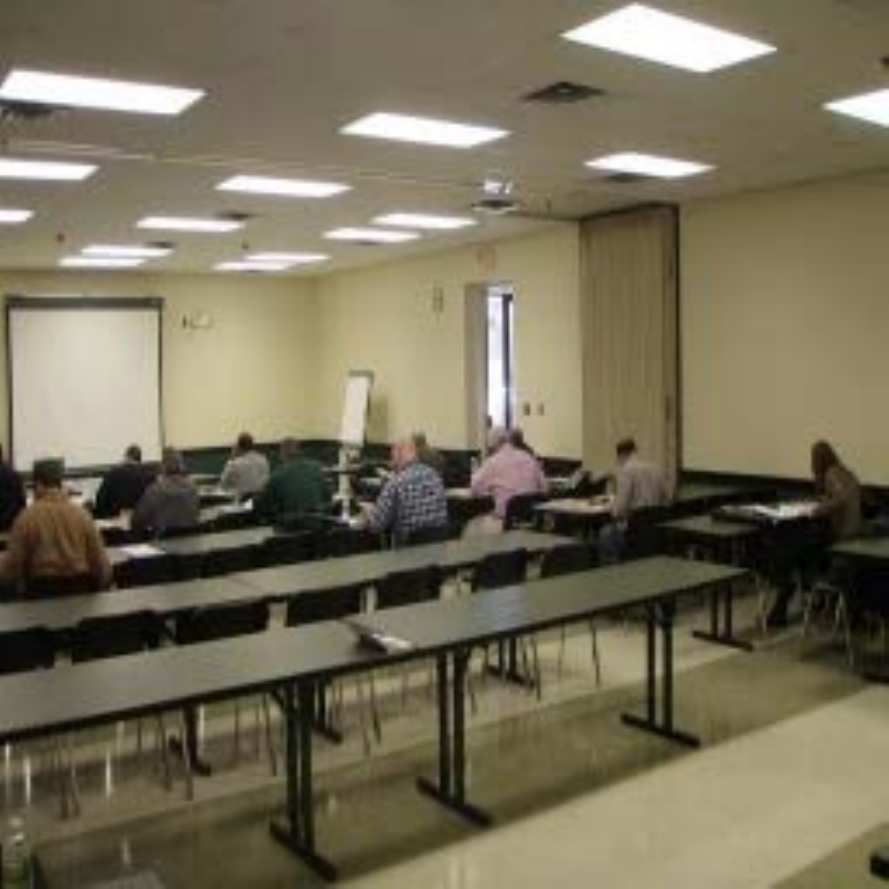}
    \end{subfigure}
    \begin{subfigure}
        \centering
        \vspace{-0.05in}
        \includegraphics[width=2.4cm]{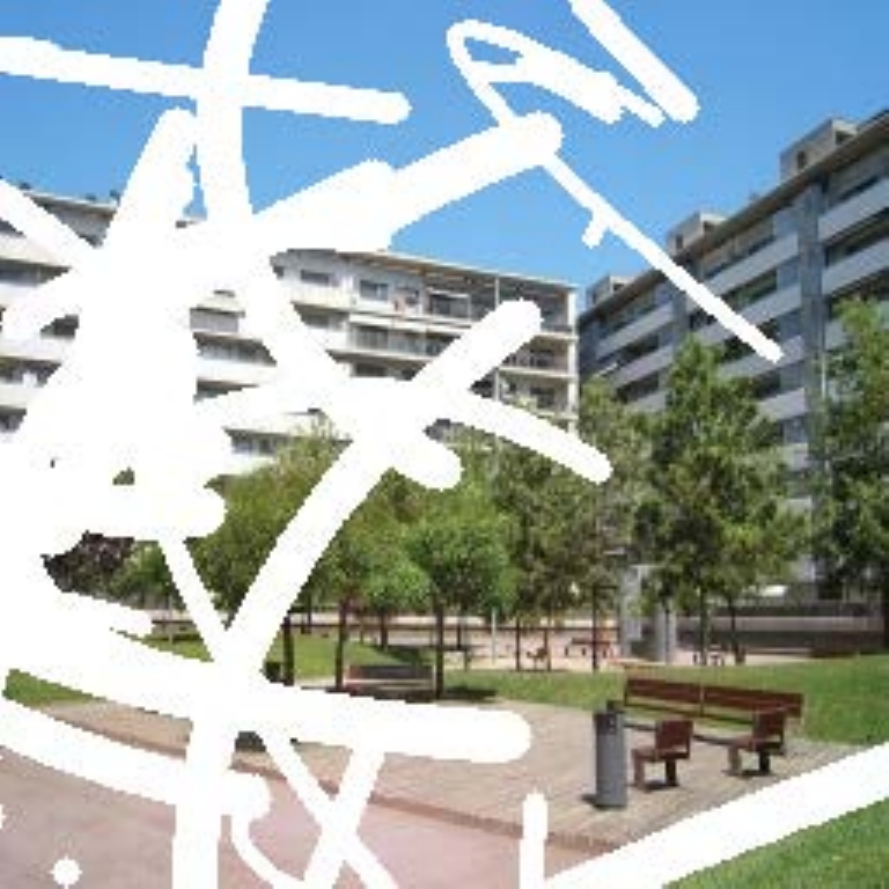}
        \includegraphics[width=2.4cm]{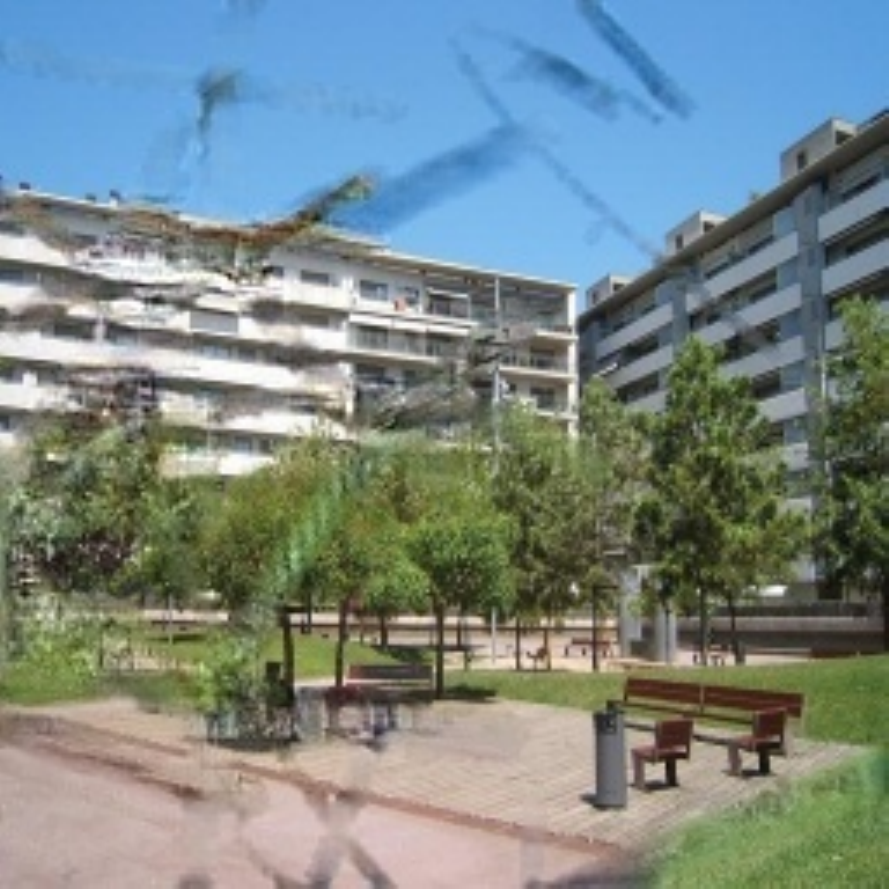}
        \includegraphics[width=2.4cm]{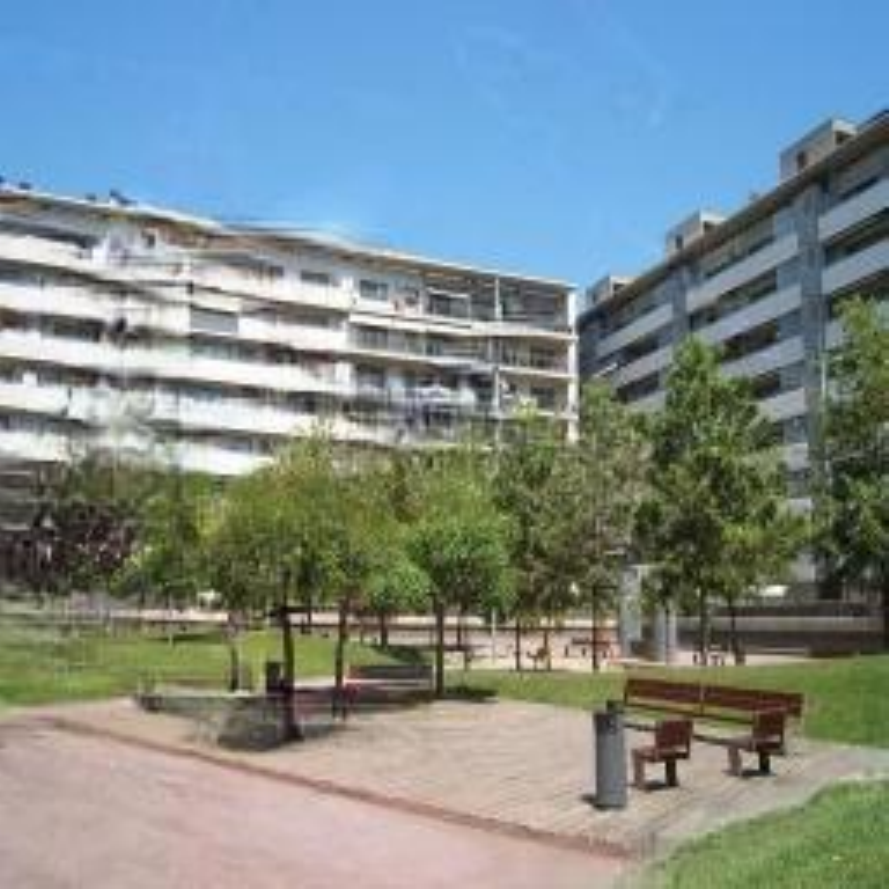}
        \includegraphics[width=2.4cm]{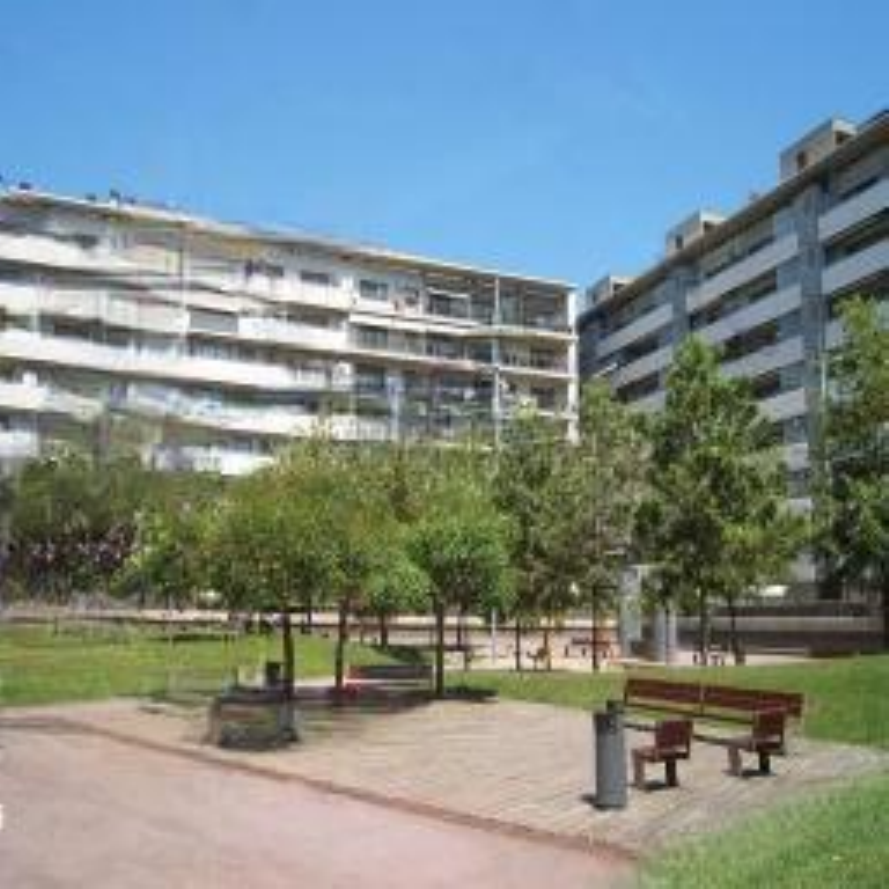}
        \includegraphics[width=2.4cm]{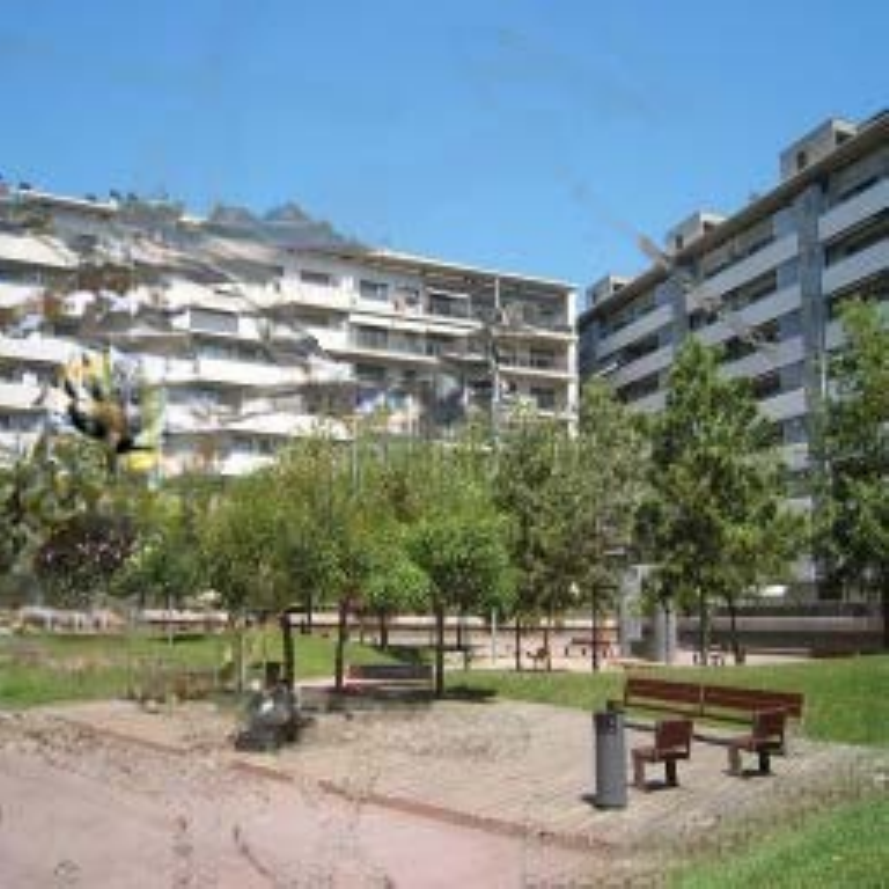}
        \includegraphics[width=2.4cm]{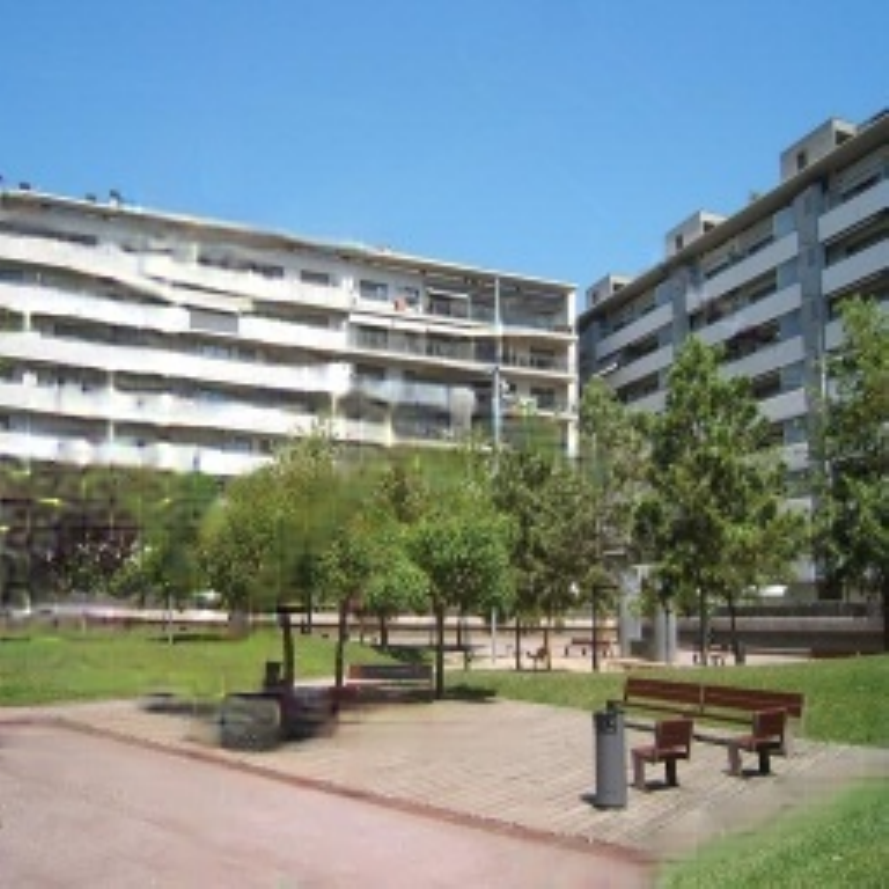}
        \includegraphics[width=2.4cm]{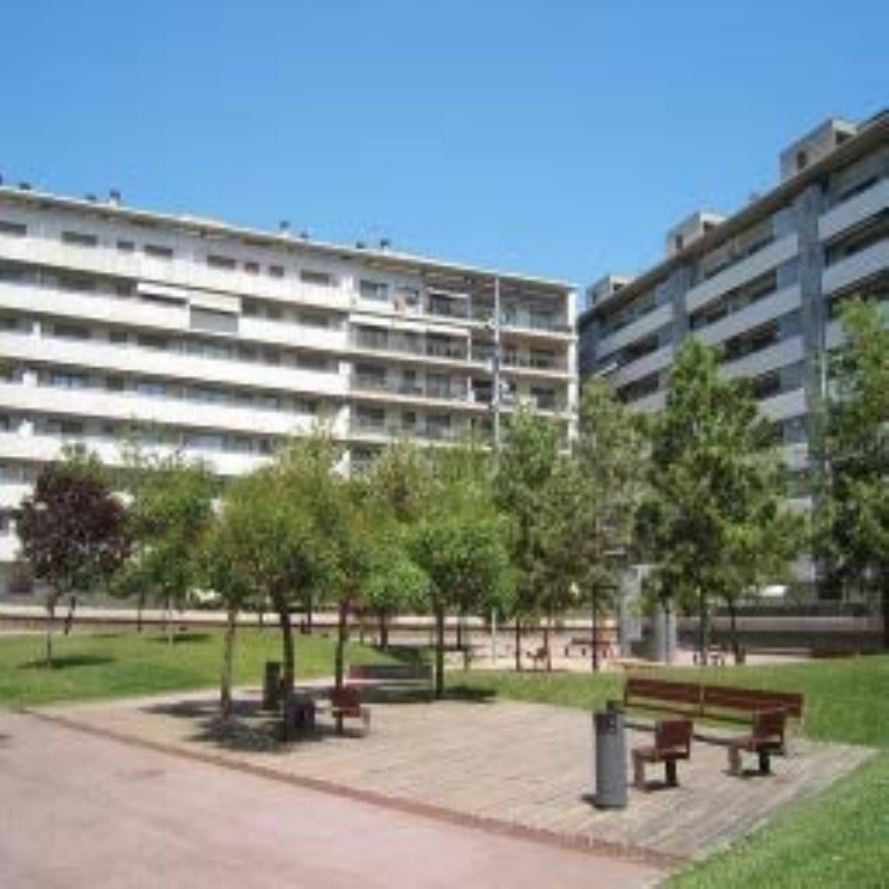}
    \end{subfigure}
    \begin{subfigure}
        \centering
        \vspace{-0.05in}
        \includegraphics[width=2.4cm]{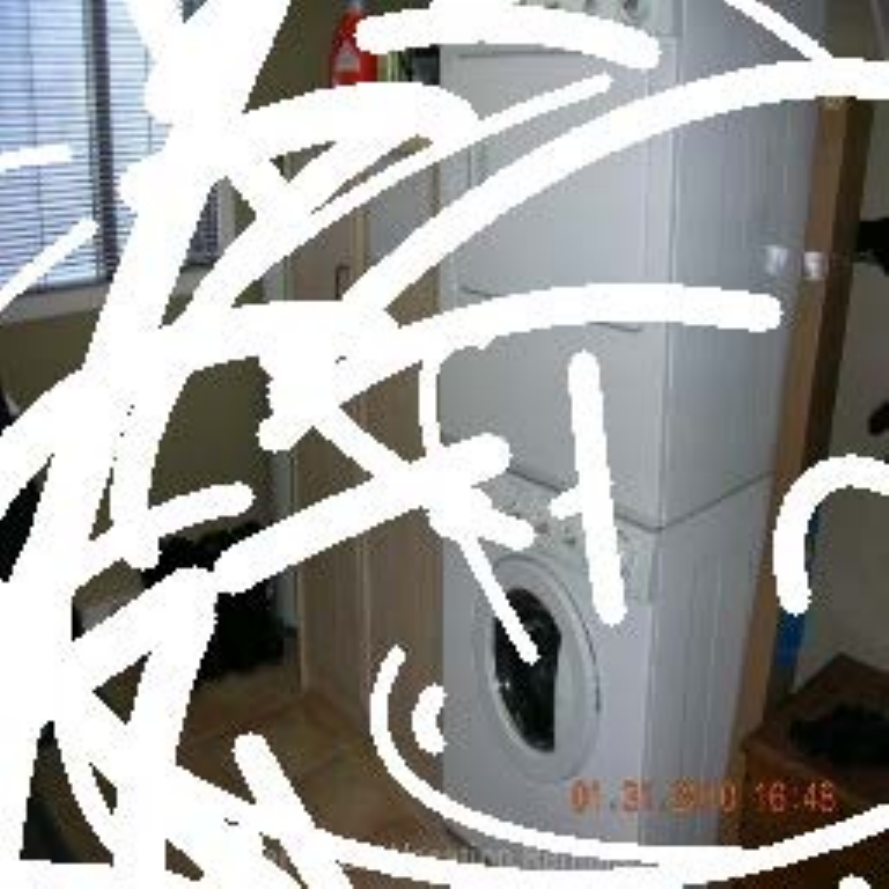}
        \includegraphics[width=2.4cm]{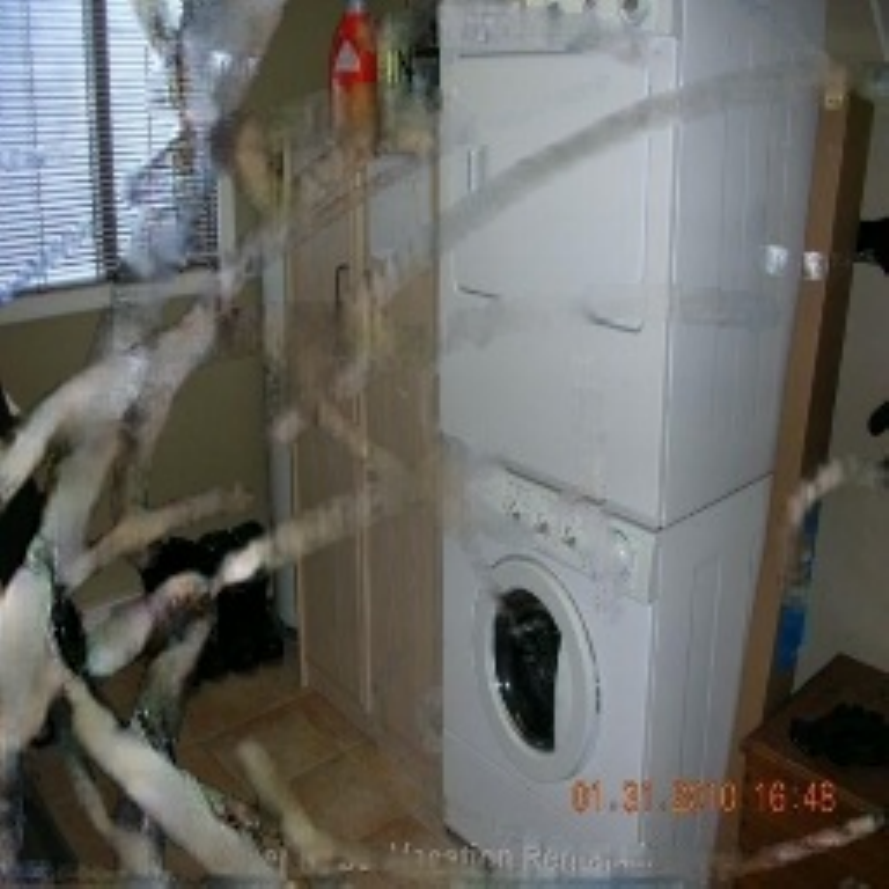}
        \includegraphics[width=2.4cm]{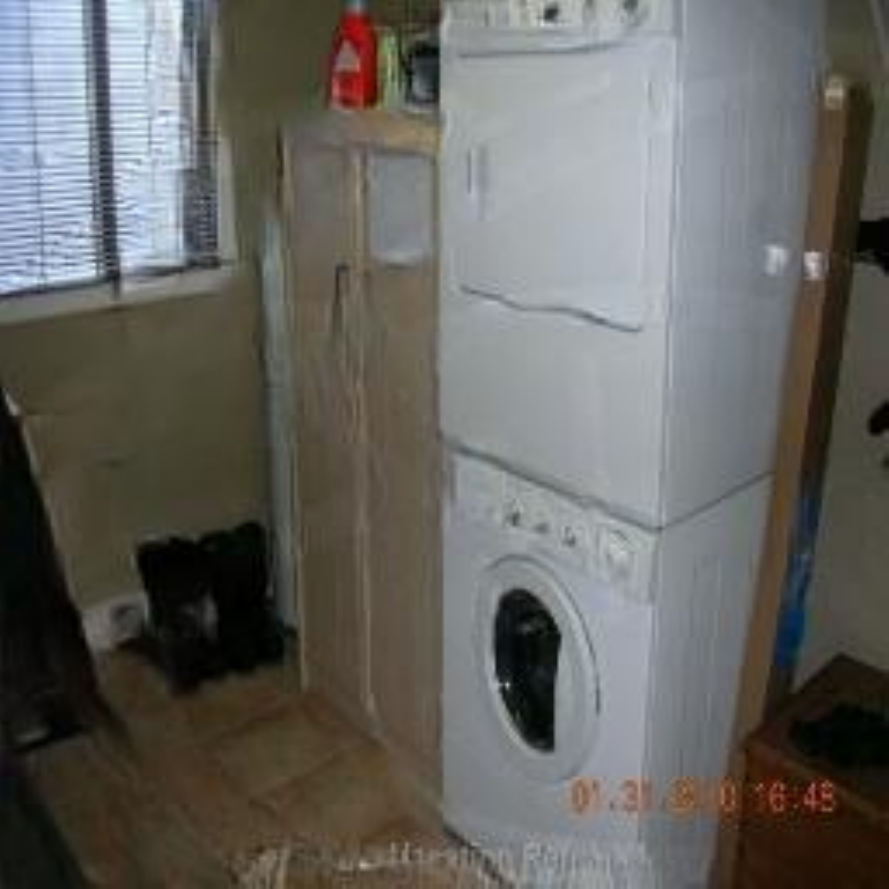}
        \includegraphics[width=2.4cm]{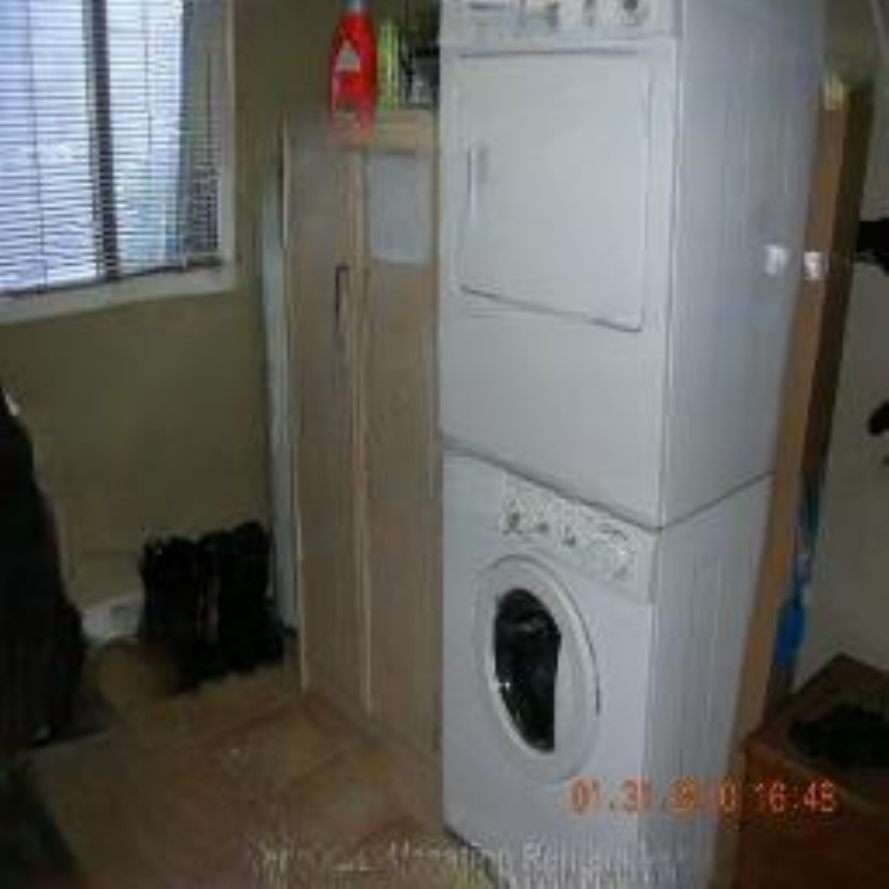}
        \includegraphics[width=2.4cm]{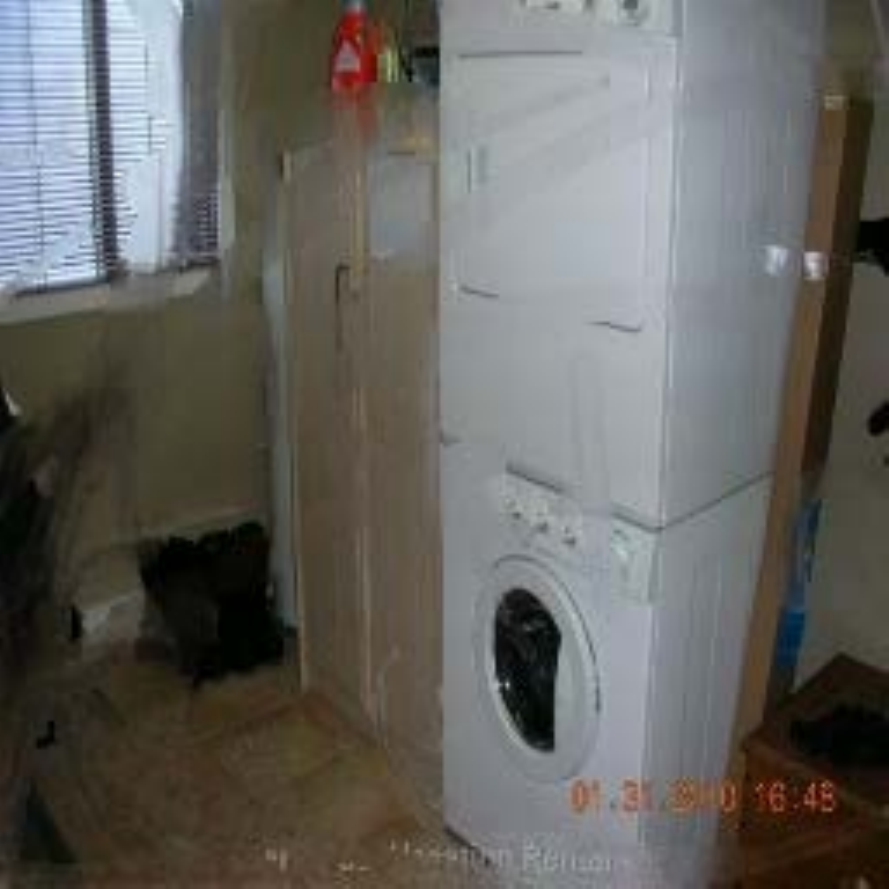}
        \includegraphics[width=2.4cm]{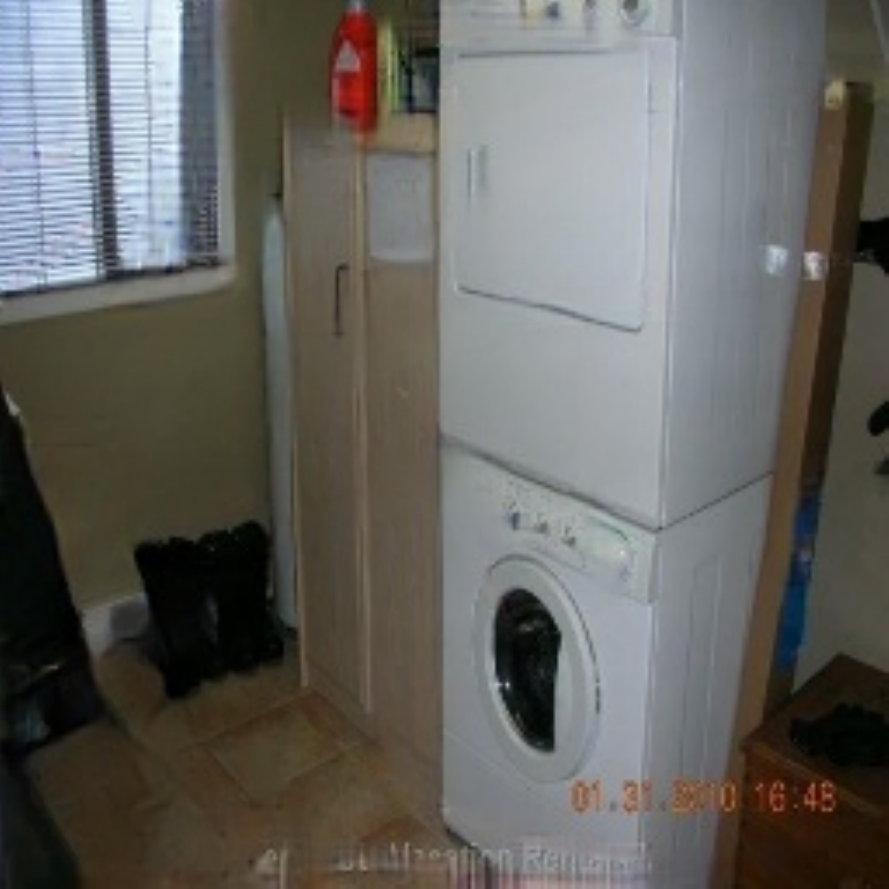}
        \includegraphics[width=2.4cm]{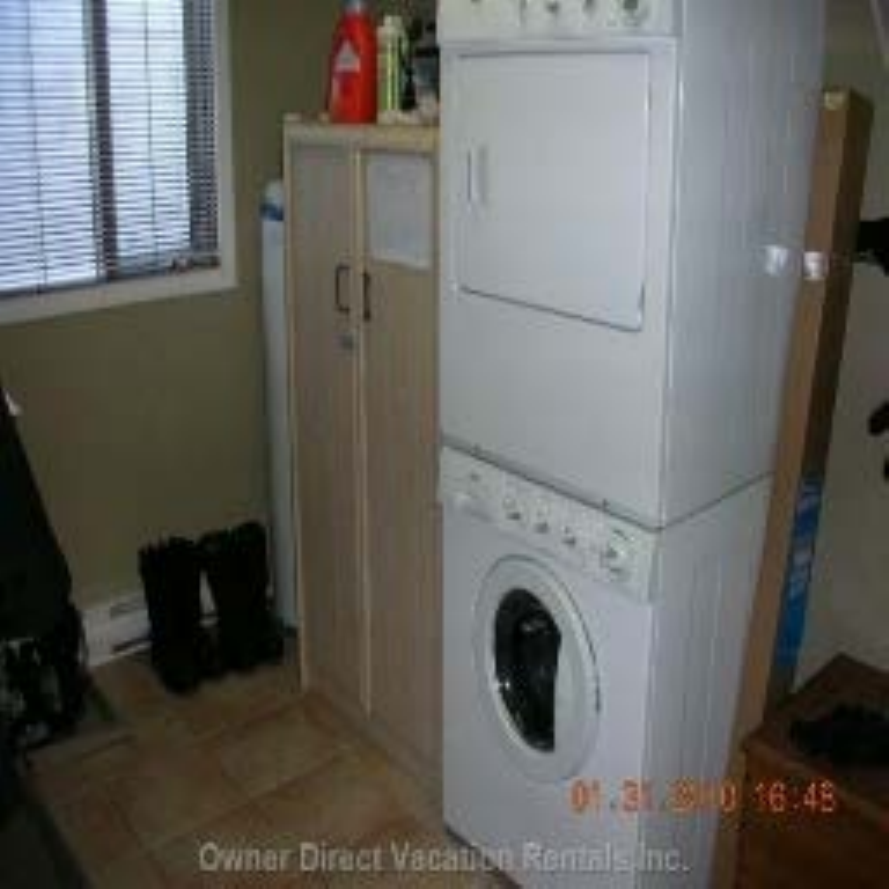}
    \end{subfigure}
    \begin{subfigure}
        \centering
        \vspace{-0.05in}
        \includegraphics[width=2.4cm]{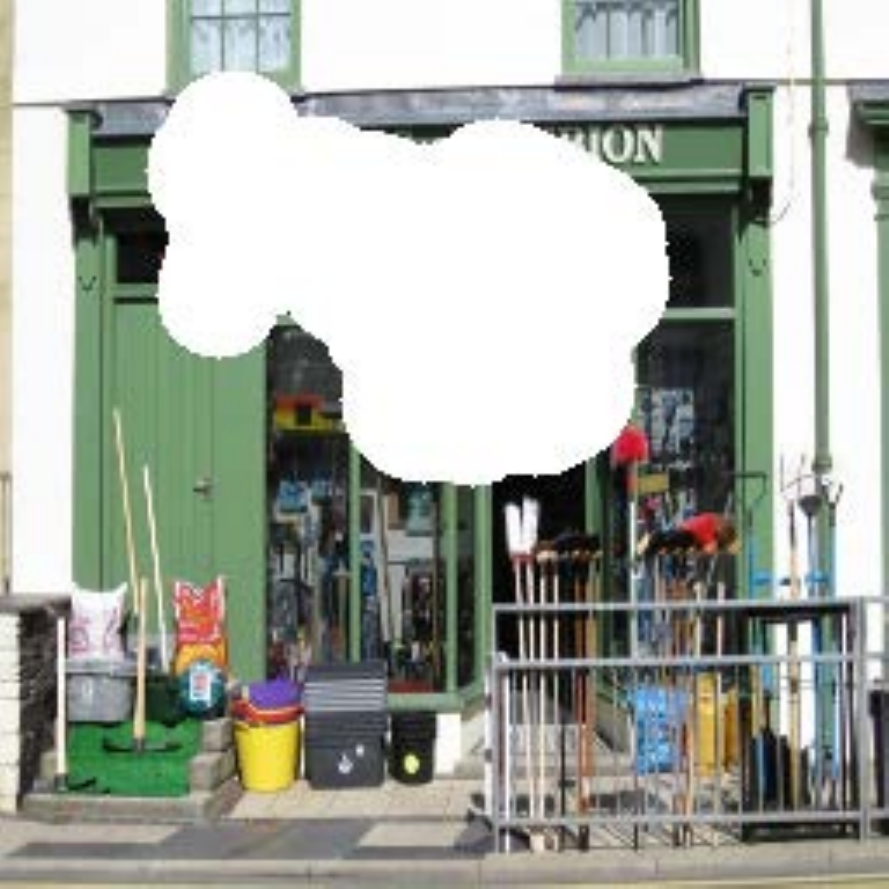}
        \includegraphics[width=2.4cm]{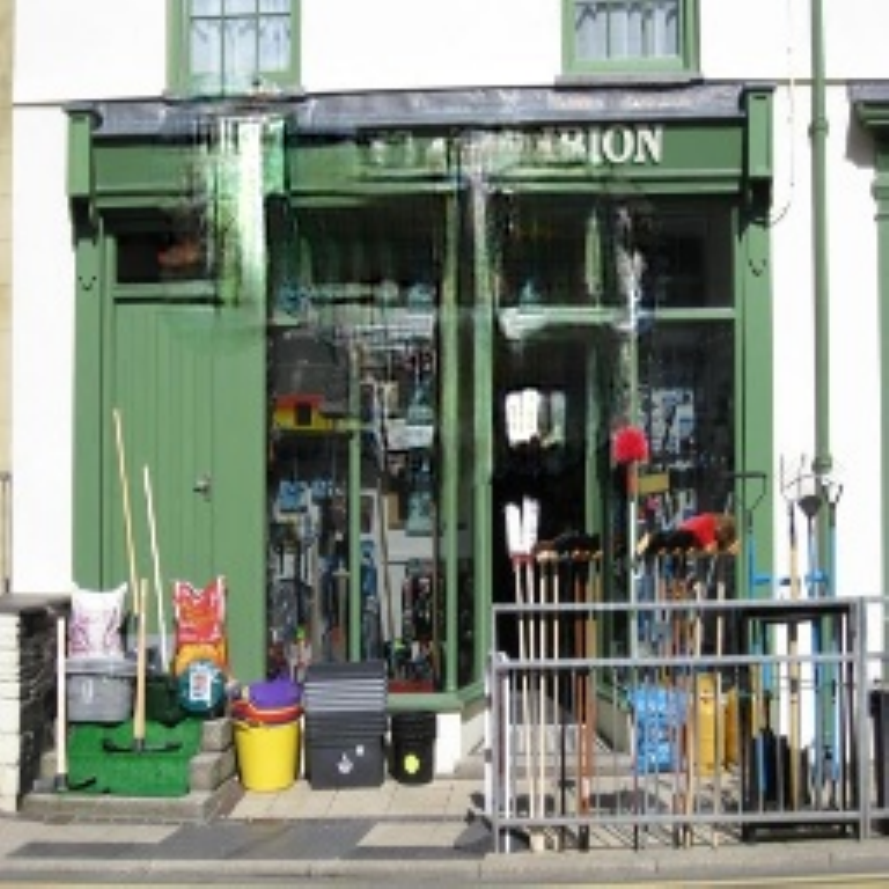}
        \includegraphics[width=2.4cm]{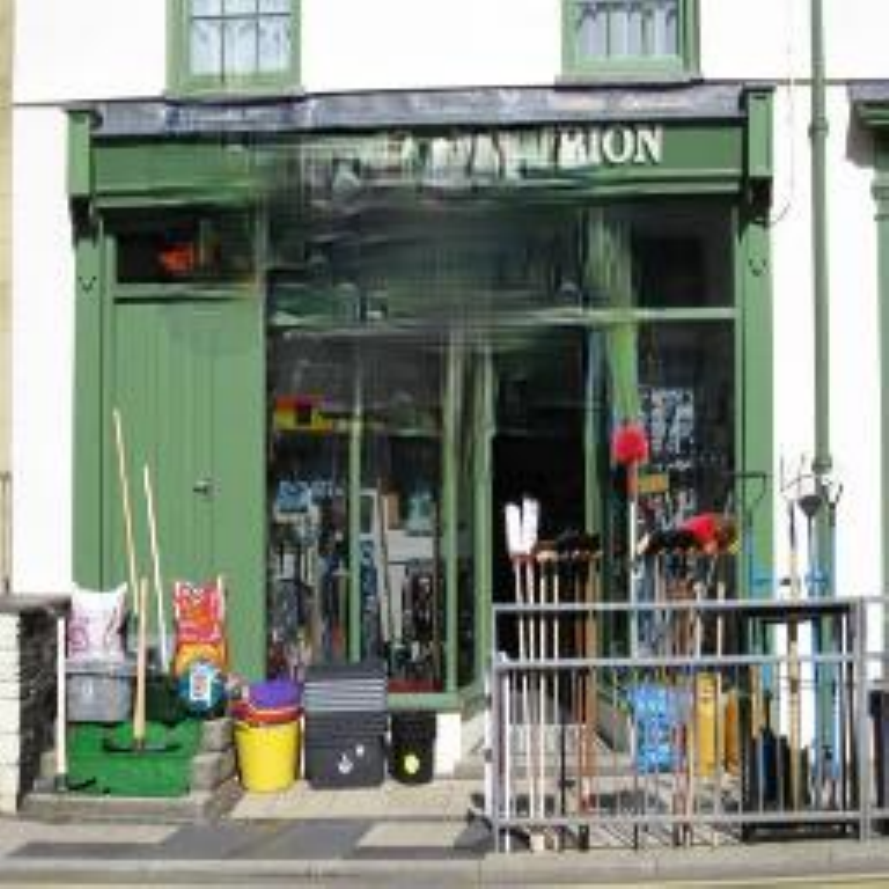}
        \includegraphics[width=2.4cm]{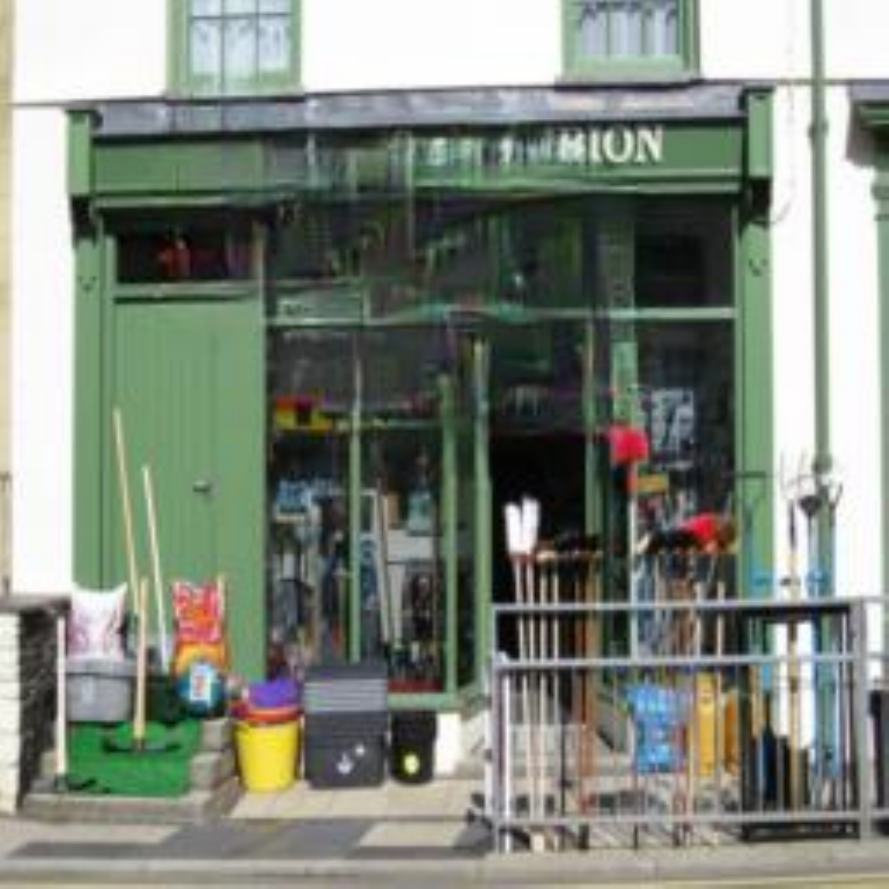}
        \includegraphics[width=2.4cm]{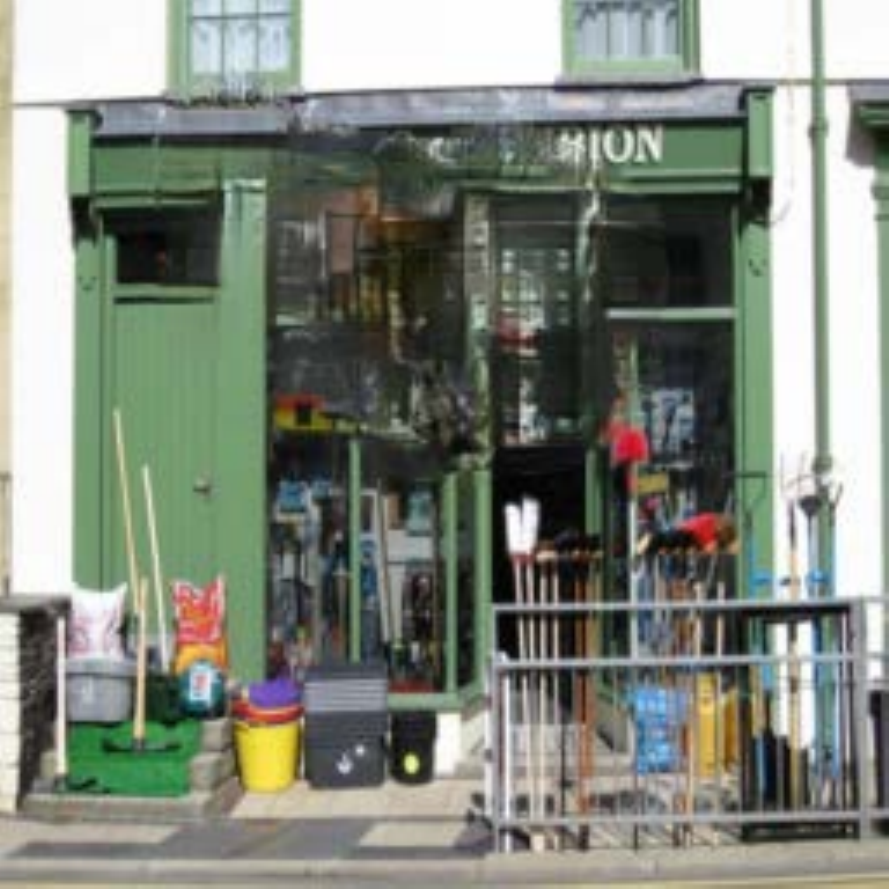}
        \includegraphics[width=2.4cm]{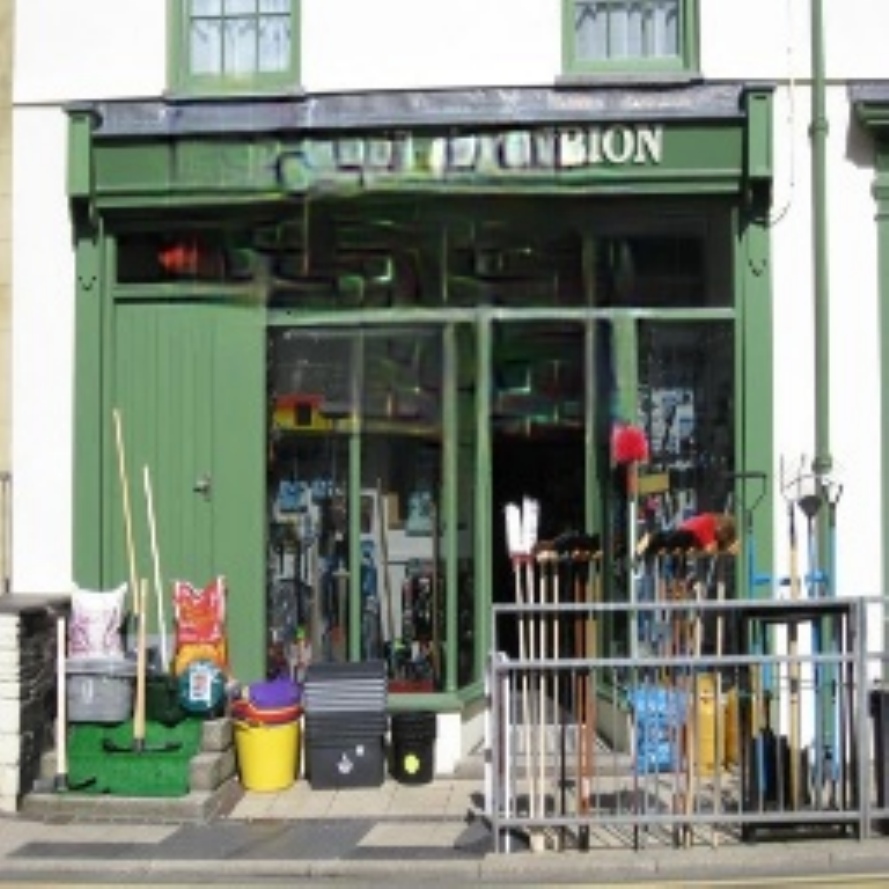}
        \includegraphics[width=2.4cm]{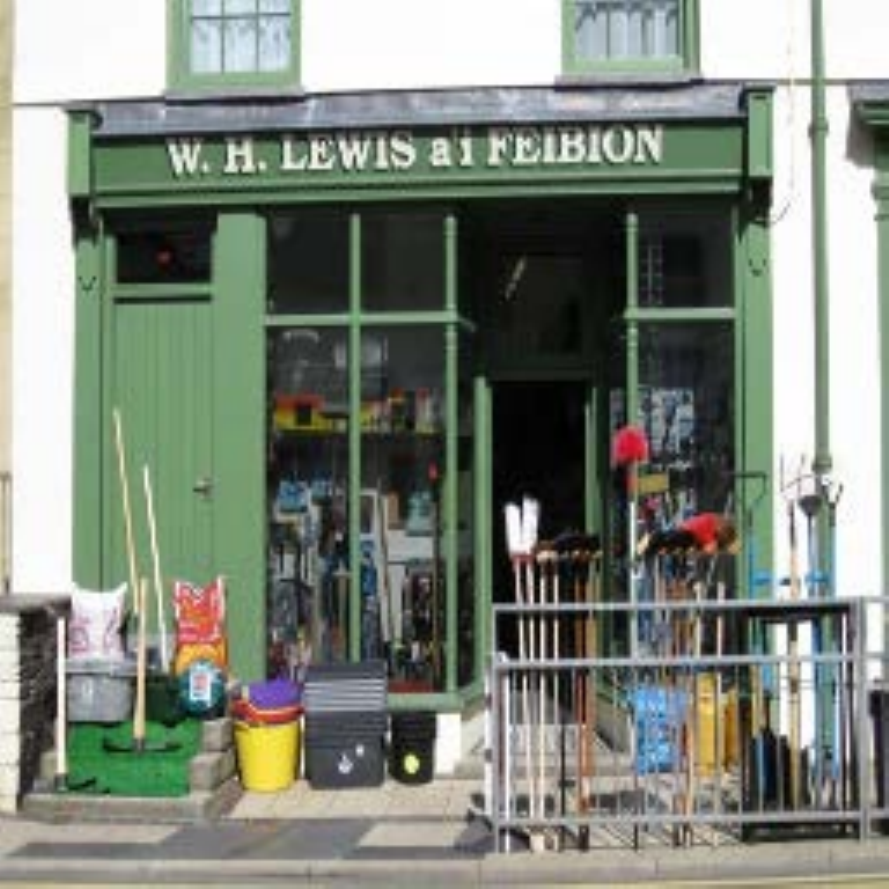}
    \end{subfigure}
    \begin{subfigure}
        \centering
        \vspace{-0.05in}
        \includegraphics[width=2.4cm]{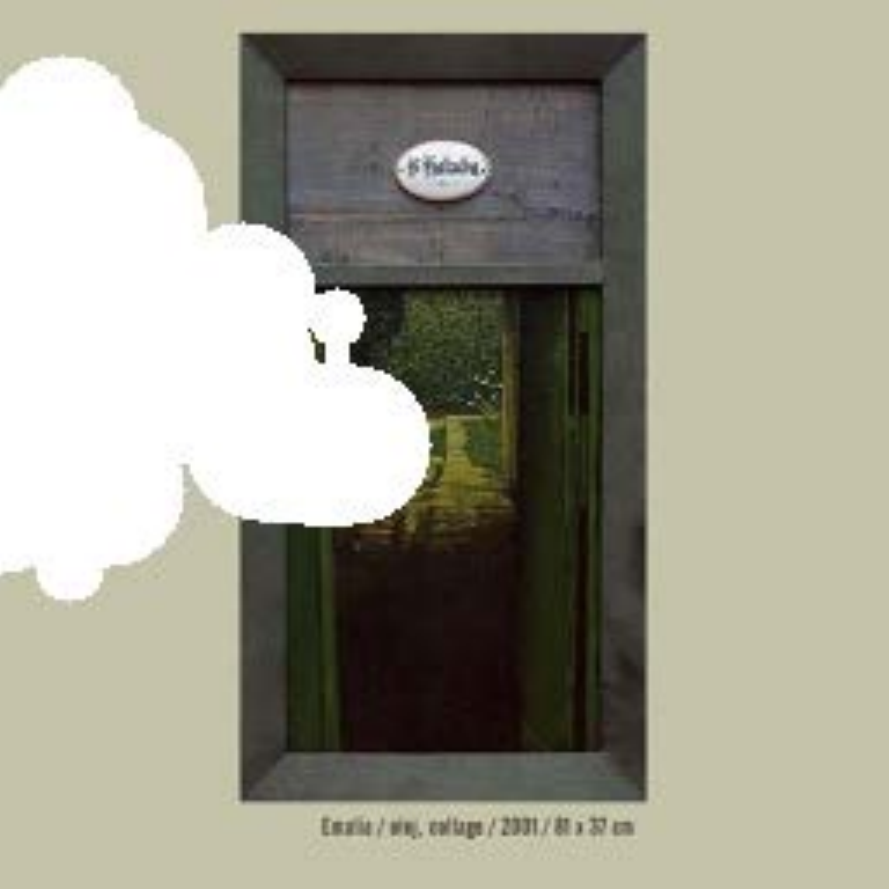}
        \includegraphics[width=2.4cm]{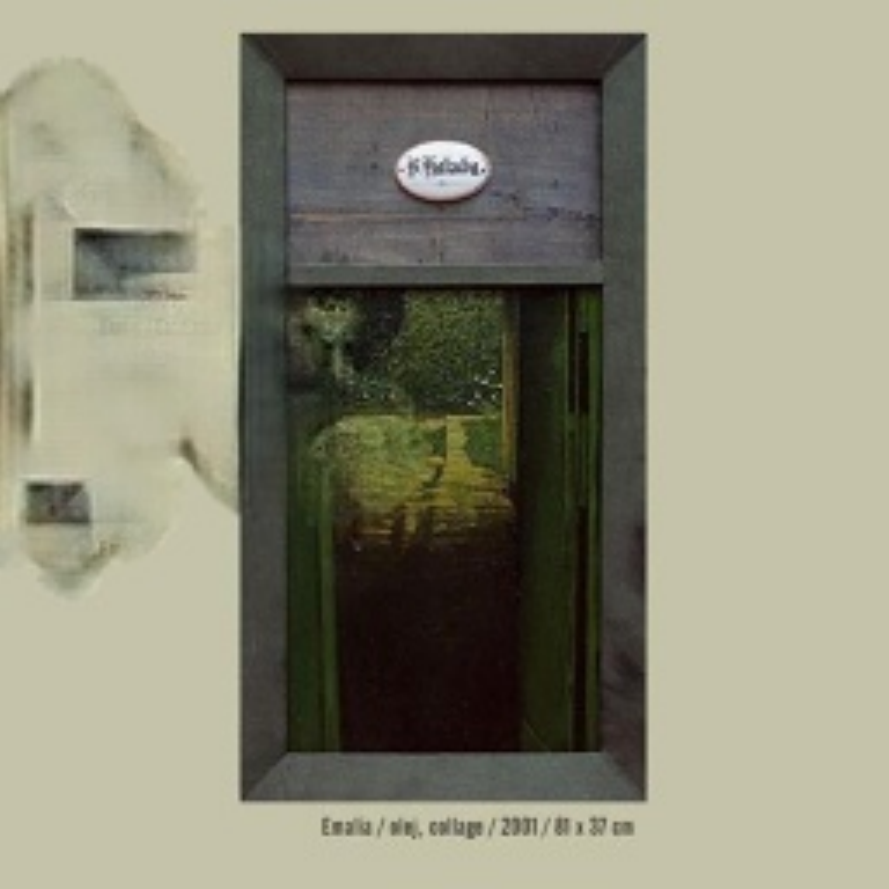}
        \includegraphics[width=2.4cm]{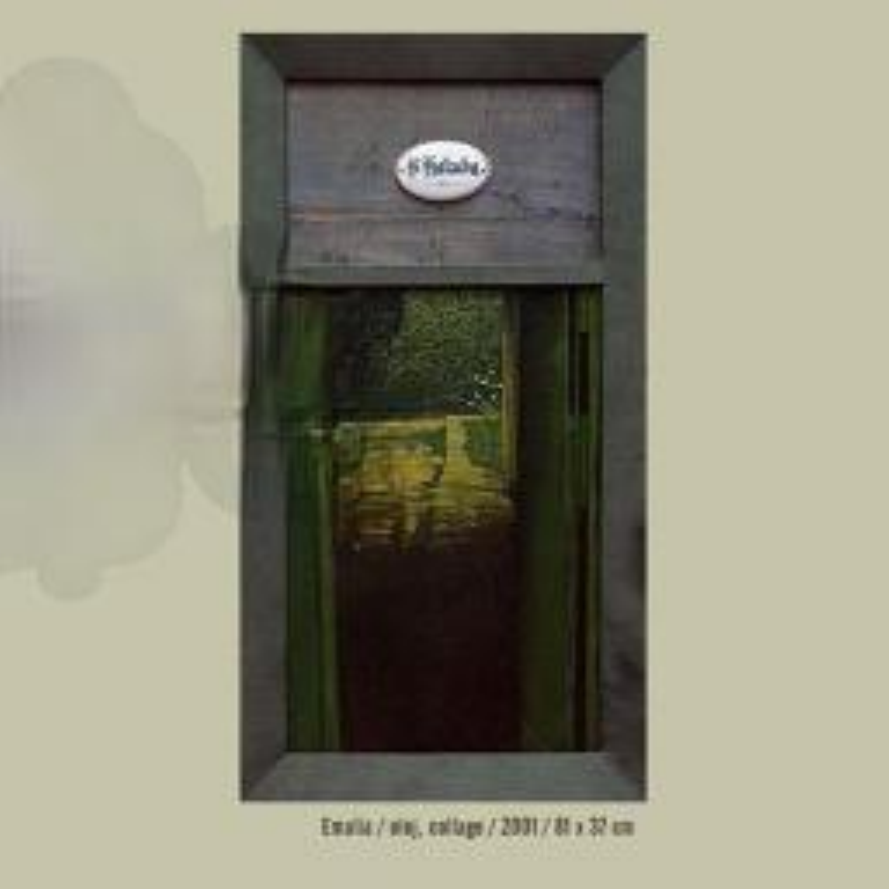}
        \includegraphics[width=2.4cm]{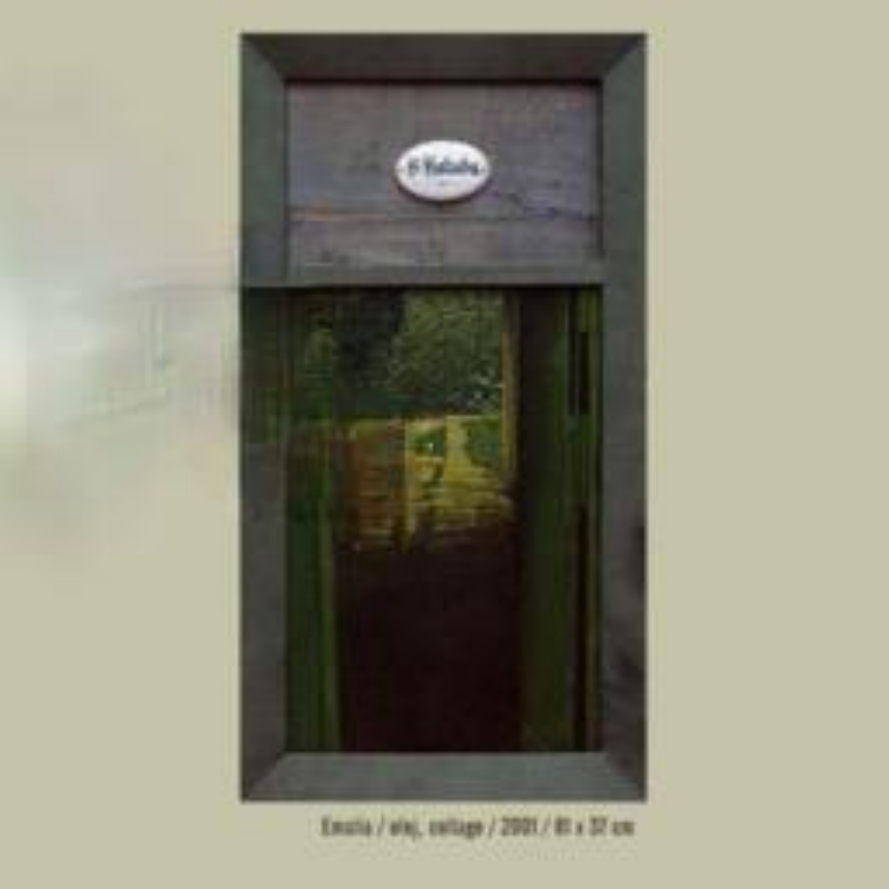}
        \includegraphics[width=2.4cm]{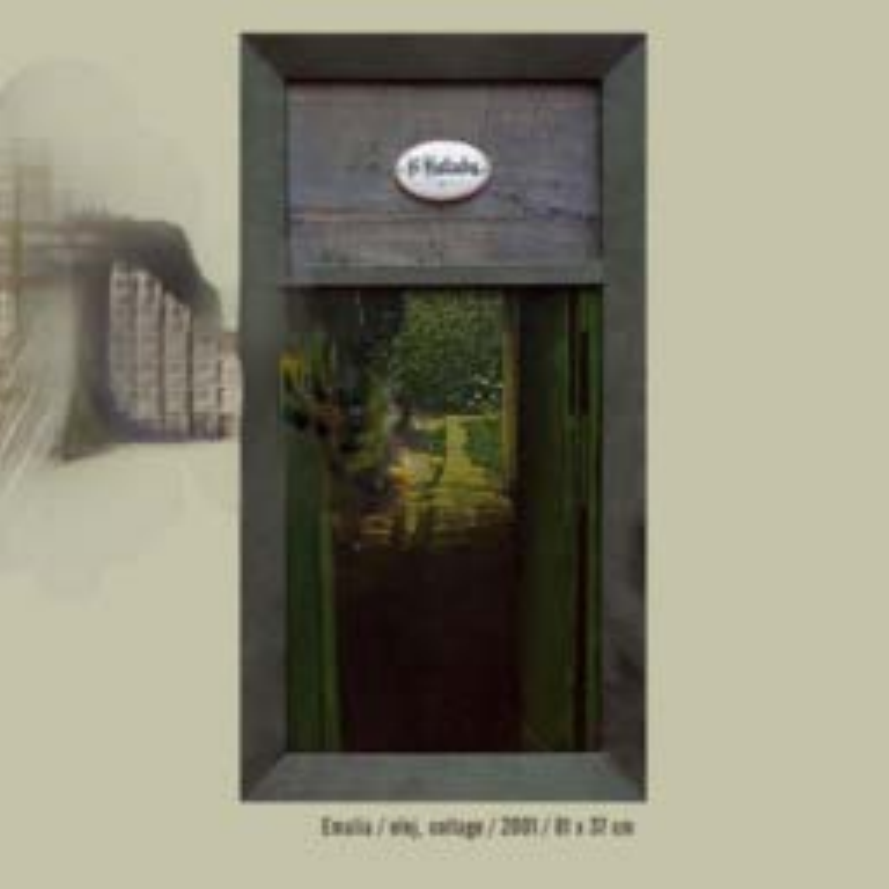}
        \includegraphics[width=2.4cm]{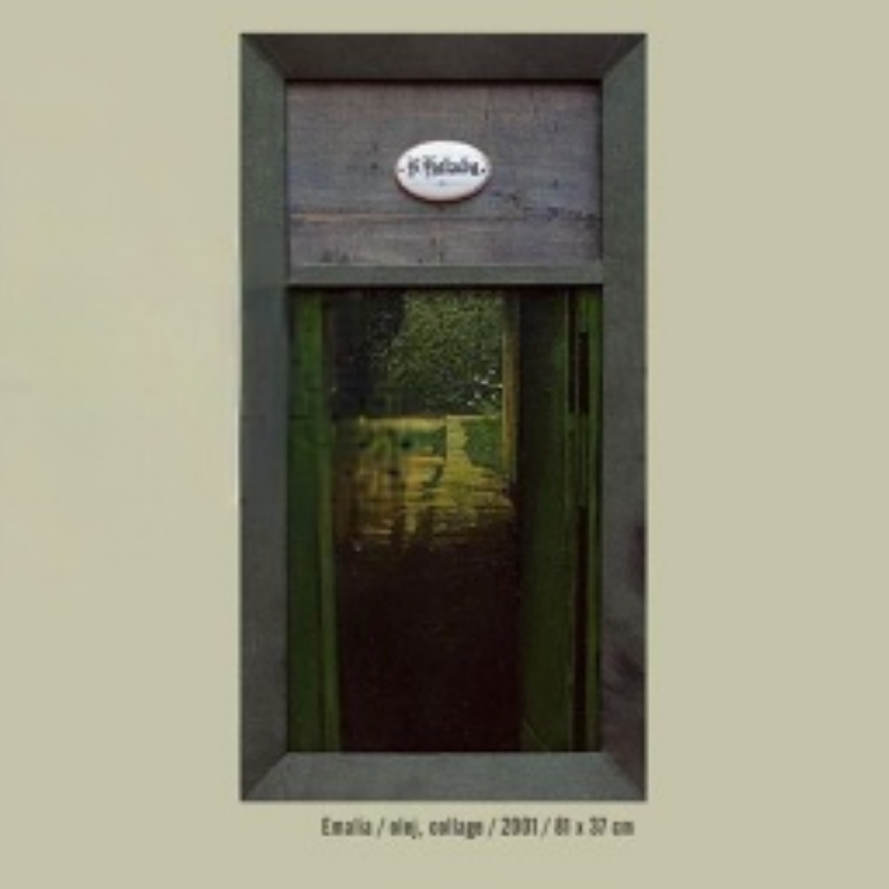}
        \includegraphics[width=2.4cm]{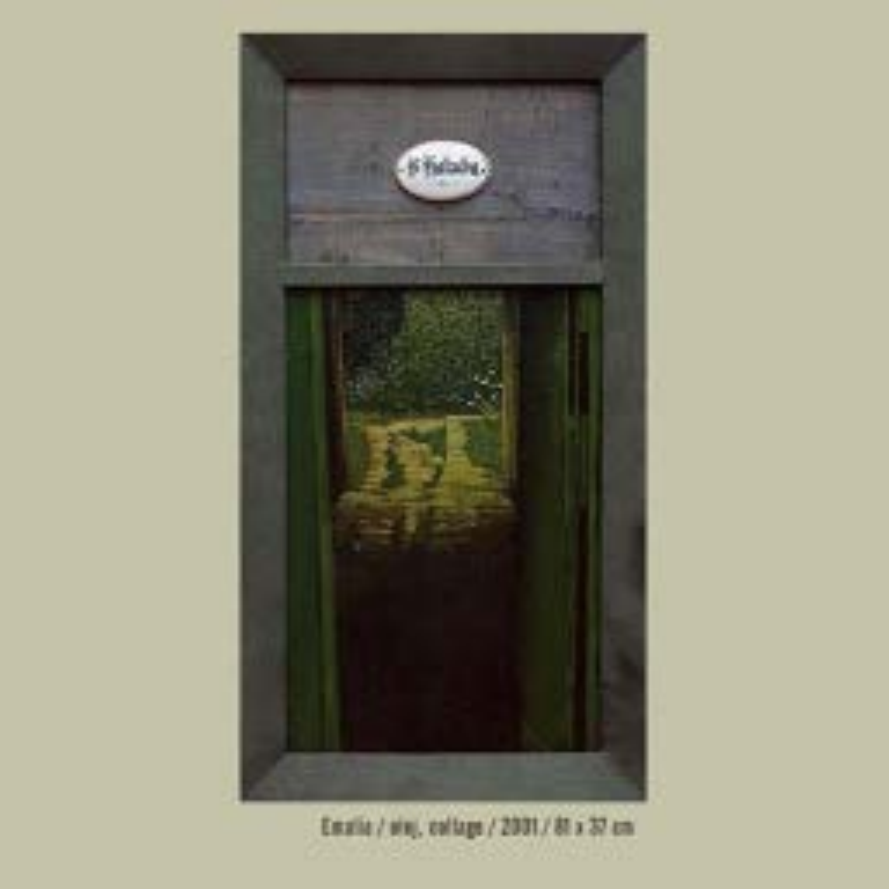}
    \end{subfigure}
    \begin{subfigure}
        \centering
        \vspace{-0.05in}
        \includegraphics[width=2.4cm]{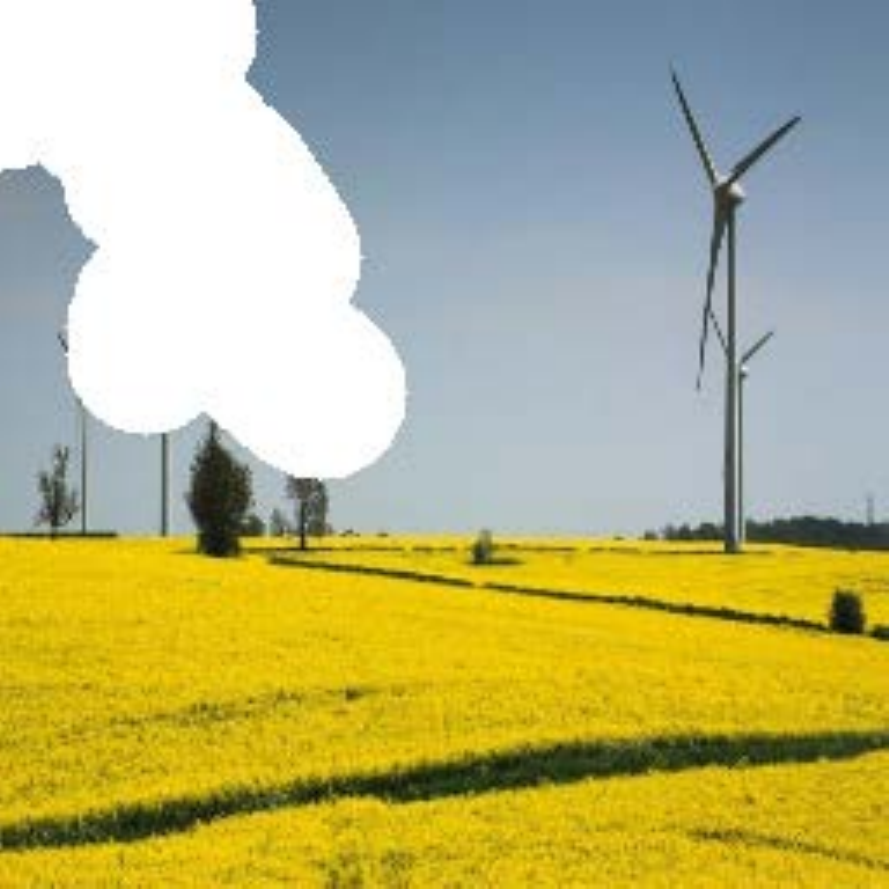}
        \includegraphics[width=2.4cm]{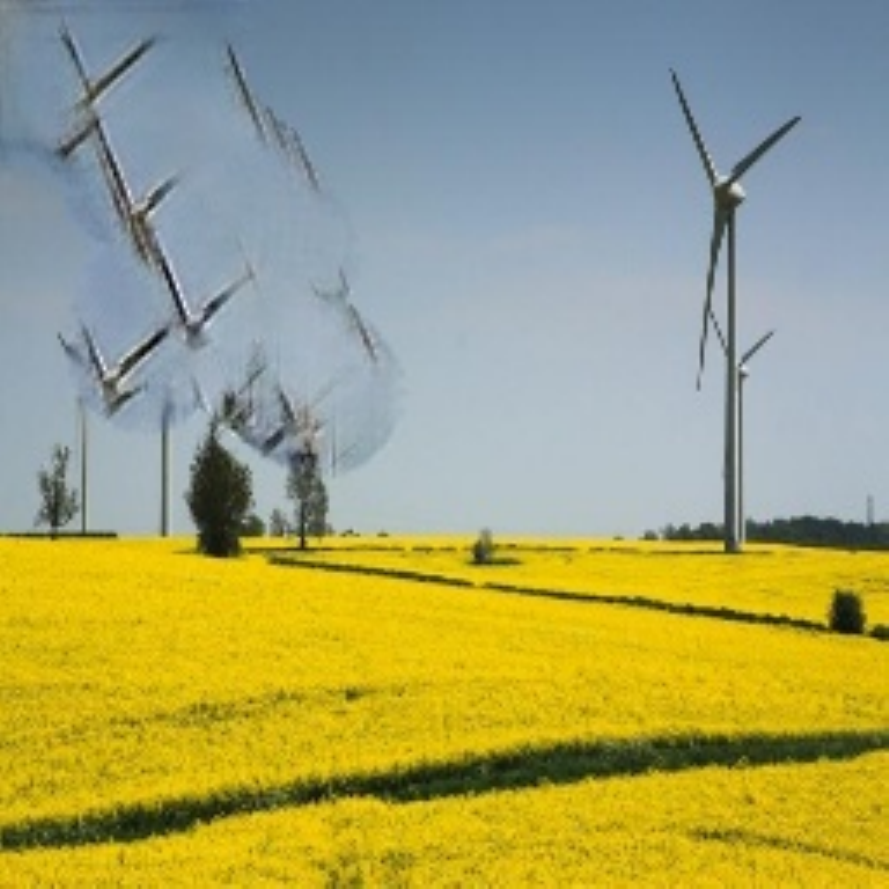}
        \includegraphics[width=2.4cm]{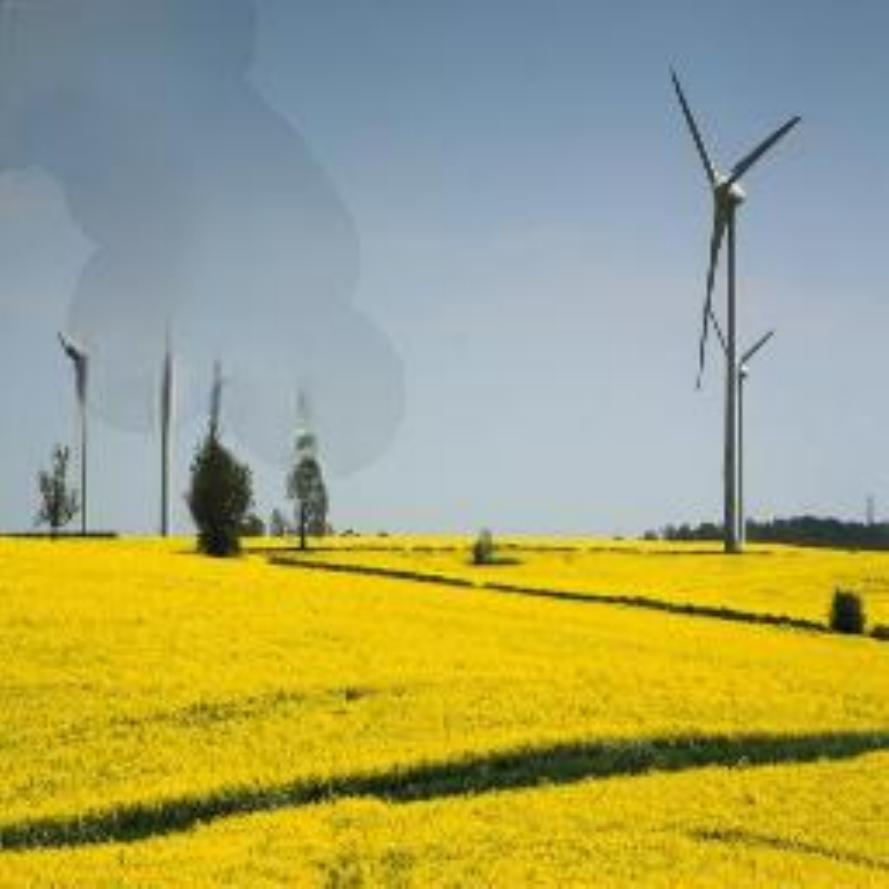}
        \includegraphics[width=2.4cm]{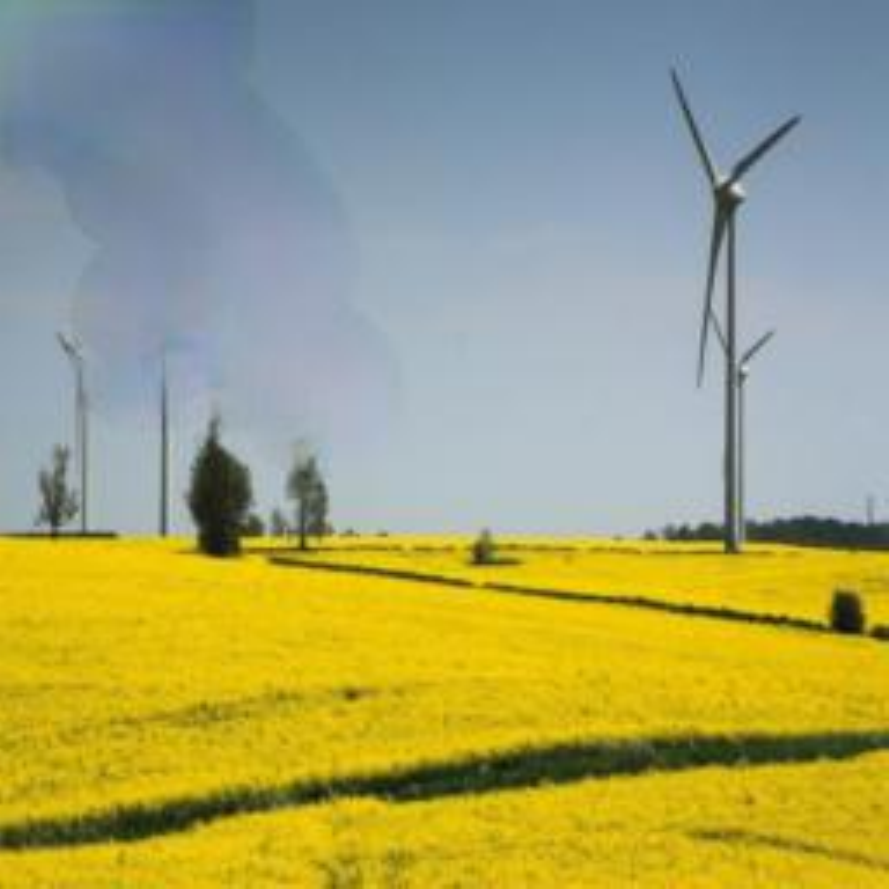}
        \includegraphics[width=2.4cm]{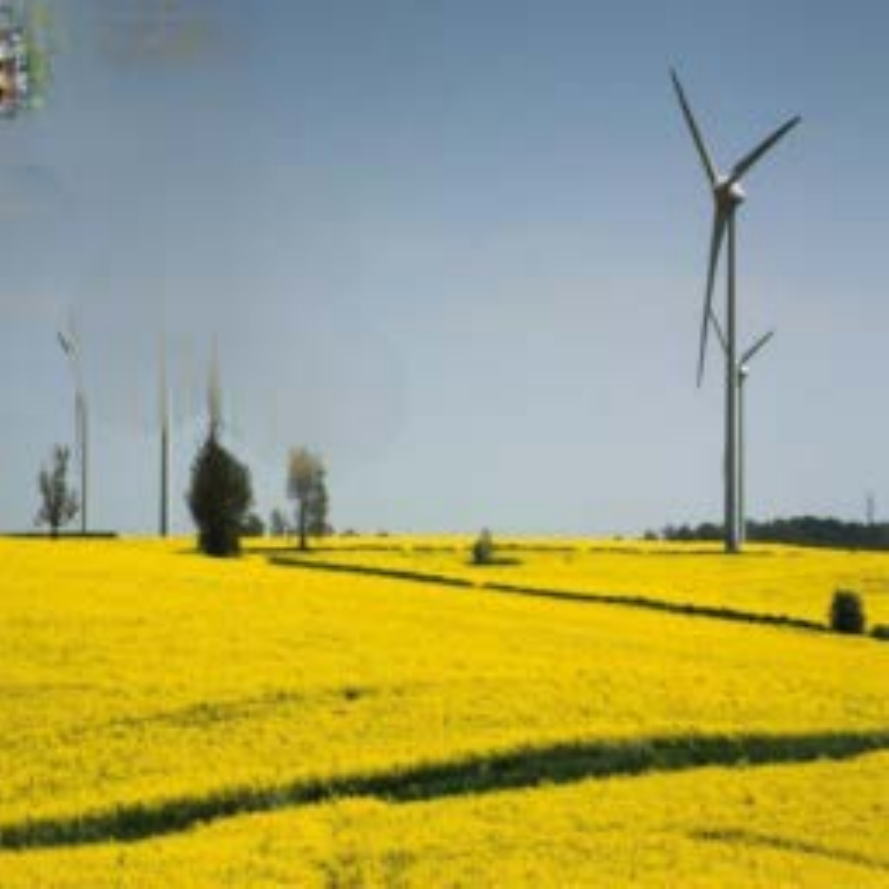}
        \includegraphics[width=2.4cm]{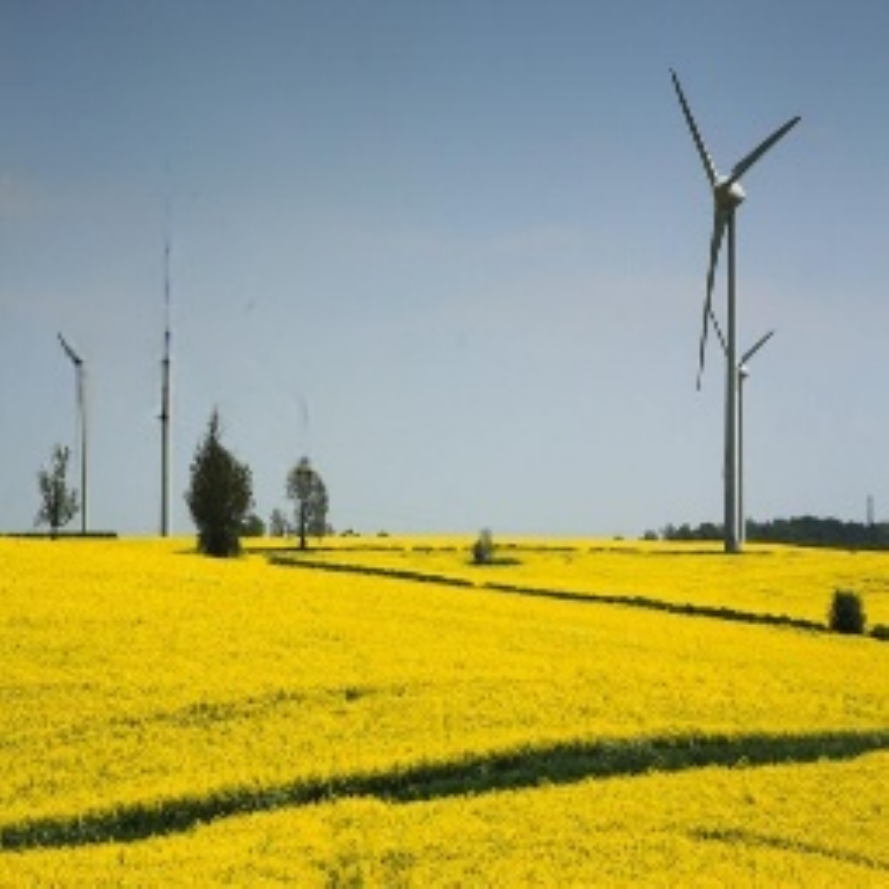}
        \includegraphics[width=2.4cm]{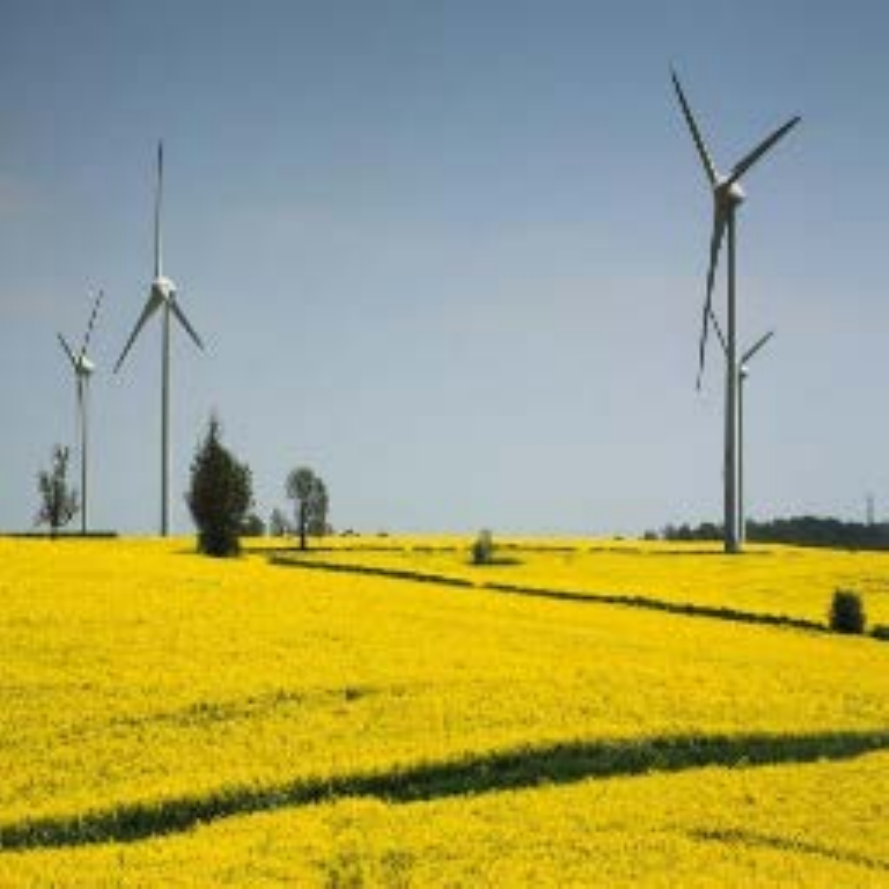}
    \end{subfigure}
    \centering
    \vspace{-0in}
    \\
    \centering
    (a) Input   \hspace{1.0cm}
    (b) CA\cite{yu2018generative}      \hspace{0.7cm}
    (c) Pconv\cite{liu2018image}  \hspace{0.6cm}
    (d) EC\cite{nazeri2019edgeconnect}      \hspace{0.6cm}
    (e) PIC\cite{Zheng2019Pluralistic}     \hspace{0.8cm}
    (f) Ours    \hspace{1.2cm}
    (g) GT      \hspace{0.5cm}
    \caption{Qualitative comparisons between different methods on places2.}
    \label{fig:new comparisons on Places2}
\end{figure*}

Figure \ref{fig:new comparisons on CelebA-HQ}, \ref{fig:new comparisons on Paris} and \ref{fig:new comparisons on Places2} show the inpainting results of different methods on several examples from CelebA-HQ, Paris StreetView, Places2, respectively, where ``GT'' stands for the ground truth images. We compare all the models both on the discontiguous and contiguous missing regions, to prove the superior generalization ability of our method in different scenarios. All the reported results are the direct outputs from trained models without using any post-processing. 

From Figure \ref{fig:new comparisons on CelebA-HQ}, we can see that CA brings strong distortions in the inpaiting images, while PConv, EC and PIC can recover the semantic information for the missing irregular regions in most cases, but still face obvious deviations from the ground truth. EC performs well when discontiguous missing regions occur, but also fails to infer the correct edge information for large holes, even filling some inappropriate semantic content into the missing regions, such as the eye-like content shown in the fourth row of Figure \ref{fig:new comparisons on CelebA-HQ}(d). For either discontiguous or contiguous missing regions, PIC could relatively better restore the missing regions on the faces. But it cannot handle the surrounding areas without distinguishing their semantic differences. Among all the methods, we can observe that our model can recover the incomplete images with more natural contents in the missing regions, i.e., the structure and detailed information for faces, which looks more consistent with existing regions and much closer to the ground truth. %For Places2 dataset, it is not clear to see the difference inside the filling region, so we zoom in on certain region to show the details. 

Similarly, for the natural scene images, as shown in Figure \ref{fig:new comparisons on Paris} and \ref{fig:new comparisons on Places2}, we can get close conclusions as that in Figure \ref{fig:new comparisons on CelebA-HQ}. For example, CA still suffers from the heavy distortions, while Pconv and EC face the inconsistency and blurry problems in the filled contents. However, here the performance of PIC shows obvious degradation, since it can hardly infer the appropriate information. The phenomenon seems more obvious on Paris dataset, which contains more complicated structure. This is mainly because it is unlikely to well approximate the distribution of complete image only guided KL divergence or adversarial loss. Different from these methods, our method can well address the severe issues with the region-wise generative adversarial learning, and thus stably generates the satisfying results on scene dataset. The superior performance further proves that our method is powerful for the generic image inpainting task.

\begin{figure}[tp!]
    \centering
    \vspace{-0in}
    \begin{subfigure}
        \centering
        \includegraphics[width=2.4cm]{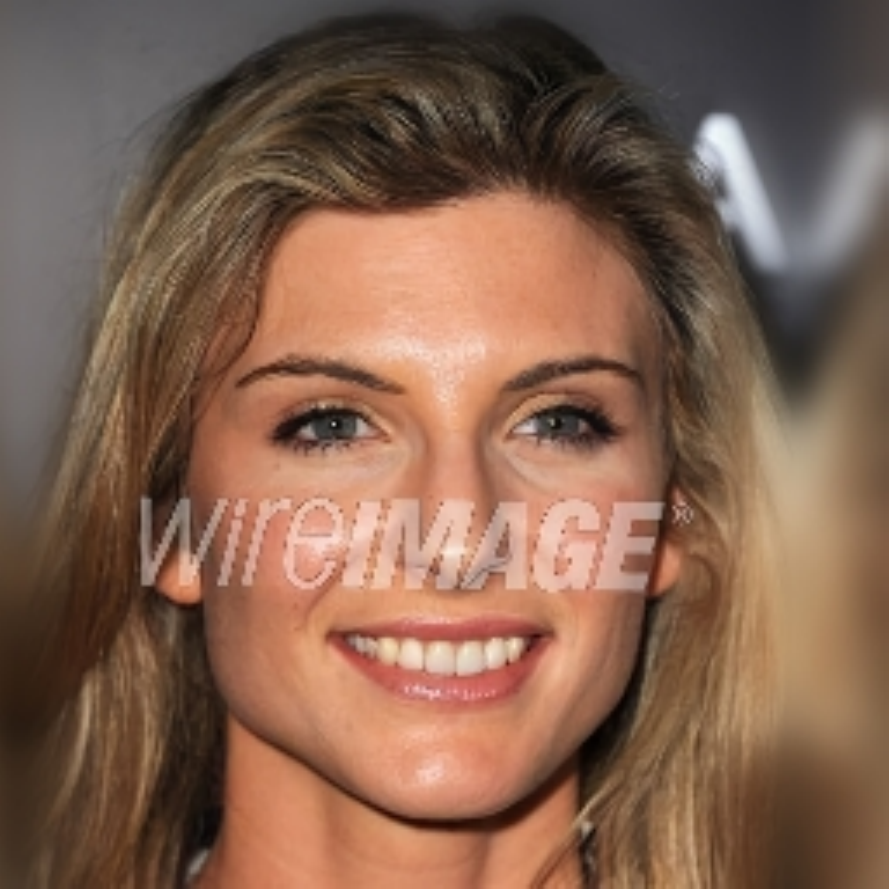}
        \includegraphics[width=2.4cm]{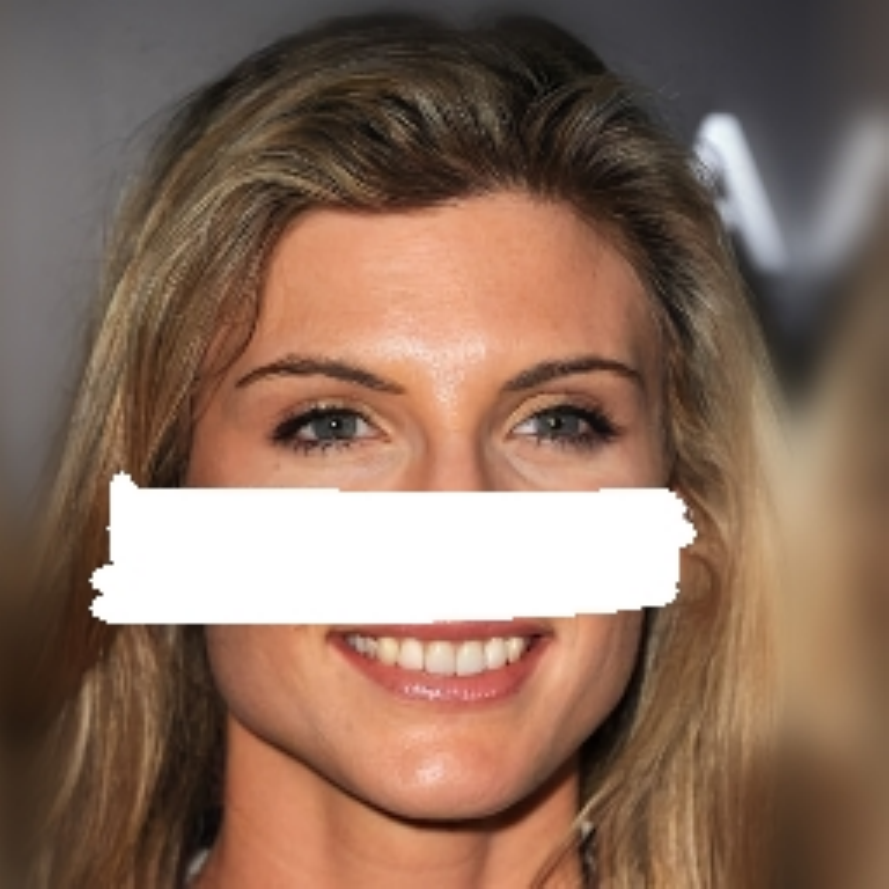}
        \includegraphics[width=2.4cm]{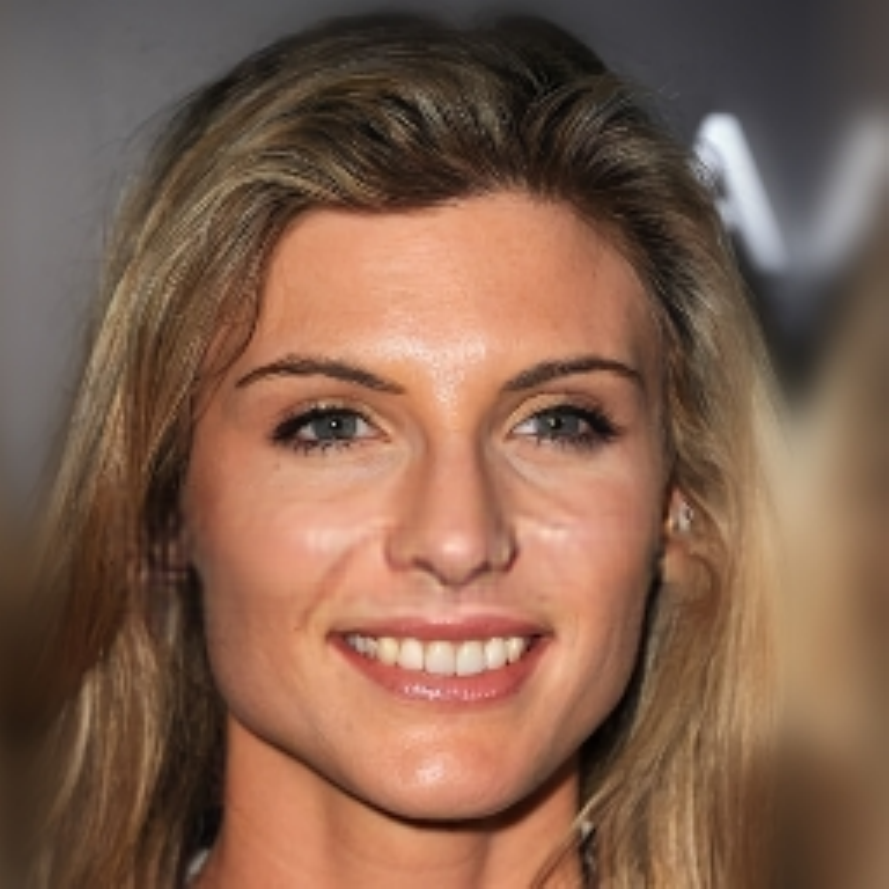}
    \end{subfigure}
    \begin{subfigure}
        \centering
        \vspace{-0.05in}
        \includegraphics[width=2.4cm]{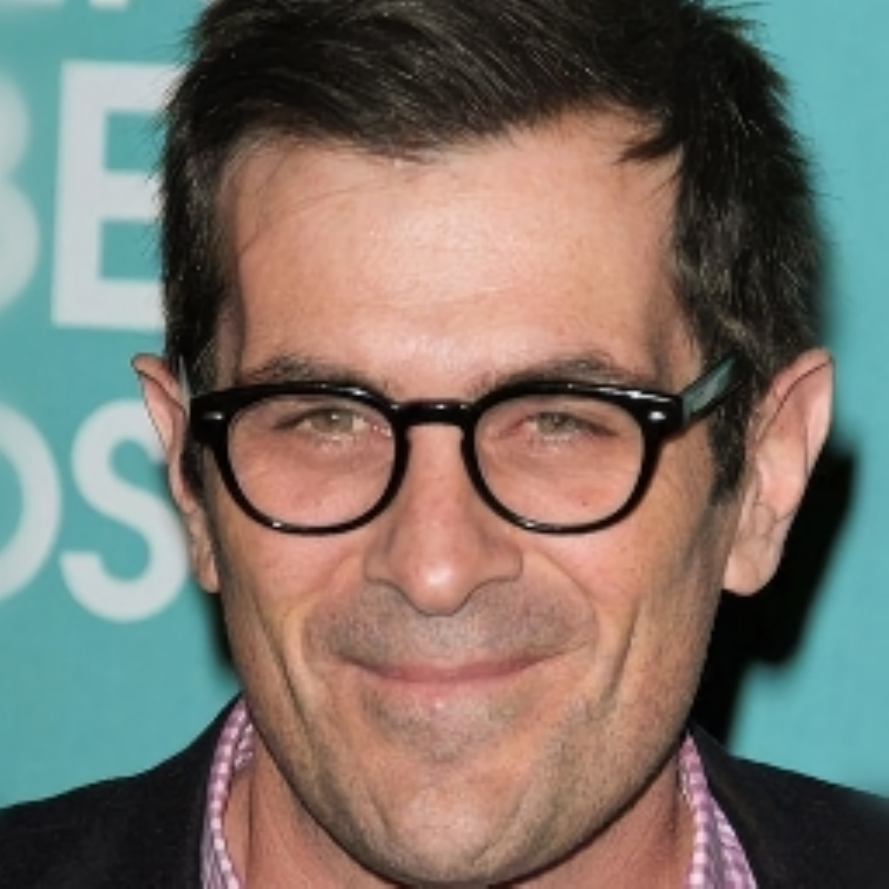}
        \includegraphics[width=2.4cm]{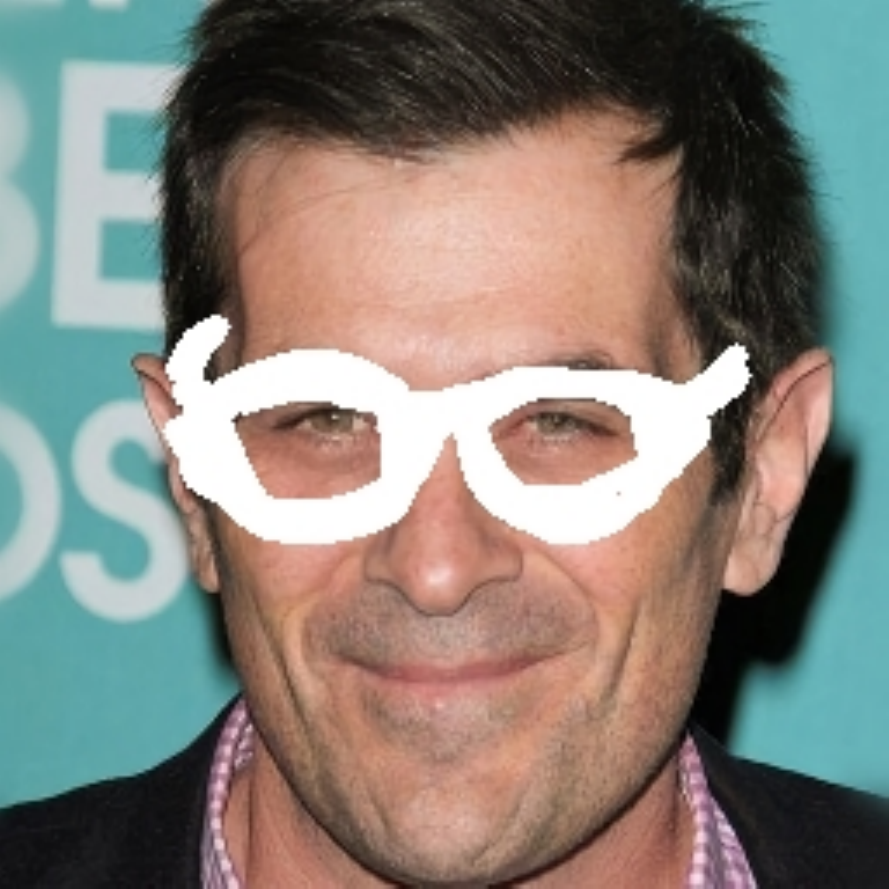}
        \includegraphics[width=2.4cm]{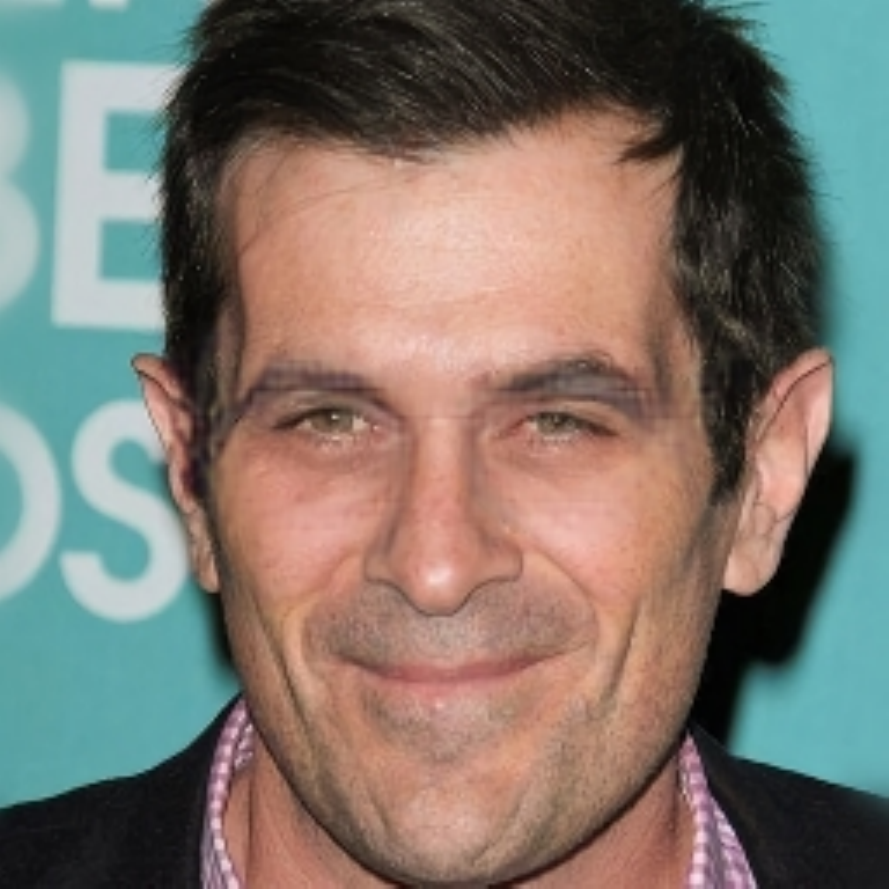}
    \end{subfigure}
    \begin{subfigure}
        \centering
        \vspace{-0.05in}
        \includegraphics[width=2.4cm]{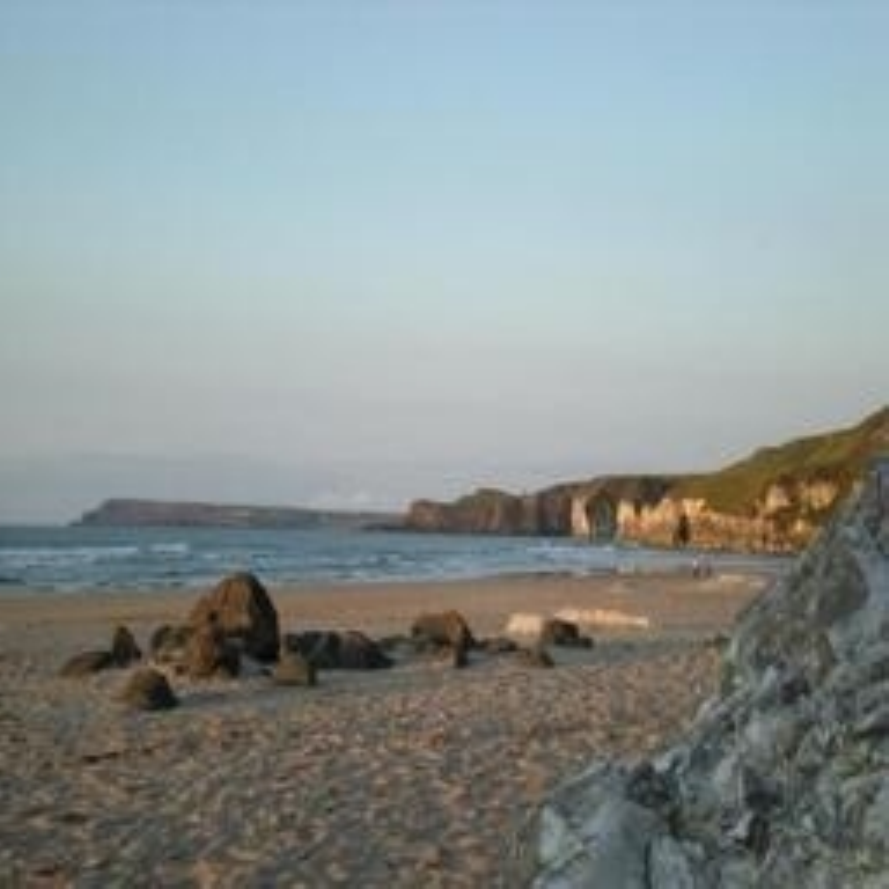}
        \includegraphics[width=2.4cm]{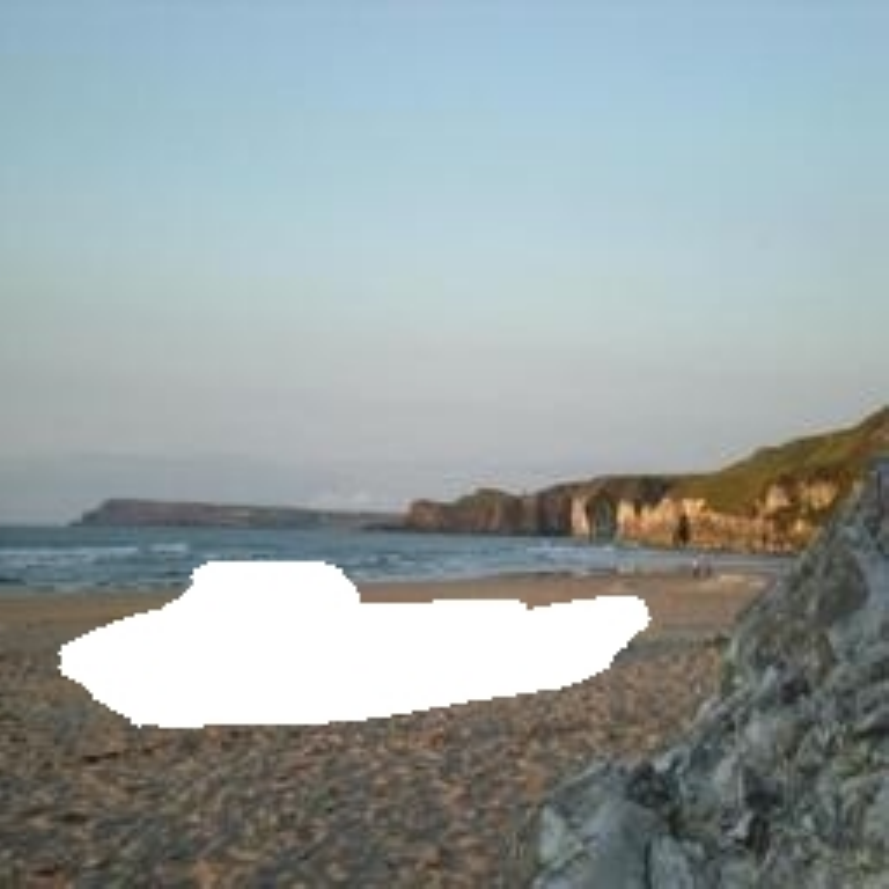}
        \includegraphics[width=2.4cm]{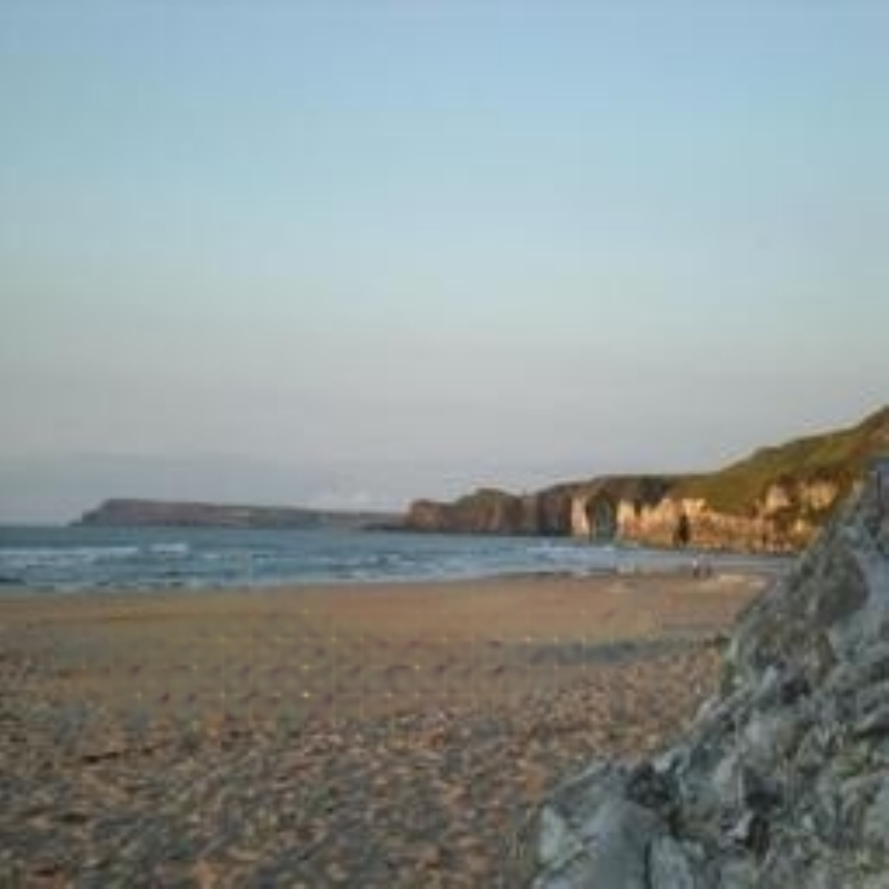}
    \end{subfigure}
    \begin{subfigure}
        \centering
        \vspace{-0.05in}
        \includegraphics[width=2.4cm]{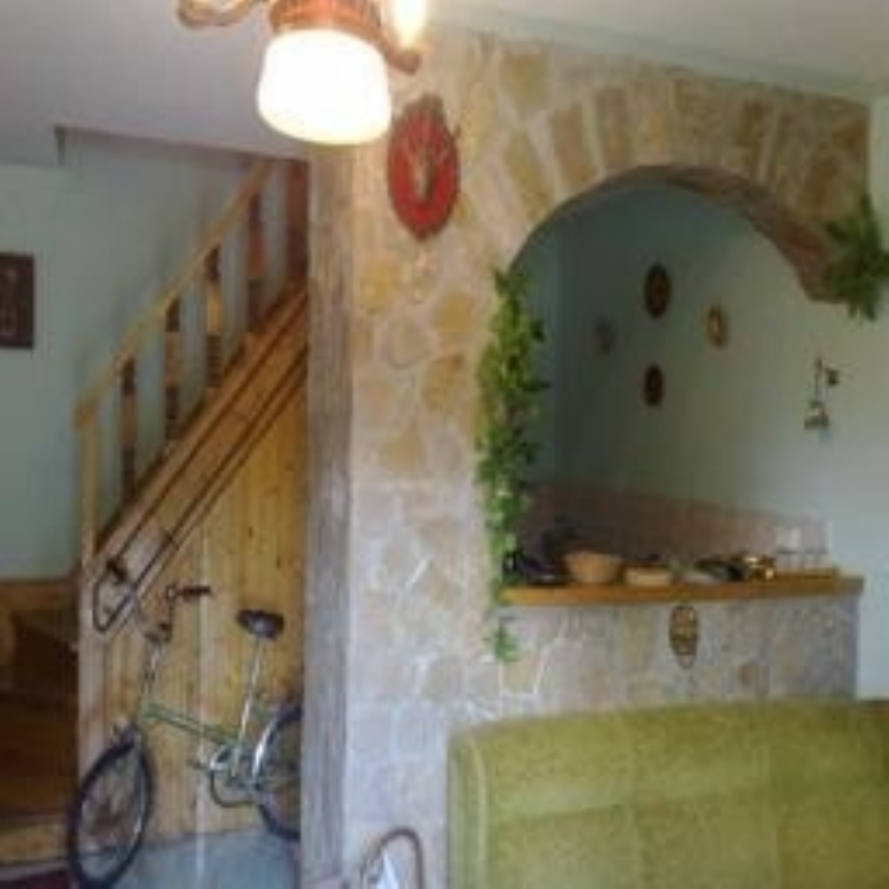}
        \includegraphics[width=2.4cm]{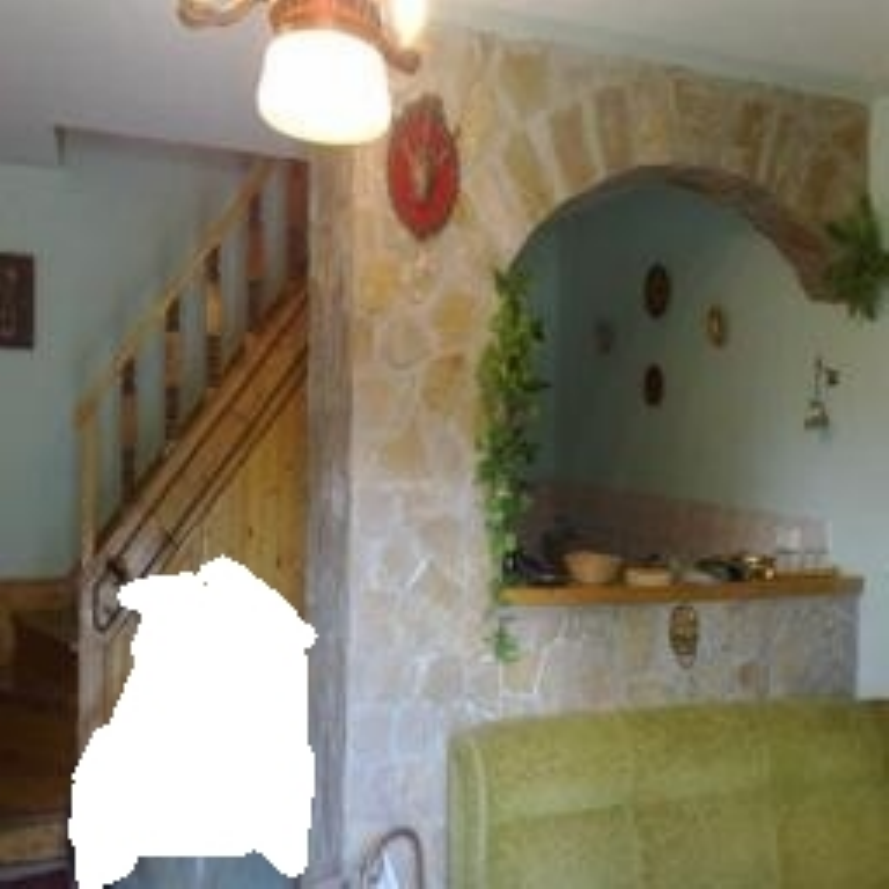}
        \includegraphics[width=2.4cm]{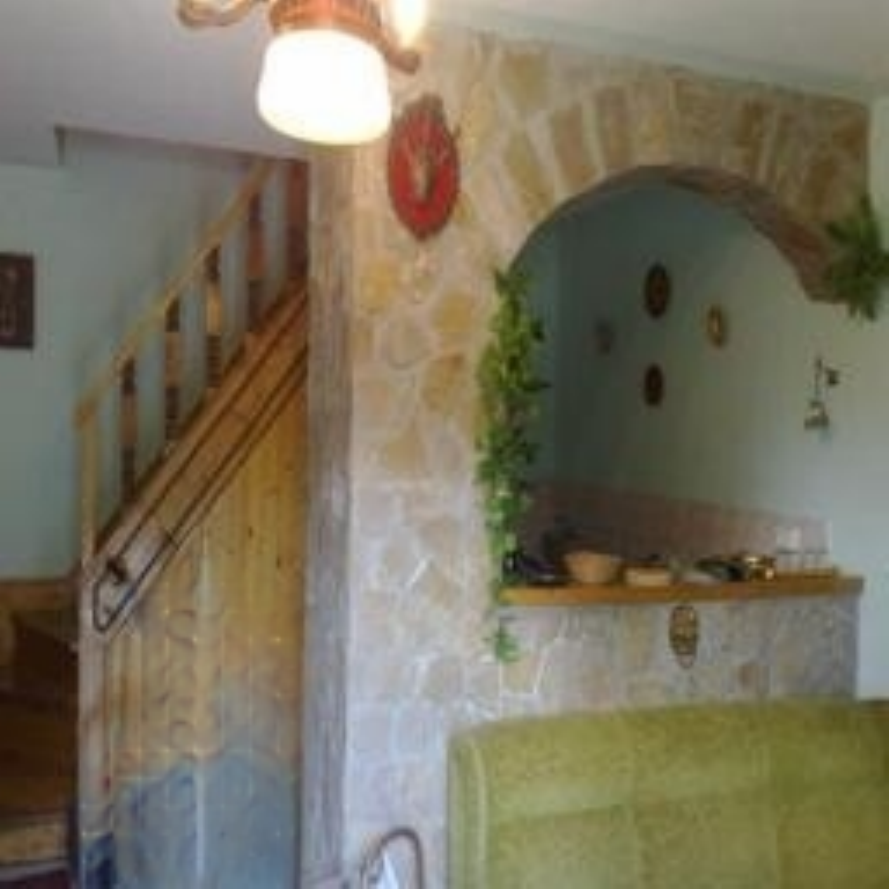}
    \end{subfigure}
    \centering
    \vspace{-0in}
    \\
    \centering
    (a) Origin  \hspace{0.9cm}
    (b) Input   \hspace{0.9cm}
    (c) Output
    \caption{Object removal results (column (c)) using our model: removing beard, watermark and kid from original images (column (a)) according to the input mask (column (b)).}
    \label{fig:remove}
    \vspace{-0.20in}
\end{figure}

\begin{table*}[htp]
    \caption{Quantitative comparisons on \textbf{contiguous} missing areas, where the \textbf{bold} indicates the best performance, and the \underline{underline} indicates the second best performance. Lower $^{\dagger}$ is better, while higher $^{\ast}$ is better.}\label{tab:comp-contiguous}
\centering
\begin{tabular}{c|c|p{6mm}p{6mm}p{6mm}p{6mm}p{6mm}|p{6mm}p{6mm}p{6mm}p{6mm}p{6mm}|p{6mm}p{6mm}p{6mm}p{6mm}p{6mm}}
    \toprule
    % column titles
    \multirow{2}{*}{ } & \multirow{2}{*}{Mask}&\multicolumn{5}{c|}{CelebA-HQ}&\multicolumn{5}{c}{Paris Street View}&\multicolumn{5}{c}{Places2}\\
    && CA & PConv & EC & PIC & ours & CA & PConv & EC & PIC & Ours & CA & PConv & EC & PIC & Ours  \\
    \midrule
    PSNR$^\ast$    
            & 0-10\% & 24.72 & 27.97 & \underline{28.87} & 26.38 & \textbf{30.01} & 25.49 & 24.98 & \underline{30.02} & 28.52 & \textbf{30.26} & 23.51 & \underline{28.26} & 25.61 & 26.16 & \textbf{29.28}\\
            & 10-20\% & 22.51 & 25.18 & \underline{26.20} & 24.33 & \textbf{26.76} & 22.645 & 23.02& \textbf{27.60} & 25.27 & \underline{27.41} & 20.48 & \underline{24.57} & 22.84 & 22.40 & \textbf{25.09}\\
            & 20-30\% & 21.67 & 23.59 & \underline{24.74} & 23.48 & \textbf{25.36} & 20.67 & 21.65 & \textbf{25.91} & 23.35 & \underline{25.46} & 18.68 & \underline{22.34} & 20.70 & 20.27 & \textbf{22.61}\\
            & 30-40\% & 20.58 & 22.45 & \underline{23.64} & 22.58 & \textbf{24.20} & 19.82 & 20.63 & \textbf{24.73} & 22.53 & \underline{24.27}  & 17.39 & \underline{20.83} & 19.11 & 18.94 & \textbf{21.11}\\
            & 40-50\% & 19.18 & 21.37 & \underline{22.57} & 21.39 & \textbf{23.02} & 18.41 & 20.16 & \textbf{23.64} & 21.53 & \underline{23.20}  & 16.34 & \underline{19.59} & 17.71 & 17.92 & \textbf{20.03}\\
    \midrule
    $\ell_1^\dagger$($10^{-3}$)
            & 0-10\% & 19.75 & 15.01 & \underline{8.60} & 17.76 & \textbf{7.37} &13.94 & 43.84 & \underline{8.50} & 11.32 & \textbf{7.46} & 26.27 & \textbf{10.00} & 27.06 & 15.50 & \underline{10.05}\\
            & 10-20\% & 28.59 & 21.78 & \underline{14.09} & 23.80 & \textbf{12.71} & 24.96 & 48.46 & \underline{14.28} & 19.87 & \textbf{13.72} & 39.76 & \underline{19.56} & 35.79 & 27.82 & \textbf{18.81}\\
            & 20-30\% & 35.09 & 28.42 & \underline{19.18} & 28.18 & \textbf{17.73} & 36.12 & 53.31 & \textbf{19.01} & 27.17 & \underline{19.54} & 53.19 & \underline{29.32} & 47.04 & 40.35 & \textbf{28.11}\\
            & 30-40\% & 43.42 & 35.53 & \underline{24.74} & 33.34 & \textbf{22.63} & 45.20 & 59.63 & \textbf{24.47} & 33.95 & \underline{25.27} & 67.12 & \underline{38.89} & 59.45 & 52.60 & \textbf{37.13}\\
            & 40-50\% & 55.72 & 42.96 & \underline{31.35} & 40.67 & \textbf{29.01} &59.67 & 63.13 & \textbf{30.89} & 42.58 & \underline{32.23} & 82.64 & \underline{49.39} & 74.91 & 65.62 & \textbf{46.50}\\
    \midrule
    $\ell_2^\dagger$($10^{-3}$)
            & 0-10\% & 4.52 & 2.52 & \underline{2.37} & 3.35 & \textbf{2.02} & 4.05 & 7.66 & \underline{1.70} & 2.78 & \textbf{1.56} & 6.27 & \underline{2.31} & 3.66 & 3.69 & \textbf{2.18}\\
            & 10-20\% & 7.02 & 3.92 & \underline{3.37} & 4.84 & \textbf{3.05} & 7.75 & 8.79 & \underline{3.25} & 5.48 & \textbf{3.19} & 11.47 & \underline{4.75} & 6.49 & 7.60 & \textbf{4.61}\\
            & 20-30\% & 8.19 & 5.03 & \underline{4.06} & 5.50 & \textbf{3.60} & 10.72 & 9.94 & \textbf{3.69} & 6.99 & \underline{4.09} & 16.45 & \underline{7.32} & 9.93 & 11.42 & \textbf{7.19}\\
            & 30-40\% & 10.10 & 6.28 & \underline{4.97} & 6.42 & \textbf{4.46} & 13.02 & 11.26 & \textbf{4.76} & 8.41 & \underline{5.17} & 21.45 & \underline{9.91} & 13.80 & 14.95 & \textbf{9.58}\\
            & 40-50\% & 13.53 & 7.97 & \underline{6.20} & 8.21 & \textbf{5.69} & 17.63 & 11.75 & \textbf{5.80} & 10.15 & \underline{6.33} & 27.03 & \underline{13.02} & 18.67 & 18.56 & \textbf{12.04}\\
    \midrule
    SSIM$^\ast$    
            & 0-10\% & 0.944 & 0.958 & \underline{0.962} & 0.930 & \textbf{0.964} & 0.941 & 0.925 & \textbf{0.958} & 0.949 & \underline{0.957} & 0.936 & \textbf{0.953} & 0.894 & 0.936 & \underline{0.950}\\
            & 10-20\% & 0.893 & 0.914 & \underline{0.923} & 0.892 & \textbf{0.927} & 0.887 & 0.880 & \textbf{0.919} & 0.902 & \underline{0.918} & 0.870 & \textbf{0.900} & 0.826 & 0.868 & \underline{0.894}\\
            & 20-30\% & 0.841 & 0.866 & \underline{0.881} & 0.855 & \textbf{0.889} & 0.822 & 0.823 & \textbf{0.874} & 0.845 & \underline{0.867} & 0.799 & \textbf{0.841} & 0.747 & 0.791 & \underline{0.830}\\
            & 30-40\% & 0.784 & 0.820 & \underline{0.839} & 0.814 & \textbf{0.850} & 0.768 & 0.771 & \textbf{0.833} & 0.797 & \underline{0.825}  & 0.729 & \textbf{0.783} & 0.665 & 0.716 & \underline{0.768}\\
            & 40-50\% & 0.714 & 0.770 & \underline{0.792} & 0.765 & \textbf{0.804} & 0.696 & 0.714 & \textbf{0.782} & 0.731 & \underline{0.773} & 0.652 & \textbf{0.722} & 0.574 & 0.634 & \underline{0.702}\\
    \midrule
    FID$^\dagger$
            & 0-10\% & 4.93 & 8.30 & 8.77 & \underline{2.76} & \textbf{1.57} & 23.94 & 30.77 & 20.09 & \underline{16.00} & \textbf{13.28} & 1.77 & 2.96 & 2.89 & \underline{1.32} & \textbf{0.79}\\
            & 10-20\% & 9.15 & 10.77 & 12.01 & \underline{4.51} & \textbf{3.35} & 41.07 & 42.39& 26.14 &\underline{24.13} &  \textbf{19.79} & 5.25 & 4.81 & 6.74 & \underline{4.61} & \textbf{2.51}\\
            & 20-30\% & 12.75 & 14.37 & 17.32 & \underline{5.45} & \textbf{5.22} & 80.59 & 80.22& \underline{42.63} & 47.26 & \textbf{41.42} & 12.21 & \underline{8.47} & 16.19 & 11.34 & \textbf{6.05}\\
            & 30-40\% & 18.00 & 18.56 & 23.11 & \textbf{7.09} & \underline{7.61} & 102.27 & 105.63 & \underline{52.15} & 57.66 & \textbf{51.89}  & 21.75 & \underline{12.76} & 30.83 & 20.13 & \textbf{10.65}\\
            & 40-50\% & 32.20 & 23.57 & 27.66 & \textbf{9.25} & \underline{10.21} & 132.18 & 139.49 & \underline{64.25} & 77.02 & \textbf{68.45} & 34.69 & \underline{18.03} & 52.96 & 32.46 & \textbf{16.21}\\
    \midrule
     \iffalse
    Perceptual$^\dagger$
            & 0-10\% & 81.00 & \textbf{39.65} & 52.36 & 115.90 & 51.59 & 87.39 & 75.52 & 66.10 & 81.88 & \textbf{62.57} & 102.87 & \textbf{41.26} & 172.83 & 106.57 & 51.84\\
            & 10-20\% & 145.76 & \textbf{70.66} & 96.79 & 156.65 & 96.56 & 150.63 & \textbf{104.75} & 107.88& 136.02 & 110.03 & 195.72 & \textbf{82.24} & 242.29 & 191.24 & 95.54\\
            & 20-30\% & 213.22 & \textbf{105.13} & 145.60 & 196.33 & 143.68 & 229.03 & \textbf{140.52} & 162.24 & 206.40 & 174.93 & 294.46 & \textbf{128.79} & 334.52 & 286.37 & 144.49\\
            & 30-40\% & 288.93 & \textbf{139.67} & 195.54 & 240.13 & 193.47 & 293.04 & \textbf{176.99} & 207.91 & 261.50 & 227.12  & 395.08 & \textbf{174.86} & 430.65 & 379.02 & 192.45\\
            & 40-50\% & 381.15 & \textbf{176.17} & 251.16 & 294.18 & 260.55 & 379.39 & \textbf{213.31} & 266.92 & 339.73 & 295.09 & 503.12 & \textbf{224.81} & 547.54 & 476.01 & 242.60
    \fi
    %\bottomrule
    \end{tabular}
    \vspace{-0.10in}
\end{table*}

\begin{table*}[hp]
    \caption{Quantitative comparisons on \textbf{discontiguous} missing areas, where the \textbf{bold} indicates the best performance, and the \underline{underline} indicates the second best performance. Lower $^{\dagger}$ is better, while higher $^{\ast}$ is better.}\label{tab:comp-discontiguous}
\centering
\begin{tabular}{c|c|p{6mm}p{6mm}p{6mm}p{6mm}p{6mm}|p{6mm}p{6mm}p{6mm}p{6mm}p{6mm}|p{6mm}p{6mm}p{6mm}p{6mm}p{6mm}}
    \toprule
    % column titles
    \multirow{2}{*}{ } & \multirow{2}{*}{Mask}&\multicolumn{5}{c|}{CelebA-HQ}&\multicolumn{5}{c}{Paris Street View}&\multicolumn{5}{c}{Places2}\\
    && CA & PConv & EC & PLU & ours & CA & PConv & EC  &PLU & ours & CA & PConv & EC  & PLU & ours  \\
    \midrule
    PSNR$^\ast$    
            & 0-10\% & 33.39 & 38.16 &\underline{38.64}& 34.24 & \textbf{40.11}& \underline{43.85} & 42.35 & 37.63 &38.25& \textbf{49.73} &  \underline{36.13} & 30.41 & 30.32 &  35.10& \textbf{42.22}\\
            & 10-20\% & 26.17 & 31.00 & \underline{31.56} & 30.64 &\textbf{32.56} & 25.90 & 29.40 & \underline{30.93} &30.37& \textbf{32.19}&  22.97 & 26.93 & 26.92 & \underline{27.42}& \textbf{29.31}\\
            & 20-30\% & 23.54 & 28.35 & \underline{28.87} & 28.48 & \textbf{29.87} & 22.89 & 26.58 & \underline{28.34} & 27.25&\textbf{29.09} &  20.26 & 24.80 & \underline{24.91} &  24.74& \textbf{26.55}\\
            & 30-40\% & 21.67 & 26.42 & \underline{27.01} &26.76& \textbf{27.95} & 21.64 & 25.11 & \underline{26.44} & 25.74 & \textbf{27.70} &  18.47 & 23.14 & \underline{23.37} & 22.89&\textbf{24.98} \\
            & 40-50\% & 20.19 & 24.88 & \underline{25.46} & 25.28& \textbf{26.40}& 19.93 & 23.36 & \underline{25.03} & 24.06& \textbf{26.21} &  17.09 &21.71 & \underline{22.06} & 21.37& \textbf{23.67} \\
    \midrule
    $\ell_1^\dagger$($10^{-3}$)
            & 0-10\% & 11.03 & \underline{2.70} & \textbf{2.11} &  9.64 &4.83 & 11.50 & \underline{3.18} & \textbf{2.39} & 3.98 &7.69 & 17.40 & 18.94 & 18.82 & \underline{4.72}& \textbf{4.70}\\
            & 10-20\% & 20.72 & 8.34 & \textbf{6.58} &  13.21 & \underline{8.33}& 22.30 & 12.92 & \textbf{7.60} &   9.84& \underline{11.75}& 32.50 & 24.49 & 24.08 &  \underline{12.65}&\textbf{11.00}\\
            & 20-30\% & 30.79 & 14.17 & \textbf{11.30} &  17.14& \underline{12.03}& 34.50 & 19.59 & \textbf{12.91} &  16.67 & \underline{16.43} & 47.76 & 30.48 & 29.62 &\underline{21.03}& \textbf{16.14}\\
            & 30-40\% & 20.57 & 23.40 & \underline{16.48} & 21.70 & \textbf{16.15} & 43.81 & 26.88 & \textbf{19.20} &  22.48& \underline{20.21} & 63.63 & 37.25 & 35.74 & \underline{30.15}&\textbf{22.30}\\
            & 40-50\% & 27.56 & 29.70 & \underline{22.30} & 26.93 & \textbf{20.68} & 58.37 & 37.49 & \underline{25.81} &  30.70& \textbf{25.24} & 80.36 & 45.23 & 42.67 & \underline{40.32}& \textbf{28.82}\\
    \midrule
    $\ell_2^\dagger$($10^{-3}$)
            & 0-10\% & 1.05 & 0.28 & \underline{0.25} &  0.49 & \textbf{0.23} & 1.12 & 0.34 & \underline{0.32} &  0.34& \textbf{0.27} & 2.20 & 1.14 & 1.17 & \underline{0.69}&\textbf{0.54}\\
            & 10-20\% & 3.08 & 0.92 & \underline{0.82} & 1.01 & \textbf{0.70}& 3.43 & 1.50 & \underline{1.07} &  1.29& \textbf{0.89}& 6.90 & 2.50 & 2.53 & \underline{2.23}& \textbf{1.65}\\
            & 20-30\% & 5.32 & 1.64 & \underline{1.48} &  1.62 & \textbf{1.23}& 6.50 & 2.57 & \underline{1.83} &  2.55 & \textbf{1.64} & 11.92 & 4.04 & \underline{4.00} & 4.07& \textbf{2.58}\\
            & 30-40\% & 7.97 & 2.51 & \underline{2.23} &  2.37 & \textbf{1.85} & 8.31 & 3.53 & \underline{3.08} &  3.52 & \textbf{2.28} & 17.34 & 5.85 & \underline{5.66} & 6.18& \textbf{4.17}\\
            & 40-50\% & 11.01 & 3.55 & \underline{3.16} &  3.31 & \textbf{2.60} & 12.29 & 5.24 & \underline{3.94} & 5.03 & \textbf{3.15}& 23.25 & 8.07 & \underline{7.58} &8.69& \textbf{5.41}\\
    \midrule
    SSIM$^\ast$
            & 0-10\% &0.967 & 0.979 & \underline{0.983} & 0.954 &\textbf{0.985}  & 0.969 & \underline{0.981} & \underline{0.981} & 0.980 & \textbf{0.986} & 0.965 & 0.924 & 0.925 & \underline{0.973}& \textbf{0.981} \\
            & 10-20\% & 0.903 & 0.940 & \underline{0.948} &  0.922 & \textbf{0.955} & 0.898 & 0.934 & \underline{0.940} &  0.933& \textbf{0.952}& 0.888 & 0.880 & 0.881 &  \underline{0.915} & \textbf{0.945}\\
            & 20-30\% & 0.838 & 0.901 & \underline{0.912} &  0.888 & \textbf{0.925} & 0.821 & 0.883 & \underline{0.899} & 0.880 & \textbf{0.914}& 0.811 & 0.834 & 0.836 &  \underline{0.854} & \textbf{0.911}\\
            & 30-40\% & 0.767 & 0.858 & \underline{0.873} & 0.849 & \textbf{0.891} & 0.757 & 0.836 & \underline{0.849} & 0.832 & \textbf{0.881} & 0.730 & 0.784 & 0.788 &  \underline{0.791}& \textbf{0.865}\\
            & 40-50\% & 0.694 & 0.813 & \underline{0.831} & 0.807 & \textbf{0.854} & 0.673 & 0.779 & \underline{0.798} & 0.768& \textbf{0.838}& 0.647 & 0.728 & 0.736 & \underline{0.724}& \textbf{0.810}\\
    \midrule
    FID$^\dagger$
            & 0-10\% & 1.26 & 1.15 & 0.94 &  \underline{0.92} & \textbf{0.75} & 25.54 & 15.58 & 7.30 & \underline{5.97} & \textbf{4.08}& 1.26 & 1.75 & 1.38 & \underline{0.81}& \textbf{0.02}\\
            & 10-20\% & 8.73 & 3.39 & 2.78 &  \underline{2.85} & \textbf{2.25} & 72.81 & 28.09 & 23.35 & \underline{20.40}& \textbf{14.05} & 8.73 & 2.10 & \underline{1.80} & 3.34& \textbf{0.13}\\
            & 20-30\% & 20.35 & 5.87 & 4.68 &  \underline{4.83} & \textbf{3.81} & 116.92 & 44.58 & 41.25 & \underline{35.65}& \textbf{25.39} & 20.35 & 2.88 & \underline{2.69} & 7.34&\textbf{0.29}\\
            & 30-40\% & 36.53 & 8.84 & 6.75 & \underline{7.23}& \textbf{5.34}& 155.64 & 57.00 & 61.94 & \underline{47.39}& \textbf{33.49}& 36.53 & \underline{4.31} & 4.36 & 13.52&\textbf{0.78}\\
            & 40-50\% & 57.60 & 12.38 & 9.60 &\underline{10.28} & \textbf{7.30}& 200.15 & 78.78 & 88.19 & \underline{66.20} & \textbf{45.06} & 57.60 & \underline{6.97} & 7.38 & 22.49&\textbf{2.67}\\
    \midrule
    \iffalse
    Perceptual$^\dagger$
            & 0-10\% & 81.58 & 48.73 & 33.07 &   & 59.60 & 95.10 & 38.37 &   & 81.58 & 128.64 & 126.98 & \\
            & 10-20\% & 220.77 & 115.00 & 92.82 &   & 178.31 & 163.90 & 108.22 &  & 220.77 & 193.84 & 192.50 & \\
            & 20-30\% & 348.93 & 176.06 & 148.02 &   & 294.37 & 230.91 & 174.80 &   & 348.93 & 258.47 & 255.98 & \\
            & 30-40\% & 471.10 & 236.14 & 201.73 &   & 382.87 & 289.68 & 241.09 &   & 471.10 & 326.36 & 321.03 & \\
            & 40-50\% & 587.90 & 295.38 & 256.34 &  & 485.33 & 353.98 & 310.27 &  & 587.90 & 401.07 & 389.19 &
    \fi
    %\bottomrule
    \end{tabular}
    \vspace{-0.10in}
\end{table*}

\subsubsection{Quantitative Results} \label{sec:quant}
Table \ref{tab:comp-contiguous}, \ref{tab:comp-discontiguous} list the results of all methods on CelebA-HQ, Paris Street View and Place2 in terms of different metrics, with respect to contiguous and discontiguous missing areas of different sizes. First, from the table we can observe that comparing to discontiguous missing areas, all the methods show obvious degradation on contiguous missing areas, which proves that it is much more difficult to restore the contiguous missing areas than discontiguous ones, which means that for the satisfying performance, the inpainting models need to be guided by well-designed regularization. Second, we can easily get the conclusion that our proposed method can achieve superior performance on both discontiguous and contiguous masks in most cases, and keep comparatively stable performance on the two types of masks. Moreover, as the missing area gradually increases, all the methods perform worse in terms of all metrics. But compared to others, our method consistently obtains the best performance in most cases, and decreases the performance much more slowly when the mask size enlarges. This means that our method can stably and robustly infer the missing contents, especially for input images with large missing regions.

In terms of PSNR, $\ell_1$ and $\ell_2$ errors, the superior performance of our method over the others further proves that our framework enjoys strong capability of generating more detailed contents for better visual quality. On discontiguous missing areas, we achieve the best performance at most cases, except for small missing areas (0-40\%) on CelebA-HQ and Paris Street View according to $\ell_1$, where EC and PConv have a narrow lead over us in certain cases. However, we achieve the best according to $\ell_2$, which means we restore the detailed information due to the correlation loss, as $\ell_2$ is sensitive to outliers. That is to say, our method could well restore the missing information in pixel-level and recover the detailed information thanks to the correlation loss. Usually EC is the second best among all the methods. However, on contiguous missing areas, it achieves comparative performance as ours on Paris Street View dataset, while PConv performs even better than EC and achieves the second best on Places2 dataset. 

Besides, in terms of FID, our method obviously achieves much more significant improvement over the state-of-the-art methods like PConv, EC and PIC, which indicates that the proposed adversarial framework can pursue more semantically meaningful contents for missing regions. PLU also shows relatively good performance than other state-of-the-art approaches according to FID, due to their guidance of KL divergence, which is only inferior to ours in most cases.

\subsection{Unwanted Object Removal}
Unwanted object removal is one of the most useful applications of image inpainting. Therefore, we also study the performance of our method in this task, and show several examples in Figure \ref{fig:remove}. Our model has the capability of removing watermark or editing images. It is obvious that the inpainting images seem very natural and harmonious, even the unwanted objects appear with complex shapes and backgrounds.

\section{Conclusion}\label{sec:5}
We propose a generic inpainting framework capable of handling the images with both contiguous and discontiguous missing areas at the same time, in an adversarial manner, where region-wise convolution layers are deployed in both generator and discriminator to locally handle the different regions, namely existing regions and missing ones, which could synthesis semantically reasonable and visually realistic inpainting results. The correlation loss is proposed to guide the generator to capture the non-local semantic relation between patches inside the image, and further provides more information to inference. We show that our proposed method is able to restore meaningful contents for missing regions and connects existing and missing regions naturally and thus significantly improves inpainting results. Furthermore, we demonstrate that our inpainting framework can edit face, clear watermarks, and remove unwanted objects in practical applications. Extensive experiments on various datasets such as faces, street views and natural scenes demonstrate that our proposed method can significantly outperform other state-of-the-art approaches in image inpainting.

\section*{Acknowledgment}
This work was supported by National Natural Science Foundation of China (61690202, 61872021), Fundamental Research Funds for Central Universities (YWF-19-BJ-J-271), Beijing Municipal Science and Technology Commission (Z171100000117022), State Key Lab of Software Development Environment (SKLSDE-2018ZX-04), and Australian Research Council Projects: FL-170100117, DE-180101438, DP-180103424, and
LP-150100671.

\textbf{}

% or

\iffalse

\appendices
\section{Proof of the First Zonklar Equation}
Appendix one text goes here.

\section{}
Appendix two text goes here.
\fi
\iffalse
\ifCLASSOPTIONcompsoc
  \section*{Acknowledgments}
\else

\fi

The authors would like to thank...
\fi

\ifCLASSOPTIONcaptionsoff
  \newpage
\fi



%\bibliographystyle{IEEEtran}
\bibliographystyle{IEEEtran}
\bibliography{ijcai19.bib}


\end{document}